\documentclass[10pt]{article} %

\usepackage[preprint]{rlj}           %

\usepackage{amssymb}            %
\usepackage{mathtools}          %
\usepackage{mathrsfs}           %
\usepackage{graphicx}           %
\usepackage{subcaption}         %
\usepackage[space]{grffile}     %
\usepackage{url}                %
\usepackage{lipsum}             %

\usepackage{hyperref}
\usepackage[dvipsnames]{xcolor}
\usepackage{wrapfig}
\usepackage{booktabs}
\usepackage{float}
\usepackage{enumitem}
\usepackage{ulem}      %
\usepackage{tabularx} %
\usepackage{array}    %
\usepackage{listings}
\usepackage{amssymb}            %
\usepackage{mathtools}          %
\usepackage{mathrsfs}           %
\usepackage{adjustbox}
\usepackage[space]{grffile}     %
\usepackage{float}
\usepackage{svg}
\usepackage{amsmath}
\usepackage{cleveref}
\usepackage{fancyhdr} 
\usepackage{algorithm}
\usepackage{algpseudocode}
\usepackage{xspace}
\usepackage{pifont}
\usepackage{caption}

\newcommand{\greencheck}{\textcolor{green}{\ding{51}}} %
\newcommand{\redx}{\textcolor{red}{\ding{55}}}         %
\newcommand{\methodName}{SegDAC\xspace}

\newcommand{\vsepimg}[1]{\hspace{4pt}\rule{0.5pt}{#1}\hspace{4pt}}
\newcommand{\rolloutlabel}[1]{\vspace{4pt}\textbf{#1}\\[2pt]}

\title{\methodName: Visual Generalization in Reinforcement Learning via Dynamic Object Tokens}

\setrunningtitle{\methodName: Visual Generalization in Reinforcement Learning via Dynamic Object Tokens}

\author{Alexandre Brown\textsuperscript{1,2}, Glen Berseth\textsuperscript{1,2}}

\emails{alexandre.brown@mila.quebec, \ glen.berseth@mila.quebec}

\affiliations{
$^{1}$\textbf{Mila Quebec AI Institute}\\
$^{2}$\textbf{Université de Montréal, Canada}\\
}

\contribution{
A transformer-based actor-critic that learns stable model-free policies from a variable-length set of object tokens whose size and content change at every timestep, robust to natural variation in token count and identity without reconstruction, auxiliary losses, or data augmentation.
}
{
Prior object-centric RL methods use fixed-size slot representations or require reconstruction objectives or data augmentation to stabilize learning \citep{10.5555/3495724.3496691, yarats2021masteringvisualcontinuouscontrol}. Robustness to natural variation in segment count and identity during online RL training has not been explicitly addressed.
}

\contribution{
A method to construct contextual per-object tokens from frozen pretrained vision models, with segment positional encoding to preserve spatial grounding without ground-truth masks or finetuning.
}
{
Prior segmentation-based RL methods either require ground-truth masks 
at setup time or operate on recomposed pixel images rather than 
structured object-level representations \citep{wang2023generalizablevisualreinforcementlearning, Chen_Xu_Liao_Ding_Zhang_Yu_Zhao_2024}.
}

\contribution{
An empirical evaluation across 8 manipulation tasks showing that \methodName 
improves over prior visual generalization methods by up to 88\% while 
matching DrQ-v2 in sample efficiency, without data augmentation or 
auxiliary losses.
}
{
Methods strong in visual generalization typically sacrifice sample 
efficiency, and vice versa \citep{almuzairee2024recipe}.
}

\contribution{
A visual generalization benchmark on 8 ManiSkill3 tasks with 12 
perturbation types across 3 difficulty levels organized by a scene 
entity taxonomy.
}
{
Prior benchmarks primarily use locomotion and simple control tasks with 
limited visual variation \citep{hansen2021stabilizingdeepqlearningconvnets,almuzairee2024recipe}.
}


\summary{
Visual reinforcement learning policies trained on pixel observations often struggle to generalize when visual conditions change at test time. Object-centric representations are a promising alternative, but most approaches use fixed-size slot representations, require image reconstruction, or need auxiliary losses to learn object decompositions. As a result, it remains unclear how to learn RL policies directly from object-level inputs without these constraints.
We propose \textbf{\methodName}, a \textbf{Seg}mentation-\textbf{D}riven \textbf{A}ctor-\textbf{C}ritic that operates on a variable-length set of object token embeddings. At each timestep, text-grounded segmentation produces object masks from which spatially aware token embeddings are extracted. A transformer-based actor-critic processes these dynamic tokens, using segment positional encoding to preserve spatial information across objects. We ablate these design choices and show that both segment positional encoding and variable-length processing are individually necessary for strong performance. We evaluate \methodName on 8 ManiSkill3 manipulation tasks under 12 visual perturbation types across 3 difficulty levels. \methodName improves over prior visual generalization methods by 15\% on easy, 66\% on medium, and 88\% on the hardest settings. \methodName matches the sample efficiency of the state-of-the-art visual RL methods while achieving improved generalization under visual changes.
}

\begin{document}

\makeCover  %
\maketitle  %

\begin{abstract}
Visual reinforcement learning policies trained on pixel observations often struggle to generalize when visual conditions change at test time. Object-centric representations are a promising alternative, but most approaches use fixed-size slot representations, require image reconstruction, or need auxiliary losses to learn object decompositions. As a result, it remains unclear how to learn RL policies directly from object-level inputs without these constraints.
We propose \textbf{\methodName}, a \textbf{Seg}mentation-\textbf{D}riven \textbf{A}ctor-\textbf{C}ritic that operates on a variable-length set of object token embeddings. At each timestep, text-grounded segmentation produces object masks from which spatially aware token embeddings are extracted. A transformer-based actor-critic processes these dynamic tokens, using segment positional encoding to preserve spatial information across objects. We ablate these design choices and show that both segment positional encoding and variable-length processing are individually necessary for strong performance. We evaluate \methodName on 8 ManiSkill3 manipulation tasks under 12 visual perturbation types across 3 difficulty levels. \methodName improves over prior visual generalization methods by 15\% on easy, 66\% on medium, and 88\% on the hardest settings. \methodName matches the sample efficiency of the state-of-the-art visual RL methods while achieving improved generalization under visual changes.\\
\textbf{Project page:} \url{https://segdac.github.io/}\\\textbf{Code:} \url{https://github.com/SegDAC/SegDAC}
\end{abstract}

\section{Introduction}
\label{sec:intro}
Visual reinforcement learning from pixel observations has made significant progress in recent years, with methods like DrQ-v2 achieving strong sample efficiency on standard benchmarks \citep{mnih2013playingatarideepreinforcement,10.5555/3305890.3305968,yarats2021masteringvisualcontinuouscontrol,lepert2025shadowleveragingsegmentationmasks}. However, policies trained on pixel observations often remain brittle when visual conditions change at test time \citep{yuan2023rlvigenreinforcementlearningbenchmark}. Small changes in background textures, lighting, or object colors can cause large drops in performance, even when the underlying task structure is unchanged. Data augmentation techniques help \citep{kostrikov2021imageaugmentationneedregularizing,yarats2021masteringvisualcontinuouscontrol,10.1109/ICRA48506.2021.9561103, almuzairee2024recipe}, but they operate on raw pixels where task-relevant and task-irrelevant information are entangled, limiting their robustness to distribution shifts beyond what the augmentation covers.

Object-centric representations offer a natural way to separate task-relevant structure from visual noise \citep{wang2023generalizablevisualreinforcementlearning,10.5555/3495724.3496691,didolkar2025ctrlo}. By representing a scene as a collection of objects rather than a grid of pixels, the policy can focus on what matters: the objects, their positions, and their relationships. Prior work in object-centric RL has explored this direction using learned decomposition methods such as Slot Attention \citep{10.5555/3495724.3496691}. However, these approaches typically rely on a fixed number of object slots, require image reconstruction as a training signal, or need auxiliary losses to learn meaningful decompositions. These constraints limit their applicability to visually complex environments, and reconstruction objectives can bias representations toward visual fidelity rather than task relevance \citep{assran2023selfsupervisedlearningimagesjointembedding}.

Recent advances in pretrained vision models have made high-quality object segmentation widely accessible without task-specific finetuning \citep{kirillov2023segment}. This opens a concrete opportunity for visual RL: rather than learning from raw pixels, a policy could reason directly over per-object representations. Yet this is less straightforward than it appears. Detected segments vary in number across timesteps as objects move, occlude, or as detection confidence fluctuates, producing a token set that is unordered, variable in length, and whose identity changes at every step. A policy must remain stable under these variations during online learning, where value estimates and gradients are already noisy \citep{haarnoja2019softactorcriticalgorithmsapplications}.

We propose \textbf{\methodName}, a \textbf{Seg}mentation-\textbf{D}riven \textbf{A}ctor-\textbf{C}ritic that addresses this question. At each timestep, text-grounded segmentation produces object masks, and spatially aware token embeddings are extracted from a pretrained vision encoder. A transformer-based actor-critic then processes this variable-length token set. We identify two design choices that are critical for stable learning: (1) segment positional encoding, which injects spatial information about each object's location into the token representation, and (2) variable-length processing without padding or truncation, which allows the policy to adapt to changes in the number of detected objects. We validate through ablations that each choice is individually necessary, with the largest effects appearing on more complex manipulation tasks.

We evaluate \methodName over eight ManiSkill3 manipulation tasks, 12 perturbation types, and 3 difficulty levels \citep{taomaniskill3}. \methodName improves over prior visual generalization methods by 15\% on easy, 66\% on medium, and 88\% on the hardest settings. Unlike these methods, \methodName also matches the sample efficiency of DrQ-v2, which does not generalize well under visual changes \citep{yuan2023rlvigenreinforcementlearningbenchmark}. Notably, \methodName achieves this without data augmentation during RL training, a technique most prior visual generalization methods depend on but that can hinder training stability and efficiency \citep{10.1109/ICRA48506.2021.9561103,almuzairee2024recipe}. 

\section{Related Work}
\label{sec:related_work}
\textbf{Data augmentation for visual RL generalization.} A common approach to visual generalization applies transformations directly to pixel observations during training. DrQ~\citep{kostrikov2021imageaugmentationneedregularizing} and DrQ-v2~\citep{yarats2021masteringvisualcontinuouscontrol} use simple image shifts and achieve strong sample efficiency, but their robustness to unseen visual conditions remains limited \citep{yuan2023rlvigenreinforcementlearningbenchmark}.SVEA~\citep{hansen2021stabilizingdeepqlearningconvnets}, SODA~\citep{10.1109/ICRA48506.2021.9561103}, and SADA~\citep{almuzairee2024recipe} build on this by stabilizing Q-targets under augmentation, improving generalization within the augmentation distribution.

\textbf{Task-relevant representations and pretrained models.} Rather than augmenting observations, a second line of work extracts representations that emphasize task-relevant information. PIE-G~\citep{NEURIPS2022_548a482d} feeds observations through a frozen pretrained encoder to produce a single global feature vector, improving generalization but compressing the entire scene into one representation and losing object-level structure. Other methods learn to mask task-irrelevant regions : SGQN~\citep{NEURIPS2022_c5ee2a08} derives saliency maps from Q-function gradients, and MaDi~\citep{grooten2023madilearningmaskdistractions} learns a soft pixel mask from the reward signal alone. SMG~\citep{zhang2024focusmattersseparatedmodels} takes a more structured approach, using separated foreground and background encoder-decoder branches with cooperative reconstruction and consistency losses to disentangle task-relevant from task-irrelevant features. These methods improve robustness to visual distractors, but all produce a single fixed-size representation regardless of scene content, discarding the compositional structure of the scene.

\textbf{Object-centric and segmentation-based representations for RL.} A third line of work reasons at the level of objects rather than pixels. Slot Attention~\citep{10.5555/3495724.3496691} and its variants decompose a scene into a fixed number of object slots through an iterative attention mechanism, typically trained with a reconstruction objective. However, the slot count must be specified in advance, reconstruction losses can bias representations toward visual fidelity rather than task relevance, and the learned decompositions do not always align with task-relevant objects \citep{assran2023selfsupervisedlearningimagesjointembedding}.
Pretrained segmentation models have been used to obtain object-level structure without learning to segment from scratch. SAM-G~\citep{wang2023generalizablevisualreinforcementlearning} uses SAM~\citep{kirillov2023segment} and DINOv2~\citep{oquab2024dinov2learningrobustvisual} features to find point correspondences from a reference image and its ground-truth mask, then produces a masked image for the RL agent.
\begin{wraptable}{rh}{0.55\textwidth}
\centering
\caption{Comparison of segmentation-based visual RL methods.}
\label{tab:compare_seg_rl_methods}
\small
\begin{tabular}{@{}lccc@{}}
\toprule
& \textbf{FTD} & \textbf{SAM-G} & \textbf{\methodName} \\
\midrule
No ground-truth masks       & \greencheck & \redx       & \greencheck \\
Dynamic object tokens       & \redx       & \redx       & \greencheck \\
Pretrained vision encoder   & \greencheck & \greencheck & \greencheck \\
Text-grounded detection     & \redx       & \redx       & \greencheck \\
No auxiliary losses         & \redx       & \greencheck & \greencheck \\
No data augmentation        & \greencheck & \redx       & \greencheck \\
\bottomrule
\end{tabular}
\vspace{-0.75em}
\end{wraptable}
This improves generalization but requires a ground-truth mask at setup time, uses data augmentation during training, and does not dynamically extract per-object features. FTD~\citep{Chen_Xu_Liao_Ding_Zhang_Yu_Zhao_2024} uses prompt-free SAM segmentation to decompose the scene into segments, scores each segment's task relevance via a learned attention mechanism, and produces a weighted combination of segment pixel crops to reconstruct an image as input to a CNN policy. However, it still requires tuning a predetermined number of segments to process per task, requires auxiliary self-supervised losses, and operates on a recomposed pixel image rather than object-level tokens. Unlike these approaches, \methodName learns directly from a variable-length set of object token embeddings constructed from representations dynamically extracted by a frozen pretrained vision pipeline, and processes them jointly through a transformer-based actor-critic with segment positional encoding, without ground-truth masks, auxiliary losses, data augmentation, or image reconstruction. We summarize the key differences in~\Cref{tab:compare_seg_rl_methods}.

\section{Method}
\label{sec:method}

\begin{figure}[htb]
  \centering
  \includegraphics[width=\linewidth]{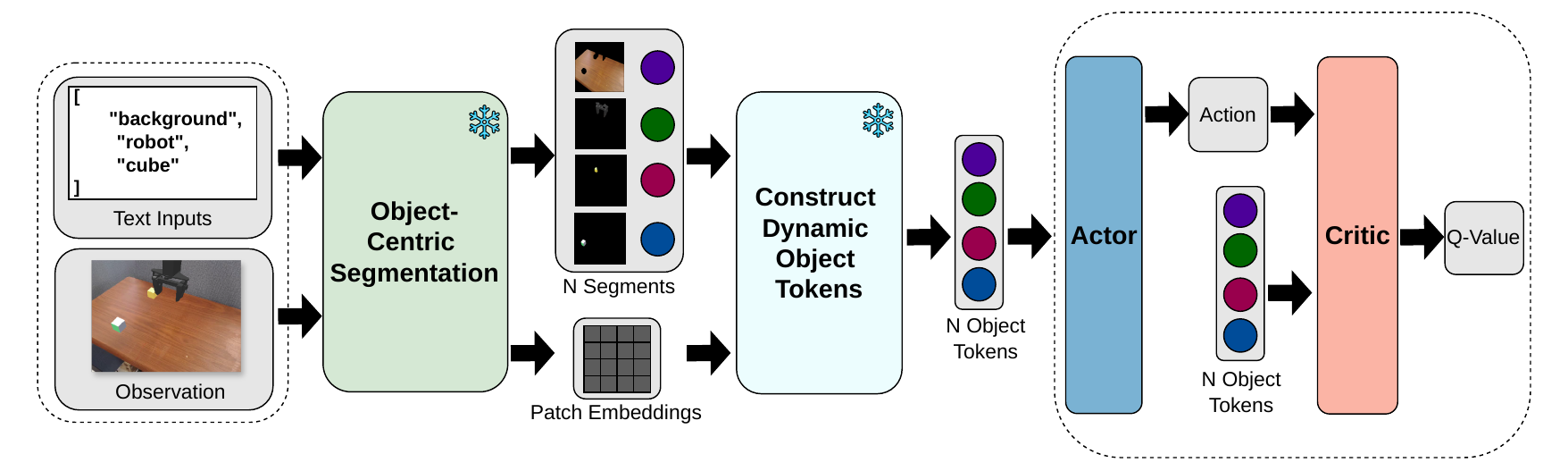}
  \caption{\methodName end-to-end pipeline from image to action/Q-value prediction.}
  \label{fig:method_overview}
\end{figure}

An overview of \methodName is shown in~\Cref{fig:method_overview}. At each time step, the agent receives a single RGB frame and a set of text inputs. A frozen segmentation pipeline decomposes the scene into a variable number of segments and produces one object token per segment (\Cref{sec:grounded_segmentation,sec:segment_embeddings}). A transformer-based actor-critic then processes these object tokens alongside proprioception to predict actions and Q-values (\Cref{sec:actor_critic}). The entire system trains with the standard SAC loss alone, without reconstruction, auxiliary objectives, data augmentation, or ground-truth masks. While we use SAC as the base RL algorithm due to its popularity in visual RL for continuous control \citep{yarats2020improvingsampleefficiencymodelfree,laskin2020reinforcementlearningaugmenteddata,kostrikov2021imageaugmentationneedregularizing,10.1109/ICRA48506.2021.9561103,hansen2021stabilizingdeepqlearningconvnets,NEURIPS2022_c5ee2a08,NEURIPS2022_548a482d,grooten2023madilearningmaskdistractions,Chen_Xu_Liao_Ding_Zhang_Yu_Zhao_2024,almuzairee2024recipe}, our framework can be adapted for other model-free algorithms by changing the output dimension of the projection head to match the desired action space.

Three design problems must be solved to make this pipeline work for online RL. First, the scene must be decomposed into semantically meaningful regions fast enough for real-time data collection. Second, each region must be represented as a compact embedding that captures both local detail and scene-level context. Third, the actor-critic must handle a set of tokens whose size and content changes at every time step and learn which tokens matter without supervision, while remaining stable for learning. The following subsections address each problem in turn.

\subsection{From Pixels to Object Masks}
\label{sec:grounded_segmentation}

\begin{figure}[t]
\centering
\includegraphics[width=\linewidth]{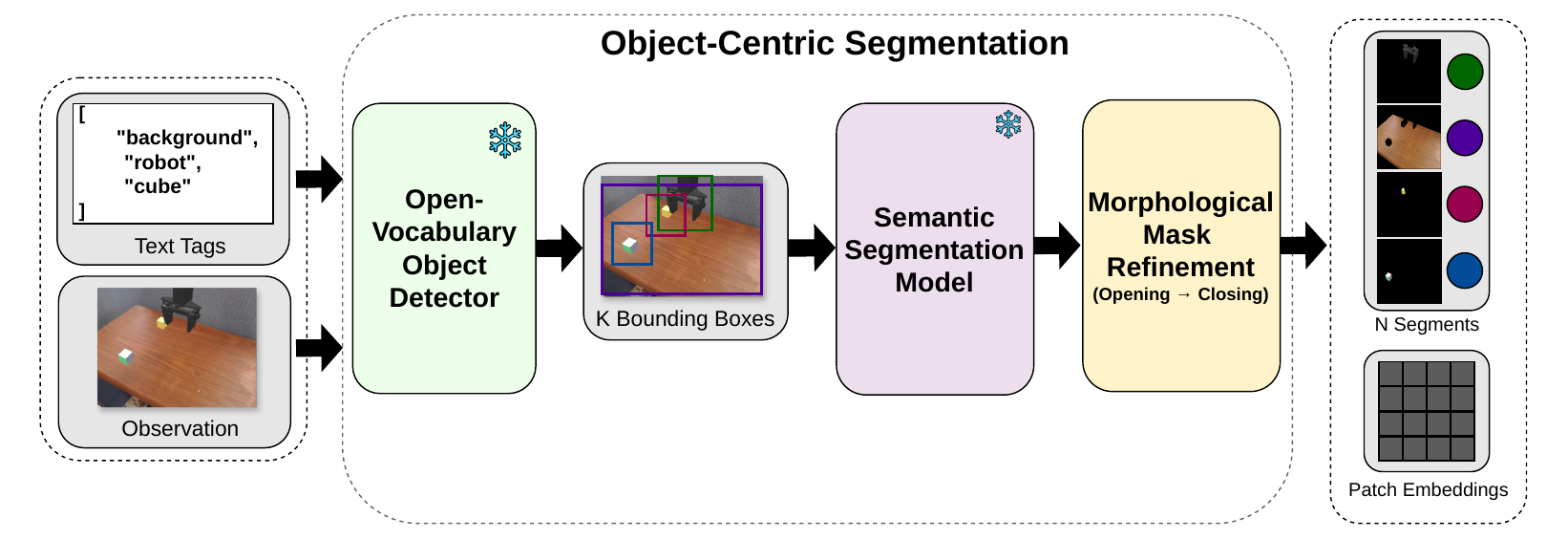}
\caption{Object-Centric Segmentation pipeline. Text concept names guide an open-vocabulary detector to propose bounding boxes, which prompt a semantic segmentation model to produce instance masks. Morphological mask refinement (opening followed by closing) then improves mask quality without the latency overhead of iterative refinement methods used in prior segmentation-based RL work. The number of output segments $N$ varies with scene content.}
\label{fig:grounded_seg}
\end{figure}

The first step of our pipeline is to decompose each frame into (sub)object-level regions that are semantically grounded and produced fast enough for online RL. Prior segmentation-based RL methods either use prompt-free segmentation, which is slow and requires heavy mask post-processing via iterative mask refinement, or rely on ground-truth masks provided offline \citep{wang2023generalizablevisualreinforcementlearning,Chen_Xu_Liao_Ding_Zhang_Yu_Zhao_2024}. We instead adopt a text-grounded approach: a lightweight open-vocabulary detector proposes bounding boxes from a short list of concept words (e.g., ``robot'', ``cube'', ``background''), and a semantic segmentation model produces masks within those boxes (\Cref{fig:grounded_seg}). Both models remain frozen throughout training. Rather than the iterative mask refinement used in prior RL segmentation work, we design a lightweight morphological mask refinement to each mask which consists of morphological opening followed by morphological closing \citep{Sreedhar_2012}, which removes small spurious artifacts and fills boundary gaps while adding negligible latency, allowing the full pipeline to remain fast enough for online RL. We provide qualitative examples and implementation details of our fast morphological mask refinement in~\Cref{appendix_morphological_mask_refinement}.

\begin{figure}[thb]
    \centering
    \includegraphics[height=0.146\linewidth]{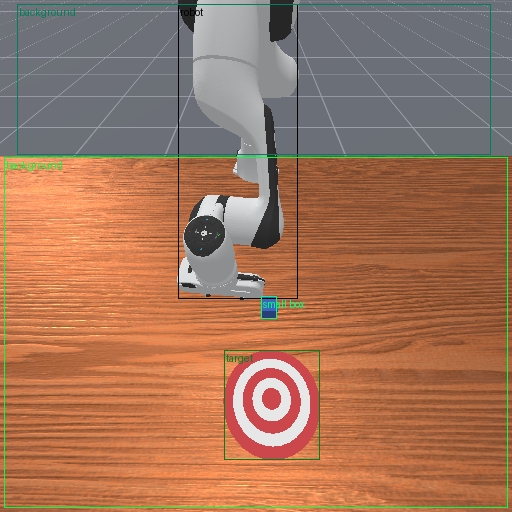}
    \includegraphics[height=0.146\linewidth]{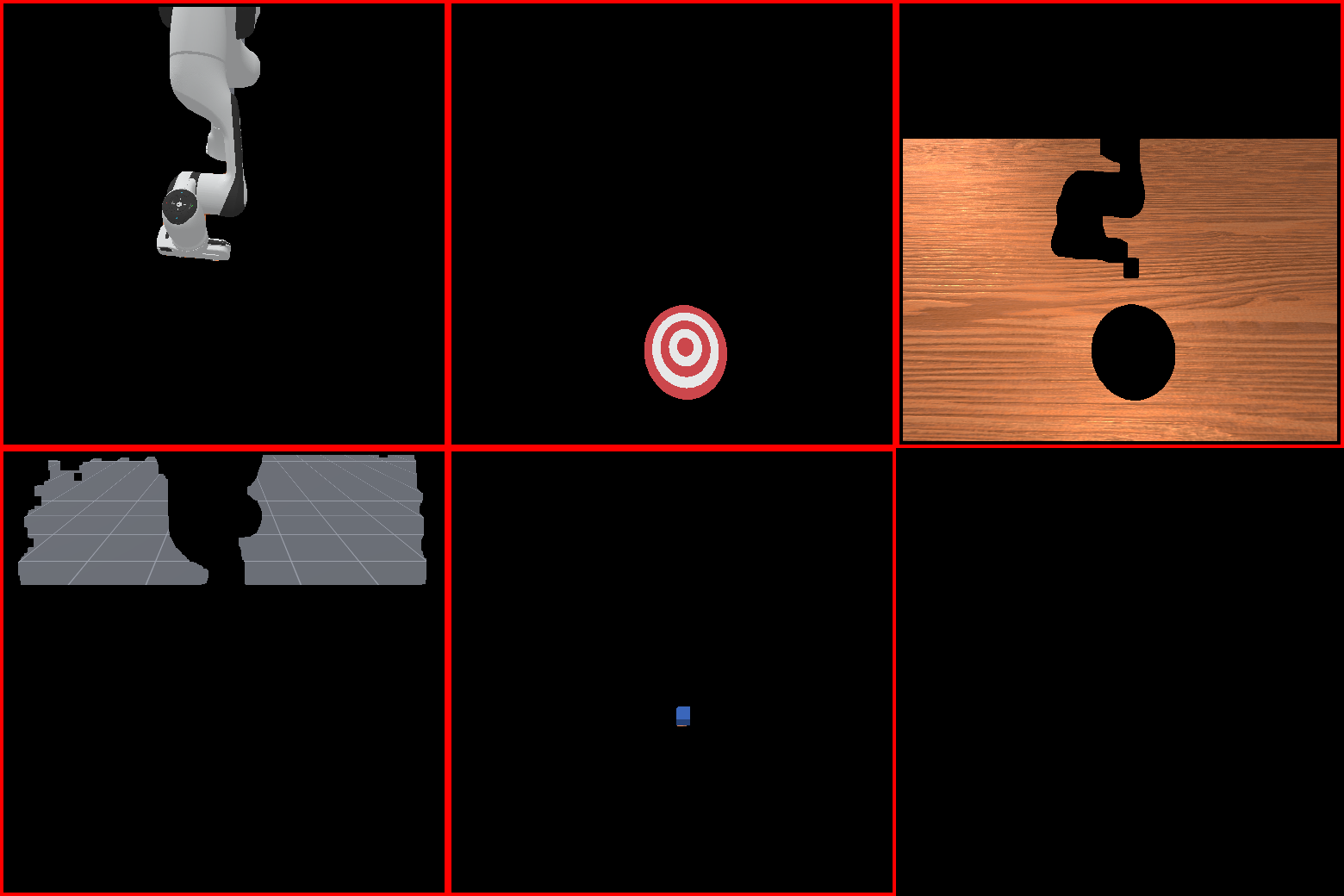}
    \includegraphics[height=0.146\linewidth]{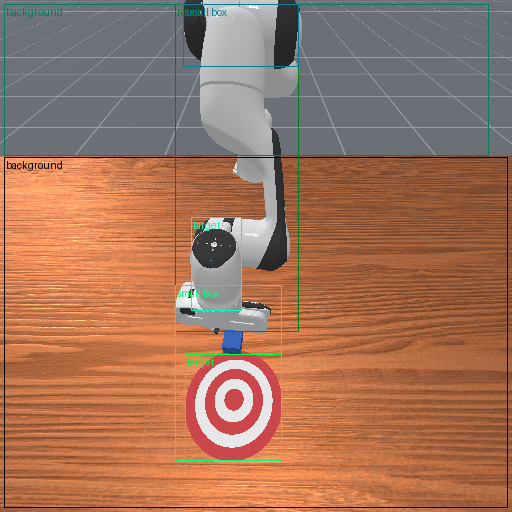}
    \includegraphics[height=0.146\linewidth]{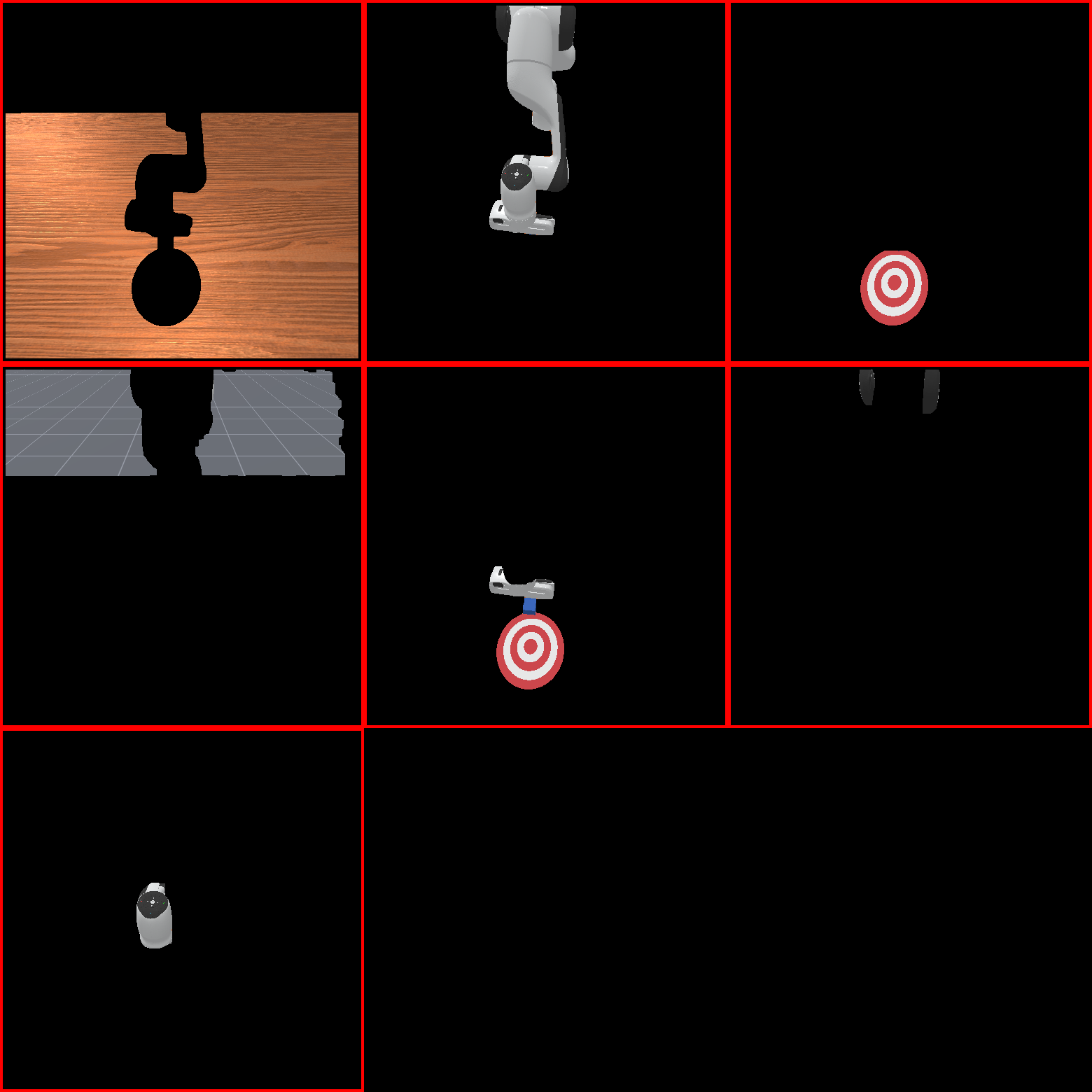}
    \includegraphics[height=0.146\linewidth]{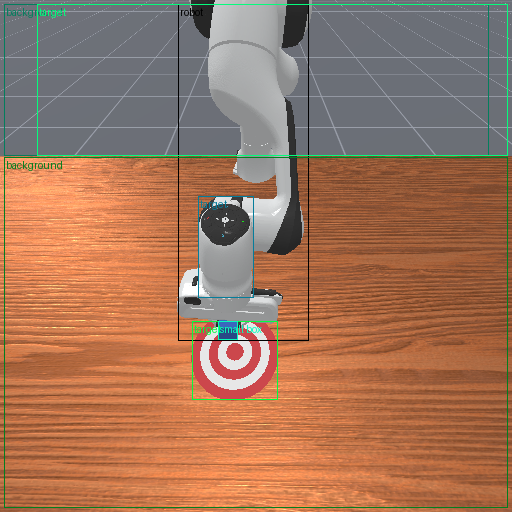}
    \includegraphics[height=0.146\linewidth]{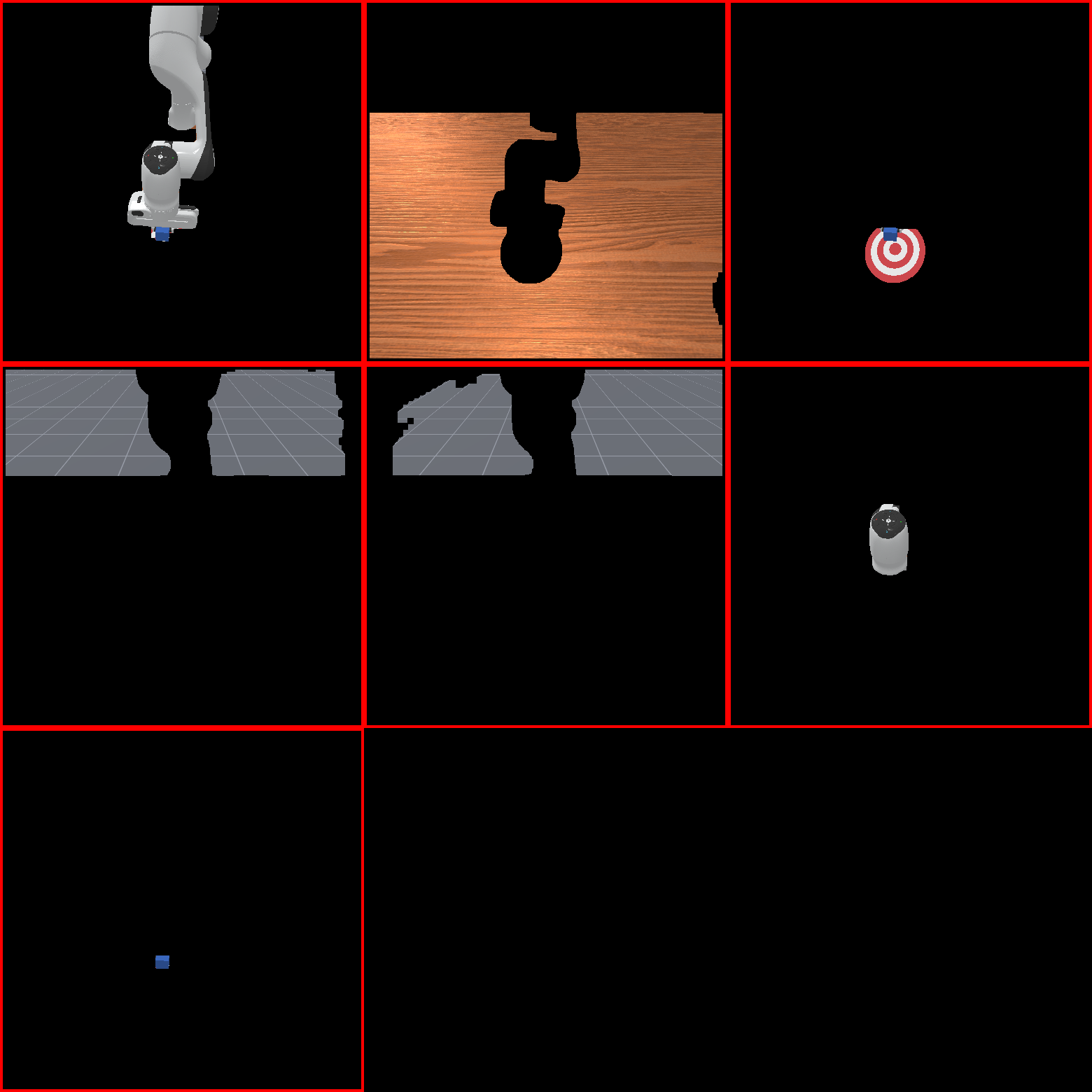}
    \caption{Examples of segments extracted by our object-centric segmentation module for the PushCube task. Left: image with extracted bounding boxes. Right: segments extracted after our lightweight post-processing.}
    \label{fig:grounded_seg_module_output_1}
\end{figure}

Concretely, we use YOLO-World~\citep{cheng2024yoloworldrealtimeopenvocabularyobject} as the detector and EfficientViT-SAM~\citep{zhang2024efficientvitsamacceleratedsegmentmodel} as the segmentation model. Achieving sufficient accuracy and throughput for online RL on $512{\times}512$ images required careful component selection, as most segmentation pipelines introduce latency incompatible with environment interaction. Importantly, incorrect class predictions from the detector do not propagate to segmentation errors, since the segmentation model uses only bounding-box coordinates as prompts rather than class labels. \Cref{fig:grounded_seg_module_output_1} shows representative segment outputs for the \textit{PushCube-v1} task.

A key property of this design is that the number of output segments $N$ varies naturally across time steps, since the detector only proposes boxes for objects it identifies in the current frame. This contrasts with slot-based methods that fix the representation count in advance and with approaches that pad to a constant maximum regardless of scene content. We ablate this choice in~\Cref{sec:ablations} and find that clamping to a fixed segment count degrades performance, particularly on tasks with high segment count variability.

The text inputs are simple concept names rather than detailed descriptions, and they can be defined per task, shared across tasks, or generated automatically by an image-tagging model~\citep{zhang2023recognizeanythingstrongimage,huang2023opensetimagetaggingmultigrained} or a VLM~\citep{zhang2025llavaminiefficientimagevideo}. We also include a generic ``background'' text input across all tasks, this was found to improve generalization by forcing the network to learn to ignore irrelevant part of the image (see~\Cref{fig:appendix_text_ablations}). Our ablations (\Cref{sec:ablations}) show that replacing all text inputs with synonyms or using a single shared vocabulary across tasks has no measurable effect on performance, confirming that \methodName does not depend on prompt engineering.

\subsection{From Masks to Contextual Object Tokens}
\label{sec:segment_embeddings}

\begin{figure}[t]
    \centering
    \includegraphics[width=1.0\linewidth]{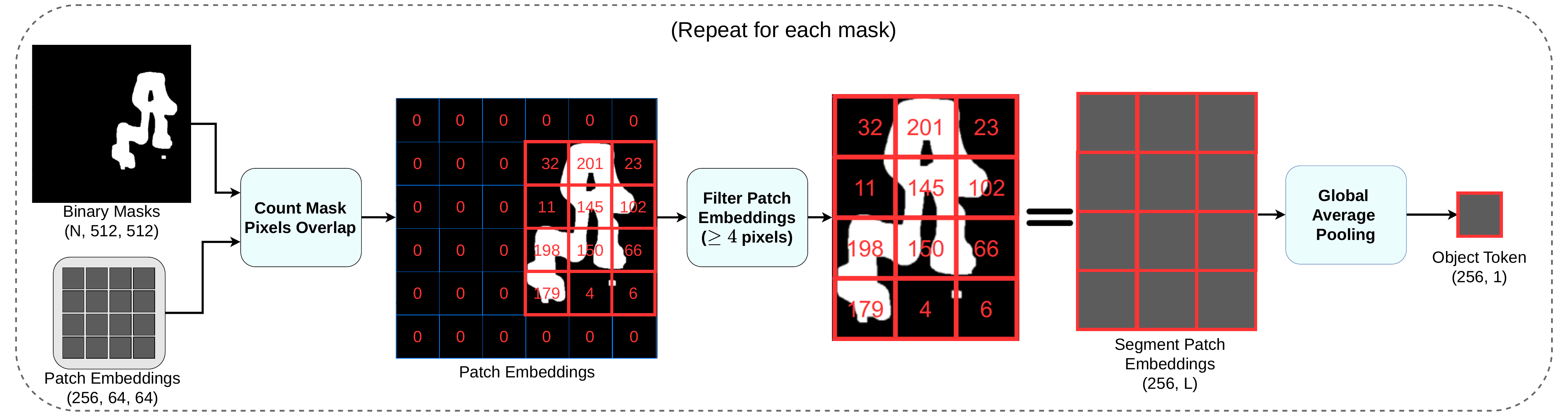}
    \caption{Dynamic object token construction. For each detected segment, 
patch embeddings with sufficient spatial overlap ($\geq 4$ pixels) are 
average-pooled into a single token. The sequence length varies 
dynamically with the number of segments at each timestep.}
    \label{fig:seg_embeds_extract_module_detailed}
\end{figure}

Given $N$ binary masks and a grid of patch embeddings from the frozen encoder, the goal is to produce one compact embedding per segment that is rich enough for policy learning. A naive approach would encode each segment independently, discarding the relationships between objects in the scene. Instead, we leverage a fundamental property of vision-transformer encoders: their patch embeddings already incorporate global context through self-attention. By pooling only the patches that overlap with each mask, we obtain object tokens that capture precise local structure while retaining scene-level context, without any additional attention or cross-segment computation (\Cref{fig:seg_embeds_extract_module_detailed}). This shared-context property is important for robustness, as we discuss in \Cref{sec:robustness}.

Concretely, for each of the $N$ masks we identify the encoder patches whose spatial footprint overlaps with the mask by at least $4$ pixels, then apply global average pooling over the selected patches to produce a single vector. This operation has no trainable parameters and adds negligible compute. Because the object tokens are computed from frozen encoder features, we store them directly in the replay buffer rather than raw images, avoiding the need to re-run the encoder during gradient updates and significantly reducing memory usage. Together with our morphological mask refinement and text-guided prompting, this design contributes to our pipeline achieving roughly $1.8\times$ the training throughput of SAM-G and $5.8\times$ that of FTD (\Cref{sec:train_speed_analysis}), making training online RL practical on a single A100 or L40s GPU ($\sim$24h).

Although we use the segmentation encoder for efficiency, reusing features already computed during mask prediction, the extraction procedure is backbone-agnostic and compatible with any vision-transformer encoder such as DINOv2~\citep{oquab2024dinov2learningrobustvisual}.

\subsection{Transformer Actor-Critic for Dynamic Object Tokens}
\label{sec:actor_critic}
\begin{figure}[t]
    \centering
    \includegraphics[width=0.49\linewidth]{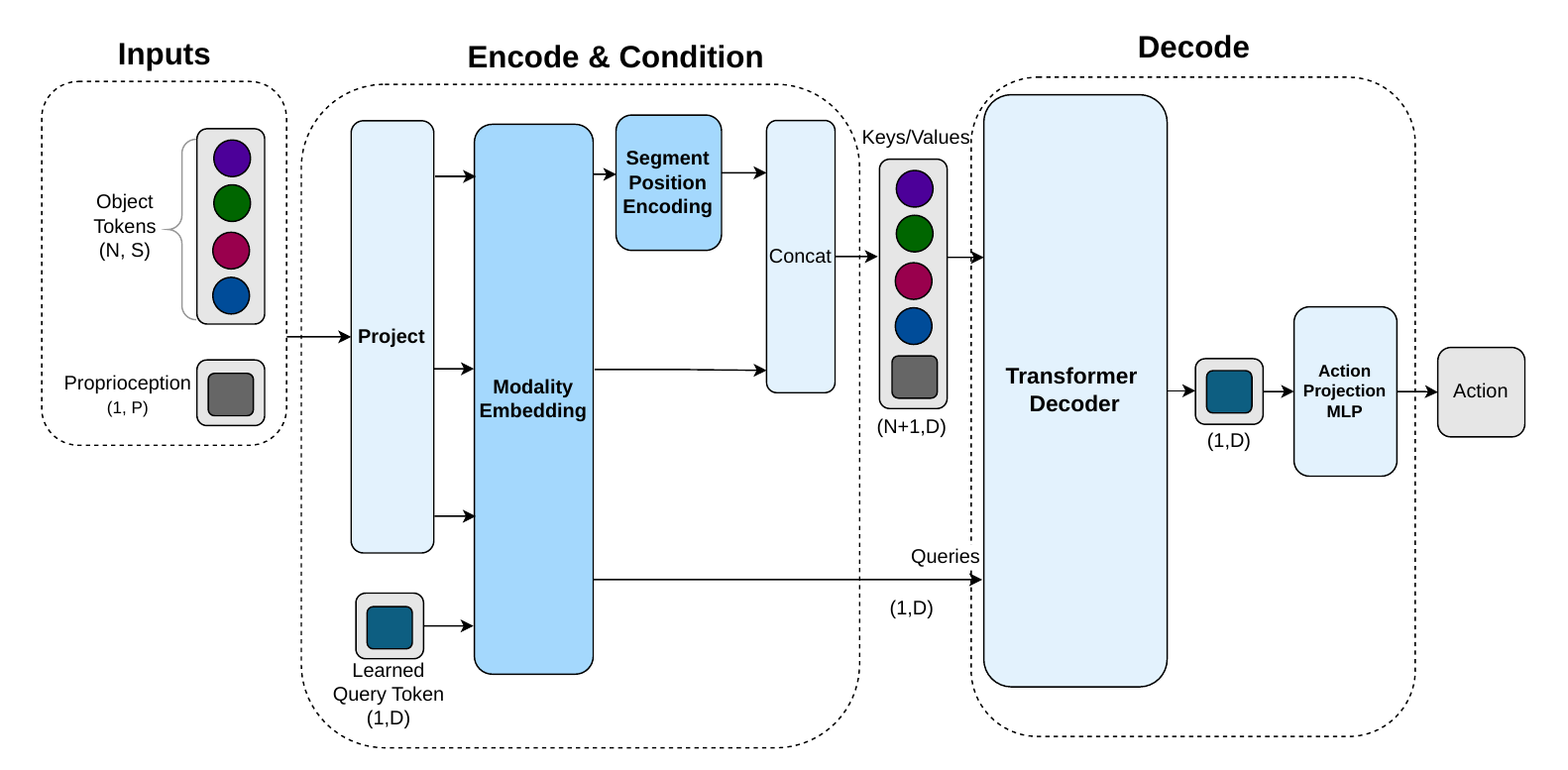}
    \includegraphics[width=0.49\linewidth]{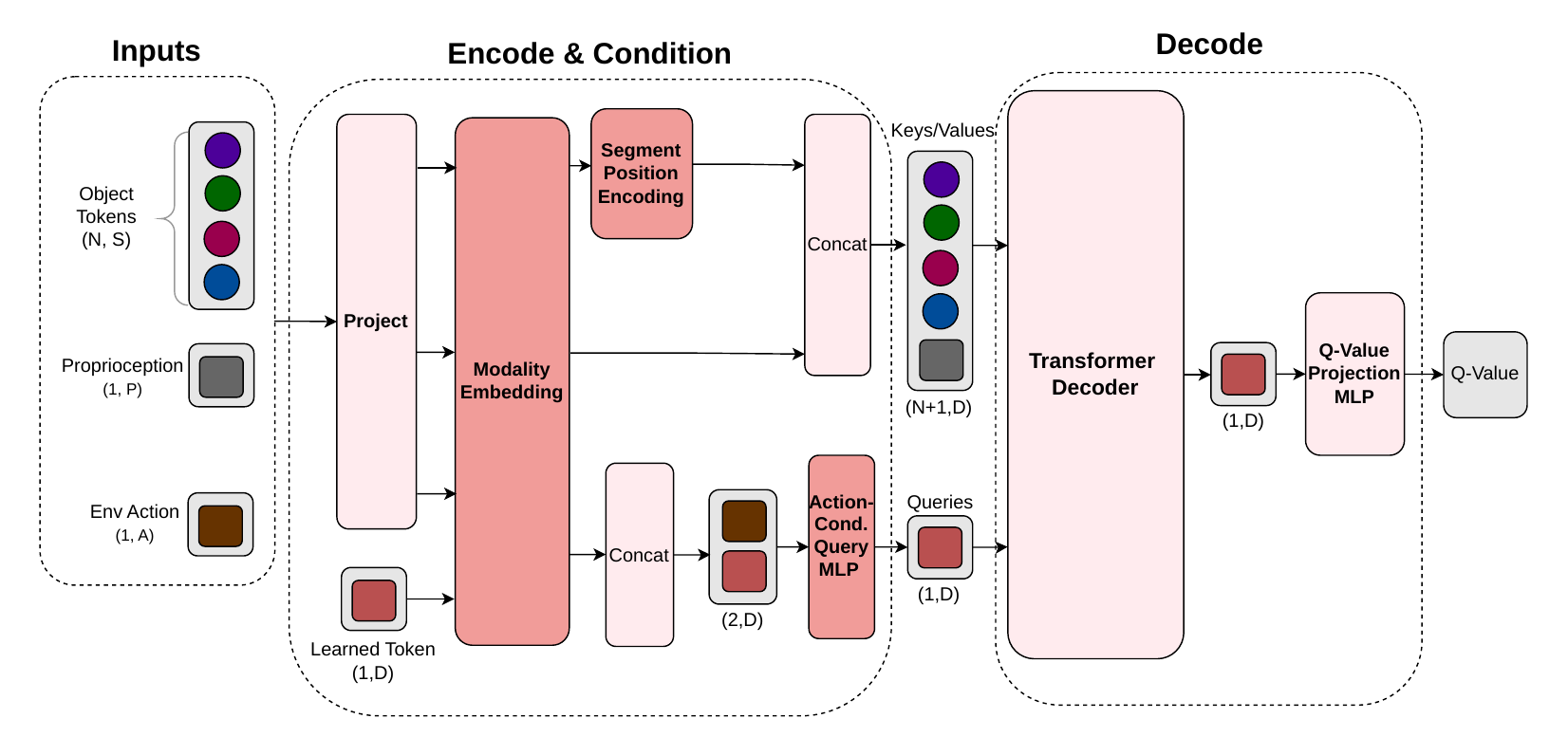}
    \caption{\methodName actor (left) and critic (right) networks. Both networks 
    project a variable-length sequence of $N$ object tokens alongside proprioception 
    into a shared latent space, conditioned with segment position encodings and 
    modality embeddings, before cross-attending via a transformer decoder. The actor 
    uses a single learned query token to produce actions, while the critic constructs 
    an action-conditioned query by concatenating a learned token with the environment 
    action prior to decoding.}
    \label{fig:actor_critic_networks_arch}
\end{figure}
The third and central design problem is building an actor-critic~\citep{NIPS1999_6449f44a} that can efficiently learn from a set of object tokens whose cardinality changes at every time step, fuse them with proprioceptive information, and produce stable Q-value and action predictions under online RL. We address this with a transformer decoder architecture~\citep{NIPS2017_3f5ee243} in which each object token is treated as an input element, inspired by how ViT treats image patches~\citep{dosovitskiy2021imageworth16x16words}.

\paragraph{Modality \& Segment position encoding.}
Inspired by BERT~\citep{devlin-etal-2019-bert}, which uses learned segment embeddings to distinguish tokens from different input sequences, we add a learned modality embedding to each token to allow the decoder to distinguish object tokens from the proprioception token and learned query tokens, segment tokens receive a learned segment positional encoding derived from their bounding-box coordinates, providing an explicit spatial reference tied to the object-centric structure of the input. Our ablation (\Cref{sec:ablations}) confirms that removing this encoding destabilizes training on harder tasks, indicating that the spatial information implicit in pretrained patch features is not sufficient for the decoder to reason about object locations.

\paragraph{Action-Conditioned Critic Query.}
In the critic, the query token is produced by an MLP that takes as input a learned token and the action: $q = \text{MLP}(q_{\text{learn}}, a)$. This allows the critic to attend over object tokens with respect to the specific action being evaluated.

\paragraph{Variable-length processing.}
Because $N$ varies across time steps, a batch of $B$ transitions yields a total of $N_{\text{total}}$ object tokens where each transition may contribute a different number. Rather than padding every sequence to a fixed number, which wastes compute on empty slots and introduces a per-task hyperparameter, we concatenate all tokens into a single packed sequence and apply an attention mask that restricts each token to attend only within its own time step. This sequence-packing strategy, inspired by recent techniques in large language models~\citep{touvron2023llama2openfoundation}, means transitions with few segments and transitions with many segments coexist in the same batch without wasted computation. There is no architectural upper bound on segment count: \methodName is limited only by available GPU memory, allowing the same model to be deployed across tasks of varying visual complexity without reconfiguration.

\paragraph{Training.}
The actor and critic each use their own transformer decoder with separate weights. Both are trained with the standard SAC objective~\citep{haarnoja2018softactorcriticoffpolicymaximum,haarnoja2019softactorcriticalgorithmsapplications} without modification. No auxiliary losses, reconstruction terms, or data augmentation are used during RL training. Because the decoder operates on a compact set of object tokens (typically 5$\sim$25 tokens per time step) rather than dense patch grids, it processes far fewer tokens than patch-based encoders, keeping the computational cost manageable despite using a full transformer architecture.

\section{Experimental Methodology}
\label{sec:experiments}
We design our experiments to answer five questions. (1)~Does learning from dynamic object tokens improve visual generalization compared to state-of-the-art pixel-based and segmentation-based methods? (2)~Can \methodName learn from dynamic object tokens in a stable manner and maintain competitive sample efficiency? (3)~Is the architecture robust to the natural variability of segments produced by pretrained  segmentation models? (4)~What is the impact of each design choice in~\Cref{sec:method} ? (5)~When the policy does fail under strong perturbations, does it degrade gracefully or collapse into erratic behavior? The following  subsections describe the benchmark, baselines, evaluation protocol, and ablation design used to address these questions. Results are presented in~\Cref{sec:results}.

\subsection{Visual Generalization Benchmark}
\label{sec:benchmark}
\begin{wrapfigure}{R}{0.4\textwidth}
    \centering
    \vspace{-0.4cm}

    \includegraphics[width=\linewidth]{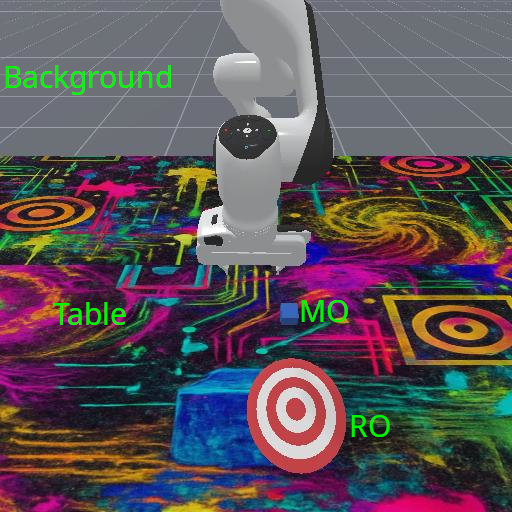}
    \caption{Hard visual perturbation in the PushCube task, with scene entities labeled using our taxonomy.}
    \label{fig:pushcubetest-v1_table_texture_test_hard}

    \vspace{-0.15cm}

    \captionof{table}{Difficulty levels.}
    \setlength{\tabcolsep}{3pt}
    \renewcommand{\arraystretch}{1.05}
    {\scriptsize
    \resizebox{\linewidth}{!}{%
    \begin{tabular}{@{}lcc@{}}
    \toprule
    \textbf{Difficulty} & \textbf{Visual change} & \textbf{Semantic perturbation} \\
    \midrule
    Easy   & \greencheck\ Low      & \redx\ None \\
    Medium & \greencheck\ Moderate & \redx\ None \\
    Hard   & \greencheck\ High     & \greencheck\ Present \\
    \bottomrule
    \end{tabular}}}

    \label{tab:vg_benchmark_levels}
\end{wrapfigure}

Existing visual RL benchmarks such as DMC-GB~\citep{hansen2021stabilizingdeepqlearningconvnets} and DMC-GB2~\citep{almuzairee2024recipe} mainly focus on locomotion tasks with relatively simple visual changes. \citet{yuan2023rlvigenreinforcementlearningbenchmark} showed that these settings do not fully capture the demands of more realistic robotic environments. To provide a more rigorous evaluation, we introduce a new visual generalization benchmark built on ManiSkill3~\citep{taomaniskill3}, covering 8 manipulation tasks across two robot embodiments (Franka Panda arm and Unitree G1 humanoid) with systematic visual perturbations organized into semantically meaningful difficulty levels.

\paragraph{Scene entity taxonomy.}
Inspired by~\citep{pumacay2024colosseumbenchmarkevaluatinggeneralization}, we define perturbations precisely, we decompose each scene into four entity types: the \emph{manipulation object} (MO), the object directly grasped or moved by the end-effector; the \emph{receiver object} (RO), an object not directly manipulated but necessary for task completion; the \emph{primary interaction surface} (table); and \emph{background elements} (walls, floor, skybox). This taxonomy enables perturbations that target specific scene components and allows us to design configurations that create controlled semantic conflicts between entities.

\paragraph{Perturbation categories.}
We define four perturbation categories, each targeting a different visual dimension: (1)~\emph{camera} (pose, field of view), (2)~\emph{lighting} (direction, color), (3)~\emph{color} (MO, RO, table, background), and (4)~\emph{texture} (MO, RO, table, background). Together these yield up to 12 perturbation types per task, depending on whether the task includes a receiver object.

\paragraph{Difficulty levels.}
Each perturbation is tested at three difficulty levels for all 8 tasks. \emph{Easy} perturbations introduce minor visual changes that remain close to the default scene, serving as a smoke test for basic robustness. \emph{Medium} perturbations produce substantial visual shifts while avoiding semantic conflicts between scene entities (e.g., no object is given the same color as another task-critical element). \emph{Hard} perturbations apply aggressive out-of-distribution changes that deliberately introduce both visual and semantic challenges, such as texturing the manipulation object with patterns that resemble the goal target or matching the table color to the cube. These are designed to stress-test perception rather than to facilitate task completion. Visual examples for all tasks and difficulty levels are provided in~\Cref{sec:difficulity_textures_examples}.

\subsection{Baselines}
\label{sec:baselines}
We compare \methodName against six baselines across three categories. \emph{Pixel-based}: SAC-AE~\citep{yarats2020improvingsampleefficiencymodelfree} (reconstruction objective, lower bound), DrQ-v2~\citep{yarats2021masteringvisualcontinuouscontrol} (state of the art in sample efficiency), MaDi~\citep{grooten2023madilearningmaskdistractions} (soft pixel masking from reward), and SADA~\citep{almuzairee2024recipe} (soft and strong data augmentation). \emph{Self-supervised}: SMG~\citep{zhang2024focusmattersseparatedmodels} (foreground-background reconstruction with consistency losses). \emph{Segmentation-based}: SAM-G~\citep{wang2023generalizablevisualreinforcementlearning} (SAM features with ground-truth masks, see~\Cref{fig:samg_g_labels}). MaDi, SADA, SMG, and SAM-G are strong visual generalization baselines and DrQ-v2 is our sample efficiency reference. All methods use official implementations, share the same training budget and evaluation protocol, with hyperparameters detailed in~\Cref{appendix_hyperparams}.

\subsection{Training and Evaluation Protocol}
\label{sec:protocol}

All agents are trained for 1M environment steps without visual perturbations, using 20 parallel GPU environments and a shared set of hyperparameters across all 8 tasks. See~\Cref{appendix_tasks_def} for more details on the tasks and their definition. For \methodName, only the text inputs differ between tasks. During training, policies are evaluated every 10K steps on unperturbed seeds using 10 rollouts across 5 seeds, with stochastic policies run in deterministic mode (taking the mean of the action distribution).

To ensure statistically robust evaluation, each perturbation-task-difficulty combination is evaluated with 50 rollouts per seed across 5 seeds, yielding one score per seed per combination. For the overall score per difficulty, the IQM is computed over $450 = 80 + 80 + 145 + 145$ scores (see \Cref{sec:scores_aggregation} for the per-category breakdown), derived from $450 \times 50 = 22{,}500$ total rollouts per difficulty. We report IQM returns with 95\% confidence intervals via stratified bootstrap, following \citet{agarwal2022deepreinforcementlearningedge}, which is robust to outlier tasks where some methods collapse entirely. Returns are normalized by each task's maximum episode length to produce values in $[0, 1]$, enabling aggregation across tasks with different horizons.

\subsection{Ablation Design}
\label{sec:ablation_design}
We isolate each design choice by removing or replacing one component at 
a time: (i)~\textbf{no segment positional encoding}, removing explicit re-injection of positional information for object tokens, (ii)~\textbf{fixed object tokens count}, clamping $N{=}5$ with zero-padding and truncation (with attention mask to prevent padding from changing the attention computation), (iii)~\textbf{global token vs object-centric tokens}, replacing per-object tokens and the transformer decoder with an MLP over a single mean-pooled vector, and (iv)~\textbf{text sensitivity}, testing synonym substitution and a shared cross-task vocabulary for the text inputs. All ablations run on 3 seeds unless stated otherwise.

\subsection{Case Studies and Failure Analysis}
\label{sec:failure_analysis}
We qualitatively compare \methodName and MaDi to characterize not just when each method fails, but how gracefully it fails. This question matters for real-world deployment, where structured failure is preferable to erratic behavior. For each method, we identify representative failure episodes under hard visual perturbations and examine whether the policy degrades smoothly or collapses entirely.

\section{Empirical Results}
\label{sec:results}
We present results organized by the five questions posed in~\Cref{sec:experiments}: visual generalization (\Cref{sec:results_generalization}), sample efficiency (\Cref{sec:results_efficiency}), robustness to segment variability (\Cref{sec:robustness}), ablations (\Cref{sec:ablations}), and failure analysis (\Cref{sec:case_study_main}). Additional analysis of the attention patterns learned over object tokens is provided in~\Cref{appendix_segment_attn_analysis}.

\subsection{Visual Generalization}
\label{sec:results_generalization}

\begin{figure}[t]
    \centering
    \includegraphics[width=1.0\linewidth]{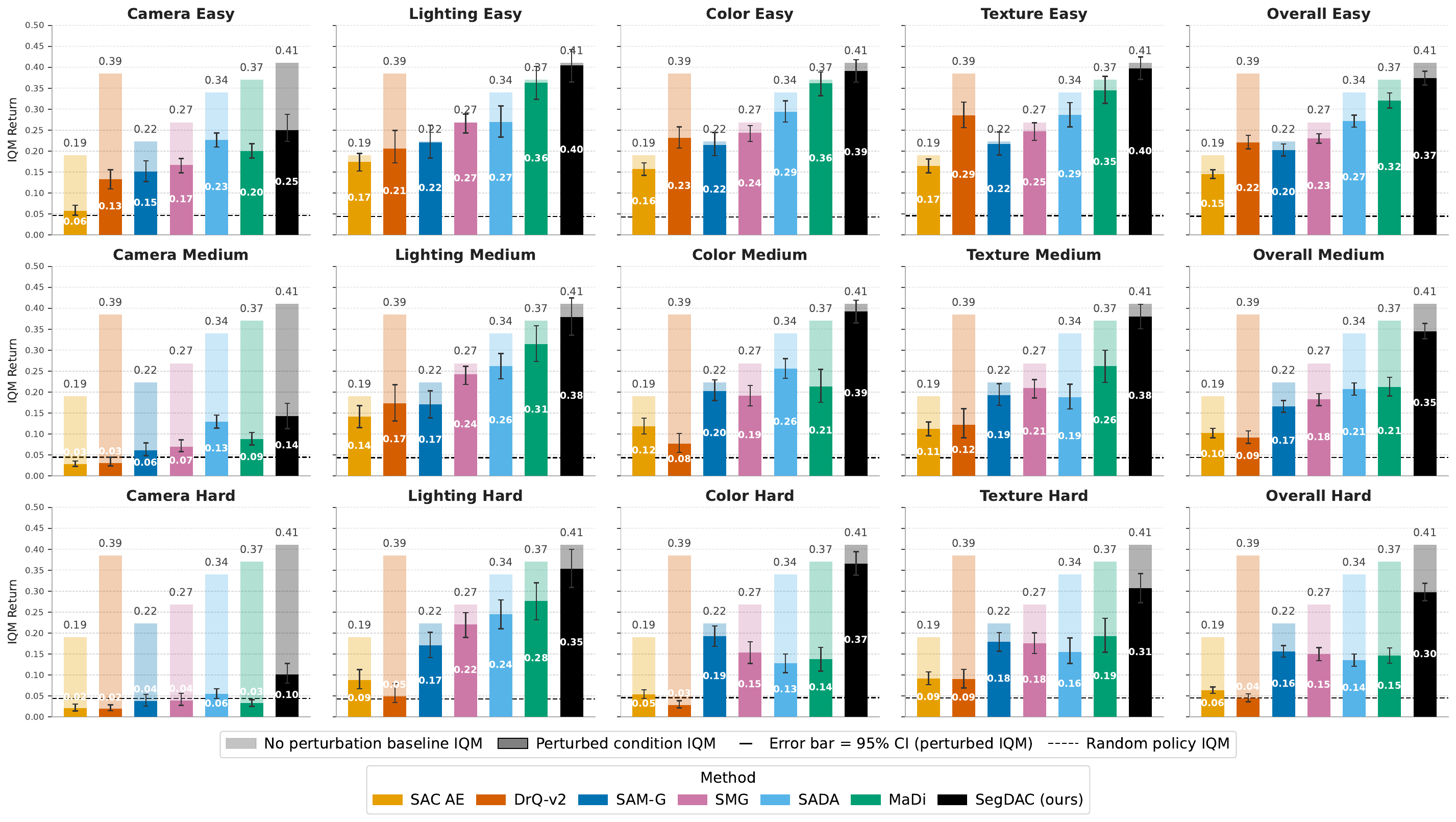}
    \caption{Visual generalization results grouped by perturbation 
category and difficulty level. Gray bars show unperturbed baseline IQM; 
colored bars show perturbed IQM with 95\% confidence intervals.}
    \label{fig:visual_generalization}
\end{figure}

The central result is in the hard setting: most baselines lose over half their unperturbed performance, with SADA and MaDi dropping by 59\% and DrQ-v2 by over 90\%. SAM-G is the only competitive baseline, yet \methodName achieves an IQM return 88\% higher. This gap reflects a representational difference: by reasoning over per-object tokens rather than a single global feature, \methodName remains effective under perturbations that fundamentally change pixel statistics but leave object-level structure intact.

Performance differences are small under easy perturbations, where all methods remain close to their unperturbed baseline, and begin to diverge at medium difficulty before separating sharply at hard (\Cref{fig:visual_generalization}). This progression confirms that the benchmark difficulty levels are well-calibrated: easy perturbations are a smoke test, and the hard setting is where the representational differences between methods matter most.

\begin{figure}[h!]
    \centering
    \begin{tabular}{cc}
        \includegraphics[width=0.2\linewidth]{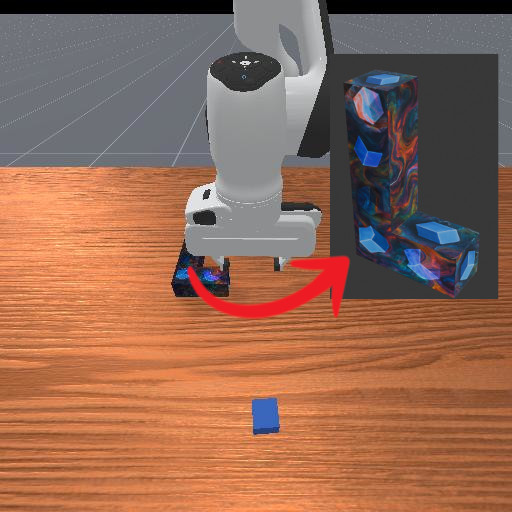}
        \includegraphics[width=0.2\linewidth]{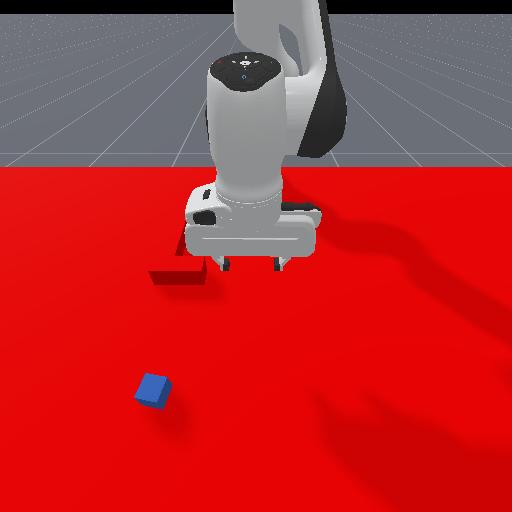}
        \includegraphics[width=0.2\linewidth]{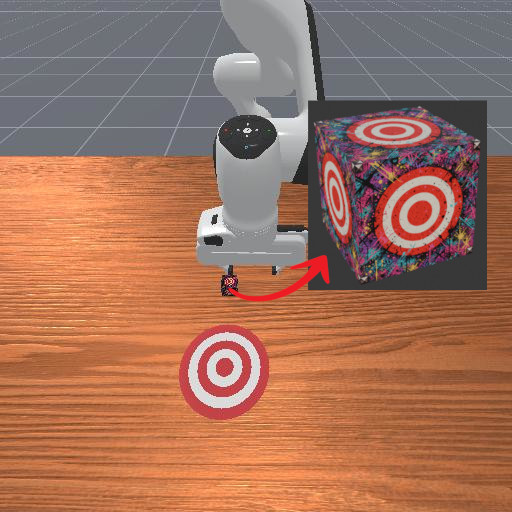}
        \includegraphics[width=0.2\linewidth]{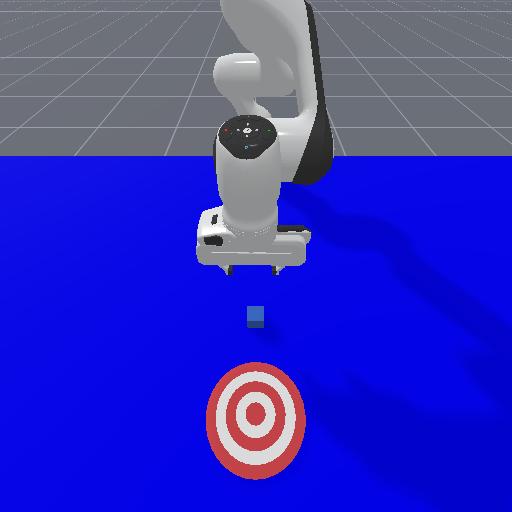}
    \end{tabular}
    \caption{\small Hard Manipulation Object (MO) Texture and Table Color for \texttt{PullCubeTool} and \texttt{PushCube}.}
    \label{fig:hard_perturbations}
\end{figure}

\paragraph{Semantic perturbations.}
Hard MO texture and hard table color perturbations introduce semantic conflicts that challenge object identity by making task-critical objects visually ambiguous (\Cref{fig:hard_perturbations}). Most baselines collapse under these conditions. \methodName achieves the highest IQM on 6 of 8 tasks under hard table color and on all 8 tasks under hard MO texture. SAM-based methods consistently outperform pixel-based methods, this confirms their inherent robustness to surface appearance changes. Complete per-perturbation-per-difficulty results are in~\Cref{appendix_vis_gen_tables}.

\begin{table}[h!]
\centering
\scriptsize
\vspace{-1.7em}
\caption{Hard Table Color Visual Generalization}
\vspace{-1.0em}
\label{tab:appendix_tablecolortest_hard}
\begin{adjustbox}{max width=\textwidth}
\begin{tabular}{l*{7}{c}}
\toprule
\textbf{Task} & \textbf{SAC AE} & \textbf{DrQ-v2} & \textbf{SAM-G} & \textbf{SMG} & \textbf{SADA} & \textbf{MaDi} & \textbf{\methodName} \\
\midrule
LiftPegUpright & 0.19$\pm$0.00 \scriptsize{(-11.7\%)} & 0.19$\pm$0.01 \scriptsize{(-60.0\%)} & \textbf{0.41$\pm$0.13 \scriptsize{(-3.2\%)}} & 0.22$\pm$0.01 \scriptsize{(-3.1\%)} & \underline{0.22$\pm$0.01 \scriptsize{(-0.5\%)}} & 0.19$\pm$0.00 \scriptsize{(-13.0\%)} & 0.38$\pm$0.14 \scriptsize{(-8.1\%)} \\
PickCube & 0.02$\pm$0.01 \scriptsize{(-88.5\%)} & 0.00$\pm$0.00 \scriptsize{(-98.9\%)} & \underline{0.09$\pm$0.01 \scriptsize{(-11.3\%)}} & 0.03$\pm$0.01 \scriptsize{(-89.9\%)} & 0.03$\pm$0.01 \scriptsize{(-92.2\%)} & 0.02$\pm$0.01 \scriptsize{(-95.2\%)} & \textbf{0.11$\pm$0.01 \scriptsize{(-67.9\%)}} \\
PokeCube & 0.02$\pm$0.01 \scriptsize{(-90.5\%)} & 0.02$\pm$0.01 \scriptsize{(-95.7\%)} & \textbf{\underline{0.27$\pm$0.06 \scriptsize{(-15.0\%)}}} & 0.09$\pm$0.02 \scriptsize{(-56.9\%)} & 0.12$\pm$0.04 \scriptsize{(-68.9\%)} & 0.03$\pm$0.03 \scriptsize{(-92.6\%)} & 0.20$\pm$0.03 \scriptsize{(-49.9\%)} \\
PullCube & 0.02$\pm$0.02 \scriptsize{(-94.0\%)} & 0.02$\pm$0.02 \scriptsize{(-95.9\%)} & \underline{0.12$\pm$0.01 \scriptsize{(-0.9\%)}} & 0.16$\pm$0.05 \scriptsize{(-59.6\%)} & 0.27$\pm$0.04 \scriptsize{(-43.9\%)} & 0.09$\pm$0.04 \scriptsize{(-81.0\%)} & \textbf{0.35$\pm$0.07 \scriptsize{(-31.1\%)}} \\
PullCubeTool & 0.03$\pm$0.02 \scriptsize{(-85.9\%)} & 0.03$\pm$0.02 \scriptsize{(-95.9\%)} & \underline{0.11$\pm$0.03 \scriptsize{(-52.5\%)}} & 0.09$\pm$0.06 \scriptsize{(-60.6\%)} & 0.02$\pm$0.03 \scriptsize{(-95.5\%)} & 0.03$\pm$0.03 \scriptsize{(-94.9\%)} & \textbf{0.14$\pm$0.04 \scriptsize{(-80.9\%)}} \\
PushCube & 0.04$\pm$0.03 \scriptsize{(-87.0\%)} & 0.04$\pm$0.03 \scriptsize{(-91.0\%)} & 0.40$\pm$0.06 \scriptsize{(-21.2\%)} & 0.05$\pm$0.02 \scriptsize{(-86.7\%)} & 0.17$\pm$0.08 \scriptsize{(-64.1\%)} & 0.06$\pm$0.03 \scriptsize{(-86.3\%)} & \textbf{\underline{0.60$\pm$0.15 \scriptsize{(+33.3\%)}}} \\
PlaceAppleInBowl & 0.00$\pm$0.01 \scriptsize{(-93.4\%)} & 0.01$\pm$0.01 \scriptsize{(-92.7\%)} & \underline{0.06$\pm$0.02 \scriptsize{(-15.9\%)}} & 0.02$\pm$0.01 \scriptsize{(-84.4\%)} & 0.04$\pm$0.05 \scriptsize{(-67.5\%)} & 0.01$\pm$0.02 \scriptsize{(-95.5\%)} & \textbf{0.25$\pm$0.09 \scriptsize{(-21.2\%)}} \\
TransportBox & \underline{0.07$\pm$0.05 \scriptsize{(+45.2\%)}} & 0.07$\pm$0.10 \scriptsize{(-73.6\%)} & 0.12$\pm$0.08 \scriptsize{(-56.3\%)} & 0.17$\pm$0.08 \scriptsize{(-37.8\%)} & 0.19$\pm$0.10 \scriptsize{(-20.7\%)} & 0.12$\pm$0.06 \scriptsize{(-55.8\%)} & \textbf{0.28$\pm$0.01 \scriptsize{(-0.6\%)}} \\
\bottomrule
\end{tabular}
\end{adjustbox}
\end{table}

\begin{table}[h!]
\centering
\scriptsize
\vspace{-1.0em}
\caption{Hard Mo Texture Visual Generalization}
\vspace{-1.0em}
\label{tab:appendix_motexturetest_hard}
\begin{adjustbox}{max width=\textwidth}
\begin{tabular}{l*{7}{c}}
\toprule
\textbf{Task} & \textbf{SAC AE} & \textbf{DrQ-v2} & \textbf{SAM-G} & \textbf{SMG} & \textbf{SADA} & \textbf{MaDi} & \textbf{\methodName} \\
\midrule
LiftPegUpright & 0.20$\pm$0.01 \scriptsize{(-5.1\%)} & 0.22$\pm$0.01 \scriptsize{(-53.9\%)} & 0.25$\pm$0.03 \scriptsize{(-39.7\%)} & \underline{0.22$\pm$0.01 \scriptsize{(+0.0\%)}} & 0.22$\pm$0.02 \scriptsize{(-1.2\%)} & 0.21$\pm$0.01 \scriptsize{(-4.4\%)} & \textbf{0.28$\pm$0.08 \scriptsize{(-32.3\%)}} \\
PickCube & 0.06$\pm$0.02 \scriptsize{(-69.8\%)} & 0.01$\pm$0.00 \scriptsize{(-97.0\%)} & \underline{0.11$\pm$0.01 \scriptsize{(+3.3\%)}} & 0.03$\pm$0.01 \scriptsize{(-90.5\%)} & 0.02$\pm$0.01 \scriptsize{(-94.4\%)} & 0.01$\pm$0.00 \scriptsize{(-98.1\%)} & \textbf{0.15$\pm$0.01 \scriptsize{(-55.2\%)}} \\
PokeCube & 0.08$\pm$0.01 \scriptsize{(-54.2\%)} & 0.03$\pm$0.01 \scriptsize{(-92.1\%)} & 0.07$\pm$0.01 \scriptsize{(-77.6\%)} & \underline{0.16$\pm$0.05 \scriptsize{(-25.1\%)}} & 0.02$\pm$0.02 \scriptsize{(-95.1\%)} & 0.04$\pm$0.09 \scriptsize{(-89.4\%)} & \textbf{0.22$\pm$0.03 \scriptsize{(-45.0\%)}} \\
PullCube & 0.12$\pm$0.04 \scriptsize{(-66.5\%)} & 0.10$\pm$0.03 \scriptsize{(-71.4\%)} & \underline{0.12$\pm$0.01 \scriptsize{(-0.7\%)}} & 0.27$\pm$0.07 \scriptsize{(-29.8\%)} & 0.10$\pm$0.04 \scriptsize{(-79.0\%)} & 0.26$\pm$0.12 \scriptsize{(-46.2\%)} & \textbf{0.43$\pm$0.05 \scriptsize{(-14.4\%)}} \\
PullCubeTool & 0.13$\pm$0.04 \scriptsize{(-40.8\%)} & 0.12$\pm$0.08 \scriptsize{(-81.2\%)} & \underline{0.16$\pm$0.05 \scriptsize{(-30.0\%)}} & 0.11$\pm$0.05 \scriptsize{(-54.3\%)} & 0.02$\pm$0.05 \scriptsize{(-94.0\%)} & 0.03$\pm$0.02 \scriptsize{(-94.6\%)} & \textbf{0.45$\pm$0.08 \scriptsize{(-38.9\%)}} \\
PushCube & 0.13$\pm$0.02 \scriptsize{(-55.9\%)} & 0.02$\pm$0.02 \scriptsize{(-94.9\%)} & 0.23$\pm$0.06 \scriptsize{(-54.3\%)} & 0.21$\pm$0.02 \scriptsize{(-38.6\%)} & 0.08$\pm$0.03 \scriptsize{(-82.7\%)} & 0.15$\pm$0.07 \scriptsize{(-67.7\%)} & \textbf{\underline{0.45$\pm$0.06 \scriptsize{(-0.4\%)}}} \\
PlaceAppleInBowl & 0.05$\pm$0.02 \scriptsize{(-32.0\%)} & 0.05$\pm$0.03 \scriptsize{(-70.3\%)} & \underline{0.06$\pm$0.03 \scriptsize{(-2.8\%)}} & 0.07$\pm$0.04 \scriptsize{(-34.4\%)} & 0.04$\pm$0.04 \scriptsize{(-66.6\%)} & 0.06$\pm$0.02 \scriptsize{(-54.9\%)} & \textbf{0.29$\pm$0.18 \scriptsize{(-7.4\%)}} \\
TransportBox & \underline{0.05$\pm$0.03 \scriptsize{(+1.6\%)}} & 0.18$\pm$0.07 \scriptsize{(-28.9\%)} & 0.27$\pm$0.01 \scriptsize{(+0.3\%)} & 0.27$\pm$0.01 \scriptsize{(+0.4\%)} & 0.23$\pm$0.09 \scriptsize{(-5.2\%)} & 0.26$\pm$0.01 \scriptsize{(-0.1\%)} & \textbf{0.28$\pm$0.01 \scriptsize{(-0.1\%)}} \\
\bottomrule
\end{tabular}
\end{adjustbox}
\end{table}

\subsection{Sample Efficiency}
\label{sec:results_efficiency}
\begin{figure}[th!]
    \centering
    \begin{subfigure}{0.6\linewidth}
        \centering
        \includegraphics[width=\linewidth]{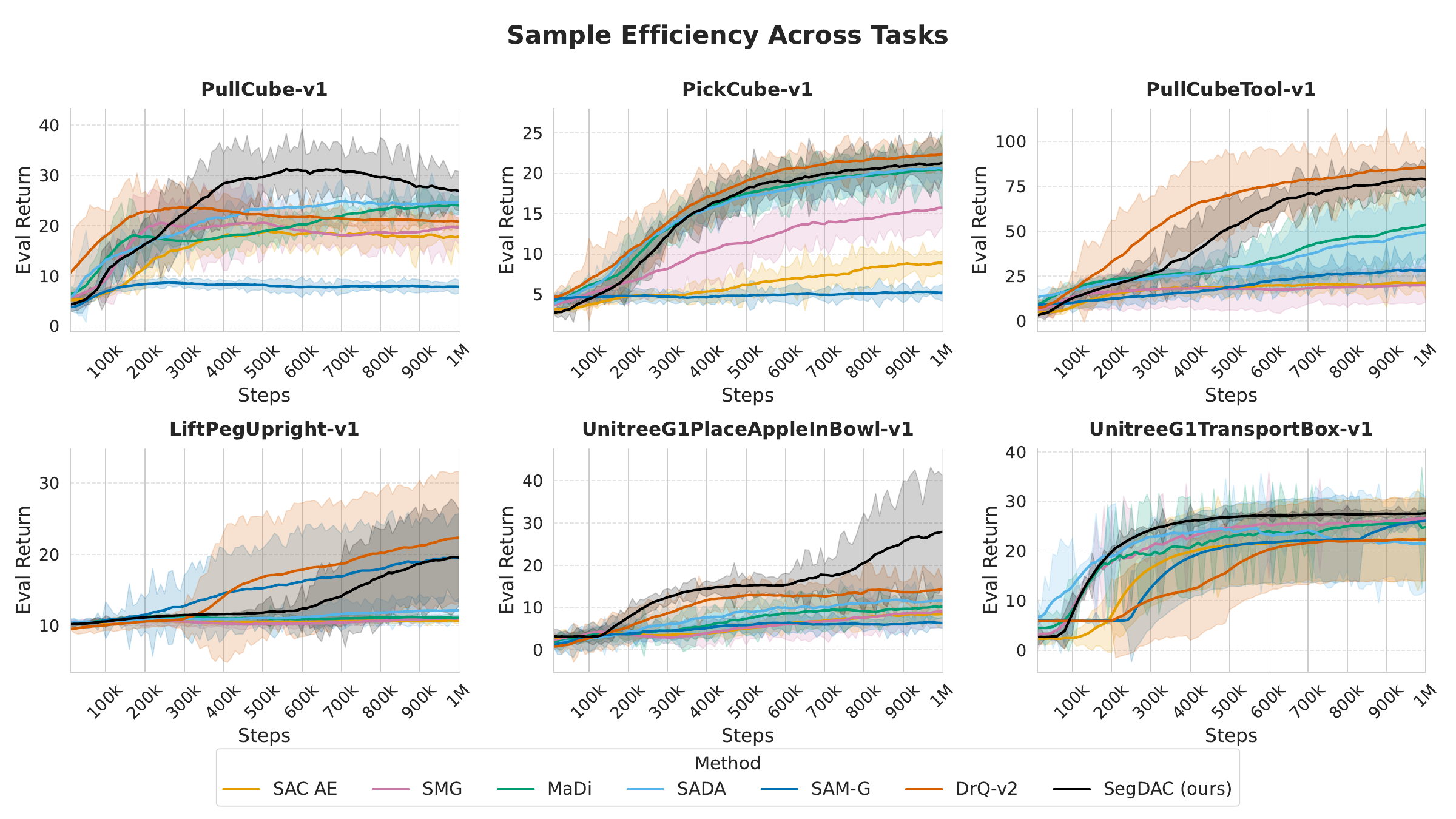}
        \caption{Per-task sample efficiency.}
    \end{subfigure}%
    \hfill
    \begin{subfigure}{0.4\linewidth}
        \centering
        \includegraphics[width=\linewidth]{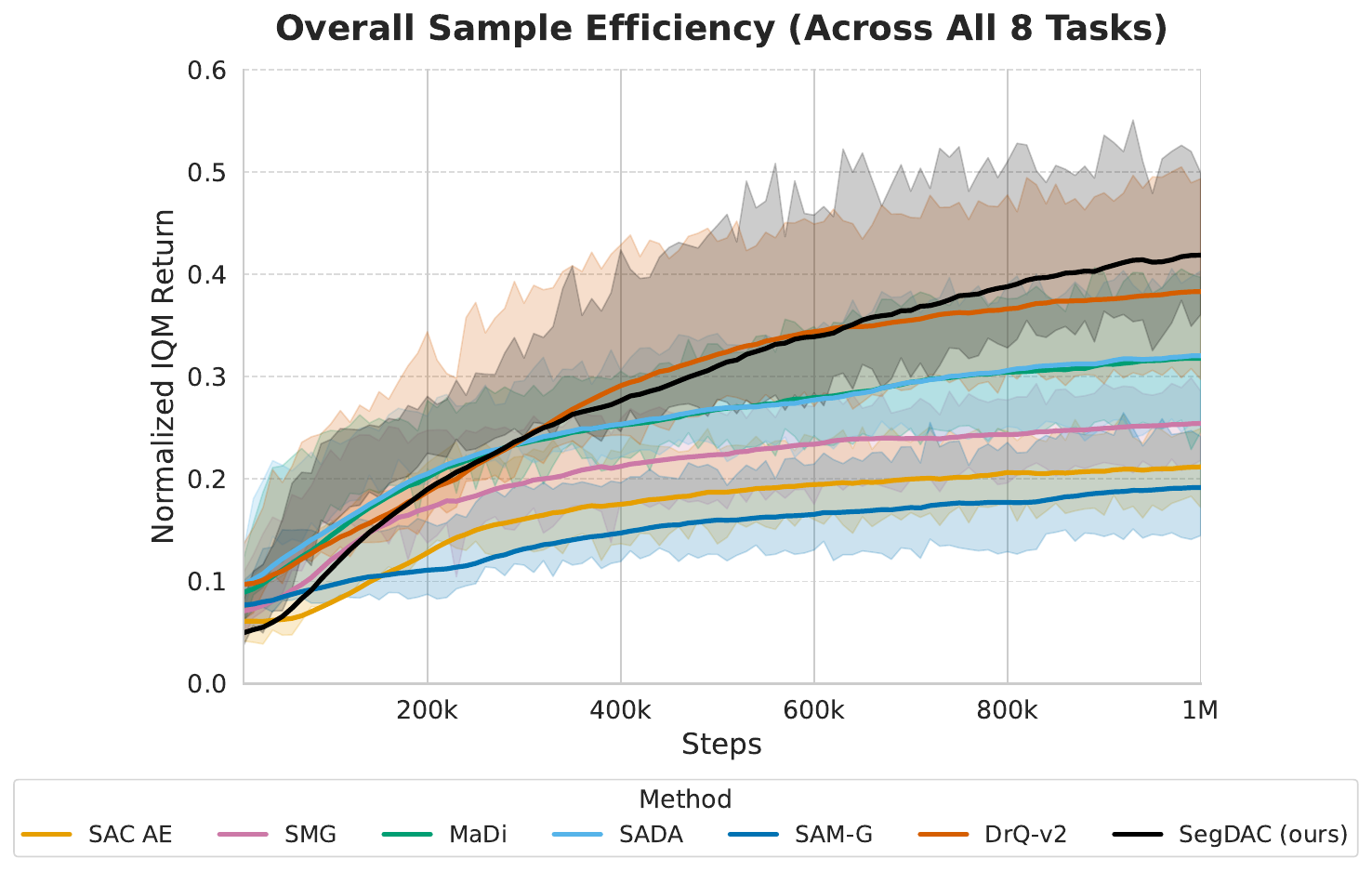}
        \caption{Aggregated IQM returns.}
    \end{subfigure}
    \caption{Sample efficiency results, aggregated using 5 seeds.}
    \label{fig:sample_efficiency}
\end{figure}
\methodName matches the sample efficiency of DrQ-v2 across all 8 tasks and outperforms it on three of eight (\Cref{fig:sample_efficiency}). This is notable because \methodName uses no data augmentation, while DrQ-v2 relies on random shifts as its primary mechanism for learning efficiency. \methodName also consistently outperforms all visual generalization baselines (MaDi, SADA, SMG, SAM-G) in sample efficiency. These methods often perform well on a subset of tasks but collapse on others: MaDi achieves competitive returns on PullCube but fails on LiftPegUpright. In contrast, \methodName maintains stable learning curves across all 8 tasks without collapsing.

This result directly addresses a known tradeoff in visual RL: methods strong in visual generalization tend to sacrifice sample efficiency, and vice versa~\citep{almuzairee2024recipe}. \methodName achieves both, suggesting that learning from dynamic object tokens provides a favourable inductive bias not just for generalization but for learning stability as well. Additional per-task curves are provided in~\Cref{sec:sample_eff_others}.

\subsection{Robustness to Segment Variability}
\label{sec:robustness}

\begin{figure}[thb]
\centering
\includegraphics[width=0.11\linewidth]{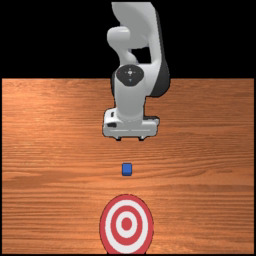}
\includegraphics[width=0.11\linewidth]{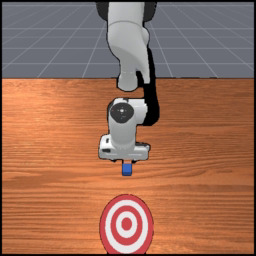}
\includegraphics[width=0.11\linewidth]{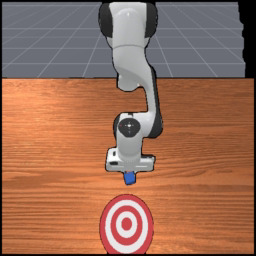}
\includegraphics[width=0.11\linewidth]{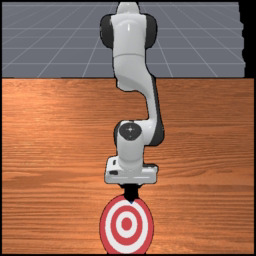}
\includegraphics[width=0.11\linewidth]{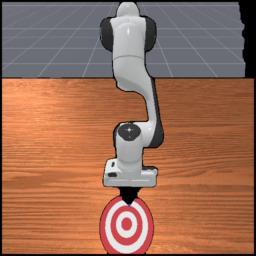}
\includegraphics[width=0.11\linewidth]{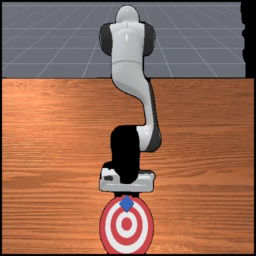}
\includegraphics[width=0.11\linewidth]{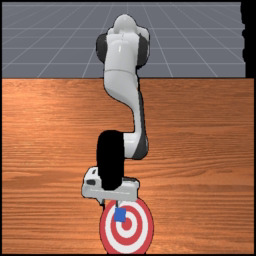}
\includegraphics[width=0.11\linewidth]{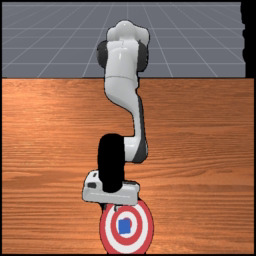}\\[2pt]
\includegraphics[width=0.10\linewidth]{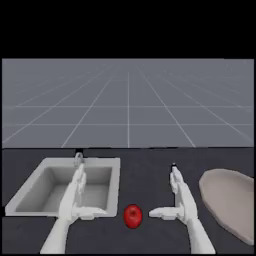}
\includegraphics[width=0.10\linewidth]{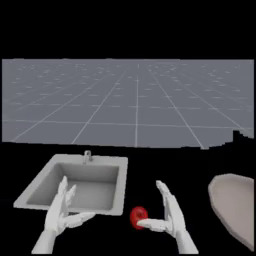}
\includegraphics[width=0.10\linewidth]{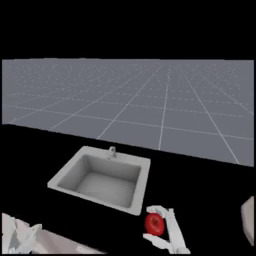}
\includegraphics[width=0.10\linewidth]{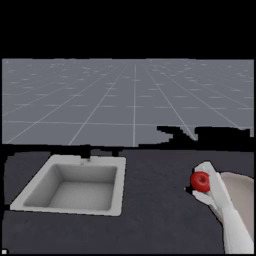}
\includegraphics[width=0.10\linewidth]{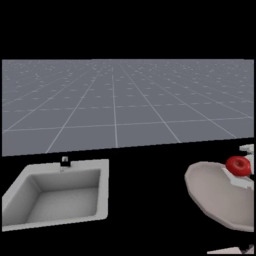}
\includegraphics[width=0.10\linewidth]{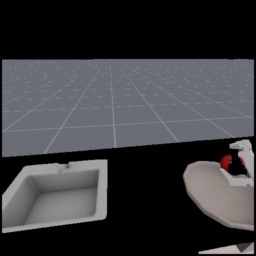}
\includegraphics[width=0.10\linewidth]{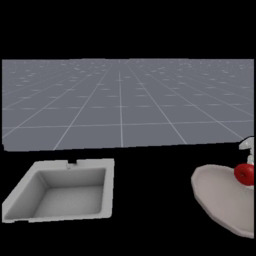}
\includegraphics[width=0.10\linewidth]{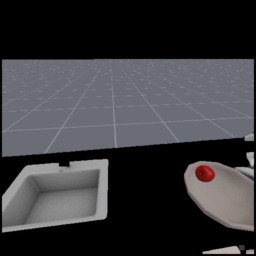}
\includegraphics[width=0.10\linewidth]{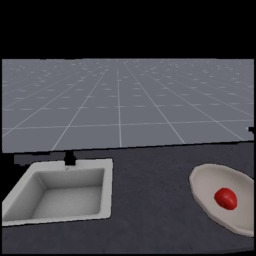}\\[2pt]
\includegraphics[width=0.11\linewidth]{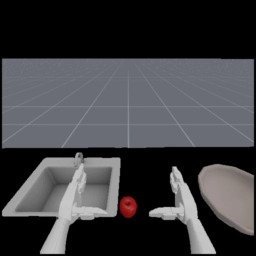}
\includegraphics[width=0.11\linewidth]{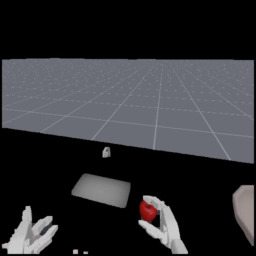}
\includegraphics[width=0.11\linewidth]{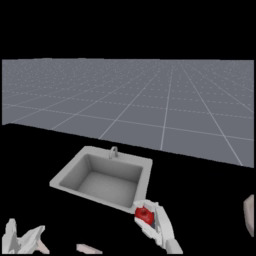}
\includegraphics[width=0.11\linewidth]{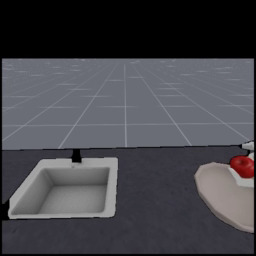}
\includegraphics[width=0.11\linewidth]{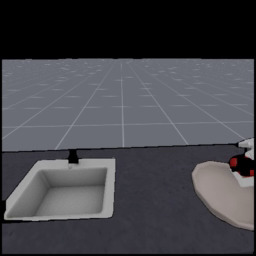}
\includegraphics[width=0.11\linewidth]{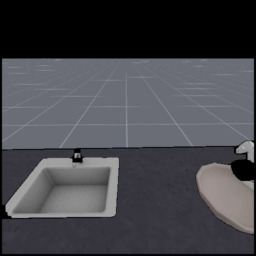}
\includegraphics[width=0.11\linewidth]{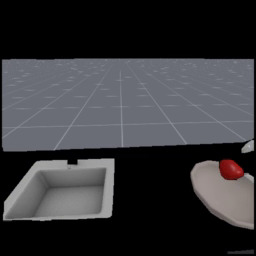}
\includegraphics[width=0.11\linewidth]{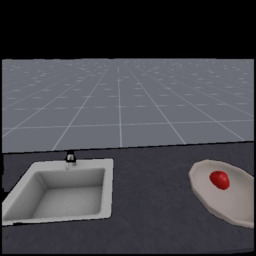}
\noindent\rule{\linewidth}{0.4pt}\\[1pt]
\includegraphics[width=0.35\linewidth, height=2.5cm, keepaspectratio]{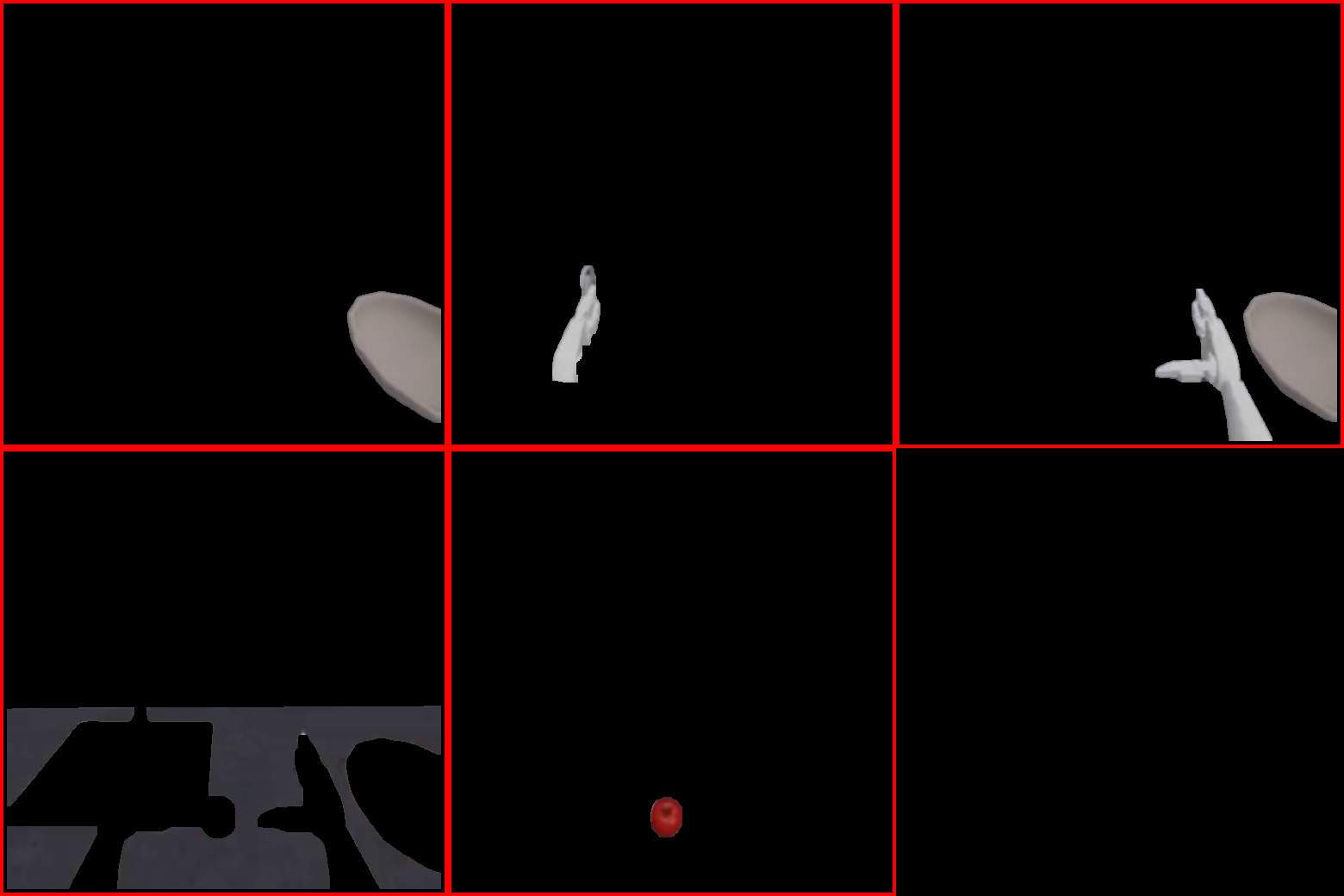}
\hspace{1em}
\includegraphics[width=0.35\linewidth, height=2.5cm, keepaspectratio]{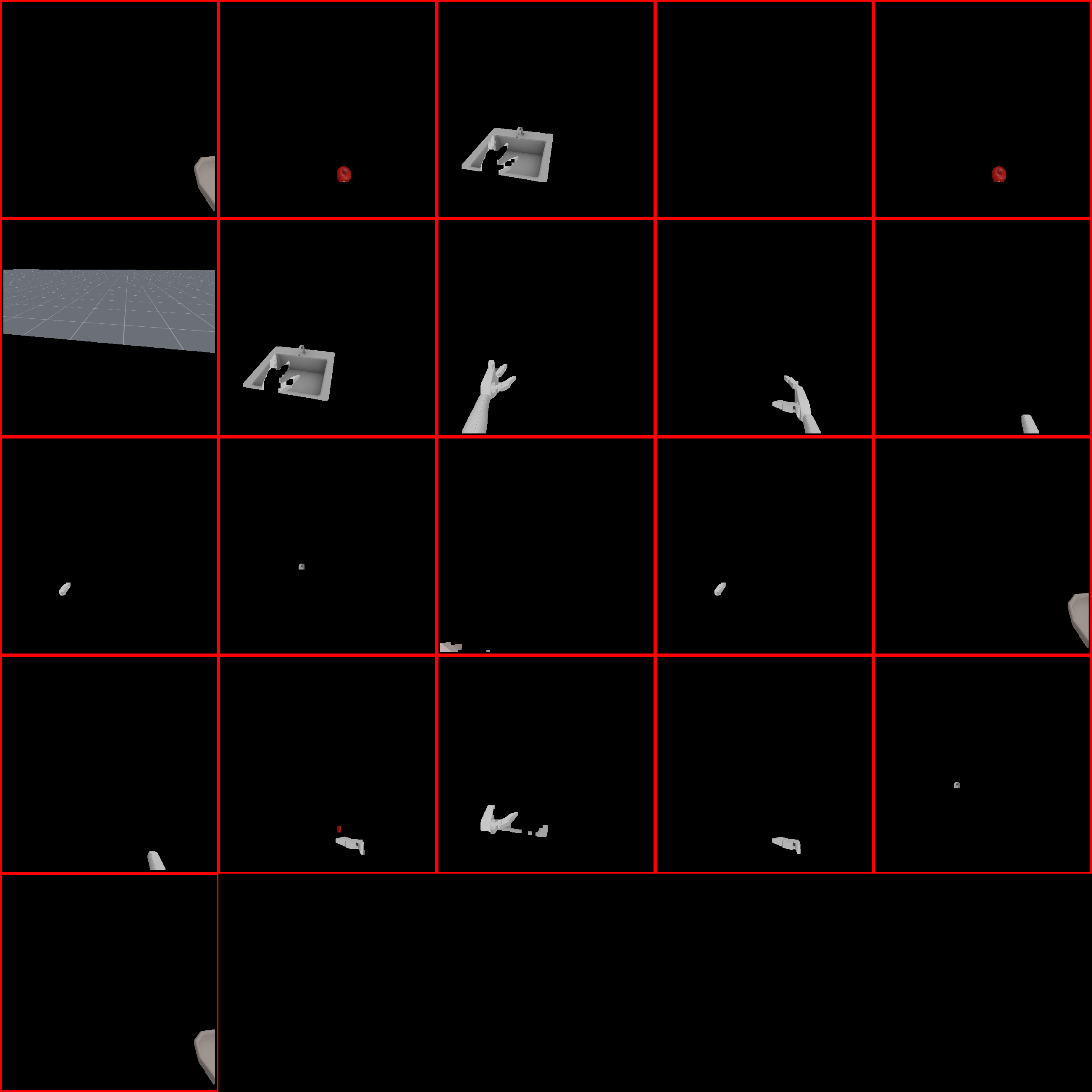}
\caption{(Top) Each row is a rollout showing variability in detected 
segments: (a)~PushCube, (b-c)~AppleInBowl.
(Bottom) Segment count within a single episode: 5 (left) vs.\ 21 (right).}
\label{fig:segment_variability}
\vspace{-1.75em}
\end{figure}
In practice, the set of detected segments varies frequently along a trajectory, both in its count and in its granularity. \Cref{fig:segment_variability} (top) illustrates this: the background may appear late, task-critical segments may temporarily disappear, and parts of the robot arm may merge or split across consecutive frames. \Cref{fig:segment_variability} (bottom) shows that within a single episode of \textit{UnitreeG1PlaceAppleInBowl}, the segment count can vary from as few as 5 to as many as 21. Despite this, \methodName maintains stable behavior and successfully completes the task even when task-critical objects temporarily vanish from the segment set. We attribute this robustness to two properties of our design. First, natural segment variability during training acts as an implicit structural augmentation: by seeing token sets that change in identity, count, and granularity at every timestep, the policy learns to reason from incomplete object information without any explicit data augmentation strategy. Second, our context-preserving pooling design means that each object token is computed from a ViT encoder that has already attended over the full scene. When a segment disappears, the remaining tokens still carry partial information about the missing region through self-attention, providing a soft redundancy. Together, these two properties are emergent benefits of our architecture and training regime.
\vspace{-0.75em}
\subsection{Ablations}
\label{sec:ablations}
\begin{figure}[t]
\centering
\includegraphics[width=\linewidth]{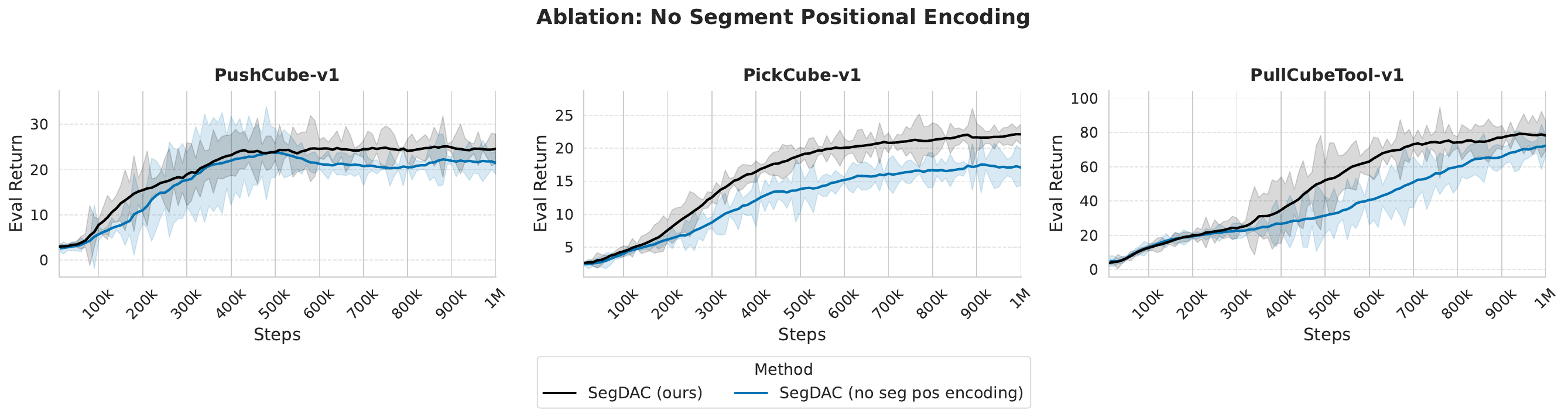}
\includegraphics[width=\linewidth]{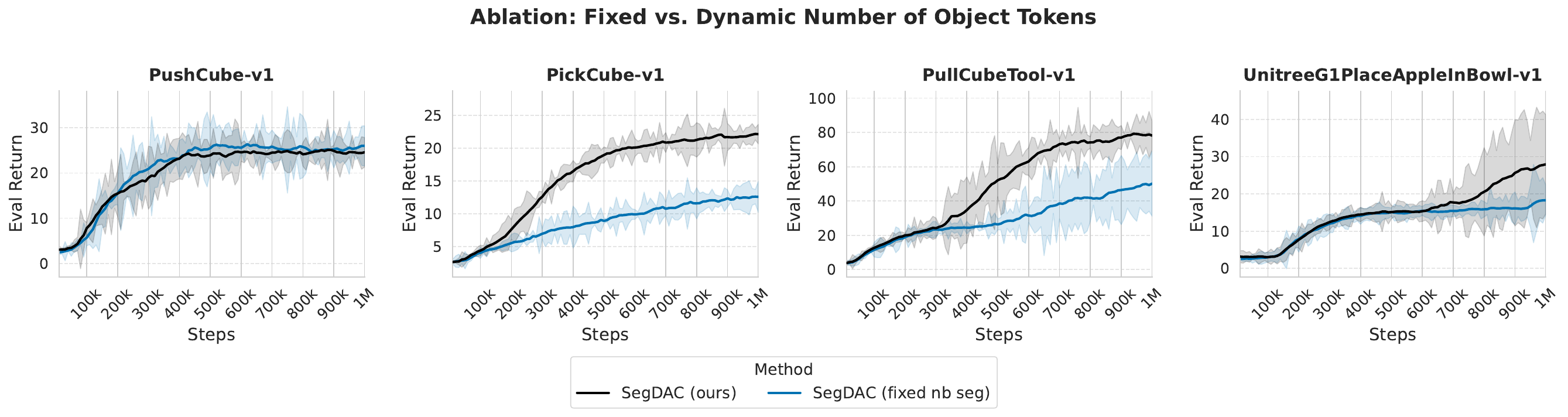}
\includegraphics[width=\linewidth]{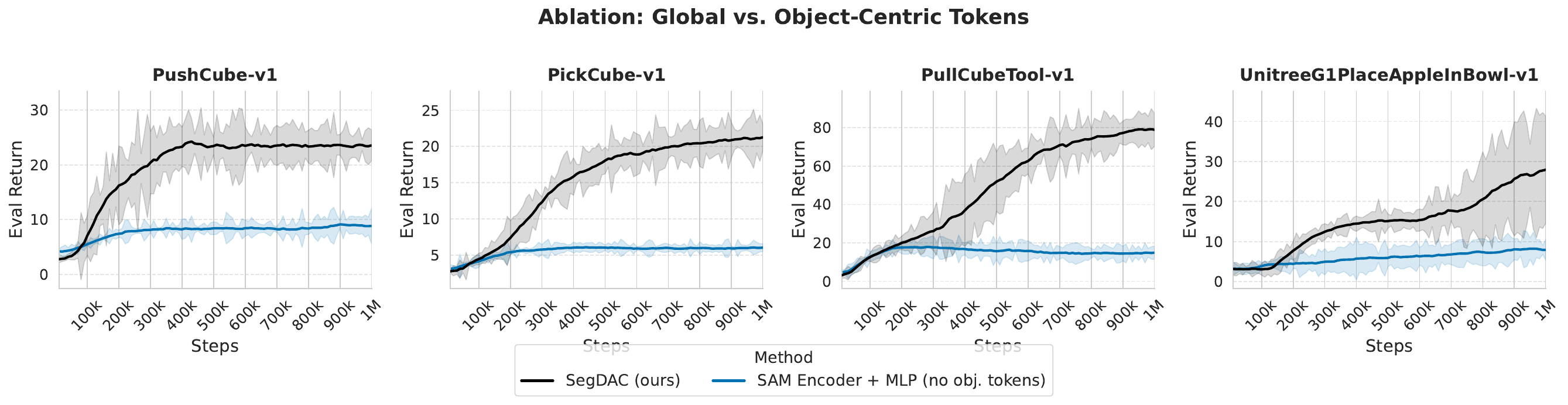}
\caption{Ablation studies. \textit{(Top)} Removing segment positional encoding consistently hurts sample efficiency. \textit{(Middle)} Fixing token count to 5 degrades performance on harder tasks. \textit{(Bottom)} Global SAM features collapse on 4 of 4 tasks shown here and 7 of 8 overall (remaining tasks in \Cref{sec:global_obj_suite}), confirming that object-level structure is the key driver of \methodName's gains. Shaded regions: $\pm$1 std (3 seeds; 5 for \textit{UnitreeG1PlaceAppleInBowl} in top, all tasks in bottom).}
\vspace{-1.5em}
\label{fig:ablations}
\end{figure}
\textbf{Segment positional encoding.} Removing our segment positional encoding does not prevent learning but consistently hurts sample efficiency: the ablated variant lags throughout training on all three tasks and shows substantially higher variance (\Cref{fig:ablations}). This confirms that pretrained patch features once merged via a mean operation do not necessarily implicitly encode the spatial grounding needed for precise manipulation. \textbf{Fixed vs.\ variable segment count.} A fixed budget of $N{=}5$ segments is competitive on the \textit{PushCube} task but degrades meaningfully on tasks with more complex manipulation and diverse segmentation (\Cref{fig:ablations}). The effect is most pronounced on \textit{UnitreeG1PlaceAppleInBowl}, where both variants track closely for the first 600K steps before diverging sharply, precisely when the policy must exploit fine-grained object location. Truncating segments is not a minor quantization error, it can discard task-critical information that cannot be recovered well when there aren't enough segments. \textbf{Global vs.\ per-object token.} Replacing per-object tokens with a single mean-pooled global feature causes consistent performance drops across 7 of 8 tasks (\Cref{fig:ablations}), showing that \methodName's gains are not purely inherited from the frozen encoder. Object-level structure must be explicitly preserved through segmentation and the transformer decoder, as strong features alone are not sufficient, even after hyperparameter tuning of the global representation baseline. \textbf{Text input sensitivity.} Synonym substitution (\Cref{sec:synonyms}) and a shared cross-task vocabulary (\Cref{sec:shared_text_inputs}) produce no measurable degradation, confirming that \methodName requires no prompt engineering. We note a slight increase in variance on one task when using a shared vocabulary, but performance remains stable across all tasks.

\subsection{Case Studies and Failure Analysis}
\label{sec:case_study_main}
Under hard perturbations, \methodName fails more gracefully than MaDi: rather than producing erratic behavior, it executes structured attempts that consistently miss by a small margin, preserving task intent. This suggests that object-centric representations maintain behavioral coherence even when generalization breaks down. Qualitative examples are provided in \Cref{sec:case_study}.

\section{Limitations and Future Directions}
\label{sec:limitations}
\methodName is evaluated on short-horizon manipulation tasks in simulation, consistent with prior model-free visual RL methods. As a memoryless policy, it cannot reason over object states across timesteps, which may limit performance on tasks requiring long-horizon planning. 2D positional encodings may also not generalize to extreme viewpoint shifts, making 3D-aware or multi-view extensions a natural next step. Finally, while the policy uses text inputs for open-vocabulary detection, it does not condition on language instructions, and extending \methodName toward instruction-following agents and real-world deployment are promising directions for future work.

\section{Conclusion}
\label{sec:conclusion}
We introduced \methodName, a transformer-based actor-critic that efficiently learns from dynamic object-centric tokens built from frozen pretrained vision models. On our ManiSkill3 visual generalization benchmark, \methodName improves over prior methods by up to 88\% on the hardest settings while matching DrQ-v2 in sample efficiency. Our ablations confirm that each design choice is individually necessary, with importance scaling with task difficulty. Reasoning over objects rather than pixels provides a favorable inductive bias for both efficient learning and robust generalization in visual RL.

\bibliography{main}
\bibliographystyle{rlj}

\beginSupplementaryMaterials

\section{Visual Generalization Benchmark Definition}
\label{appendix_vis_gen}

\subsection{Visual Perturbation Categories}
\begin{table}[H]
\centering
\begin{tabular}{|l|l|}
\hline
\textbf{Category} & \textbf{Perturbations} \\
\hline
Camera            & camera pose, camera FOV \\
\hline
Lighting          & lighting direction, lighting color \\
\hline
Color             & MO color, RO color, table color, ground color \\
\hline
Texture           & MO texture, RO texture, table texture, ground texture \\
\hline
\end{tabular}
\caption{Perturbation categories and their corresponding visual perturbation tests.}
\label{tab:appendix_cateogyr_perturbations}
\end{table}
Table~\ref{tab:appendix_cateogyr_perturbations} lists the visual perturbation categories used in Figure~\ref{fig:visual_generalization}. 

\subsection{Model Selection}
For all methods, we tested the final model (for each seed) after 1 million training steps. Early stopping had no significant effect on results, so using the final checkpoint ensured consistency and simplicity across tests.

\subsection{Scores Aggregation}
\label{sec:scores_aggregation}
Scores are computed using normalized returns, where each return is divided by the maximum number of steps for the task, yielding values in $[0,1]$.
To compute benchmark returns, we first evaluate baseline performance on the unperturbed tasks by running 50 rollouts per task and per model seed. Each average return (over 50 rollouts) is computed for all $m = 8$ tasks and $n = 5$ seeds, yielding $m \times n = 40$ scores. We then compute the IQM and 95\% confidence intervals using stratified bootstrap with 50,000 replications, following \citet{agarwal2022deepreinforcementlearningedge}. 

For individual perturbation results (e.g., Table~\ref{tab:appendix_camerafovtest_easy}), we set $m = 1$ (a single task) and $n = 5$ (seeds), with each score averaged over 50 rollouts. For category-level scores (e.g., Camera Easy, Lighting Easy), $m$ is the total number of perturbation-task combinations in that category. For instance, both the Camera and Lighting categories include two perturbations (Pose/FOV and Direction/Color, respectively) applied to 8 tasks, resulting in $m = 2 \times 8 = 16$.

Some perturbation categories involve a receiving object (RO), such as RO Texture and RO Color in the Texture and Color categories. Tasks that do not include a RO (e.g., LiftPegUpright) cannot be evaluated on these specific perturbations, reducing $m$ by one for those tasks. As a result, the total number of perturbation-task pairs varies across categories depending on RO availability. For both Texture and Color, we obtain $m = 29$ after summing over all tasks while accounting for the presence or absence of ROs. A summary of the number of valid perturbation-task pairs per category, accounting for RO presence, is shown in Table~\ref{tab:perturbation_tests_per_category}.

For the overall scores, we aggregated results across all tasks and all perturbations for each difficulty level, yielding a total of $80 + 80 + 145 + 145 = 450$ test cases per difficulty. We then computed the IQM over these 450 results, providing a robust and reliable evaluation.

\begin{table}[H]
\centering
\begin{tabular}{lc}
\hline
\textbf{Perturbation Category} & \textbf{\# Tests ($m \times n$)} \\
\hline
Camera & $16 \times 5 = 80$ \\
Lighting & $16 \times 5 = 80$ \\
Color & $29 \times 5 = 145$ \\
Texture & $29 \times 5 = 145$ \\
\hline
\end{tabular}
\caption{Number of test scores per visual perturbation category, \textbf{for each test} we perform \textbf{50 rollouts} to compute performance metrics.}
\label{tab:perturbation_tests_per_category}
\end{table}

\subsection{Difficulty Textures Examples}
\label{sec:difficulity_textures_examples}
To illustrate the impact of benchmark difficulty on scene appearance, we include Blender renders of selected visual perturbations and full-scene previews for each task. Not all Blender renders are shown, the complete set is available at \url{https://github.com/SegDAC/SegDAC}. The textures and colors are also task-specific, for example, in the Pick Cube task, the default cube is red and the target is green. Accordingly, the texture perturbations vary in difficulty: the easy texture resembles metallic red (similar to the default), the medium texture is unrelated to either red or green, and the hard texture uses green dots resembling the target to induce confusion. 
\begin{figure}[ht]
    \centering
    \begin{subfigure}[b]{0.24\linewidth}
        \includegraphics[width=\linewidth]{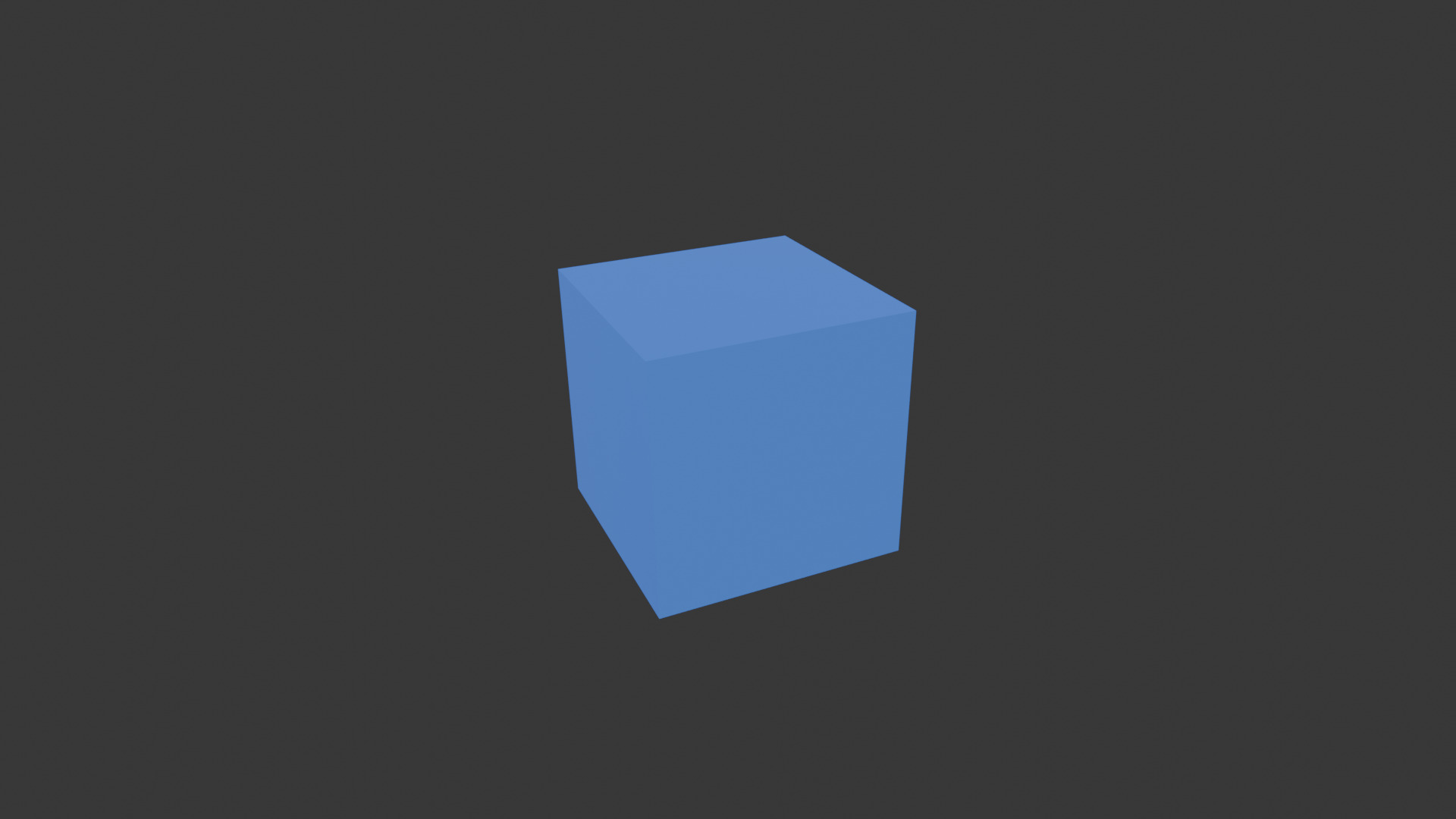}
        \caption{Default Cube}
    \end{subfigure}
    \hfill
    \begin{subfigure}[b]{0.24\linewidth}
        \includegraphics[width=\linewidth]{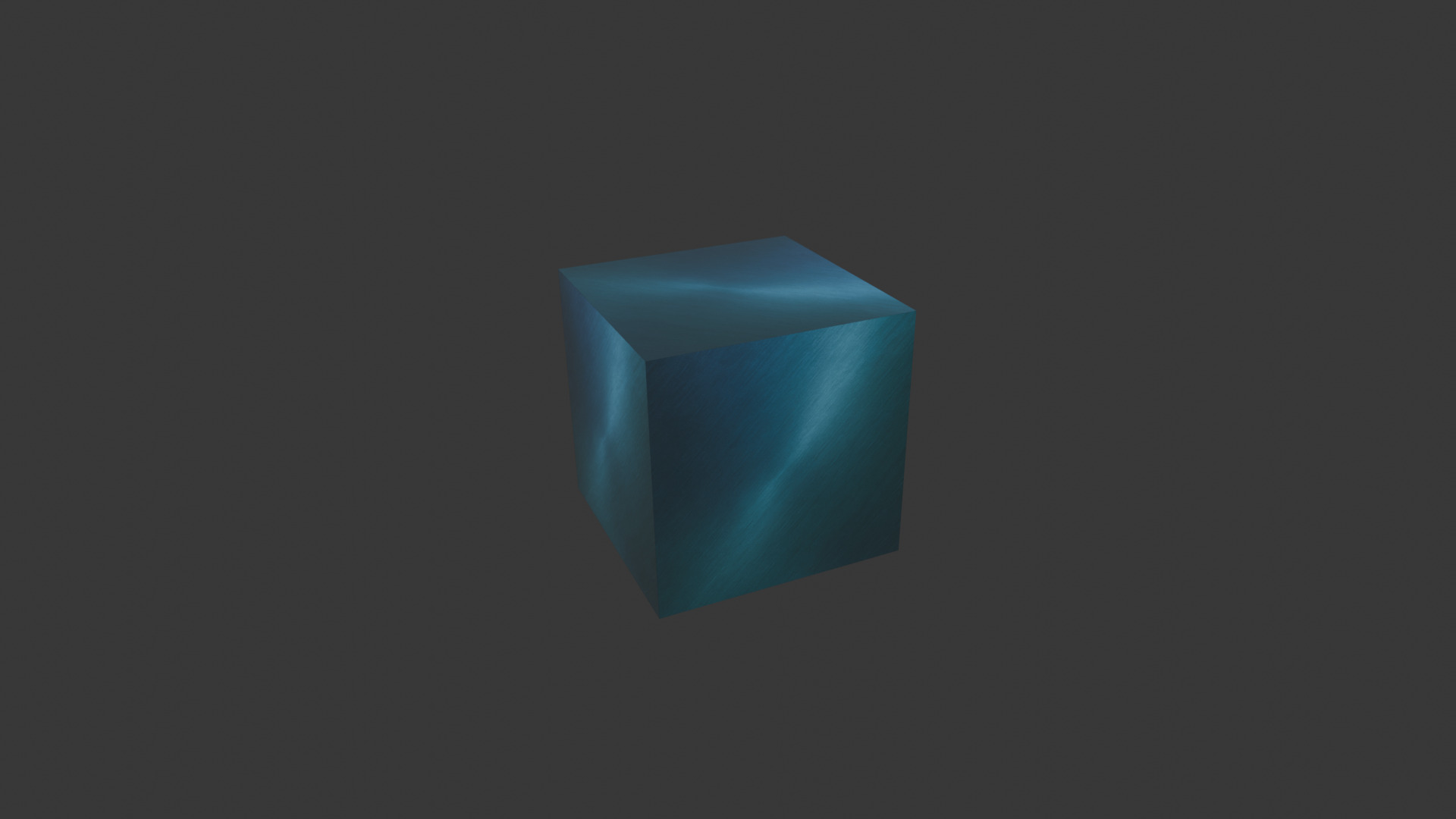}
        \caption{Easy Cube}
    \end{subfigure}
    \hfill
    \begin{subfigure}[b]{0.24\linewidth}
        \includegraphics[width=\linewidth]{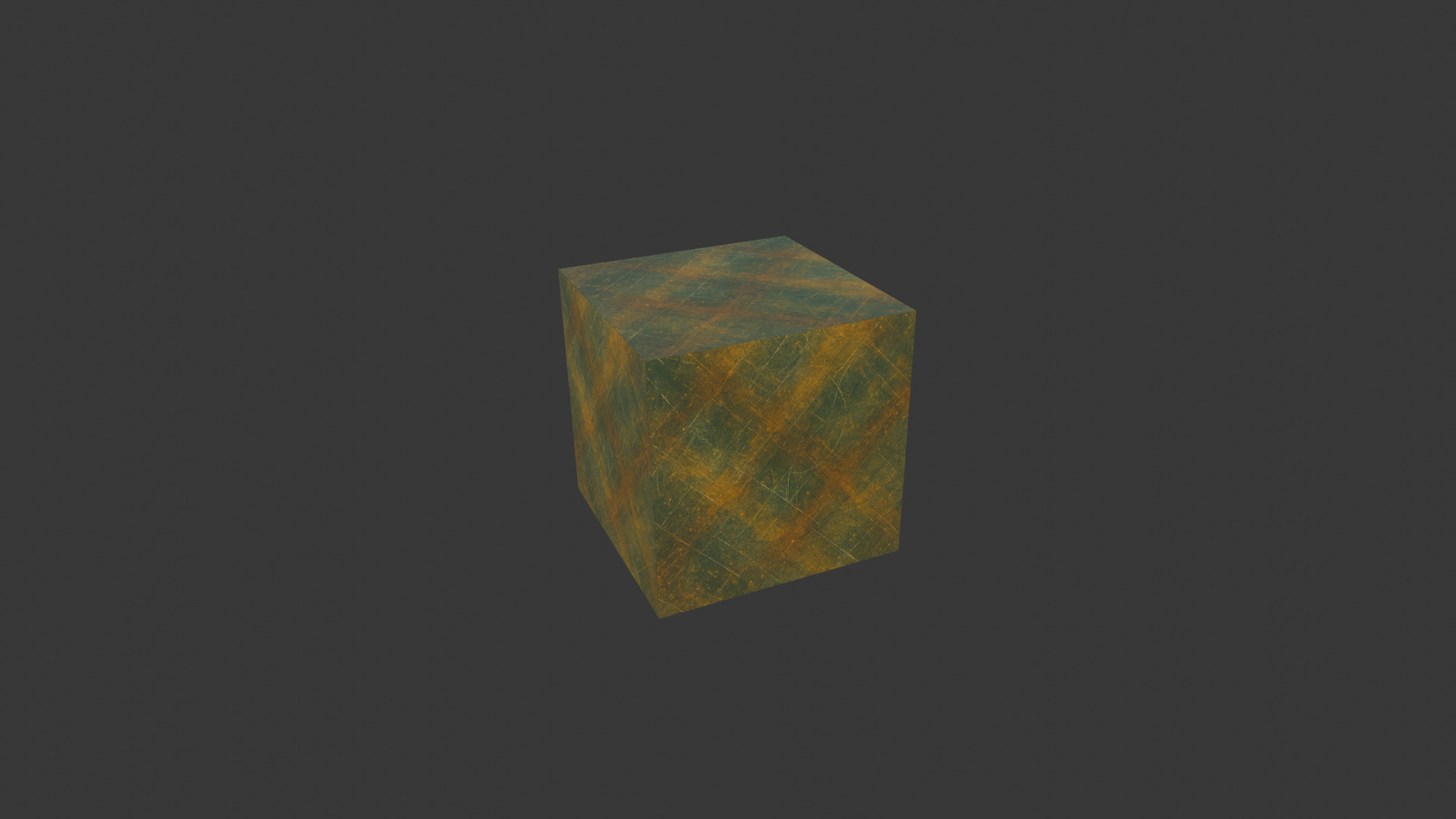}
        \caption{Medium Cube}
    \end{subfigure}
    \hfill
    \begin{subfigure}[b]{0.24\linewidth}
        \includegraphics[width=\linewidth]{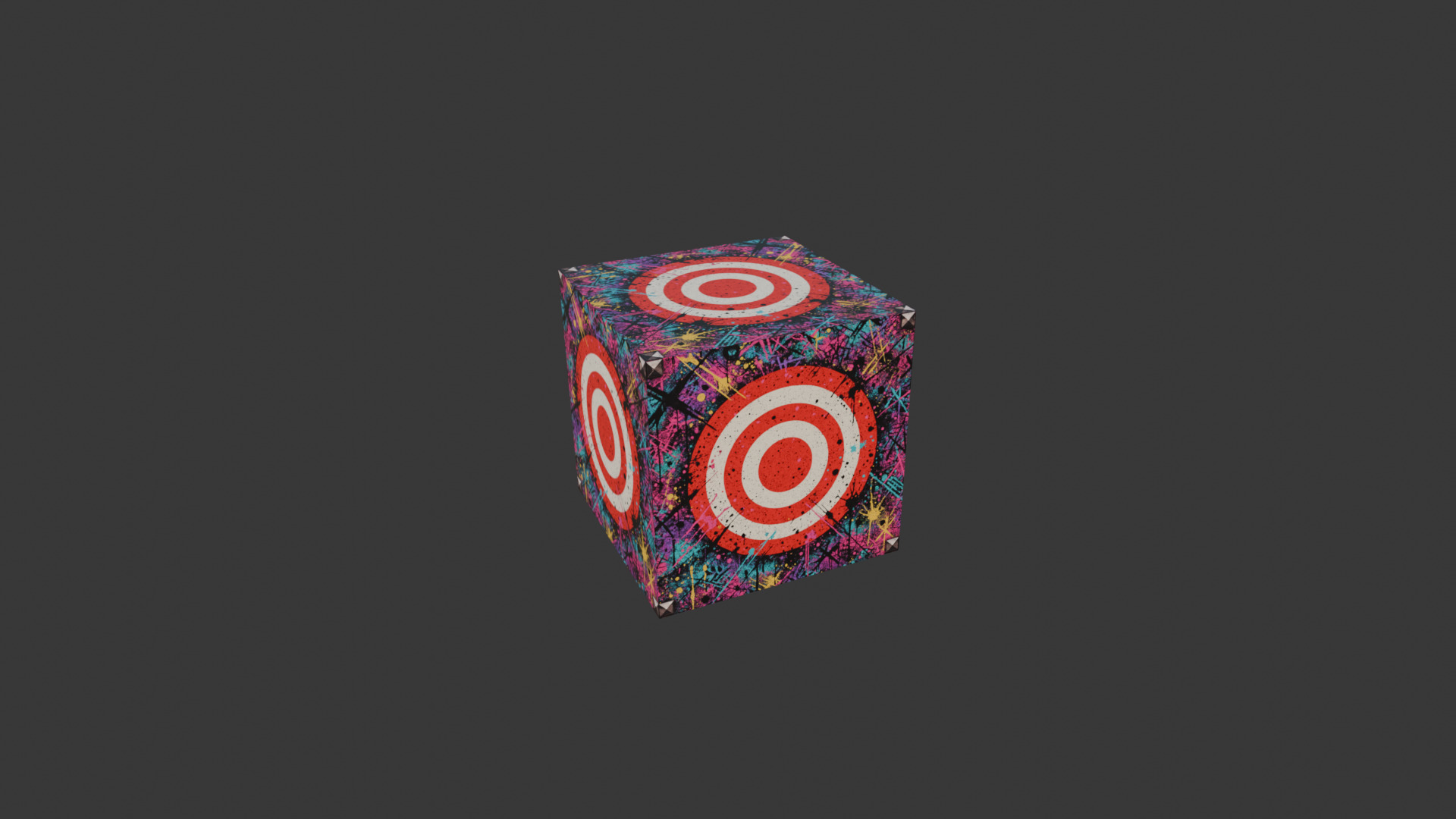}
        \caption{Hard Cube}
    \end{subfigure}
    \caption{Blender renders for the cube, used in tasks like push/pull/poke cube.}
    \label{fig:appendix_visual_gen_blender_renders_cube}
\end{figure}
Figure~\ref{fig:appendix_visual_gen_blender_renders_cube} shows how cube appearance changes with perturbation difficulty. The default cube (used when no perturbations are applied) is a solid light blue. The easy texture remains blue with minor variations, such as darker shades and simple patterns, maintaining high visual similarity with the default cube. The medium texture introduces green and yellow-brown tones with slightly more complex patterns, it does not share a lot (if any) visual similarity with the default cube while avoiding conflicts with other scene elements (e.g: robot arm, table, or target). The hard texture is visually and semantically disruptive: it features a chaotic mix of colors and complex patterns, with each cube face displaying a target identical to the task’s actual target. This creates a semantic clash, making it harder for the policy to recognize the cube. These examples demonstrate that our difficulty levels correspond to meaningful and progressively more challenging visual changes.

\begin{figure}[H]
    \centering
    \begin{subfigure}[b]{0.24\linewidth}
        \centering
        \includegraphics[width=1.9cm,height=1.9cm]{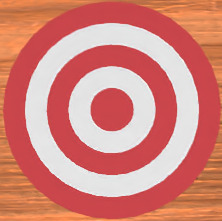}
        \caption{Default Target}
    \end{subfigure}
    \hfill
    \begin{subfigure}[b]{0.24\linewidth}
        \includegraphics[width=\linewidth]{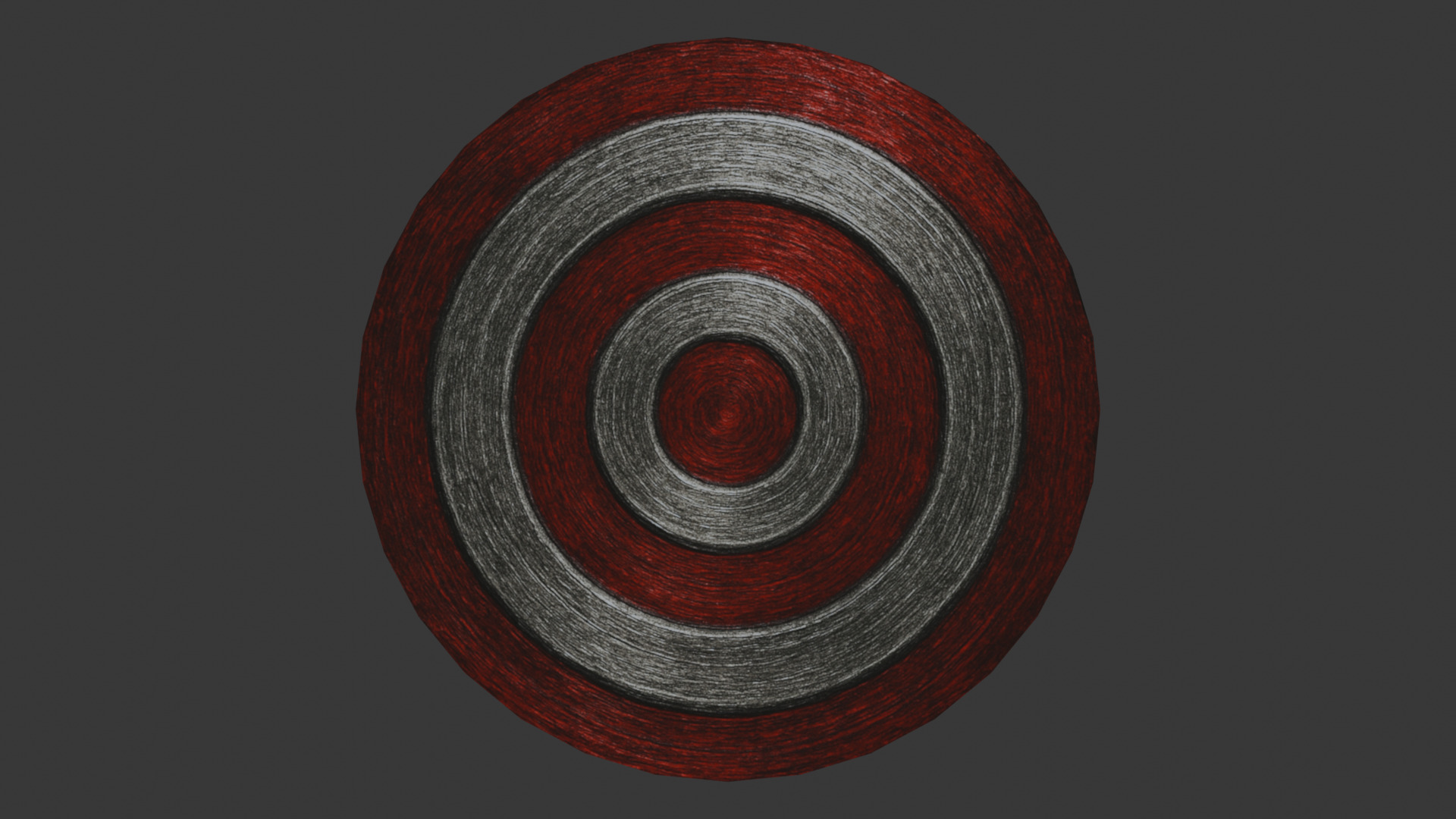}
        \caption{Easy Target}
    \end{subfigure}
    \hfill
    \begin{subfigure}[b]{0.24\linewidth}
        \includegraphics[width=\linewidth]{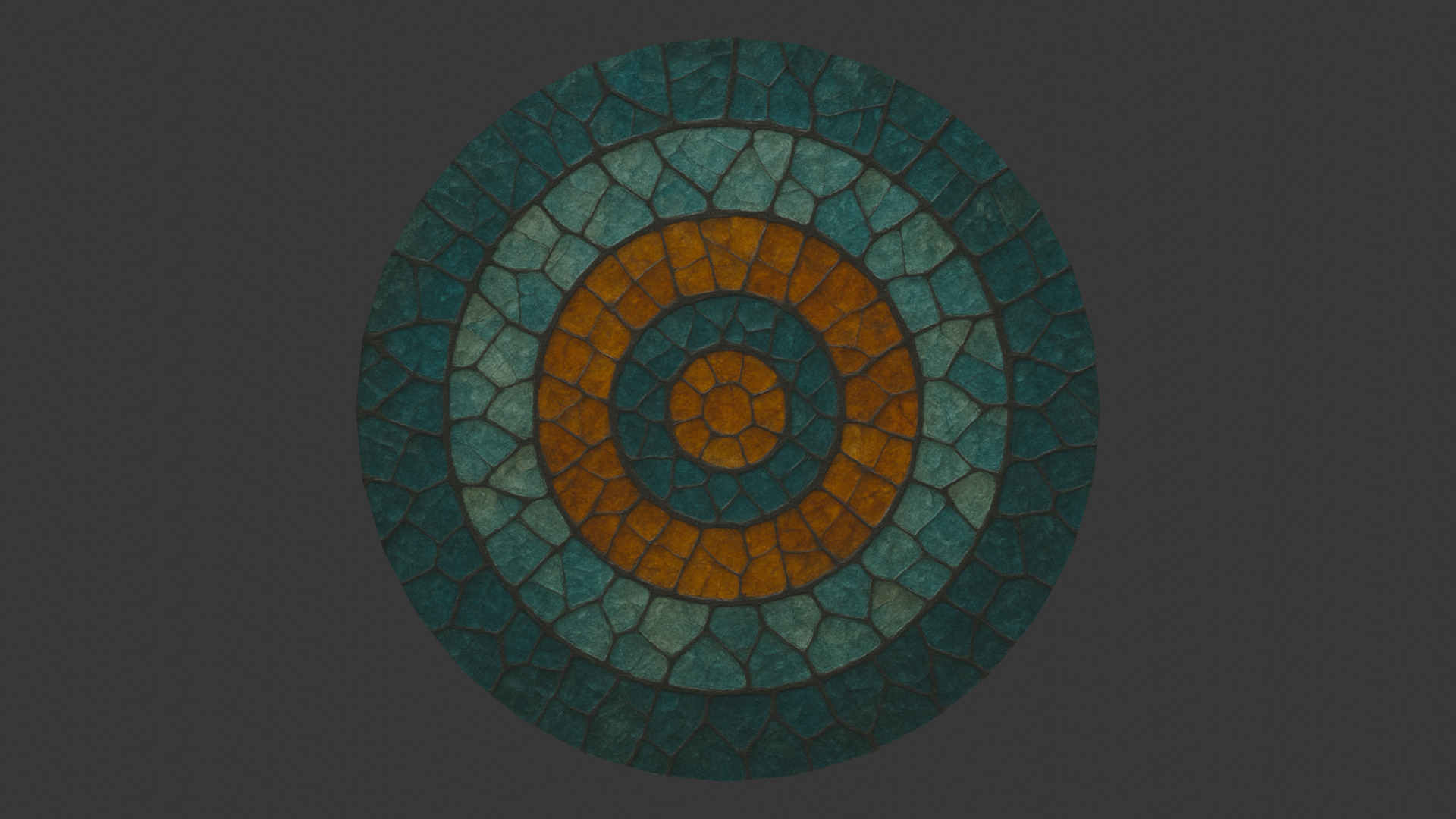}
        \caption{Medium Target}
    \end{subfigure}
    \hfill
    \begin{subfigure}[b]{0.24\linewidth}
        \includegraphics[width=\linewidth]{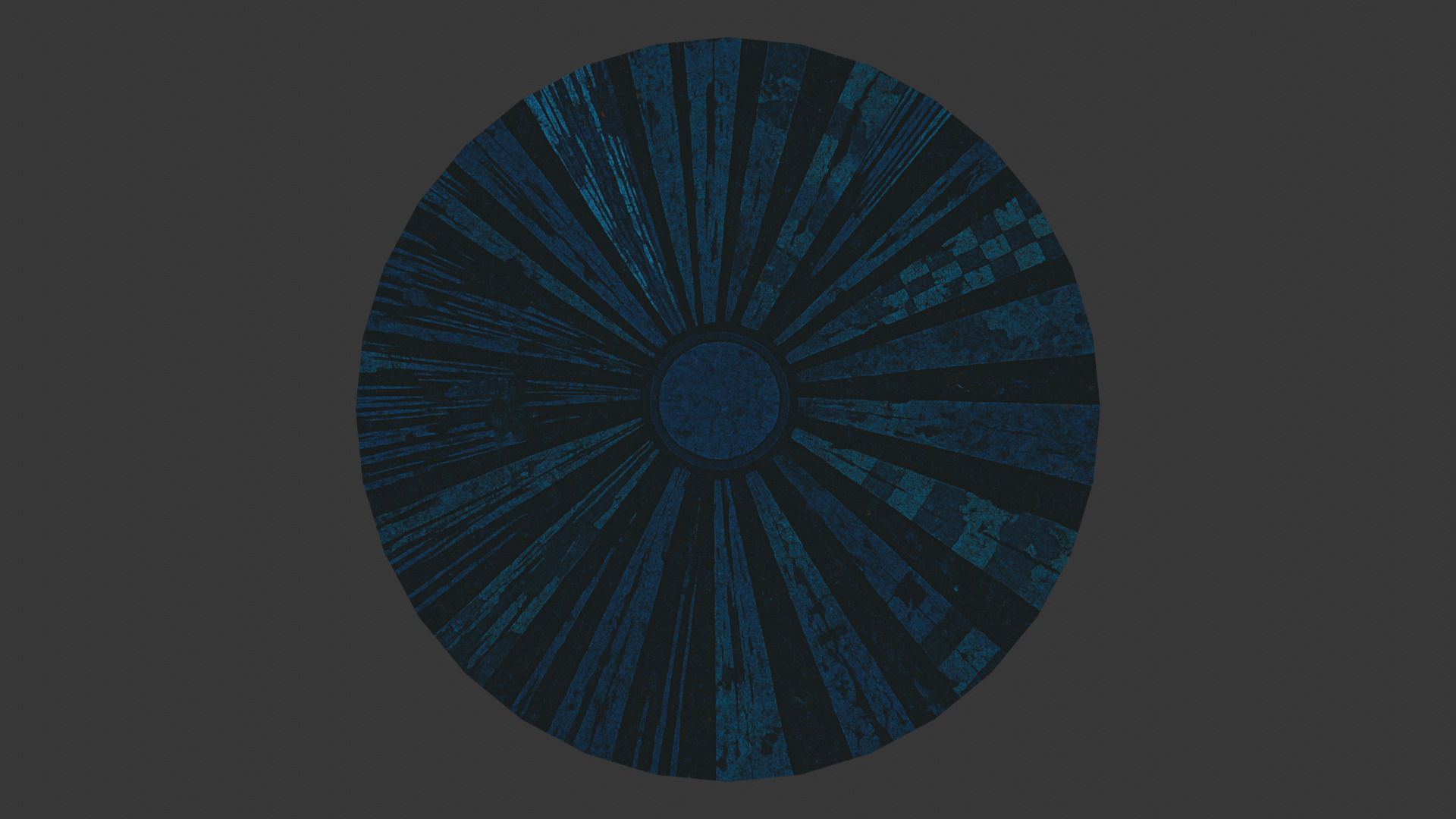}
        \caption{Hard Target}
    \end{subfigure}
    \caption{Blender renders for the target.}
    \label{fig:appendix_visual_gen_blender_renders_target}
\end{figure}
Figure~\ref{fig:appendix_visual_gen_blender_renders_target} illustrates the different target textures across difficulty levels. The easy texture closely resembles the default, with red and white nested rings and a slight metallic finish to introduce subtle variation. The medium texture uses entirely different colors and more complex patterns, but retains the same nested ring structure and number of rings. The hard texture is significantly different: it replaces the nested rings with inward-pointing stripes, resembling a dartboard. The color scheme also changes drastically (e.g., blue instead of red), making the target visually and semantically harder to recognize.

\begin{figure}[H]
    \centering
    \begin{subfigure}[b]{0.24\linewidth}
        \centering
        \includegraphics[width=\linewidth]{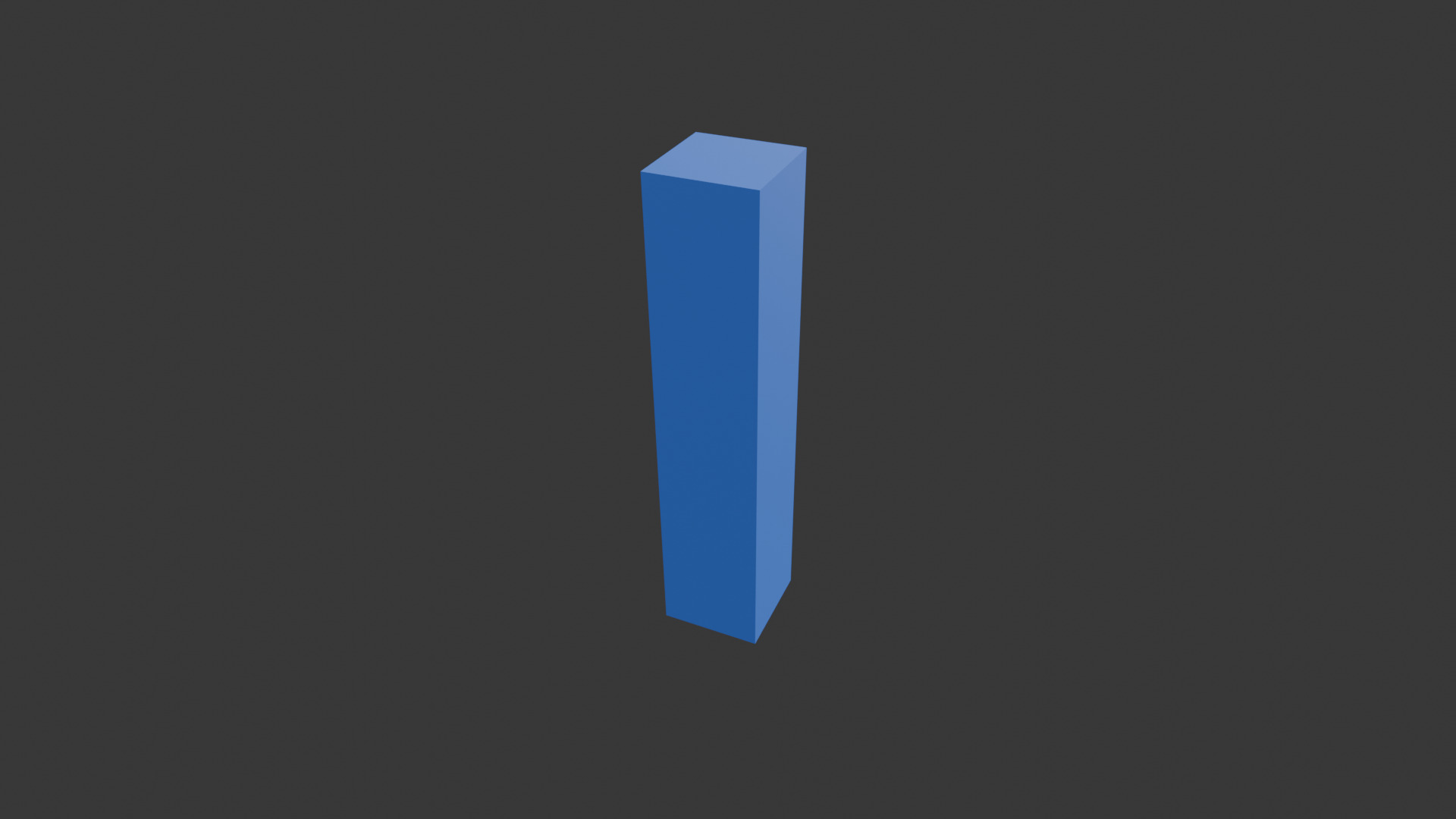}
        \caption{Default Peg}
    \end{subfigure}
    \hfill
    \begin{subfigure}[b]{0.24\linewidth}
        \includegraphics[width=\linewidth]{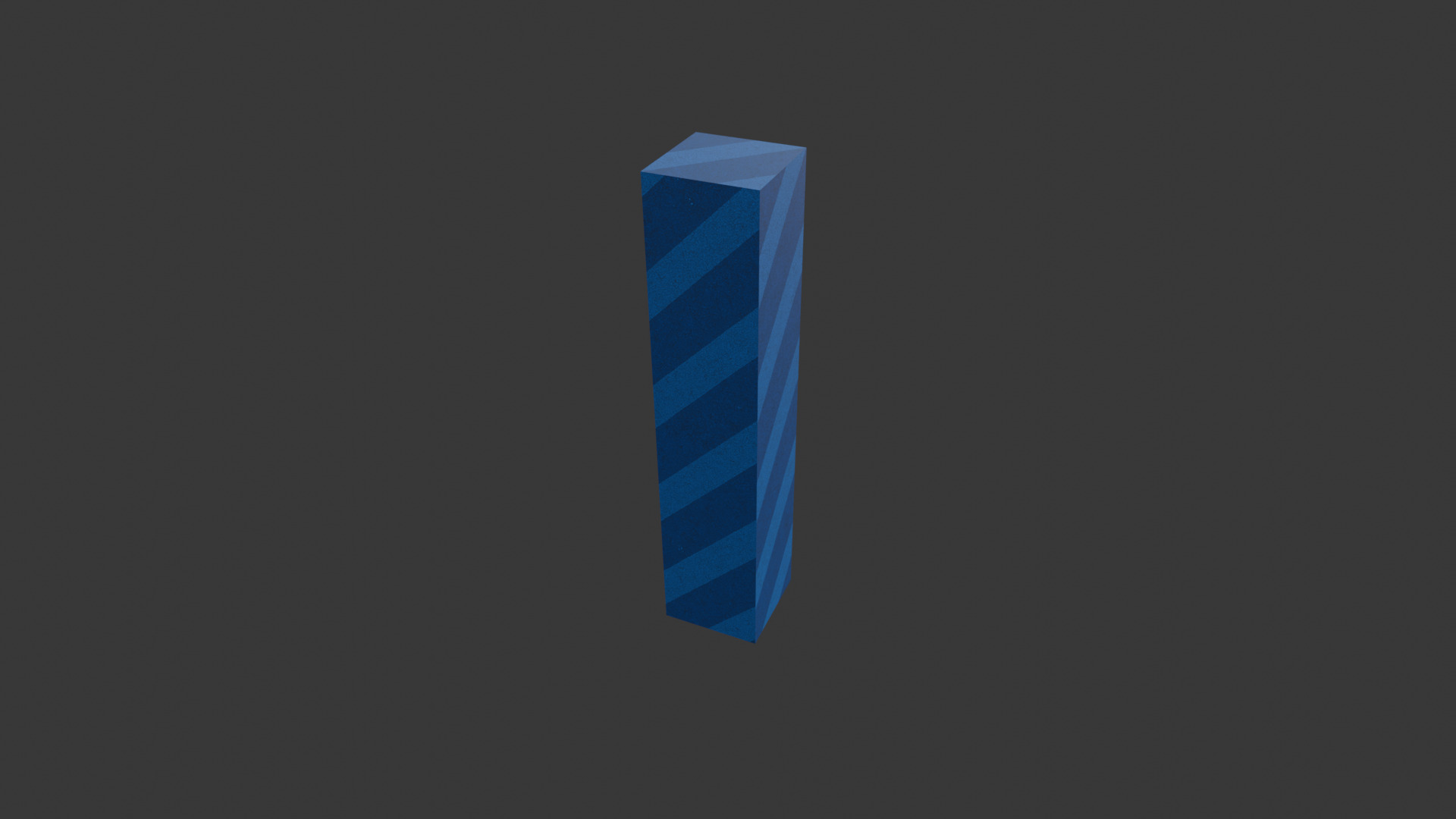}
        \caption{Easy Peg}
    \end{subfigure}
    \hfill
    \begin{subfigure}[b]{0.24\linewidth}
        \includegraphics[width=\linewidth]{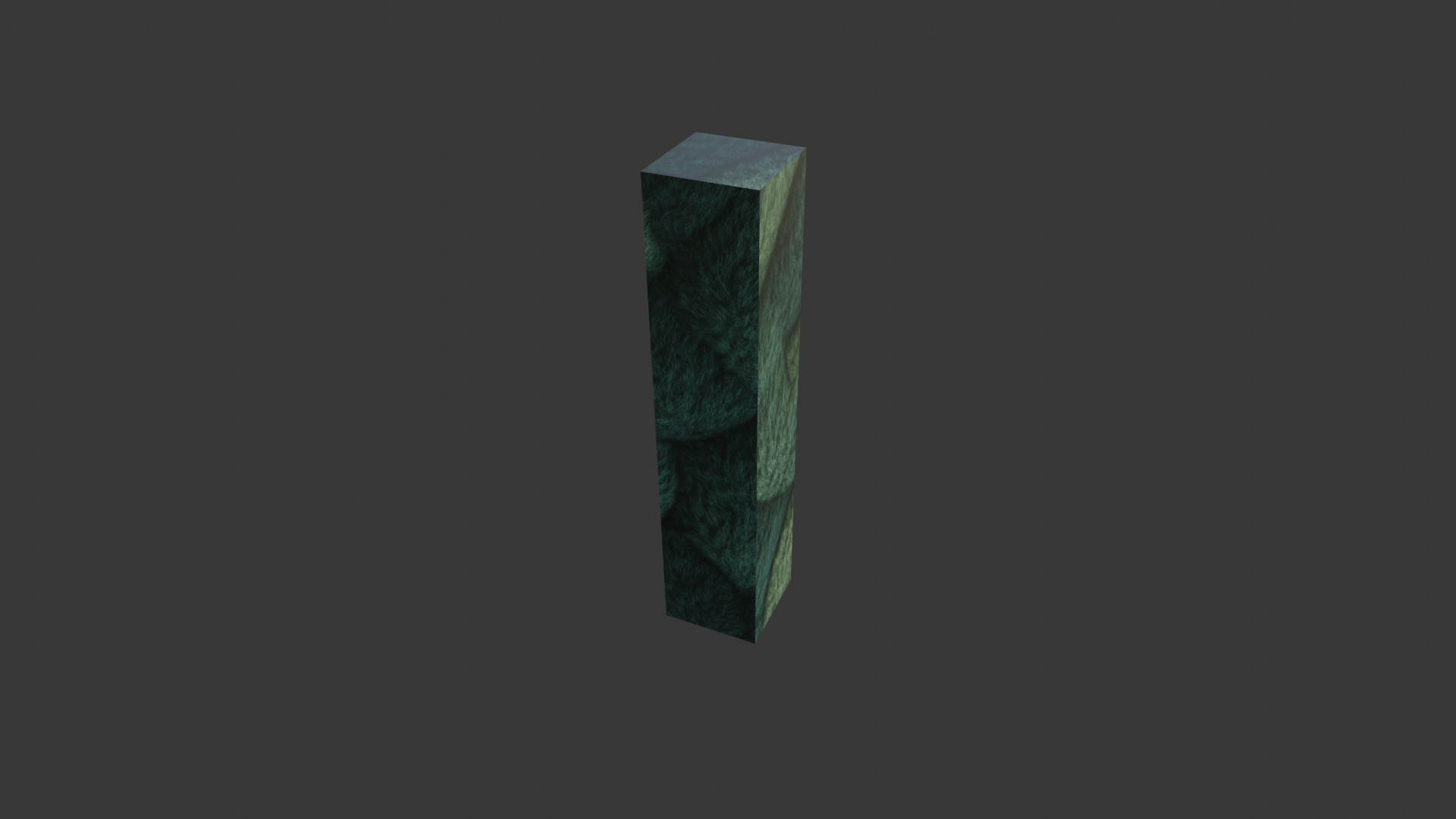}
        \caption{Medium Peg}
    \end{subfigure}
    \hfill
    \begin{subfigure}[b]{0.24\linewidth}
        \includegraphics[width=\linewidth]{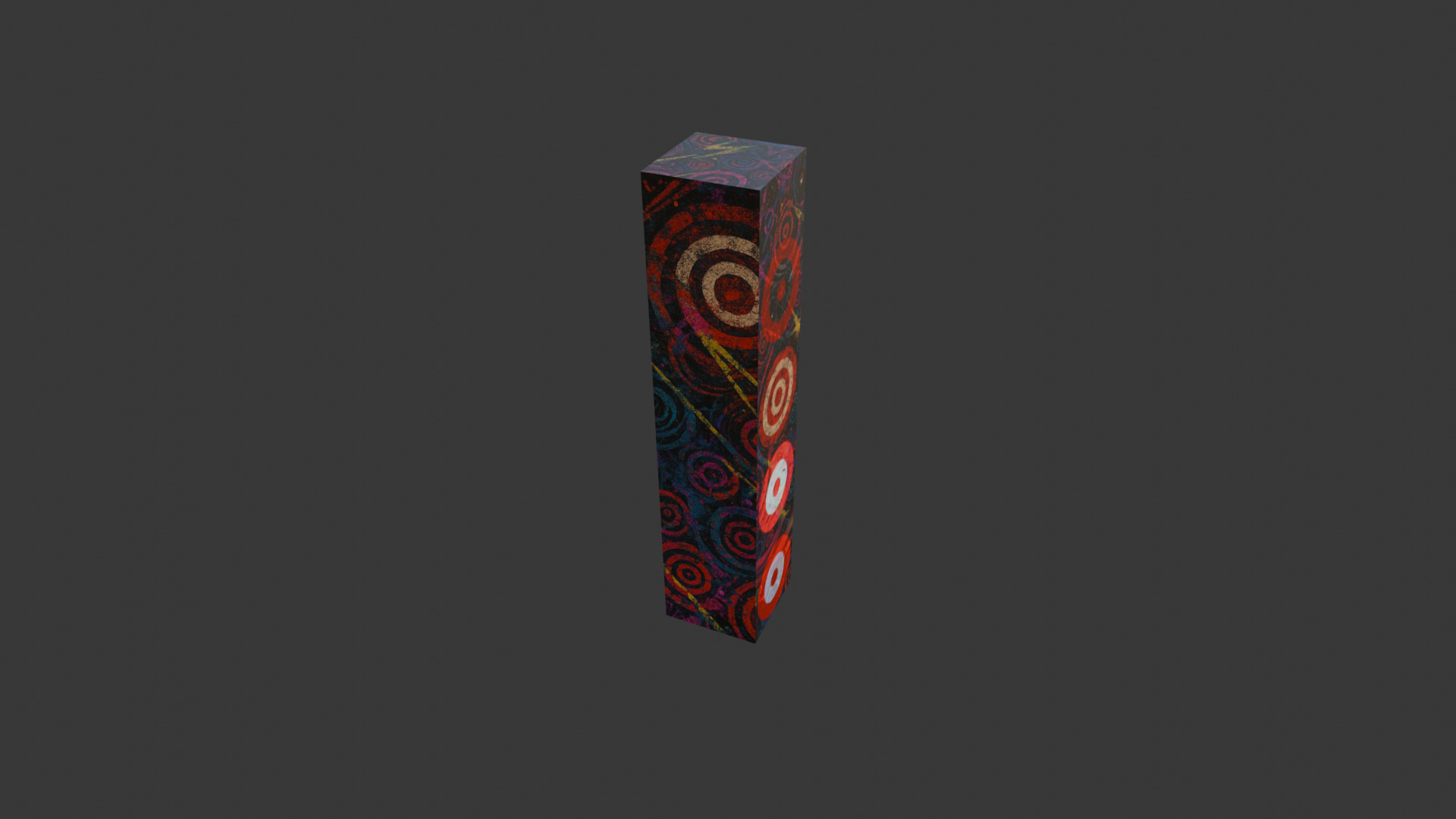}
        \caption{Hard Peg}
    \end{subfigure}
    \caption{Blender renders for the peg used in the poke cube task.}
    \label{fig:appendix_visual_gen_blender_renders_peg}
\end{figure}
Figure~\ref{fig:appendix_visual_gen_blender_renders_peg} shows the peg textures across difficulty levels. While multiple tasks include a peg, their textures may differ, the renders shown here are for the poke cube task. The easy texture shares strong visual similarity with the default view, as both are blue. The medium texture uses green, a neutral color that does not resemble any other object in the scene, and features a more complex pattern. The hard texture is visually and semantically distinct, with no similarity to the default. It includes repeated target symbols similar to those used for the task, increasing ambiguity and making the object harder to identify correctly.

\begin{figure}[H]
    \centering
    \begin{subfigure}[b]{0.24\linewidth}
        \centering
        \includegraphics[width=\linewidth]{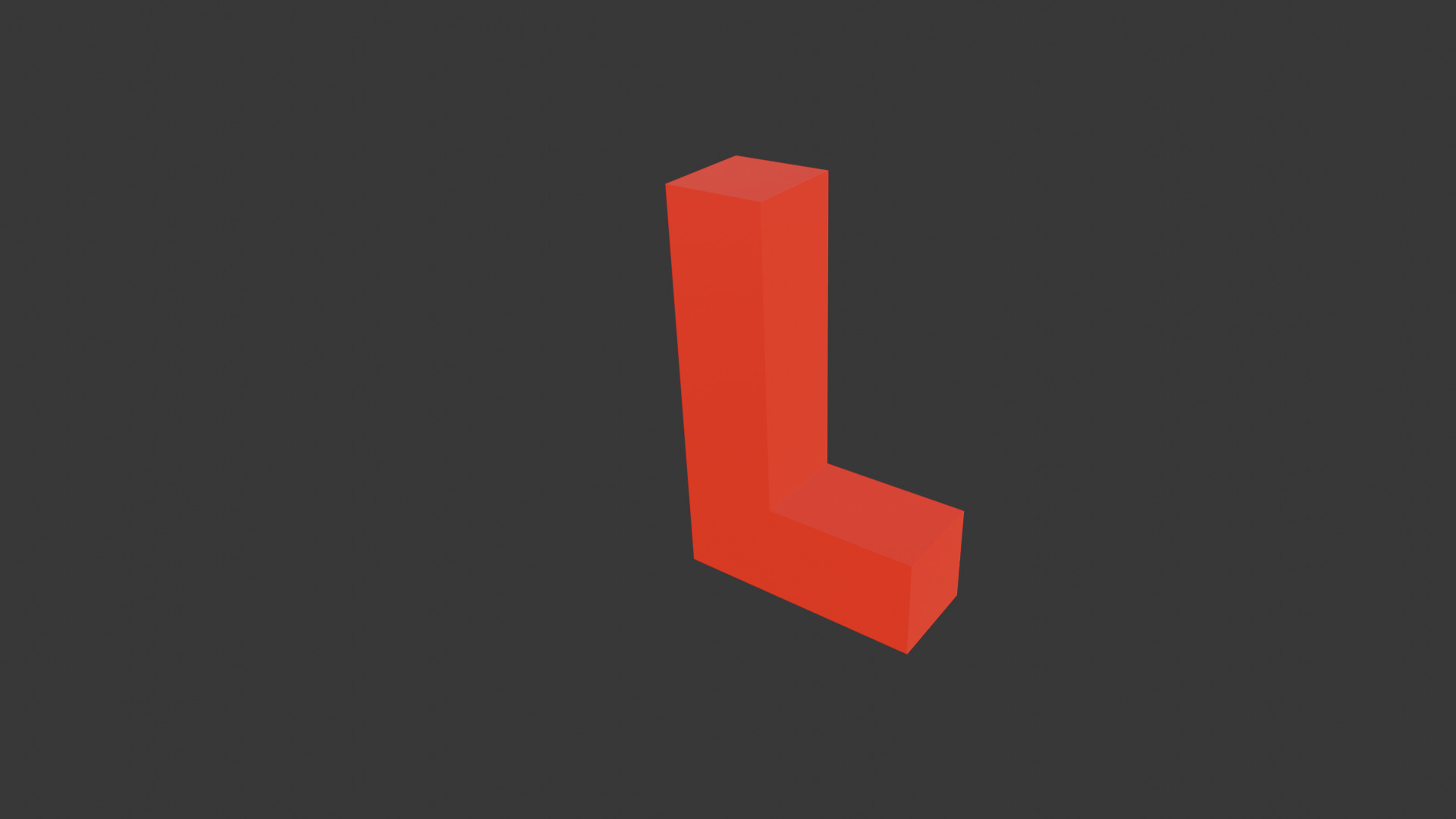}
        \caption{Default Tool}
    \end{subfigure}
    \hfill
    \begin{subfigure}[b]{0.24\linewidth}
        \includegraphics[width=\linewidth]{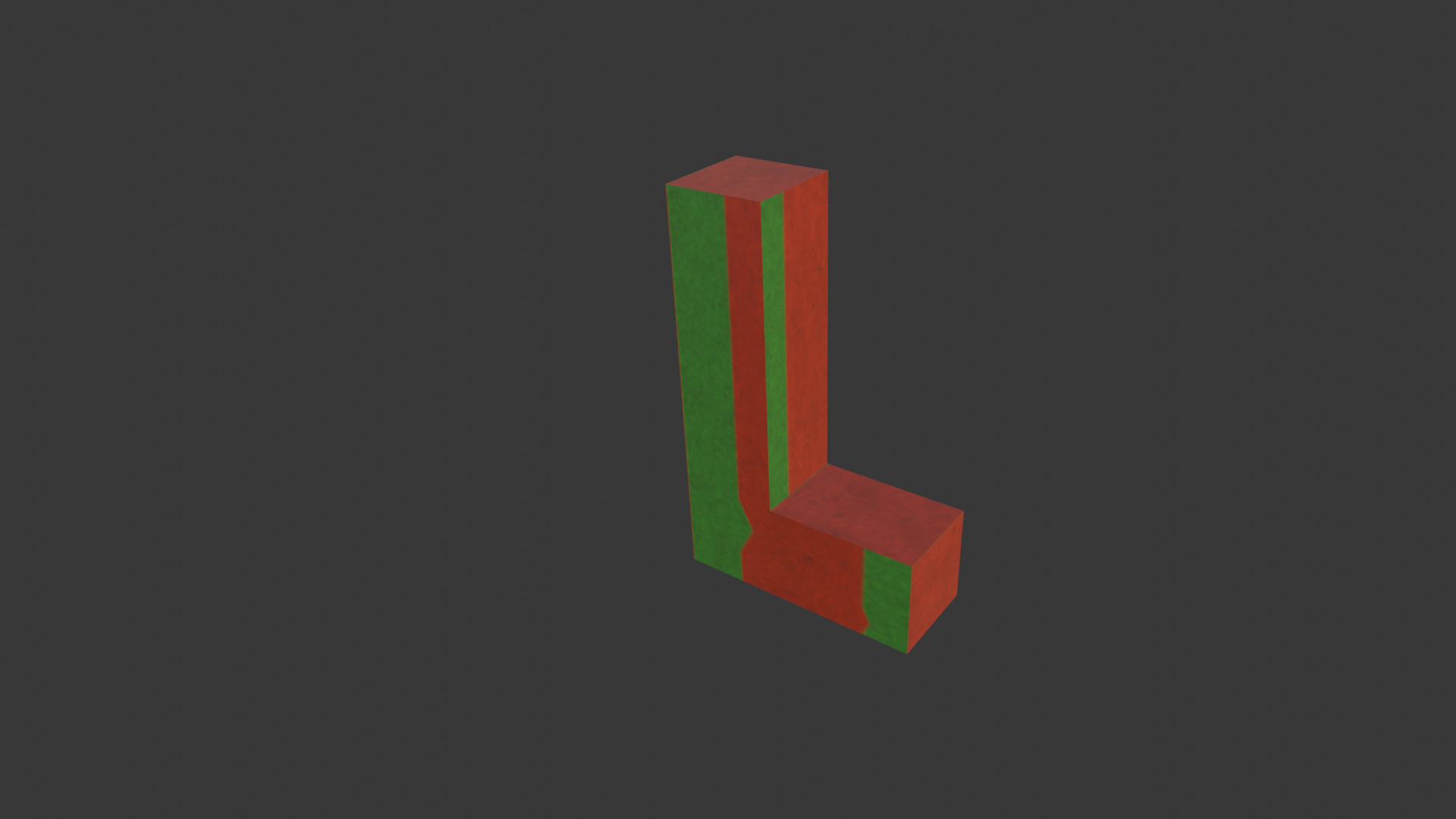}
        \caption{Easy Tool}
    \end{subfigure}
    \hfill
    \begin{subfigure}[b]{0.24\linewidth}
        \includegraphics[width=\linewidth]{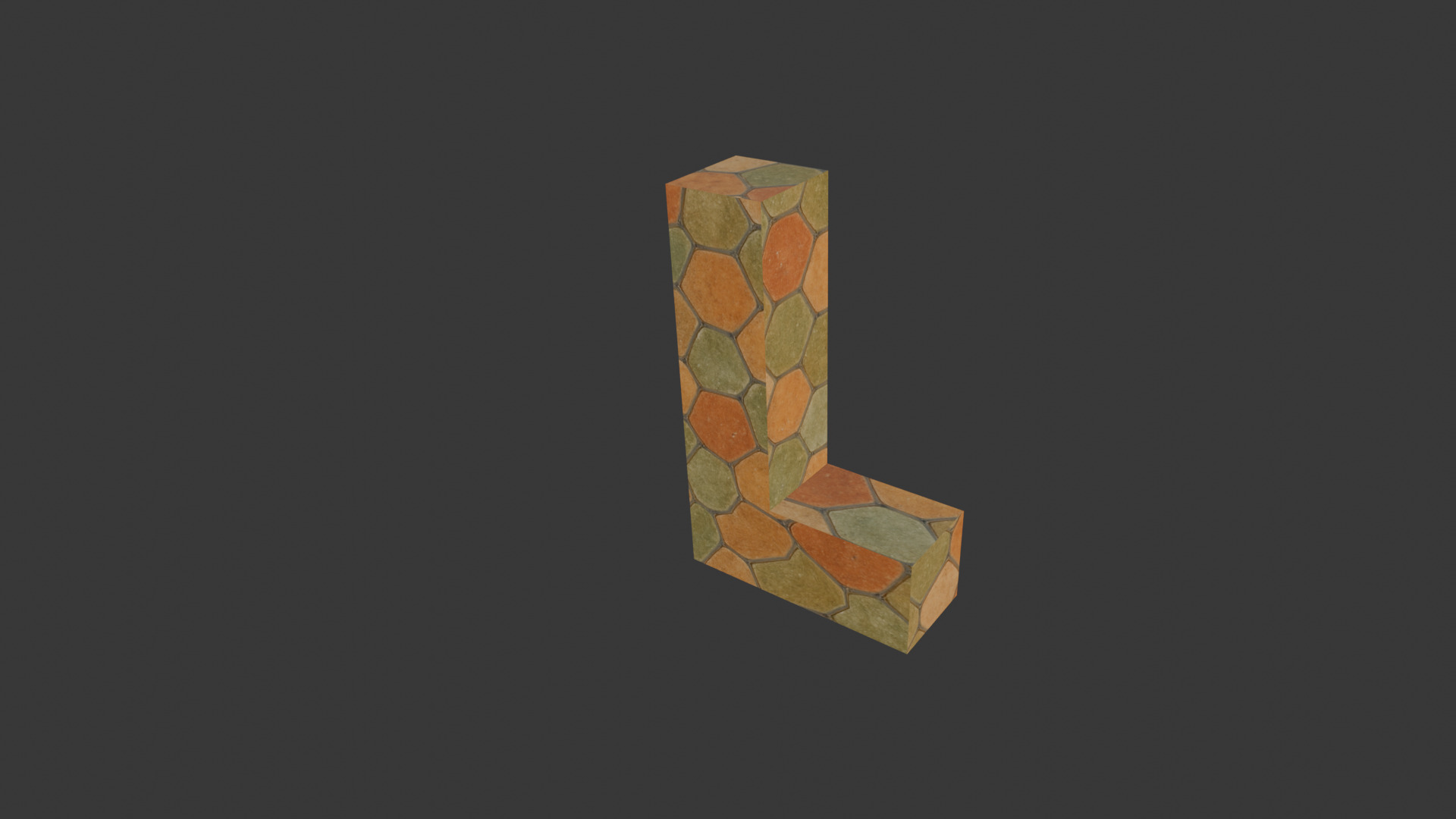}
        \caption{Medium Tool}
    \end{subfigure}
    \hfill
    \begin{subfigure}[b]{0.24\linewidth}
        \includegraphics[width=\linewidth]{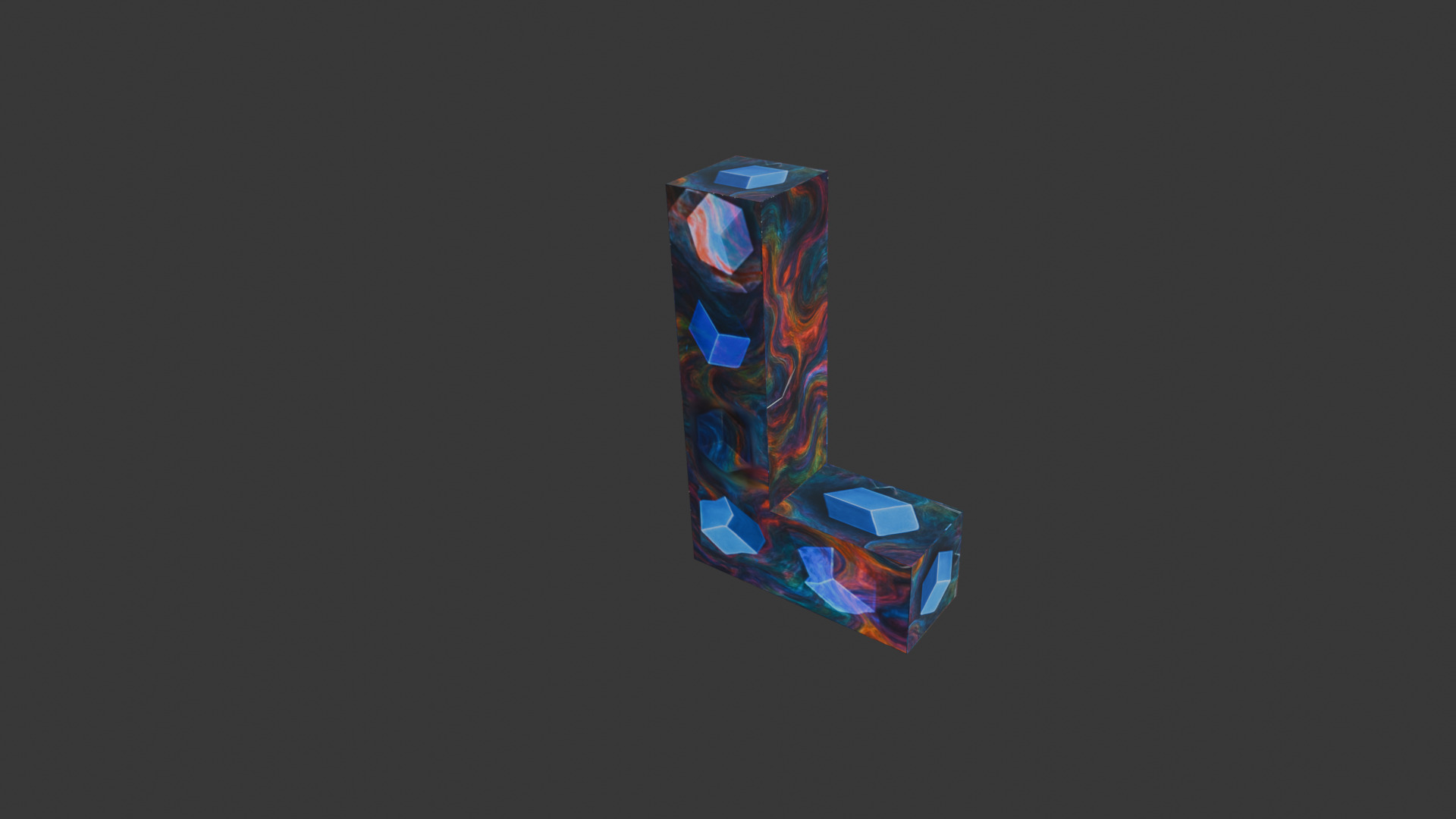}
        \caption{Hard Tool}
    \end{subfigure}
    \caption{Blender renders for the l-shaped tool used in the pull cube tool task.}
    \label{fig:appendix_visual_gen_blender_renders_l_shaped_tool}
\end{figure}

\begin{figure}[H]
    \centering
    \begin{subfigure}[b]{0.24\linewidth}
        \centering
        \includegraphics[width=\linewidth]{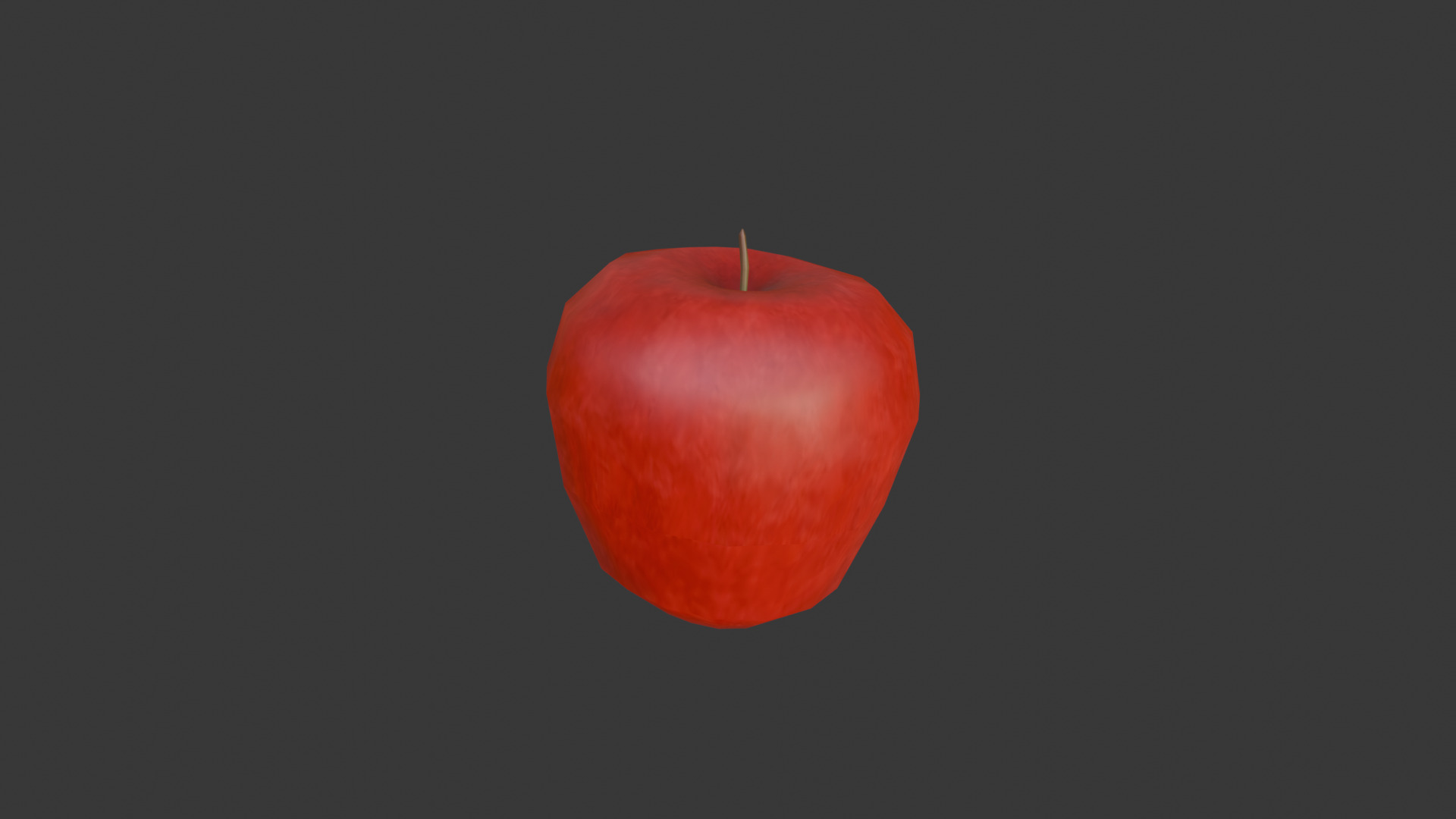}
        \caption{Default Apple}
    \end{subfigure}
    \hfill
    \begin{subfigure}[b]{0.24\linewidth}
        \includegraphics[width=\linewidth]{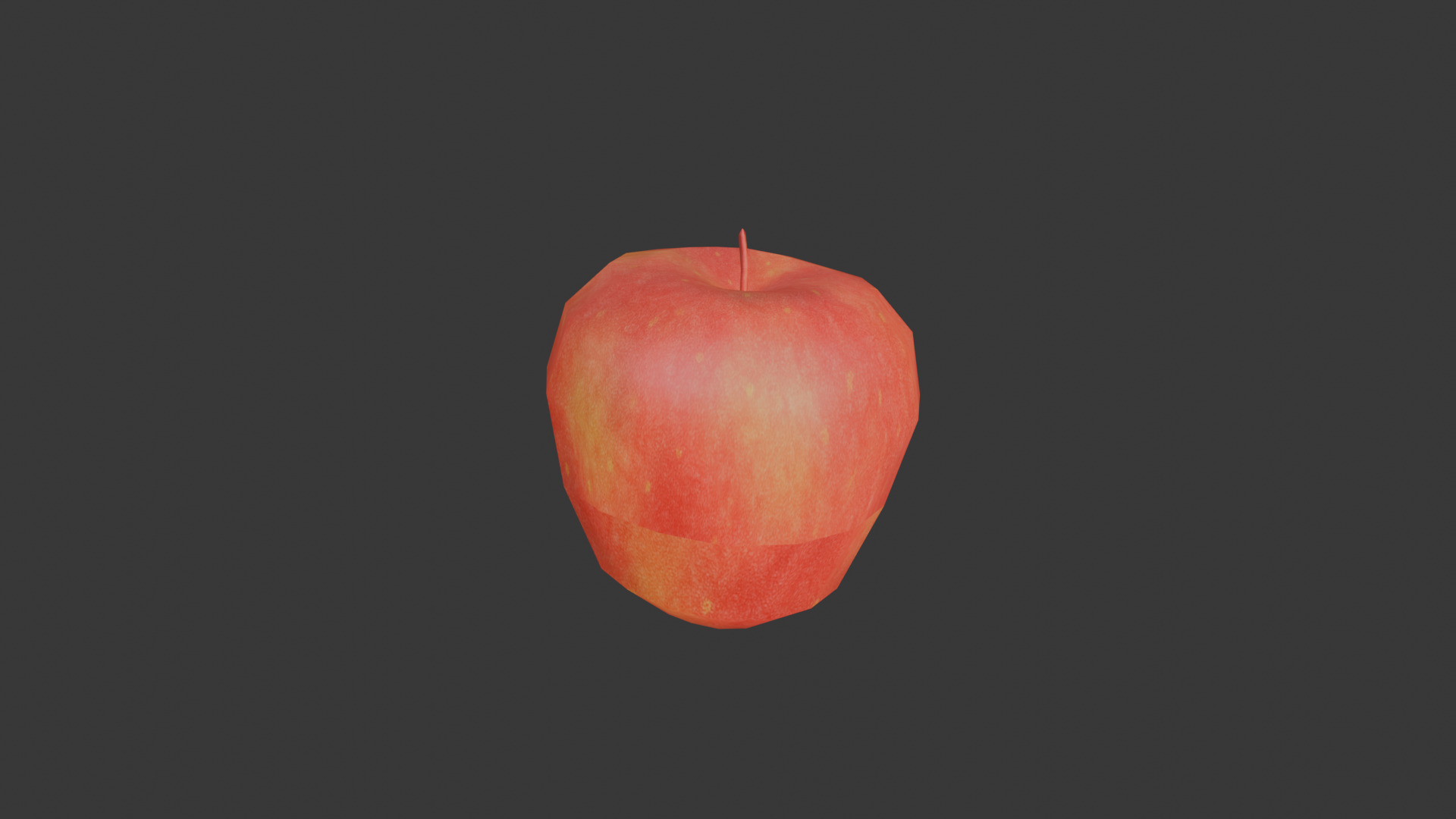}
        \caption{Easy Apple}
    \end{subfigure}
    \hfill
    \begin{subfigure}[b]{0.24\linewidth}
        \includegraphics[width=\linewidth]{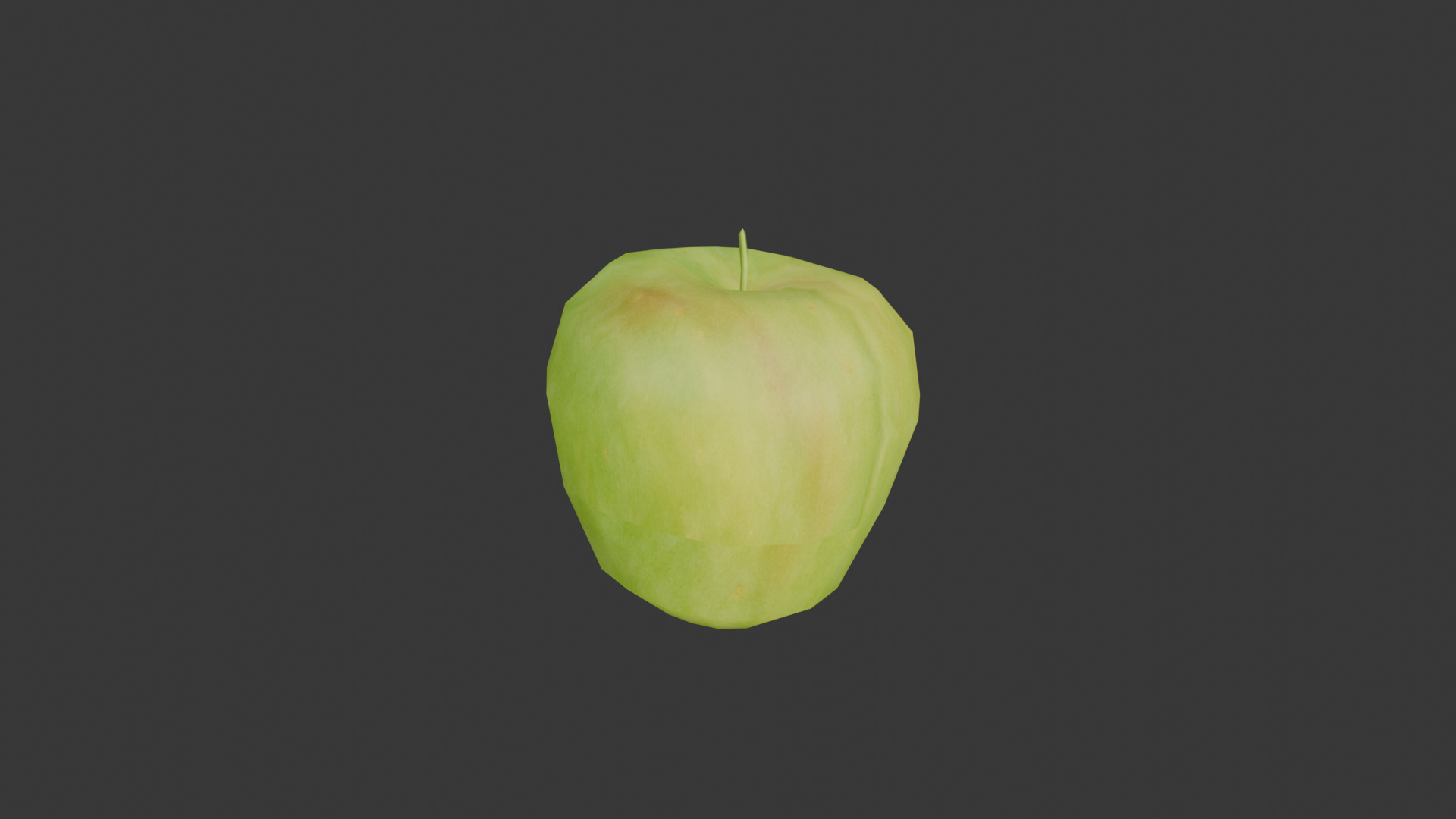}
        \caption{Medium Apple}
    \end{subfigure}
    \hfill
    \begin{subfigure}[b]{0.24\linewidth}
        \includegraphics[width=\linewidth]{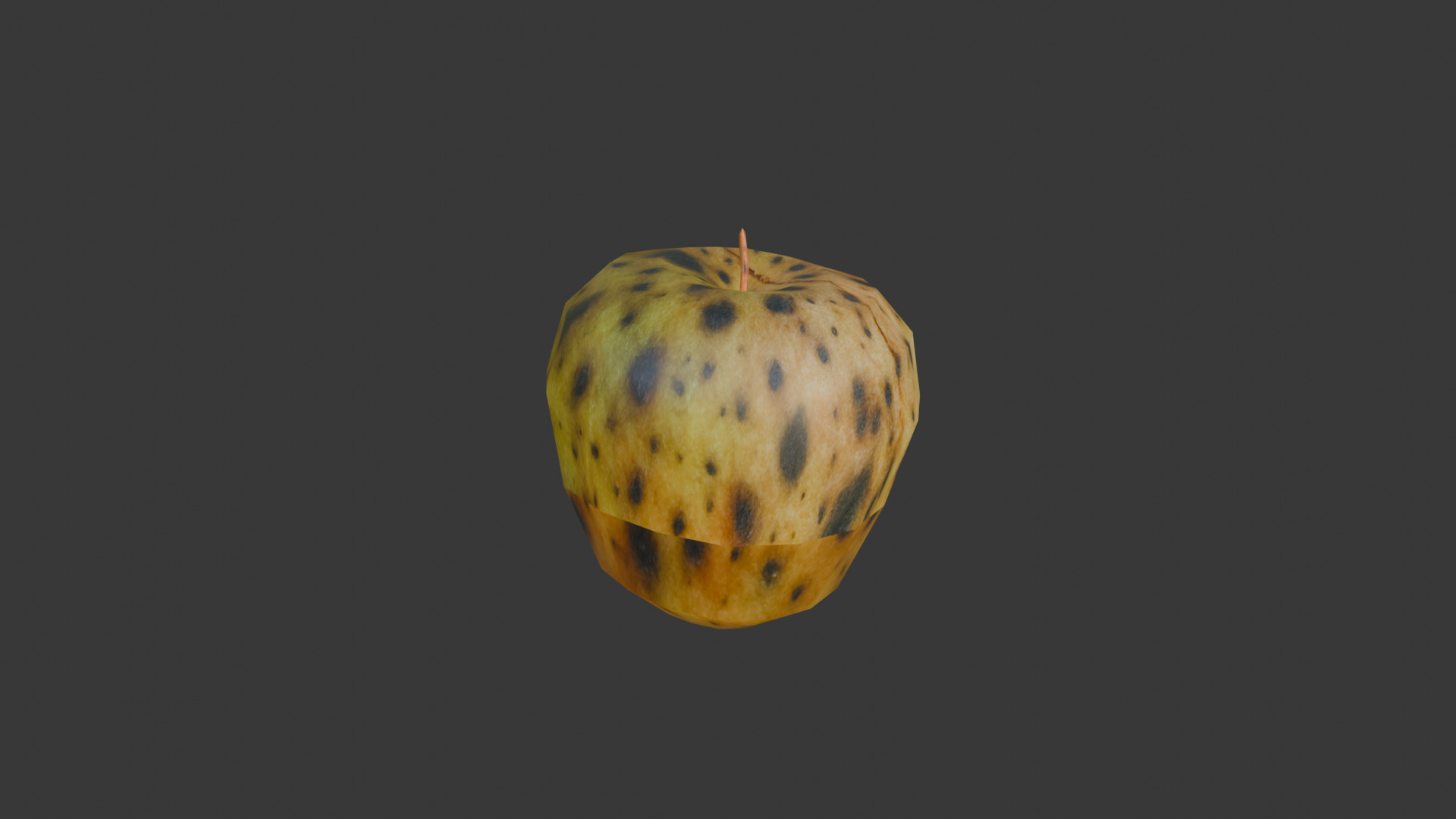}
        \caption{Hard Apple}
    \end{subfigure}
    \caption{Blender renders for the apple used in the place apple in the bowl task.}
    \label{fig:appendix_visual_gen_blender_renders_apple}
\end{figure}
Figure~\ref{fig:appendix_visual_gen_blender_renders_apple} shows the apple textures across difficulty levels. The default apple is a simple red apple. The easy texture is a lighter red with faint orange areas, maintaining strong visual similarity. The medium texture is green, representing a common apple variety, but differing in color. The hard texture shows a brownish, rotten apple that is both visually distinct and semantically less typical, representing a rare case that may be underrepresented in real-world data or training distributions.

\begin{figure}[H]
    \centering
    \begin{subfigure}[b]{0.24\linewidth}
        \centering
        \includegraphics[width=\linewidth]{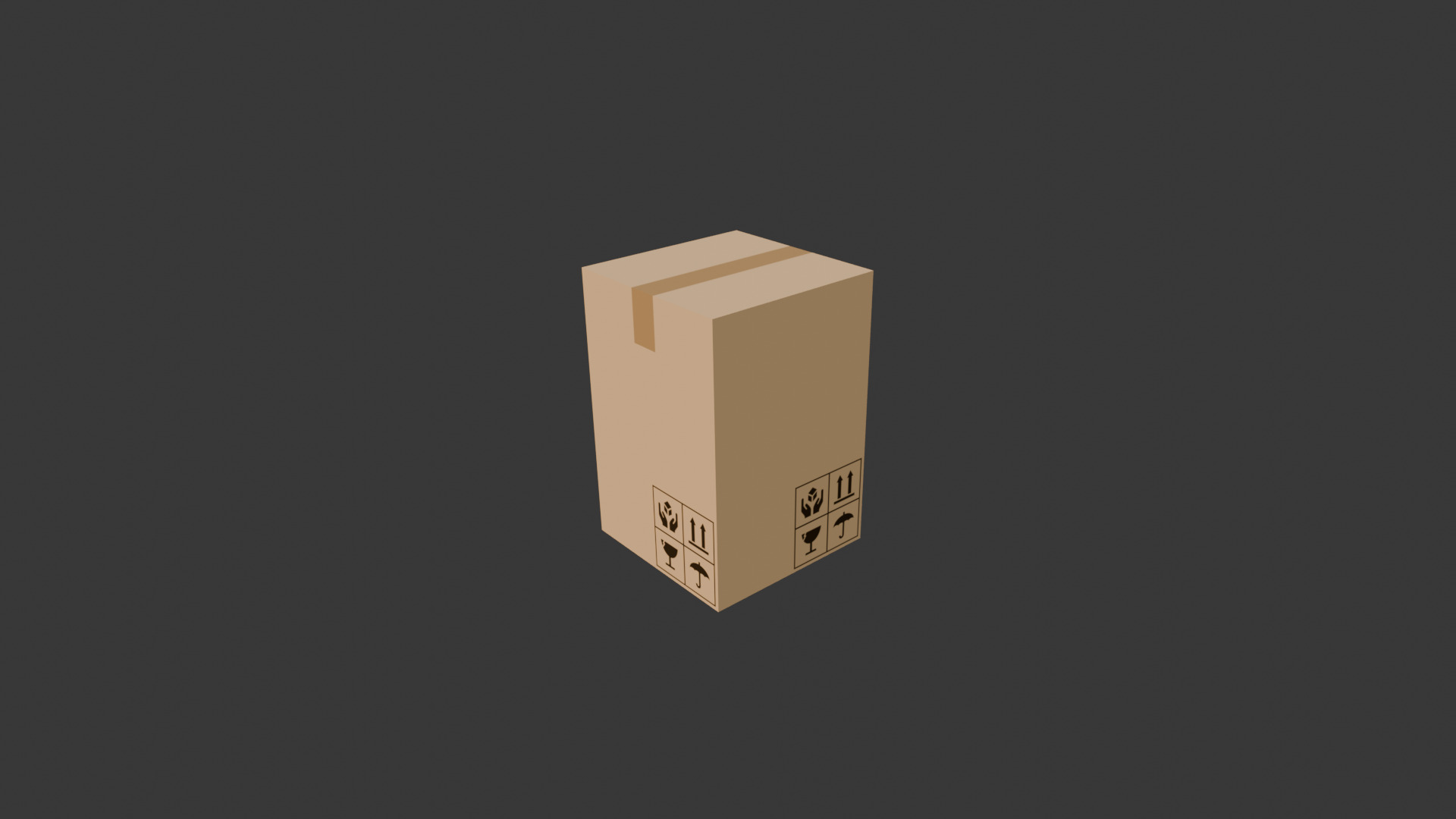}
        \caption{Default Box}
    \end{subfigure}
    \hfill
    \begin{subfigure}[b]{0.24\linewidth}
        \includegraphics[width=\linewidth]{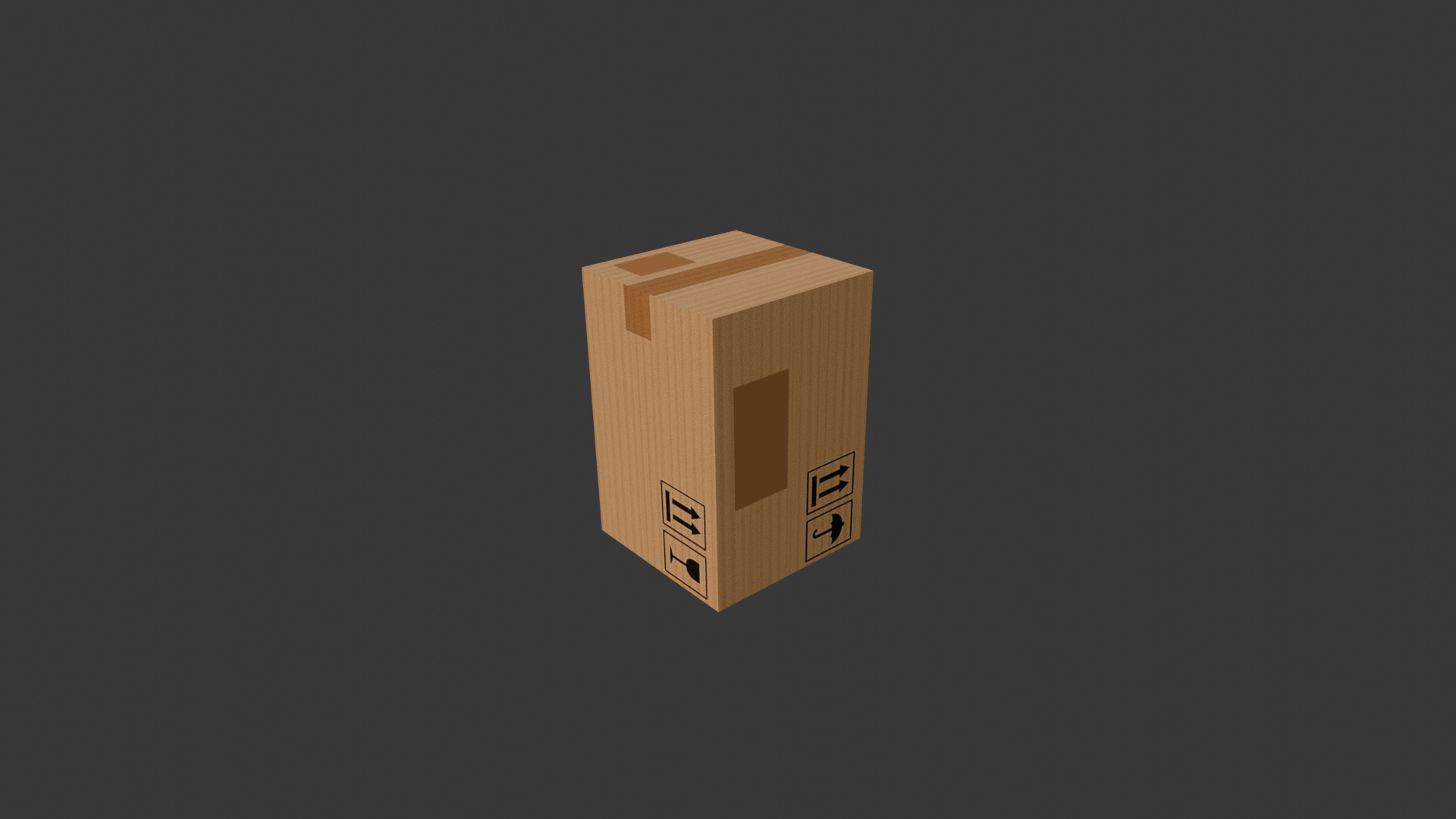}
        \caption{Easy Box}
    \end{subfigure}
    \hfill
    \begin{subfigure}[b]{0.24\linewidth}
        \includegraphics[width=\linewidth]{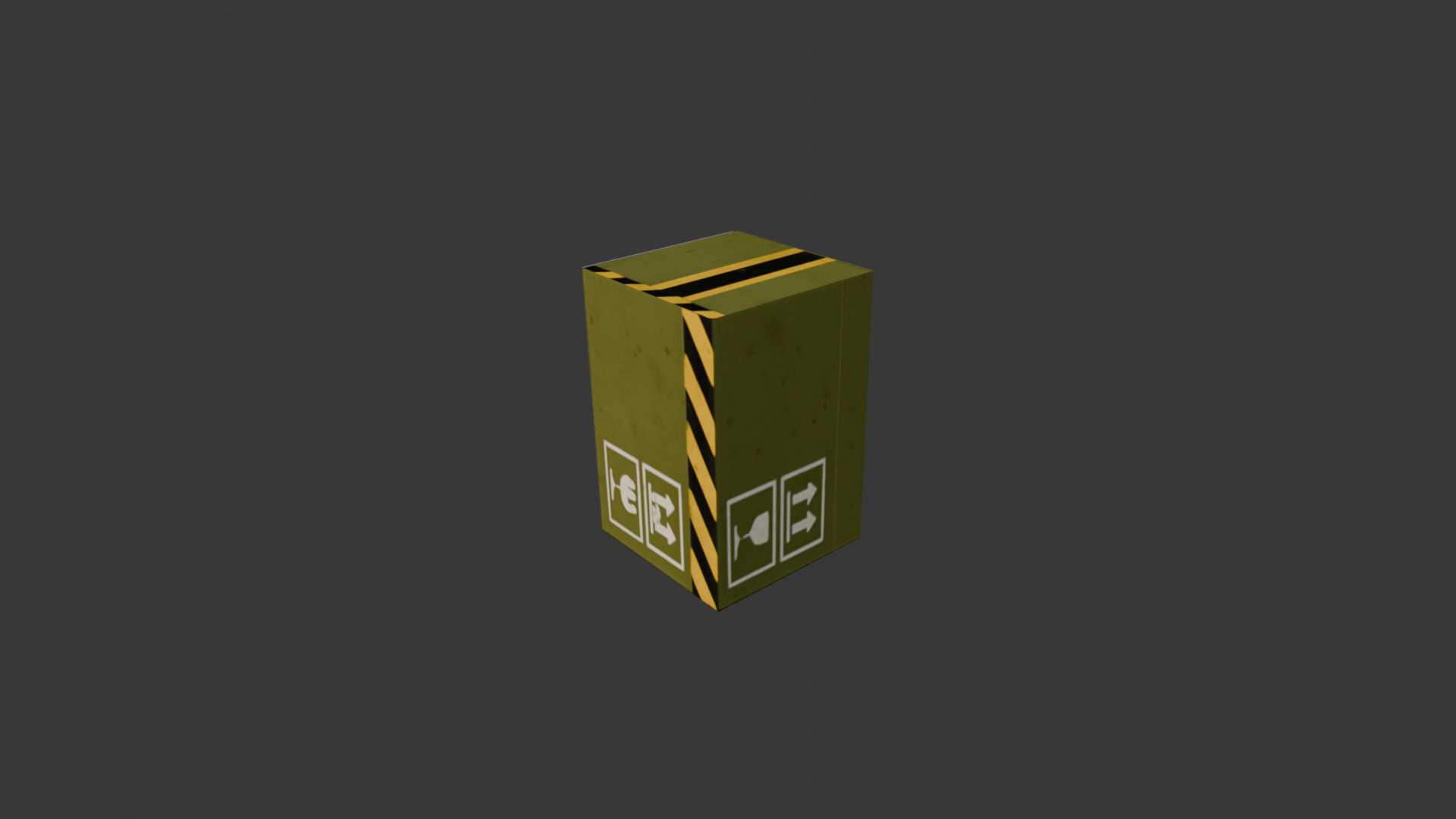}
        \caption{Medium Box}
    \end{subfigure}
    \hfill
    \begin{subfigure}[b]{0.24\linewidth}
        \includegraphics[width=\linewidth]{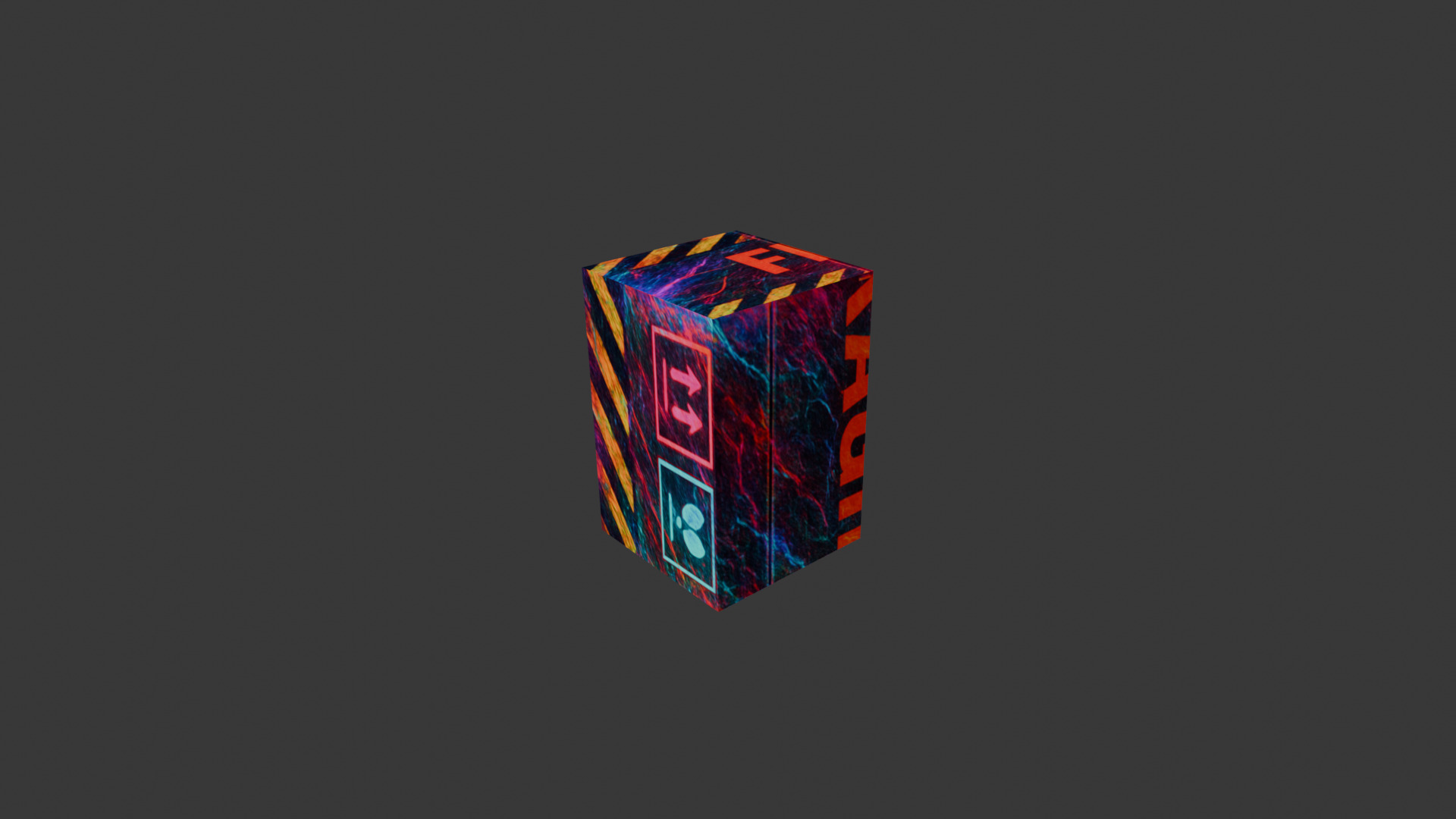}
        \caption{Hard Box}
    \end{subfigure}
    \caption{Blender renders for the box used in the transport box task.}
    \label{fig:appendix_visual_gen_blender_renders_box}
\end{figure}

\begin{figure}[H]
    \centering
    \includegraphics[width=1.00\linewidth]{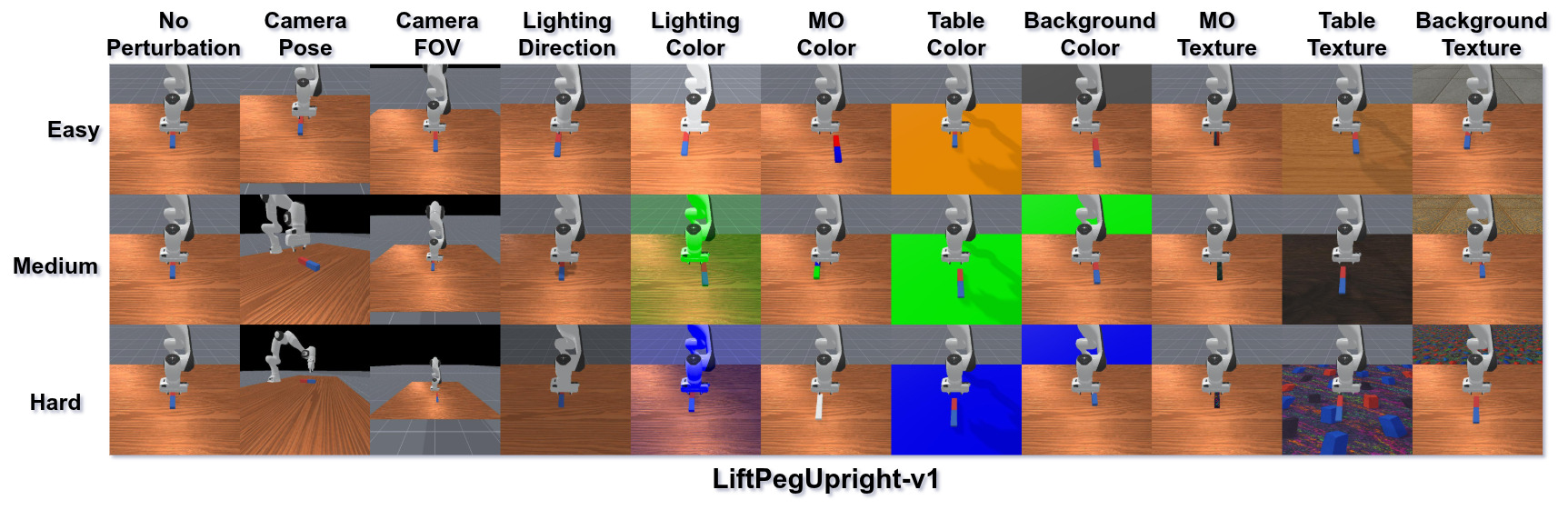}
    \includegraphics[width=1.00\linewidth]{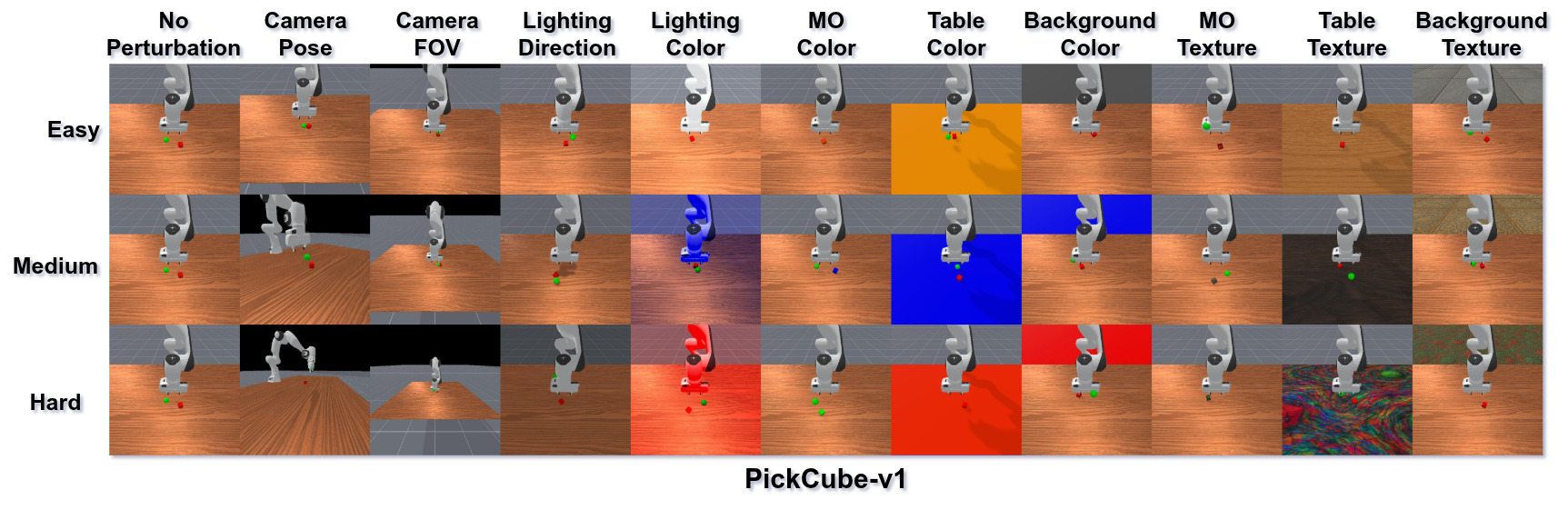}
    \includegraphics[width=1.00\linewidth]{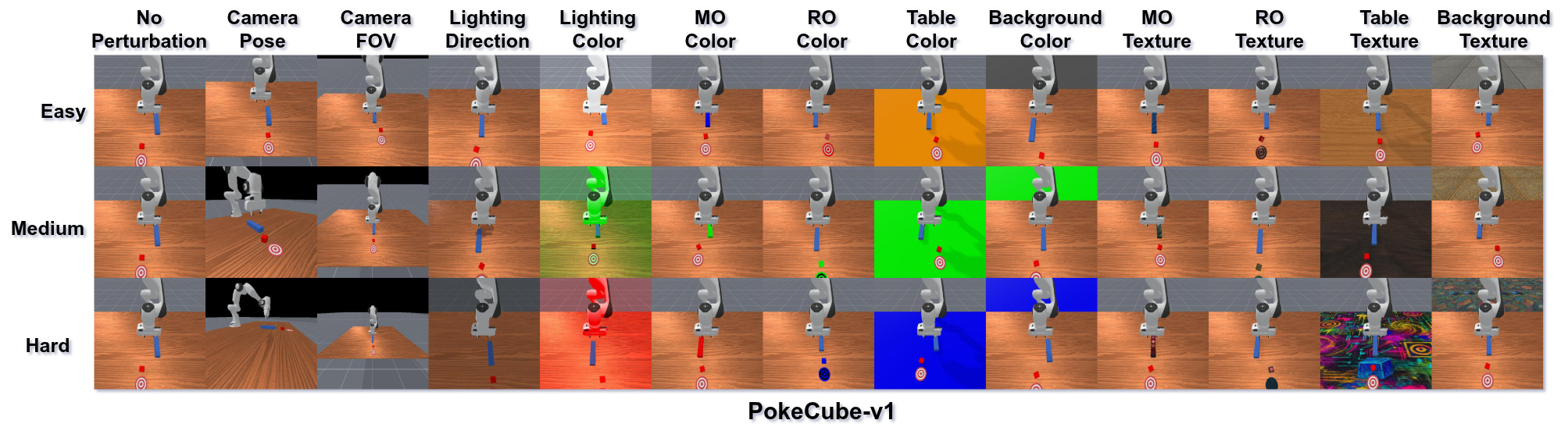}
    \includegraphics[width=1.00\linewidth]{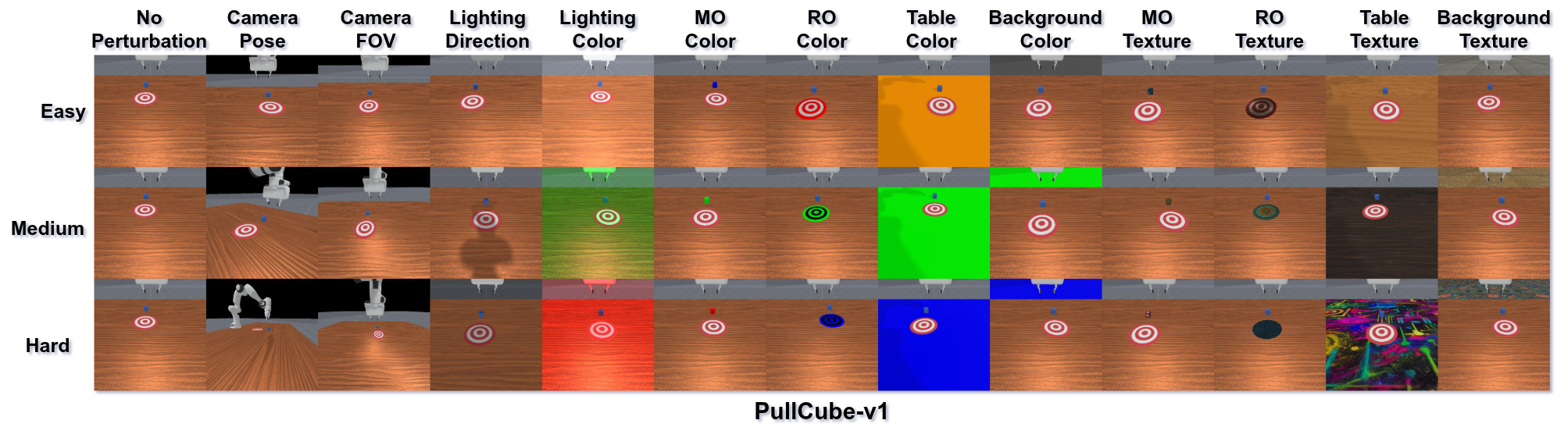}
    \caption{(Part 1) Visual perturbations for all tasks and difficulties.}
    \label{fig:appendix_visual_tasks_perturb}
\end{figure}
\begin{figure}[H]
    \centering
    \includegraphics[width=1.00\linewidth]{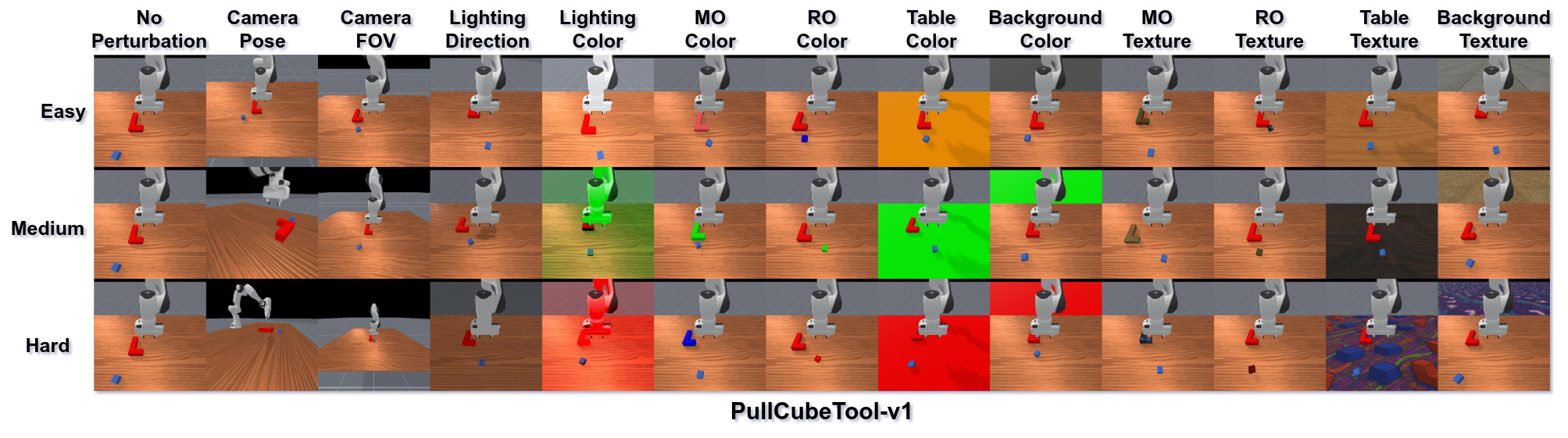}
    \includegraphics[width=1.00\linewidth]{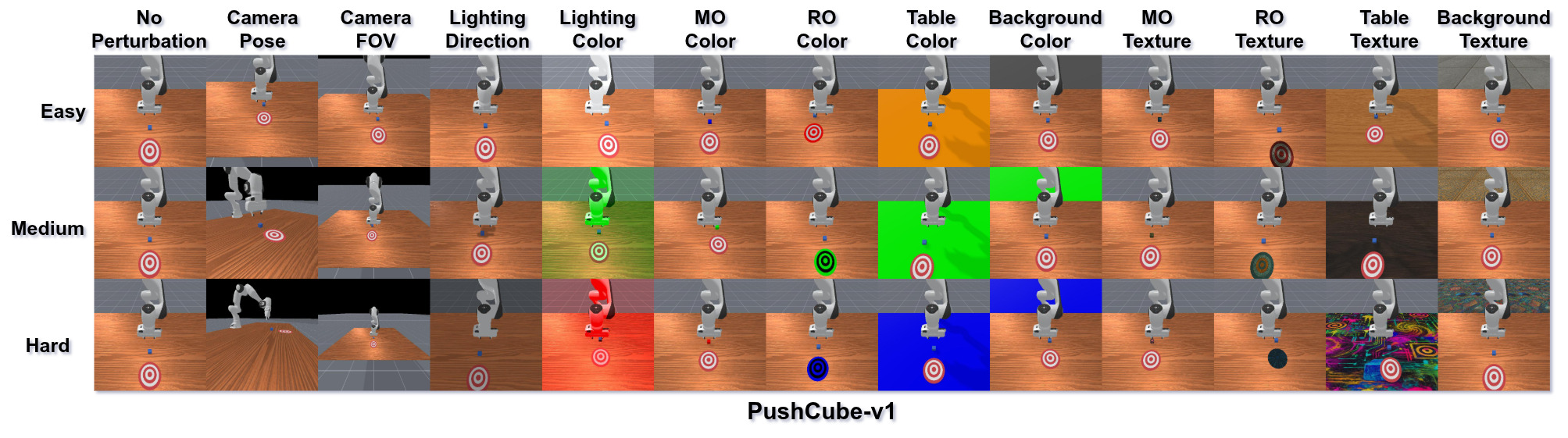}
    \includegraphics[width=1.00\linewidth]{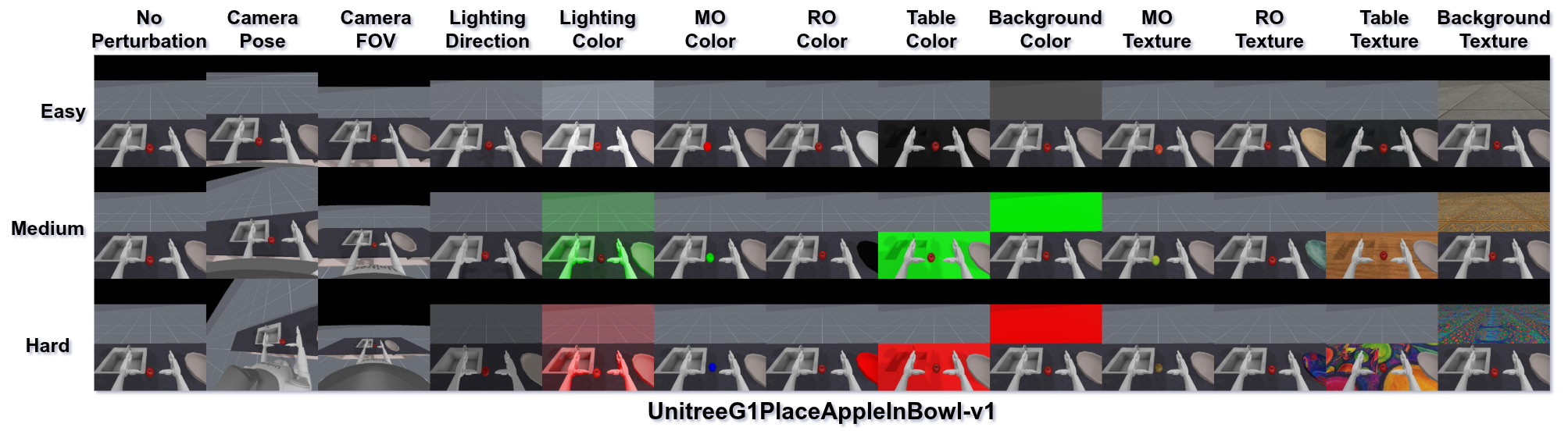}
    \includegraphics[width=1.00\linewidth]{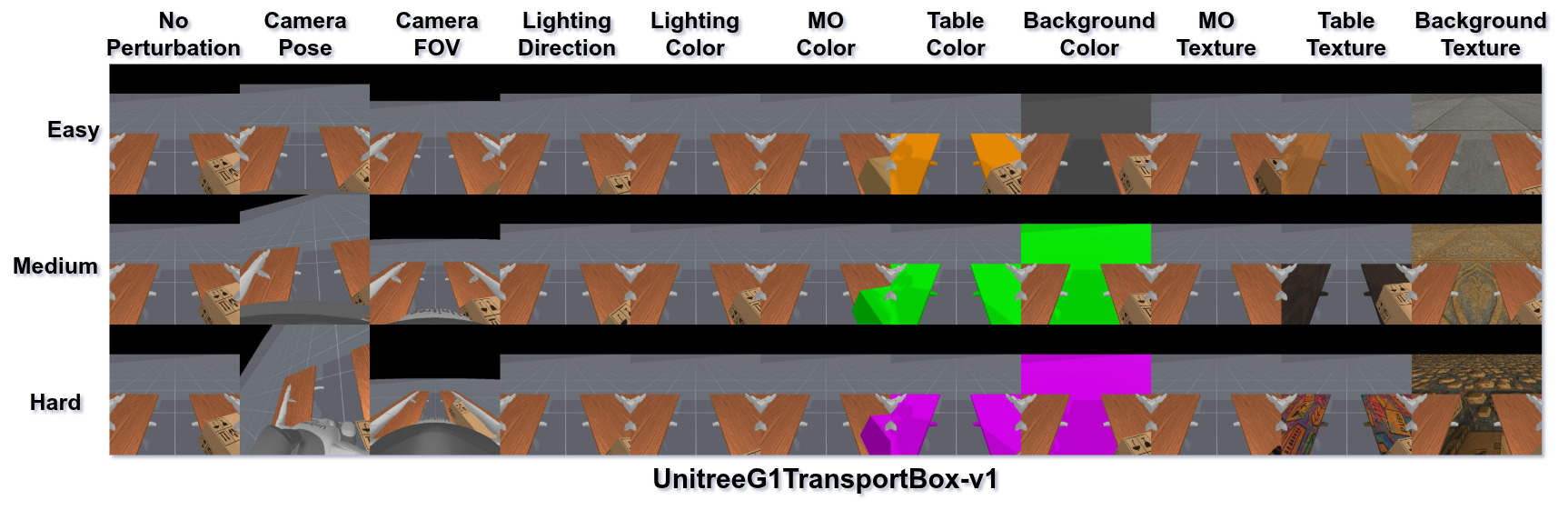}
    \caption{(Part 2) Visual perturbations for all tasks and difficulties.}
    \label{fig:appendix_visual_tasks_perturb_2}
\end{figure}
Figures~\ref{fig:appendix_visual_tasks_perturb} and~\ref{fig:appendix_visual_tasks_perturb_2} show all visual perturbations for each task alongside their default (unperturbed) views. As difficulty increases, observations become more out-of-distribution and challenging. Notably, our camera pose perturbations are more aggressive than those typically used in other benchmarks, making them especially difficult for policies to handle. While no benchmark is perfect, we believe that exposing policies to strong perturbations provides a clearer understanding of their true robustness and generalization capabilities.

\subsection{Case Study \& Failure Cases}
\label{sec:case_study}
Below are qualitative rollout examples comparing \methodName to MaDi across various tasks and perturbations. We also include cases with camera pose perturbations, as they were the most challenging for all methods.  

\begin{figure}[H]
  \centering
  \rolloutlabel{MaDi}
  \includegraphics[height=2.1cm]{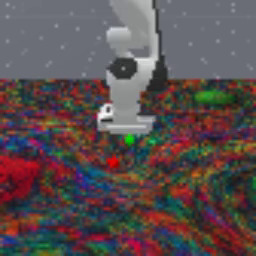}
  \includegraphics[height=2.1cm]{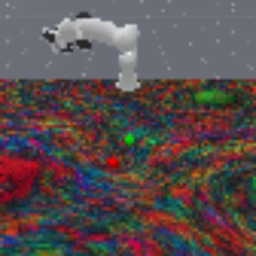}
  \vsepimg{2.1cm}
  \includegraphics[height=2.1cm]{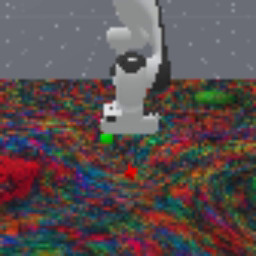}
  \includegraphics[height=2.1cm]{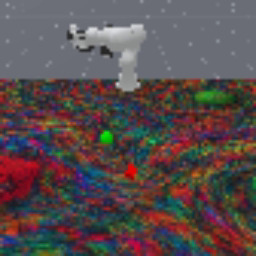}
  
  \rolloutlabel{\methodName}
  \includegraphics[height=2.1cm]{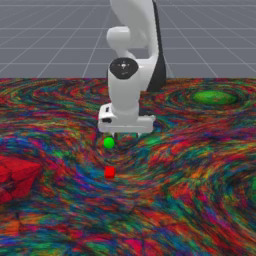}
  \includegraphics[height=2.1cm]{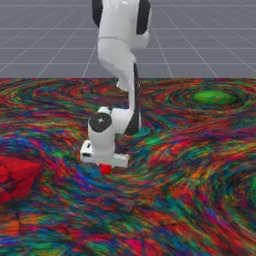}
  \includegraphics[height=2.1cm]{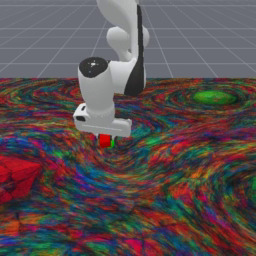}
  \vsepimg{2.1cm}
  \includegraphics[height=2.1cm]{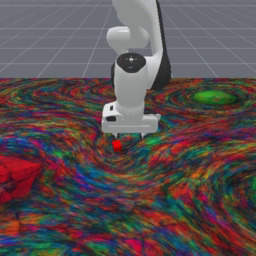}
  \includegraphics[height=2.1cm]{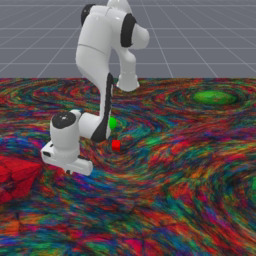}
  \includegraphics[height=2.1cm]{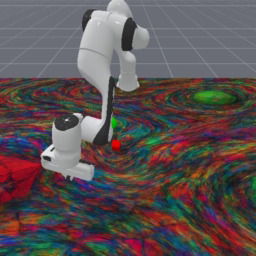}

  \caption{2 Rollouts for \textit{PickCube} - Hard Table Texture (top: MaDi, bottom: \methodName).}
  \label{fig:pick_cube_hard_texture}
\end{figure}

\begin{figure}[H]
  \centering
  \begin{tabular}{@{}c@{}}
    \textbf{MaDi}\\
    \includegraphics[height=2.2cm]{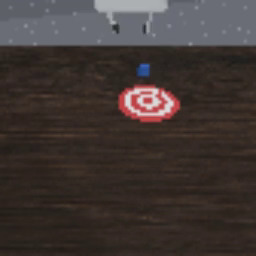}
    \includegraphics[height=2.2cm]{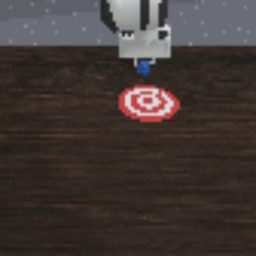}
    \includegraphics[height=2.2cm]{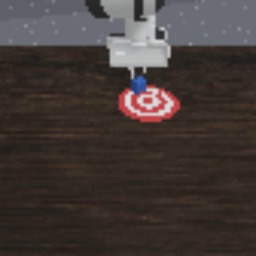}\\[3pt]
    \textbf{\methodName}\\
    \includegraphics[height=2.2cm]{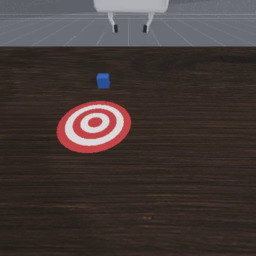}
    \includegraphics[height=2.2cm]{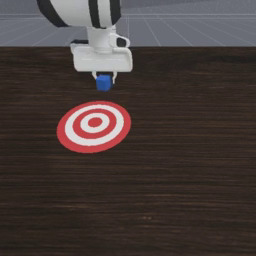}
    \includegraphics[height=2.2cm]{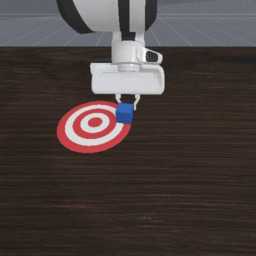}\\
  \end{tabular}
  \caption{Rollout for \textit{PullCube} - Medium Table Texture (top: MaDi, bottom: \methodName).}
  \label{fig:pull_cube_medium_texture}
\end{figure}
Figure~\ref{fig:pull_cube_medium_texture} shows that both MaDi and \methodName are able to successfully complete the Pull Cube task under the medium table texture perturbation (success is reaching the first white ring), regardless of the target's initial position. This highlights that, in some cases, policies can still succeed under medium-level perturbations and do not always collapse when faced with moderate visual changes.

\begin{figure}[H]
  \centering
  \begin{tabular}{@{}c@{}}
    \textbf{MaDi}\\
    \includegraphics[height=2.2cm]{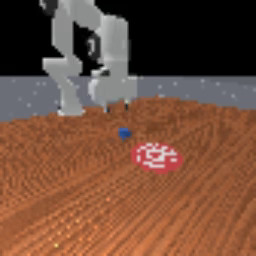}
    \includegraphics[height=2.2cm]{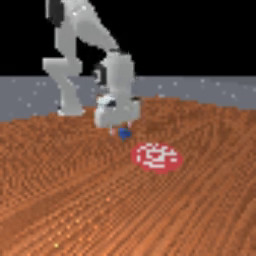}
    \includegraphics[height=2.2cm]{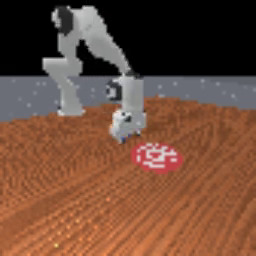}
    \includegraphics[height=2.2cm]{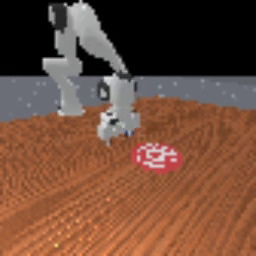}
    \includegraphics[height=2.2cm]{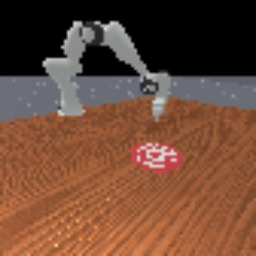}
    \includegraphics[height=2.2cm]{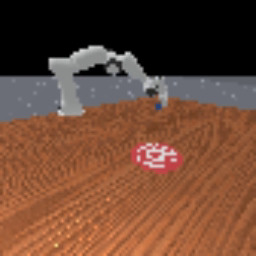}
    \includegraphics[height=2.2cm]{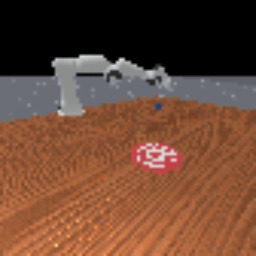}\\[3pt]
    \textbf{\methodName}\\
    \includegraphics[height=2.2cm]{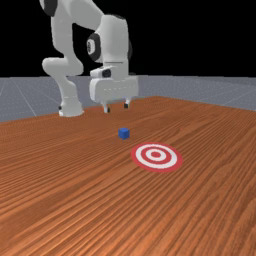}
    \includegraphics[height=2.2cm]{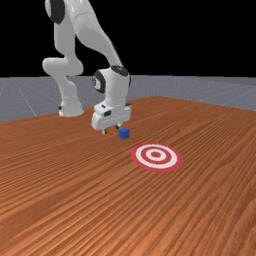}
    \includegraphics[height=2.2cm]{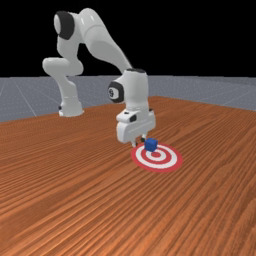}
    \includegraphics[height=2.2cm]{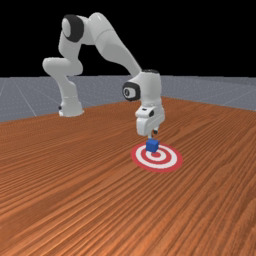}\\
  \end{tabular}
  \caption{Rollout for \textit{PushCube} - Medium Camera Pose (top: MaDi, bottom: \methodName).}
  \label{fig:push_cube_medium_camera_pose}
\end{figure}
Figure~\ref{fig:push_cube_medium_camera_pose} shows a rollout of the Push Cube task under the medium camera pose perturbation for both MaDi and \methodName. MaDi initially behaves reasonably by reaching toward the cube, but fails to position the arm correctly. This leads to accumulating errors and eventually to unstable actions that push the cube away from the target. In contrast, \methodName handles the perturbation more effectively and often succeeds, showing more stable and goal-directed behavior.

\begin{figure}[H]
  \centering
  \begin{tabular}{@{}c@{}}
    \textbf{MaDi}\\
    \includegraphics[height=2.2cm]{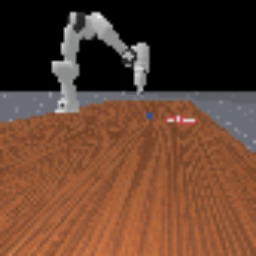}
    \includegraphics[height=2.2cm]{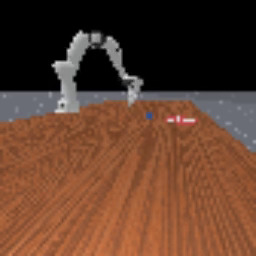}
    \includegraphics[height=2.2cm]{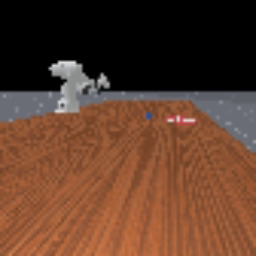}\\[3pt]
    \textbf{\methodName}\\
    \includegraphics[height=2.2cm]{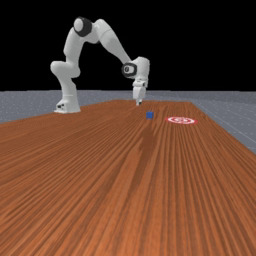}
    \includegraphics[height=2.2cm]{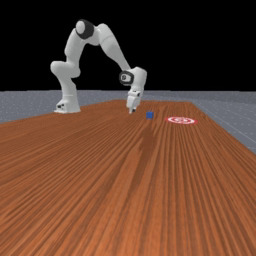}
    \includegraphics[height=2.2cm]{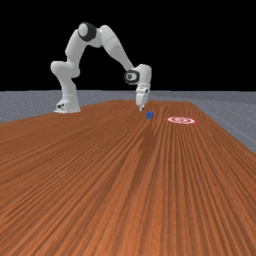}\\
  \end{tabular}
  \caption{Rollout for \textit{PushCube} - Hard Camera Pose (top: MaDi, bottom: \methodName).}
  \label{fig:push_cube_hard_camera_pose}
\end{figure}
Figure~\ref{fig:push_cube_hard_camera_pose} illustrates the difficulty of the hard camera pose perturbation for both policies. MaDi consistently collapses and produces erratic, non-sensical actions. While \methodName does not succeed either, it shows more structured behavior, making small back-and-forth motions in an attempt to push the cube. However, it lacks the precision needed and repeatedly misses the cube by a small margin.

\begin{figure}[H]
  \centering
  \begin{tabular}{@{}c@{}}
    \textbf{MaDi}\\
    \includegraphics[height=2.2cm]{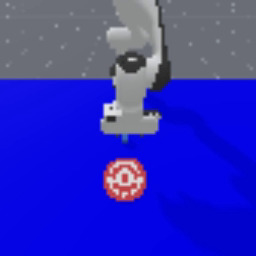}
    \includegraphics[height=2.2cm]{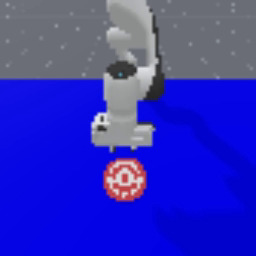}
    \includegraphics[height=2.2cm]{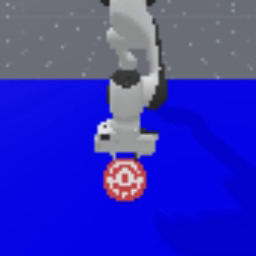}
    \includegraphics[height=2.2cm]{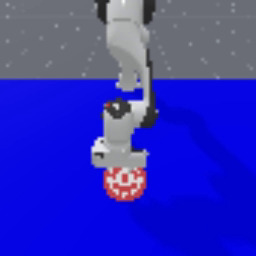}
    \includegraphics[height=2.2cm]{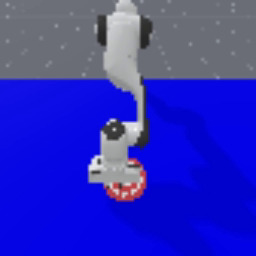}\\[3pt]
    \textbf{\methodName}\\
    \includegraphics[height=2.2cm]{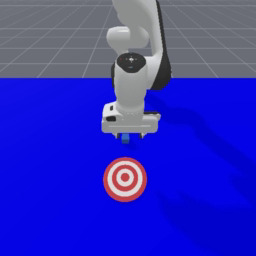}
    \includegraphics[height=2.2cm]{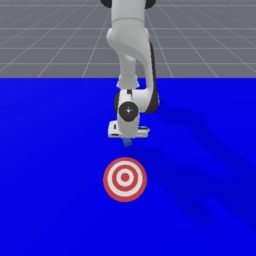}
    \includegraphics[height=2.2cm]{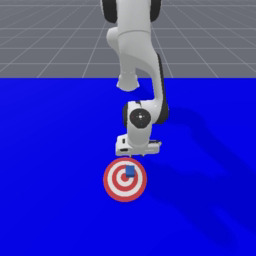}\\
  \end{tabular}
  \caption{Rollout for \textit{PushCube} - Hard Table Color (top: MaDi, bottom: \methodName).}
  \label{fig:push_cube_hard_color}
\end{figure}
Figure~\ref{fig:push_cube_hard_color} shows a rollout of the Push Cube task under the hard table color perturbation. In this case, the table is blue, the same color as the cube, which appears to confuse MaDi. The policy behaves as if the cube is already between the gripper and moves toward the target, while the actual cube remains at its initial position. This issue does not occur with \methodName, which accurately identifies the cube and pushes it to the center of the target. This highlights the strength of object-centric approaches. While MaDi reasons at the pixel level, \methodName decomposes the scene into segments and leverages bounding box coordinates of the cube segment to guide the policy more effectively.

\section{Ablations}
\label{appendix_ablations}

\subsection{Text Inputs Ablation}
\label{appendix_text_prompts_ablation}
\subsubsection{Does Learning To Ignore Background Improve Visual Generalization?}
\begin{figure}[h]
    \centering
    \includegraphics[width=1.00\linewidth]{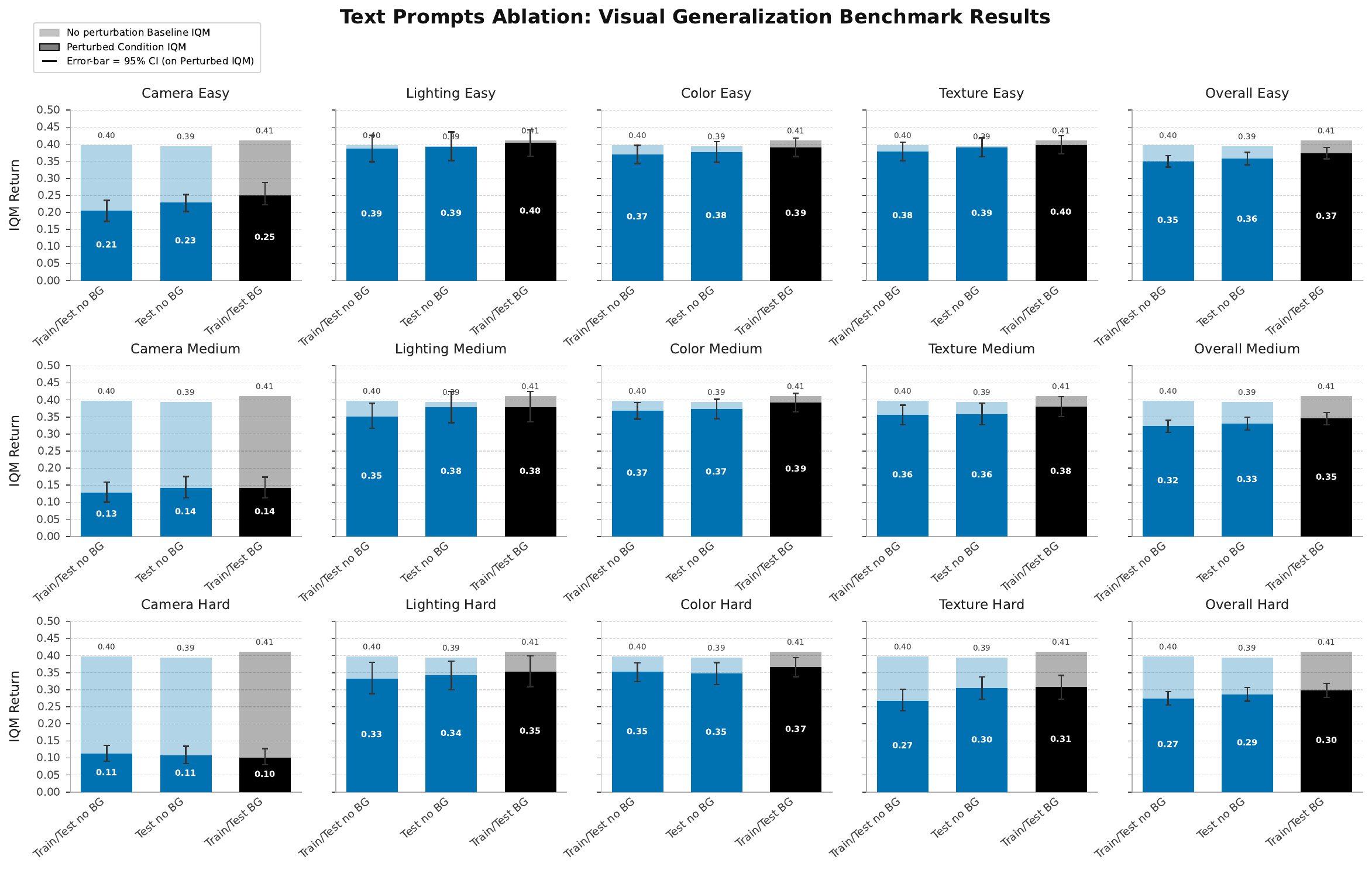}
    \caption{Visual generalization results when including or excluding the "background" text tags.}
    \label{fig:appendix_text_ablations}
\end{figure}
We conducted an ablation to evaluate whether including the "background" text tag improves visual generalization in \methodName. In the "Train/Test no BG" setting (Figure~\ref{fig:appendix_text_ablations}), the model is trained and tested without the "background" tag. For example, in the Push Cube task, the input tags would be ["robot", "gripper", "small box", "target"] instead of ["background", "robot", "gripper", "small box", "target"]. In the "Test no BG" setting, the model is trained with the background tag but tested without it. The final configuration, shown in black in Figure~\ref{fig:appendix_text_ablations}, uses the background tag during both training and testing and corresponds to the final version of \methodName.

The goal of this ablation is to determine whether training with background segments encourages the model to ignore irrelevant information more effectively, or if excluding them allows it to focus more directly on task-relevant objects. While YOLO-World and SAM can still introduce irrelevant segments even without the background tag, our experiments show that including the tag leads to a noticeably larger number of irrelevant segments.

Results in Figure~\ref{fig:appendix_text_ablations} show that all configurations perform similarly, but "Train/Test BG" achieves about $\sim9\%$ higher IQM returns. This suggests that including the background tag improves generalization by encouraging the model to focus less on distractions.
\newpage

\subsubsection{Text Inputs Sensitivity}
\label{sec:synonyms}
\begin{figure}[h]
    \centering
    \includegraphics[width=0.32\linewidth]{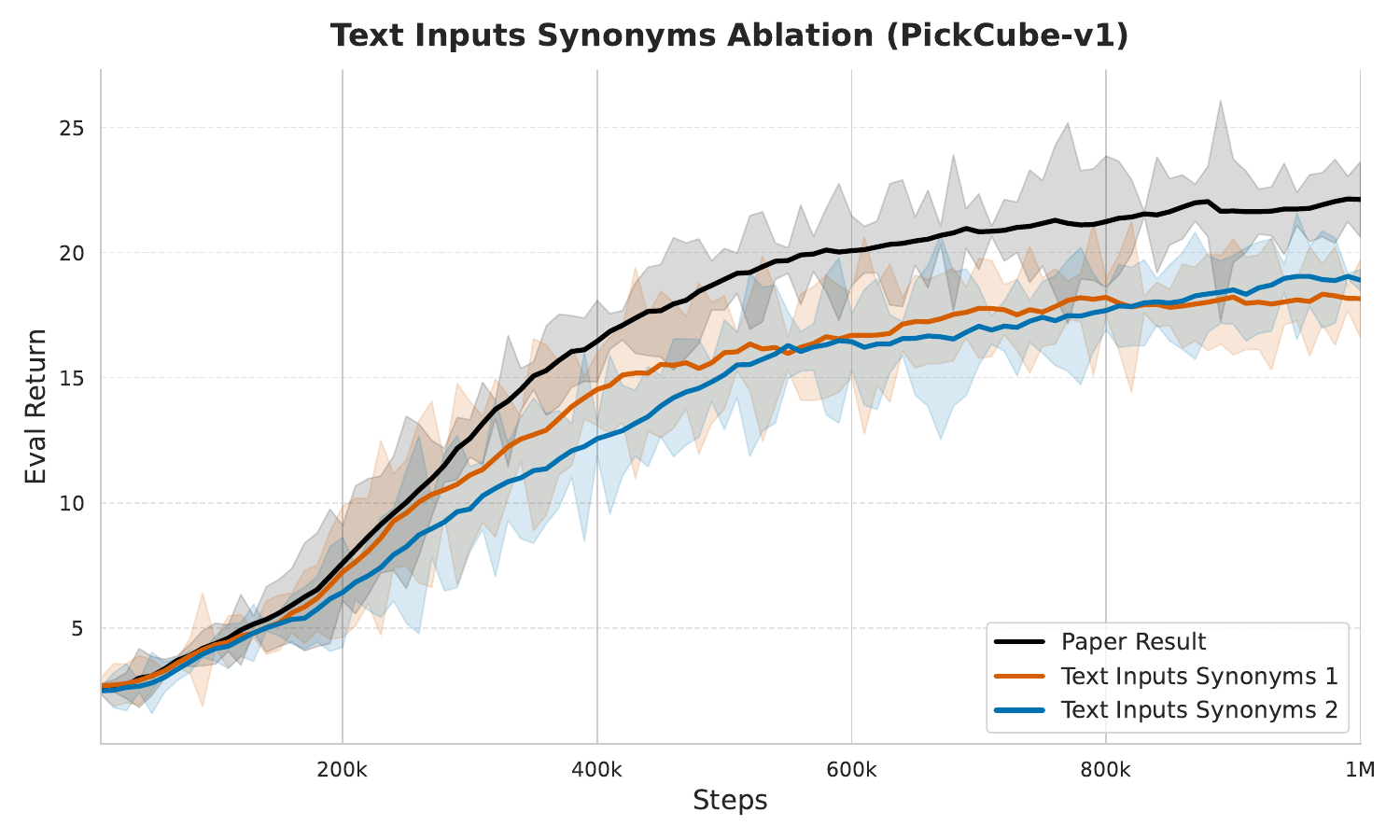}
    \includegraphics[width=0.32\linewidth]{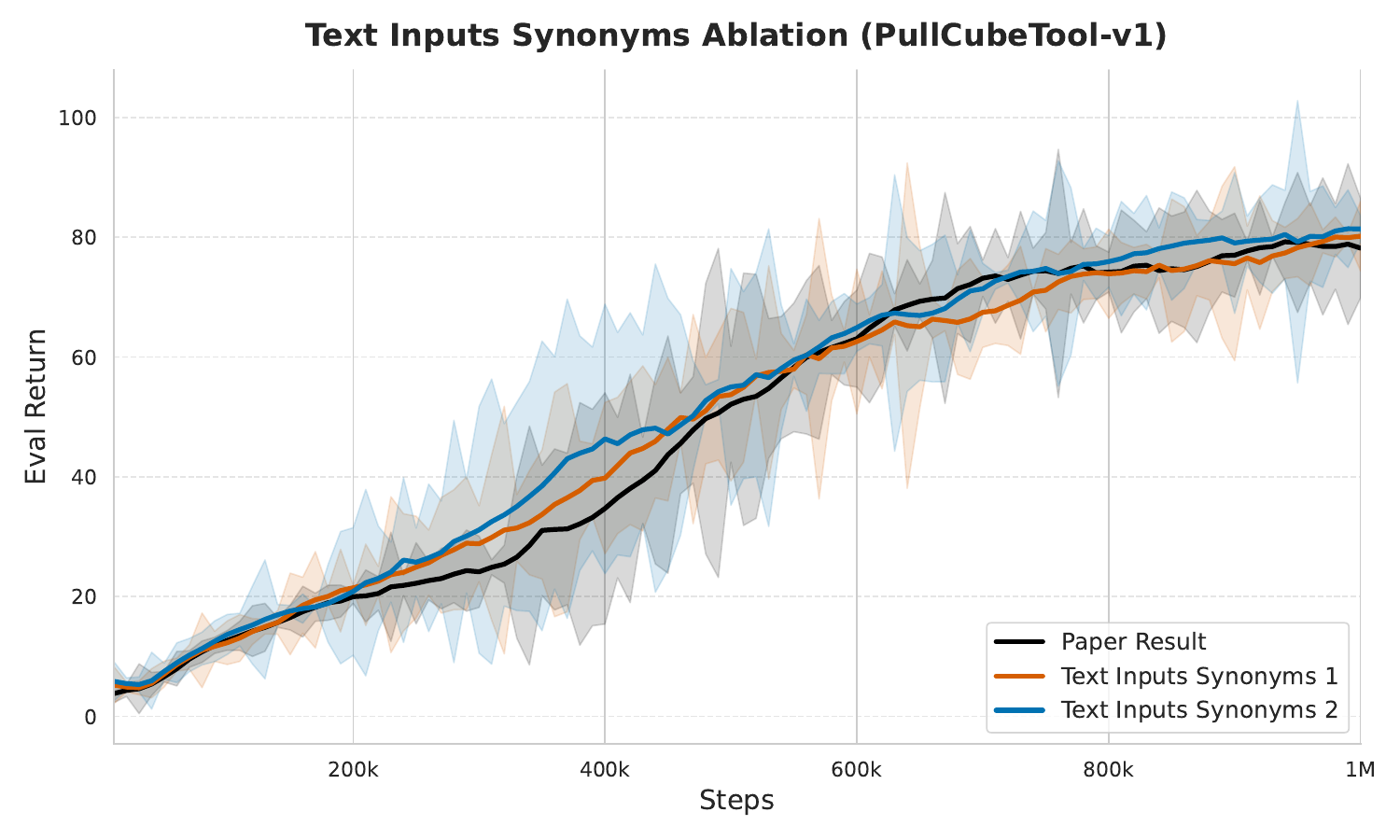}
    \includegraphics[width=0.32\linewidth]{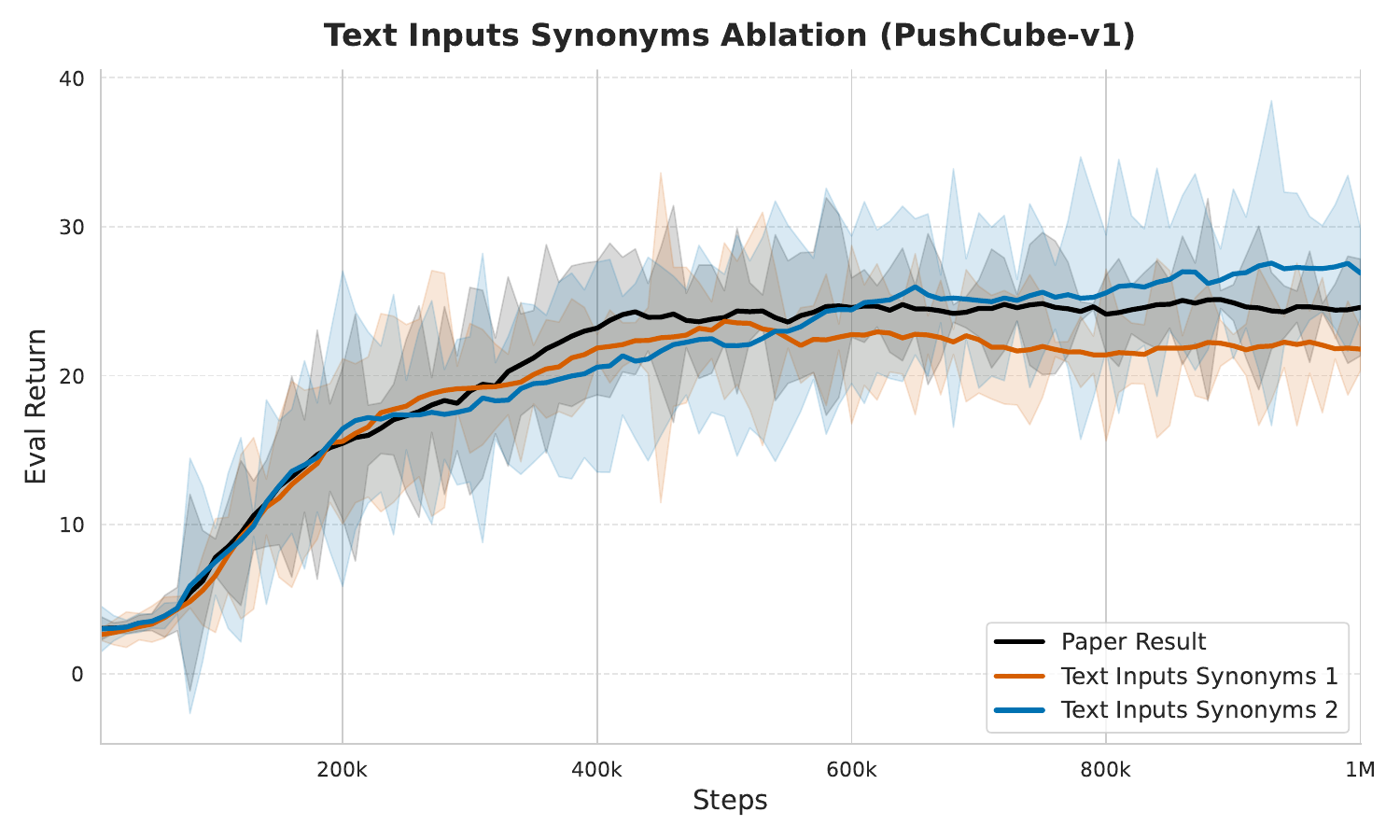}
    \caption{Sample efficiency when using different text inputs synonyms (using 3 seeds).}
\label{fig:synonyms_ablation}
\end{figure} 

\begin{table}[h]
\centering
\vspace{-0.75em}
\scriptsize
\caption{Text inputs used for the synonym ablation for each task.}
\label{text_inputs_synonym_table}

\resizebox{\linewidth}{!}{%
\begin{tabular}{l l l l}
\toprule
\textbf{Text Input Set} &
\textbf{Pick Cube} &
\textbf{Pull Cube Tool} &
\textbf{Push Cube} \\
\midrule

\textbf{Default} &
background, robot, gripper, small box, sphere &
background, robot, gripper, small box, block &
background, robot, gripper, small box, target \\

\textbf{Synonyms 1} &
background, robot, manipulator, box, target &
background, robot, manipulator, box, tool, handle &
background, robot, manipulator, box, goal \\

\textbf{Synonyms 2} &
background, bot, end-effector, cube, circle &
background, robot, end effector, cube, peg, stick &
background, bot, arm, cube, bullseye \\

\bottomrule
\end{tabular}
}%
\end{table}

Figure \ref{fig:synonyms_ablation} shows the sample efficiency curves when training with different sets of synonyms for the text inputs used by the YOLO-World detector, the exact text inputs are shown in table \ref{text_inputs_synonym_table}. Across all tasks we observe that the overall learning trend remains stable. Some tasks show small increases or decreases in absolute performance, but the shape of the curve is consistent and no collapse occurs. This indicates that \methodName is not sensitive to the exact choice of words used to describe the objects. The method does not require heavy prompt engineering or tuning of phrasing, which is often a concern in work relying on large language models. YOLO-World operates on short, simple concept words and ignores terms that do not match the frame. Together these results suggest that the text-guided segmentation component of \methodName is stable and easy to use in practice. 

\subsubsection{Shared Text Inputs}
\label{sec:shared_text_inputs}
\begin{figure}[h]
    \centering
    \includegraphics[width=0.32\linewidth]{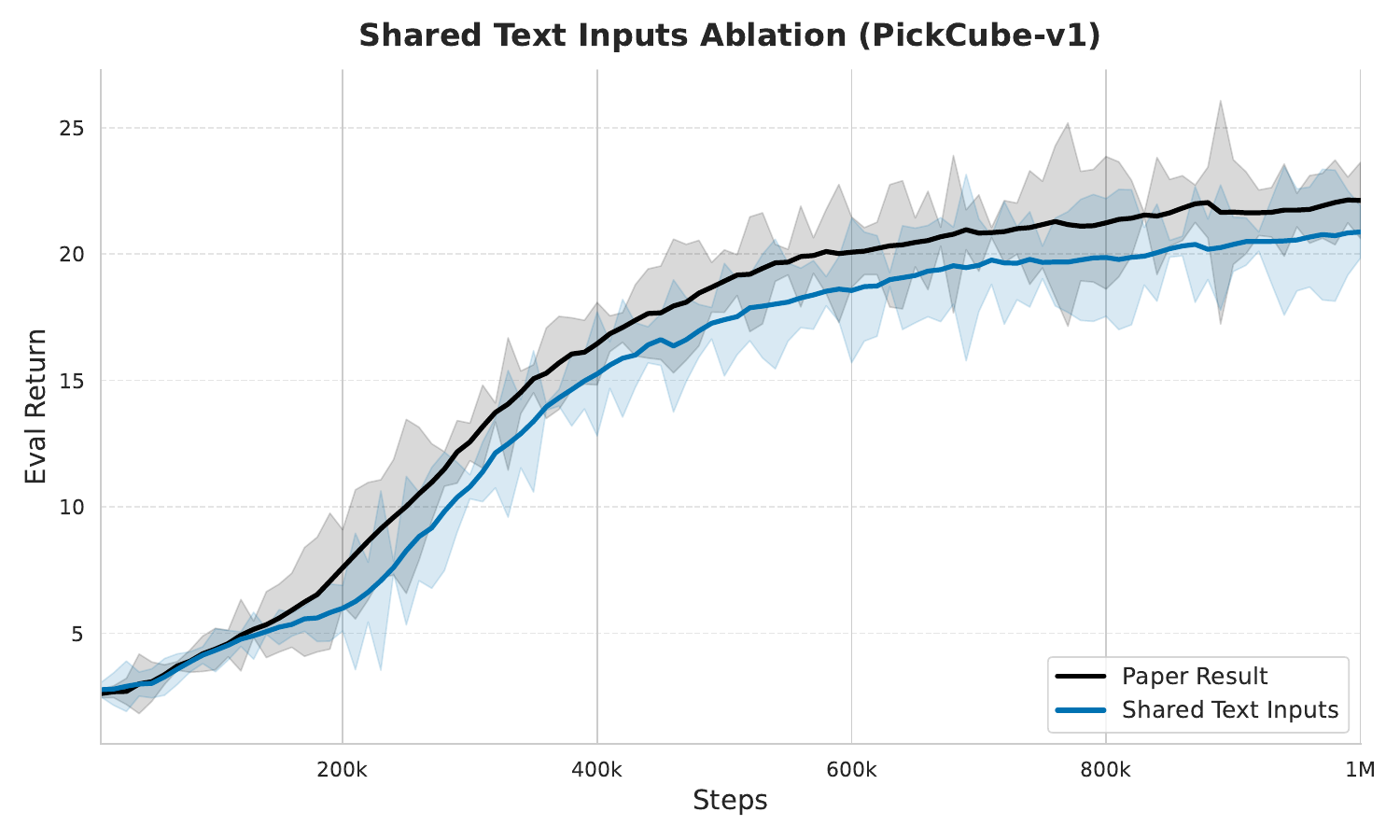}
    \includegraphics[width=0.32\linewidth]{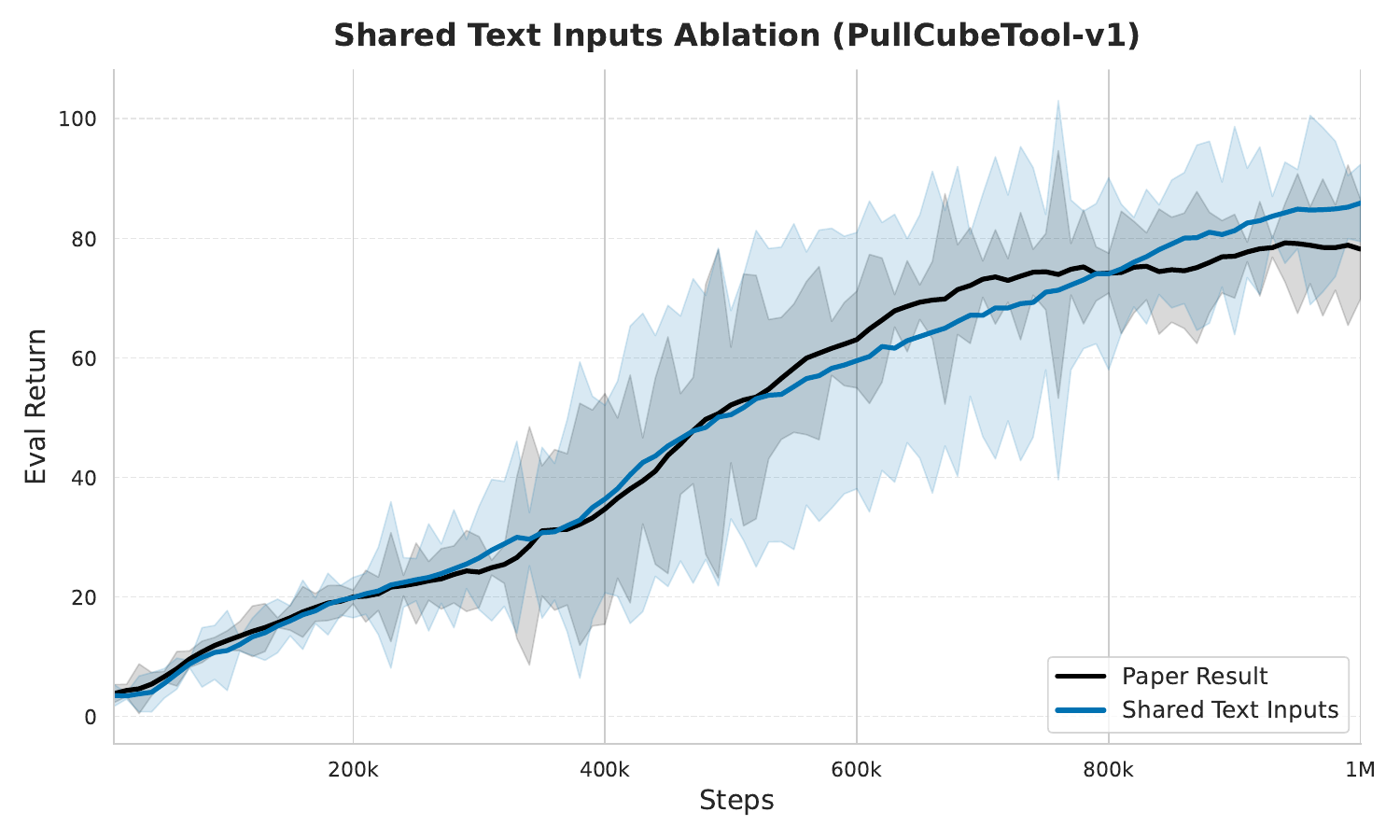}
    \includegraphics[width=0.32\linewidth]{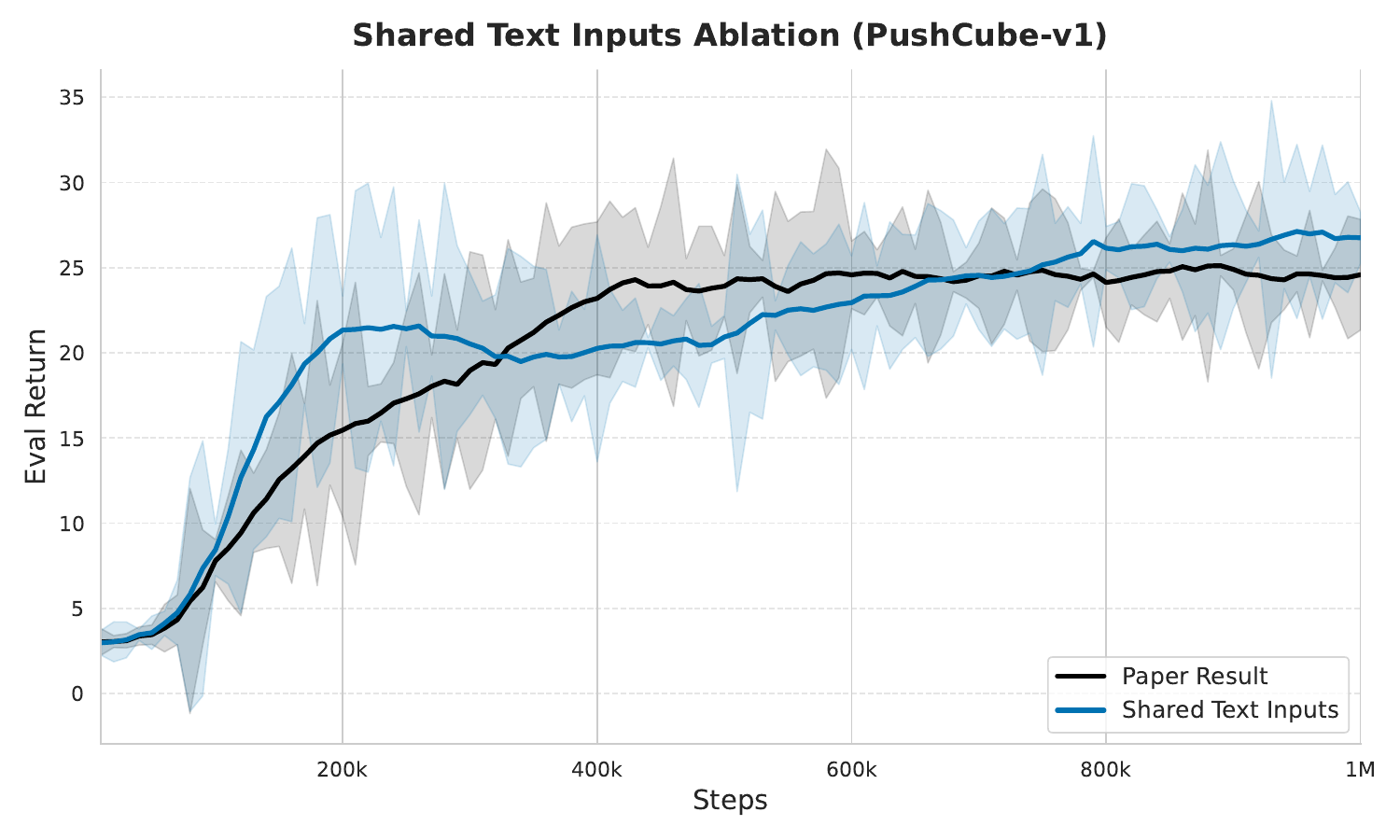}
    \caption{Sample efficiency when using a shared list of text inputs for all tasks (using 3 seeds).}
\label{fig:shared_text_inputs_ablation}
\end{figure} 

\begin{table}[h]
\centering
\caption{Comparison of the shared text inputs versus the task-specific default lists.}
\label{shared_text_inputs_table}
\scriptsize
\begin{tabularx}{\linewidth}{l X}
\toprule
\textbf{Configuration} & \textbf{Text Inputs} \\
\midrule
\textbf{Shared Text Inputs (Ablation)} & background, robot, robot hand, gripper, box, target, block, bar, apple, bowl, sphere, wooden peg \\
\midrule
\textbf{Pick Cube (Default)} & background, robot, gripper, small box, sphere \\
\textbf{Pull Cube Tool (Default)} & background, robot, gripper, small box, block \\
\textbf{Push Cube (Default)} & background, robot, gripper, small box, target \\
\bottomrule
\end{tabularx}
\end{table}
We evaluate the robustness of \methodName by using a single, shared vocabulary across all tasks. This contrasts with the default setting where each task uses a specific set of text inputs. Table \ref{shared_text_inputs_table} details the specific text inputs used. Although we evaluated this ablation on three specific tasks, the shared text inputs includes concepts required for all eight tasks presented in this paper. This design simulates a realistic scenario where a single shared vocabulary is used across a diverse set of tasks/environments.

Figure \ref{fig:shared_text_inputs_ablation} presents the sample efficiency results. We observe that performance remains stable across all tasks when using the shared list. The shape of the learning curves is consistent with the baseline and no performance collapse occurs. This suggests that a single general vocabulary can be reused across different tasks.

We attribute this stability to two main factors. First, the text tags in YOLO-World are used to score potential regions. If a text input describes an object that is not visible in the scene, the detector generally does not output a bounding box. Consequently, the number of generated segments is primarily determined by the image content rather than the size of the text vocabulary. Second, while the object detector may produce a slightly different number of bounding boxes when the text inputs change, \methodName is explicitly trained to handle a variable number of segments. This ensures the policy remains robust to fluctuations in the count or granularity of detected objects.

Furthermore, the number of resulting SAM masks remains modest (typically 5 to 25) even with a larger vocabulary. This keeps the computational cost significantly lower than a typical transformer encoder that would need to process thousands of patch embeddings. For these experiments, we maintained the same low object detection threshold used in the main results and did not perform specific parameter tuning for the shared text inputs setting.

\subsection{Global vs Object-Centric Tokens}
\label{sec:global_obj_suite}
To assess whether object-centric representations are more effective than global image features, we compared \methodName to an SAC baseline that uses a fixed-length global image representation. This baseline, referred to as SAC SAM Encoder, computes the mean of SAM's patch embeddings over the entire image, without using segmentation. The resulting vector is then processed by an MLP with the same number of layers as the projection head used in \methodName to predict actions or Q-values.

\begin{figure}[ht]
    \centering\scriptsize
    \begin{tabular}{@{}cccc@{}}
      \includegraphics[width=1.0\linewidth]{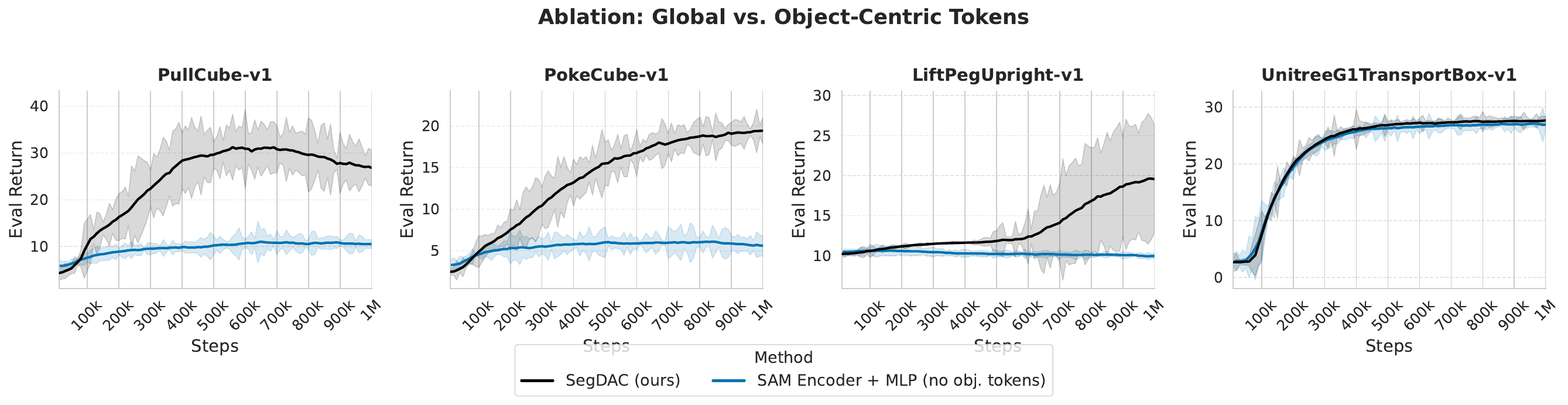} 
    \end{tabular}
    \caption{Global vs.\ object-centric token ablation on remaining 4 tasks.}
    \label{fig:appendix_ablation_sac_sam_encoder}
\end{figure}

Figure~\ref{fig:appendix_ablation_sac_sam_encoder} shows the remaining results. Overall \methodName outperforms SAC SAM Encoder on 7 out of 8 tasks and ties on the remaining one. This highlights a key limitation of directly using pre-trained encoders like SAM for global image representations in online visual RL. Even after attempting hyperparameter tuning for SAC SAM Encoder, we observed no improvement in performance.

\section{Tasks Definitions}
\label{appendix_tasks_def}
\begin{figure}[H]
\centering

\begin{subfigure}{0.23\textwidth}
  \includegraphics[width=\linewidth]{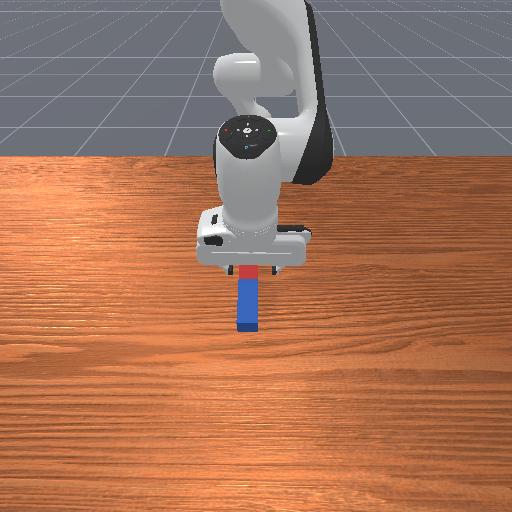}
  \caption{LiftPegUpright-v1}
\end{subfigure}
\hfill
\begin{subfigure}{0.23\textwidth}
  \includegraphics[width=\linewidth]{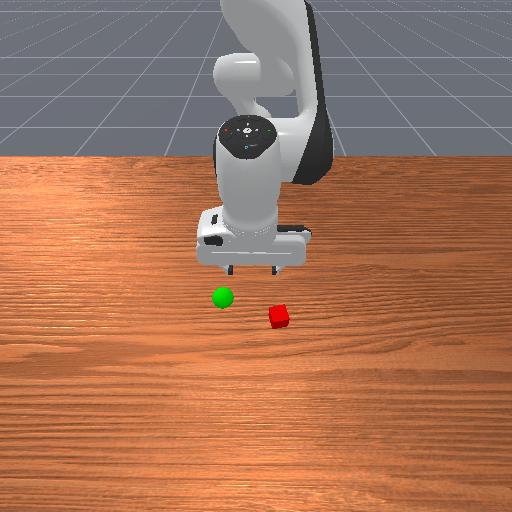}
  \caption{PickCube-v1}
\end{subfigure}
\hfill
\begin{subfigure}{0.23\textwidth}
  \includegraphics[width=\linewidth]{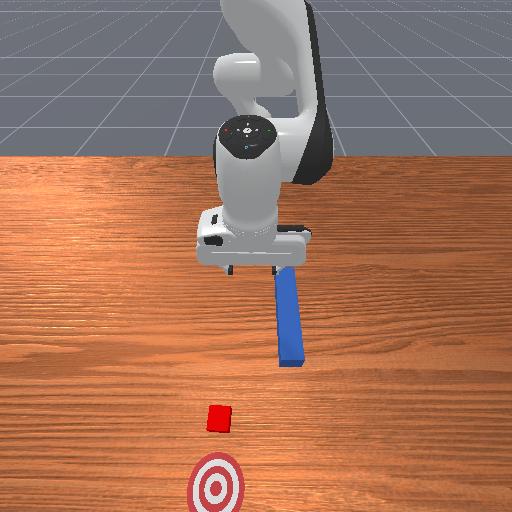}
  \caption{PokeCube-v1}
\end{subfigure}
\hfill
\begin{subfigure}{0.23\textwidth}
  \includegraphics[width=\linewidth]{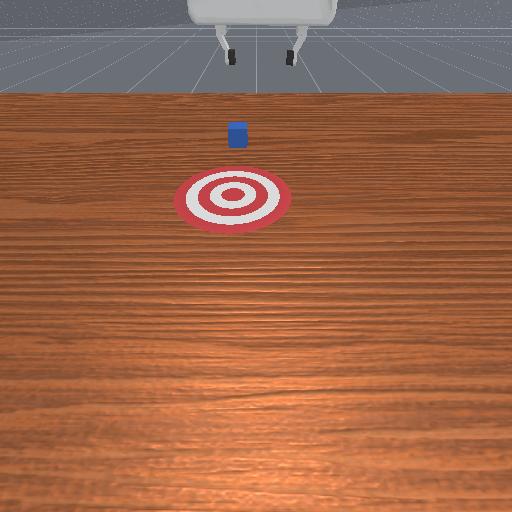}
  \caption{PullCube-v1}
\end{subfigure}

\vspace{0.5em}

\begin{subfigure}{0.23\textwidth}
  \includegraphics[width=\linewidth]{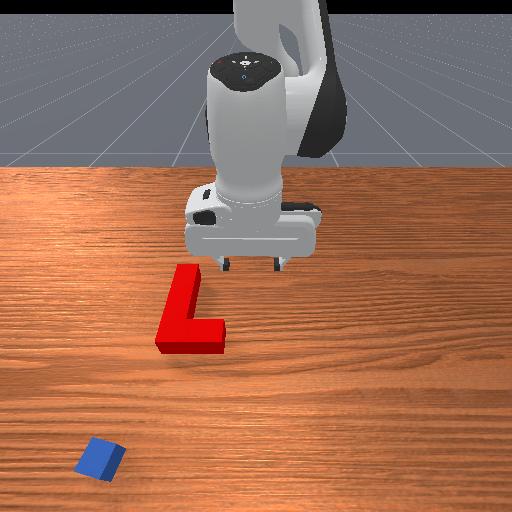}
  \caption{PullCubeTool-v1}
\end{subfigure}
\hfill
\begin{subfigure}{0.23\textwidth}
  \includegraphics[width=\linewidth]{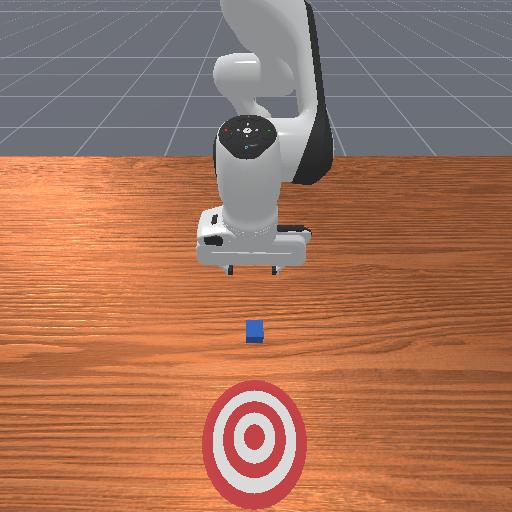}
  \caption{PushCube-v1}
\end{subfigure}
\hfill
\begin{subfigure}{0.23\textwidth}
  \includegraphics[width=\linewidth]{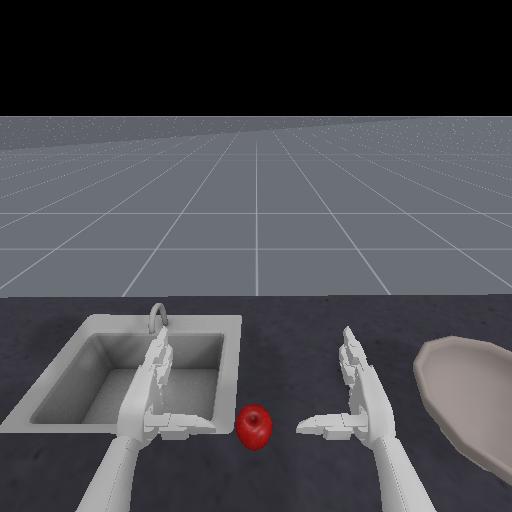}
  \caption{\scalebox{0.6}{UnitreeG1PlaceAppleInBowl-v1}}
\end{subfigure}
\hfill
\begin{subfigure}{0.23\textwidth}
  \includegraphics[width=\linewidth]{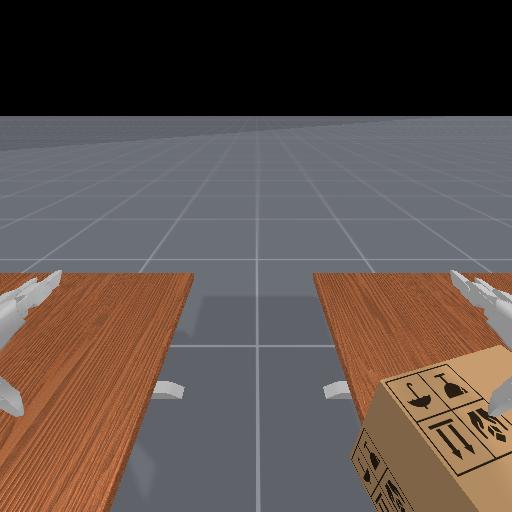}
  \caption{\scalebox{0.6}{UnitreeG1TransportBox-v1}}
\end{subfigure}

\caption{Default (no perturbation) scenes for each of the 8 tasks used for training.}
\label{fig:task_default_views}
\end{figure}
To train \methodName and the baselines, we used 8 tasks from ManiSkill3 (version \texttt{mani\_skill==3.0.0b21}) with the default normalized dense reward. The tasks and their corresponding configurations are summarized in table~\ref{tab:task_configs} and visualized in figure~\ref{fig:task_default_views}.

\begin{table}[H]
\centering
\begin{tabular}{lcc}
\hline
\textbf{Task Name} & \textbf{Max Episode Length} & \textbf{Control Mode} \\
\hline
LiftPegUpright-v1 & 50 & PD EE Delta Pos \\
PickCube-v1* & 50 & PD EE Delta Pos \\
PokeCube-v1 & 50 & PD EE Delta Pos \\
PullCube-v1 & 50 & PD EE Delta Pos \\
PullCubeTool-v1 & 100 & PD EE Delta Pos \\
PushCube-v1 & 50 & PD EE Delta Pos \\
UnitreeG1PlaceAppleInBowl-v1 & 100 & PD Joint Delta Pos \\
UnitreeG1TransportBox-v1 & 100 & PD Joint Delta Pos \\
\hline
\end{tabular}
\caption{Task configurations used for training \methodName and baselines.}
\label{tab:task_configs}
\end{table}

\noindent
* For the PickCube-v1 task, we created a modified version that makes the visual target sphere visible. This allows vision-based policies to leverage goal information. We achieved this by commenting out the following line in the original task code:
\begin{verbatim}
# self._hidden_objects.append(self.goal_site)
\end{verbatim}

\section{Hyperparameters}
\label{appendix_hyperparams}
Each method was trained using its own hyperparameter configuration. Once a configuration was found to perform well for a given method, it was kept fixed across all tasks. For the baselines, we used the official implementations provided by the authors and used their default hyperparameters as starting points. In most cases, these settings were appropriate without modification. The only adjustment we made was to the discount factor, which we set to 0.80 for all methods, as we found this value worked better with ManiSkill3 and is also used in the framework’s official baselines.

For \methodName, text tags were provided manually per task by a human. However, automatic tagging is also feasible using a VLM or image text tagging models such as RAM++ \citep{zhang2023recognizeanythingstrongimage}. We opted for manual tags to reduce the scope of this research and to reduce computational overhead.

Note that due to computational constraints, we did not perform an extensive hyperparameter search for \methodName, and we did not tune the parameters of YOLO-World or SAM.

Tables~\ref{tab:segdac_hyperparams} and \ref{tab:baseline_hyperparams} summarize the key hyperparameters used for \methodName and the baselines. We report only the most relevant parameters for reproducibility, omitting architectural defaults that are standard for each algorithm. For the complete list of hyperparameters, readers are invited to check the official code repository \url{https://github.com/SegDAC/SegDAC}.

\begin{table}[H]
\centering
\caption{Text tags used by \methodName for each task.}
\label{tab:text_tags_per_task}
\begin{adjustbox}{width=\textwidth}
\begin{tabular}{p{4.5cm} p{9.5cm}}
\toprule
\textbf{Task} & \textbf{Text Tags} \\
\midrule
LiftPegUpright-v1 & background, robot, gripper, rectangle bar, wooden peg \\
PickCube-v1 & background, robot, gripper, small box, sphere \\
PushCube-v1 & background, robot, gripper, small box, target \\
PullCube-v1 & background, robot, gripper, small box, target \\
PokeCube-v1 & background, robot, gripper, small box, bar, target \\
PullCubeTool-v1 & background, robot, gripper, small box, block \\
\scriptsize UnitreeG1PlaceAppleInBowl-v1 & background, robot, robot hand, gripper, apple, bowl \\
UnitreeG1TransportBox-v1 & background, robot, robot hand, gripper, box \\
\bottomrule
\end{tabular}
\end{adjustbox}
\end{table}

The list of text tags used by \methodName for each task is provided in Table~\ref{tab:text_tags_per_task}.

\begin{table}[H]
\centering
\begin{tabular}{lc}
\hline
\textbf{Hyperparameter} & \textbf{\methodName} \\
\hline
Actor learning rate & $3 \times 10^{-4}$ \\
Critic learning rate & $5 \times 10^{-4}$ \\
Entropy learning rate & $3 \times 10^{-4}$ \\
Optimizer & Adam \\
Gamma (discount) & 0.80 \\
Target update rate ($\tau$) & 0.01 \\
Actor update frequency & 1 \\
Critic update frequency & 1 \\
Target networks update frequency & 2 \\
Image resolution & 512 \\
Min log std & -10 \\
Max log std & 2 \\
Embedding dim & 128 \\
Transformer decoder layers & 6 \\
Transformer decoder heads & 8 \\
Transformer decoder hidden size & 1024 \\
Transformer dropout & 0.0 \\
Projection head & ResidualMLP \\
Projection head hidden layers & 4 \\
Projection head hidden size & 256 \\
Projection head norm & LayerNorm \\
Projection head pre-normalize input & true \\
Projection head activation & ReLU \\
YOLO-World & yolov8s-worldv2 \\
YOLO-World Confidence Threshold & 0.0001 \\
YOLO-World IoU Threshold & 0.01 \\
EfficientViT SAM & efficientvit-sam-l0 \\
Mask post-processing kernel size & 9 \\
Segment Embedding Min Pixel & 4 \\
Replay Buffer Max Size & 1,000,000 \\
\hline
\end{tabular}
\caption{Key hyperparameters used for training \methodName.}
\label{tab:segdac_hyperparams}
\end{table}
Note that in \methodName, the actor and critic each use their own transformer decoder and projection head, but they share the same architecture and hyperparameters. For clarity, we omit separate entries for each and report the shared configuration.
\begin{table}[H]
\centering
\begin{adjustbox}{width=\textwidth}
\begin{tabular}{lcccccc}
\hline
\textbf{Hyperparameter} & \textbf{SADA} & \textbf{MaDi} & \textbf{DrQ-v2} & \textbf{SAC-AE} & \textbf{SAM-G} & \textbf{SMG} \\
\hline
Actor Learning Rate & $5 \times 10^{-4}$ & $1 \times 10^{-3}$ & $1 \times 10^{-4}$ & $1 \times 10^{-4}$ & $1 \times 10^{-4}$ & $1 \times 10^{-3}$ \\
Critic Learning Rate & $5 \times 10^{-4}$ & $1 \times 10^{-3}$ & $1 \times 10^{-4}$ & $1 \times 10^{-3}$ & $1 \times 10^{-4}$ & $1 \times 10^{-3}$ \\
Entropy Learning Rate & $5 \times 10^{-4}$ & $1 \times 10^{-4}$ & -- & $1 \times 10^{-4}$ & -- & $1 \times 10^{-4}$ \\
Target Update Rate ($\tau$) & 0.01 & 0.01 & 0.01 & 0.01 & 0.01 & 0.01 \\
Actor Update Frequency & 2 & 2 & 2 & 1 & 4 & 2 \\
Critic Update Frequency & 2 & 1 & 2 & 1 & 4 & 2 \\
Target Networks Update Frequency & 2 & 2 & 2 & 2 & 4 & 2 \\
Replay Buffer Size & 1,000,000 & 1,000,000 & 1,000,000 & 1,000,000 & 1,000,000 & 1,000,000 \\
Gamma (discount) & 0.80 & 0.80 & 0.80 & 0.80 & 0.80 & 0.80 \\
Image Resolution & 84 & 84 & 84 & 84 & 84 & 84 \\
MLP Projection Layers & 4 & 3 & 4 & 3 & 4 & 3 \\
MLP Features Dim & 1024 & 1024 & 1024 & 1024 & 1024 & 1024 \\
Frame Stack & 3 & 3 & 3 & 3 & 3 & 3 \\
N-step Return & -- & -- & 3 & -- & 3 & -- \\
Optimizer & Adam & Adam & Adam & Adam & Adam & Adam \\
\hline
\end{tabular}
\end{adjustbox}
\caption{Key hyperparameters for baseline methods. Note that SAM-G doesn't support batching so we had to change the update frequency to match the update-to-data ratio of 0.25 used by every other methods.}
\label{tab:baseline_hyperparams}
\end{table}

\section{Morphological Mask Refinement}
\label{appendix_morphological_mask_refinement}
\begin{figure}[H]
    \centering

    \begin{subfigure}{0.3\linewidth}
        \includegraphics[width=\linewidth]{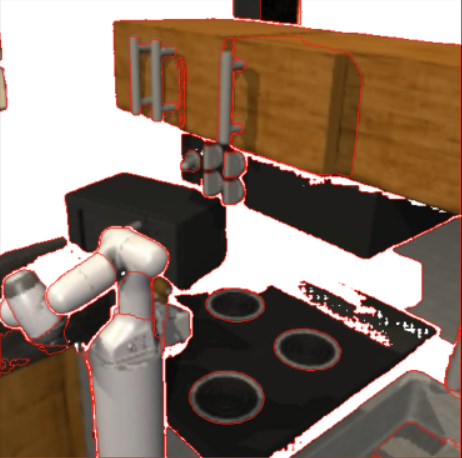}
        \caption{Raw SAM segments stitched into 1 frame.}
    \end{subfigure}
    \quad
    \begin{subfigure}{0.3\linewidth}
        \includegraphics[width=\linewidth]{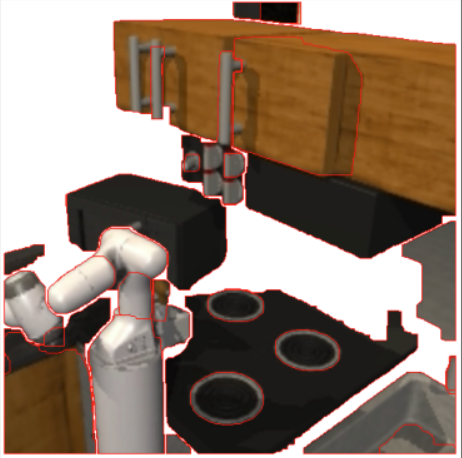}
        \caption{After post-processing.}
    \end{subfigure}

    \vspace{1em}

    \begin{subfigure}{0.3\linewidth}
        \includegraphics[width=\linewidth]{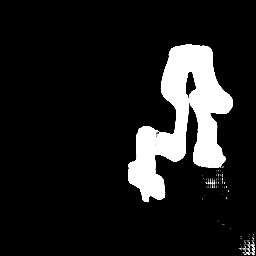}
        \caption{Raw binary mask.}
    \end{subfigure}
    \quad
    \begin{subfigure}{0.3\linewidth}
        \includegraphics[width=\linewidth]{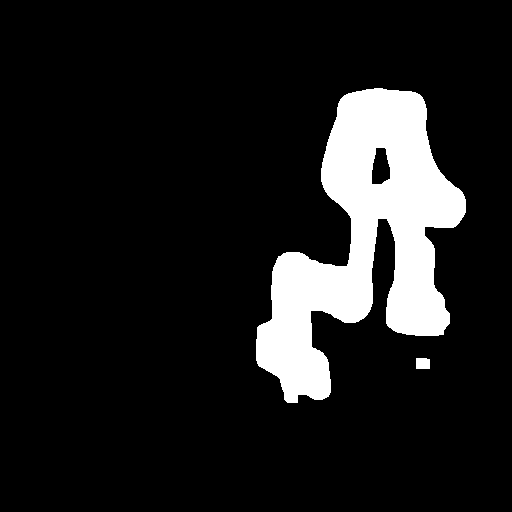}
        \caption{Post-processed mask.}
    \end{subfigure}

    \caption{Effect of our lightweight post-processing on SAM outputs. Top: stitched overlays. Bottom: single mask.}
    \label{fig:appendix_seg_post_process}
\end{figure}
To improve the quality of segments predicted by SAM, prior work such as SAM \citep{kirillov2023segment}, SAM-G \citep{wang2023generalizablevisualreinforcementlearning}, and PerSAM \citep{zhang2023personalizesegmentmodelshot} typically rely on heavy post-processing pipelines, often involving iterative mask refinement. In contrast, we adopt a lightweight and efficient approach based on morphological opening and closing operations \citep{Sreedhar_2012}. These classical image processing techniques are fast and effective at removing small artifacts such as holes or sprinkles commonly found in raw segmentation masks. The impact of this post-processing is illustrated in Figure~\ref{fig:appendix_seg_post_process}.

Our implementation of this post-processing procedure is written in pure PyTorch, as shown below:
\begin{lstlisting}[language=Python]
import torch
import torch.nn.functional as F

...
    def post_process_masks(self, masks: torch.Tensor, kernel_size: int) -> torch.Tensor:
        """
        masks: (N,1,H,W)
        """
        opened = self.apply_morphological_opening(
            masks.to(torch.float32), kernel_size=kernel_size
        )
        return self.apply_morphological_closing(opened, kernel_size=kernel_size).to(
            torch.uint8
        )

    def apply_morphological_opening(
        self, masks: torch.Tensor, kernel_size: int
    ) -> torch.Tensor:
        eroded = self.apply_erosion(masks, kernel_size)
        return self.apply_dilation(eroded, kernel_size)

    def apply_erosion(self, masks: torch.Tensor, kernel_size: int) -> torch.Tensor:
        return -self.max_pool2d_same_dim(-masks, kernel_size=kernel_size)

    def apply_dilation(self, masks: torch.Tensor, kernel_size: int) -> torch.Tensor:
        return self.max_pool2d_same_dim(masks, kernel_size=kernel_size)

    def max_pool2d_same_dim(self, masks: torch.Tensor, kernel_size: int):
        stride = 1
        dilation = 1
        pad_h_top, pad_h_bottom = self.compute_padding(
            kernel_size, stride, dilation)
        pad_w_left, pad_w_right = self.compute_padding(
            kernel_size, stride, dilation)

        padded_input = F.pad(
            masks, (pad_w_left, pad_w_right, pad_h_top, pad_h_bottom))

        return F.max_pool2d(
            padded_input, kernel_size=kernel_size, stride=stride, dilation=dilation
        )

    def compute_padding(self, kernel_size, stride=1, dilation=1):
        padding_total = max(0, (kernel_size - 1) * dilation - stride + 1)
        pad_before = padding_total // 2
        pad_after = padding_total - pad_before
        return pad_before, pad_after

    def apply_morphological_closing(
        self, masks: torch.Tensor, kernel_size: int
    ) -> torch.Tensor:
        dilated = self.apply_dilation(masks, kernel_size)
        return self.apply_erosion(dilated, kernel_size)
\end{lstlisting}
\newpage
\section{Dynamic Object Token Construction Details}
\label{appendix_seg_embeds_extract}

\begin{figure}[H]
    \centering
    \includegraphics[width=0.85\linewidth]{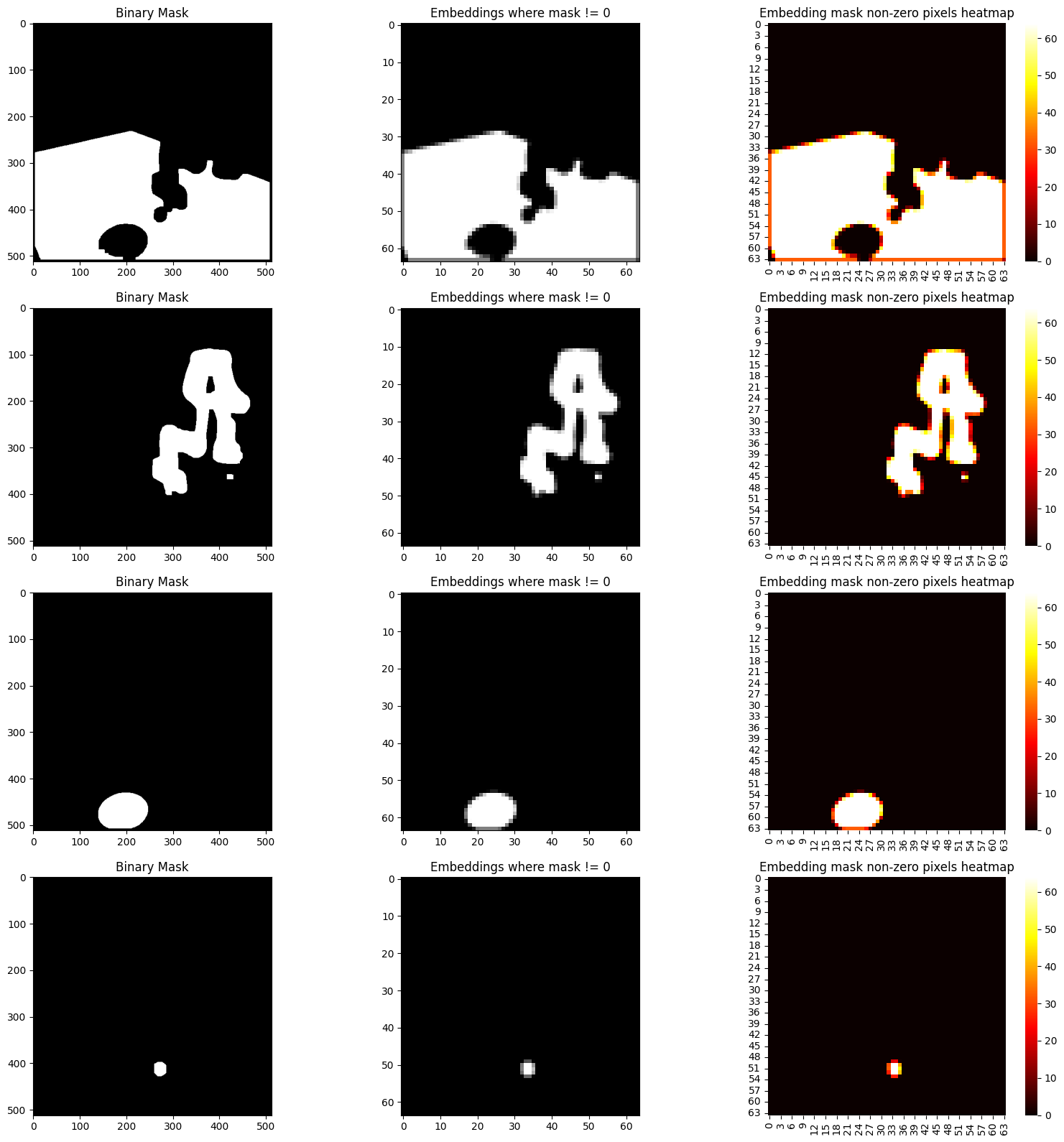}
    \caption{Left: binary mask for the selected segment. Middle: SAM encoder patch grid, showing only patches whose spatial footprint overlaps the mask. Right: per-patch mask–overlap heatmap, defined as the number (or fraction) of mask pixels inside each patch. Boundary patches have lower overlap (red) because they straddle multiple segments, while interior patches approach full coverage (white). This heatmap reflects spatial overlap only, not the contextual information carried by the patch embeddings.}
    \label{fig:sam_enc_seg_extract_non_zero_pixels_count}
\end{figure} 

Figure~\ref{fig:sam_enc_seg_extract_non_zero_pixels_count} illustrates the relationship between the binary masks produced by our grounded segmentation module (left), the corresponding \textbf{overlapping} patch embeddings (middle), and a heatmap showing the pixel count per patch (right). Most patch embeddings have a high pixel count, while those near the mask edges tend to overlap less exclusively with the segment.

\section{Segment Attention Analysis}
\label{appendix_segment_attn_analysis}
\begin{figure}[ht]
    \centering
    \vspace{-0.45cm}
    \includegraphics[width=1.00\linewidth]{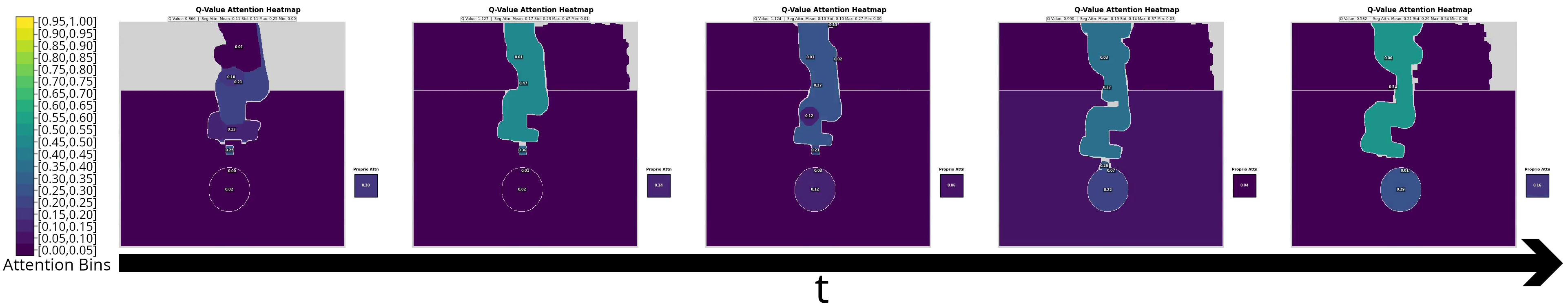}
    \caption{Critic attention on visual segments and proprioception during Q-value prediction.}
    \vspace{-0.35cm}
    \label{fig:attention_analysis}
\end{figure}

\Cref{fig:attention_analysis} shows the critic's attention distribution 
over segments and proprioception during Q-value prediction. Without 
history or frame stacking, the model adapts its attention dynamically 
over the course of a trajectory: it focuses on the robot arm and 
manipulation object at the start, then shifts toward the goal after 
grasping. The model does not maintain uniform attention on the robot 
throughout. In some tasks, it places more weight on proprioceptive 
features, while in others it emphasizes the robot arm. The goal is often 
ignored initially, especially when the first sub-task is to reach or 
grasp the object, and attention reallocates to the goal once the object 
is secured.

\begin{figure}[H]
    \centering
    \includegraphics[width=0.32\linewidth]{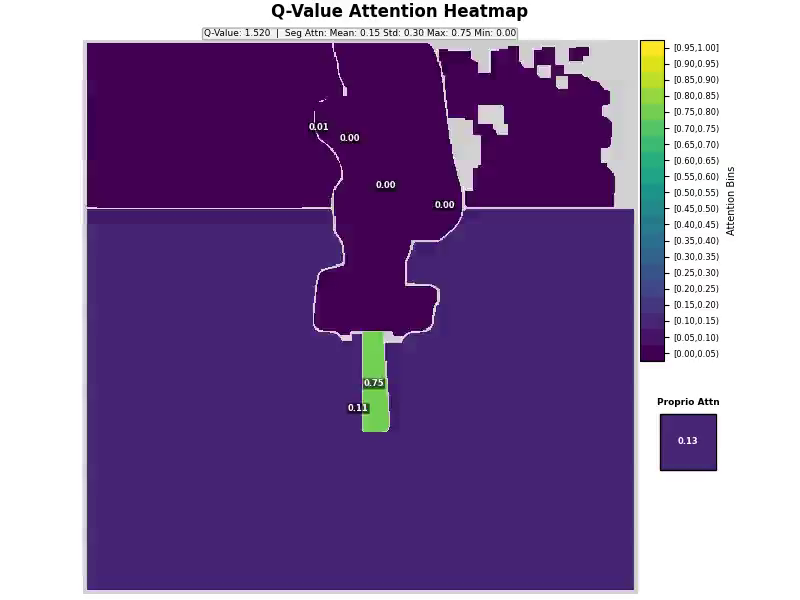}
    \includegraphics[width=0.32\linewidth]{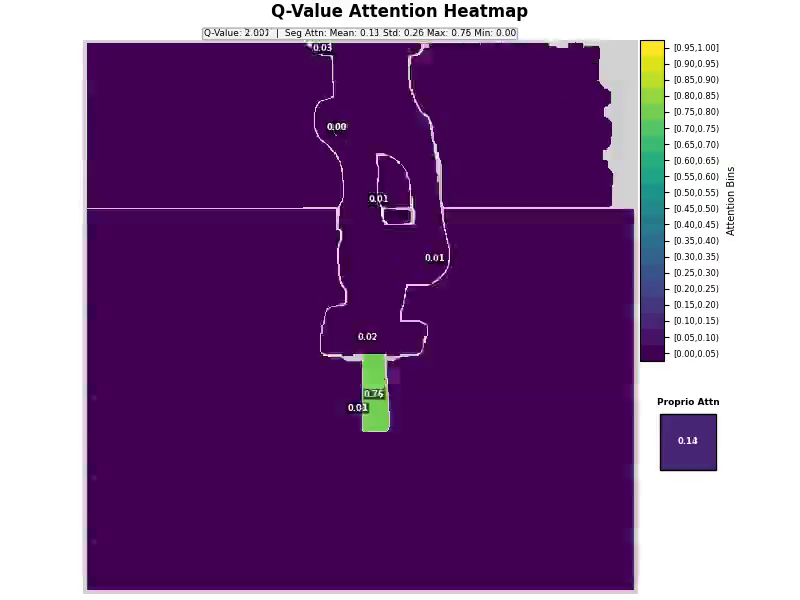}
    \includegraphics[width=0.32\linewidth]{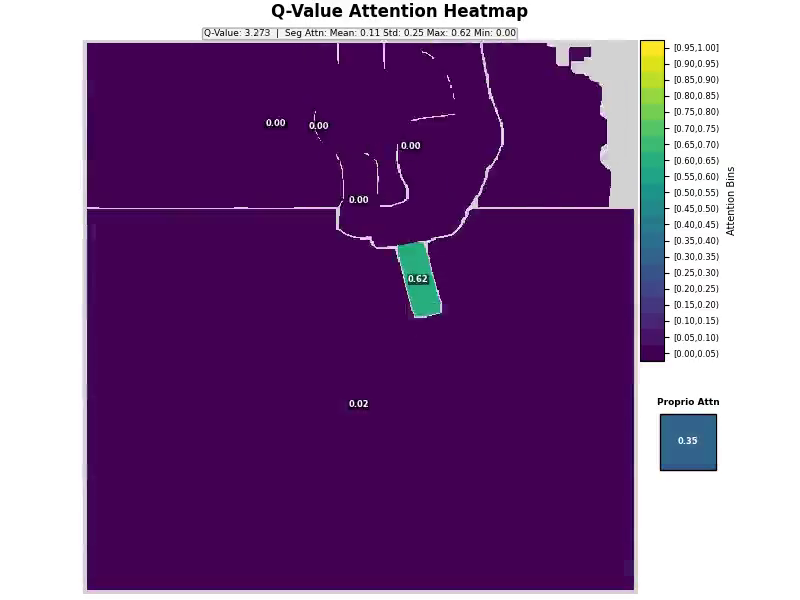}
    \includegraphics[width=0.32\linewidth]{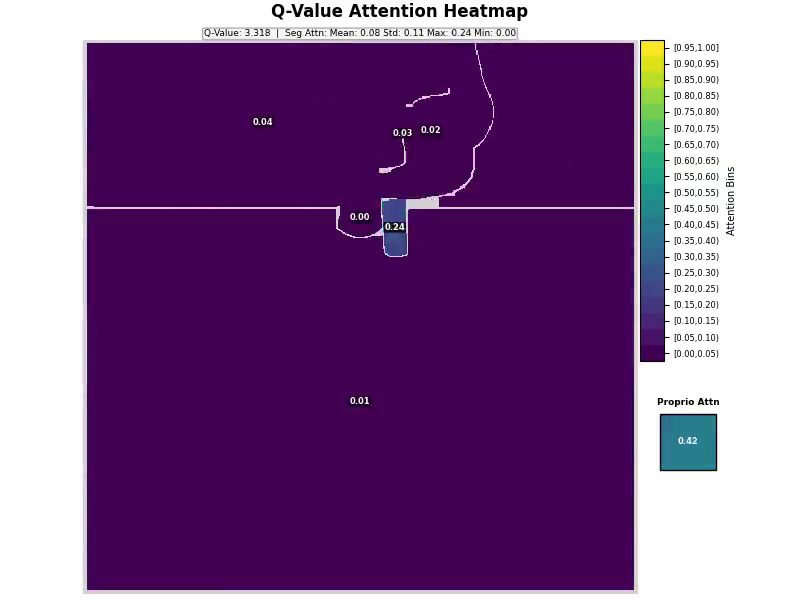}
    \includegraphics[width=0.32\linewidth]{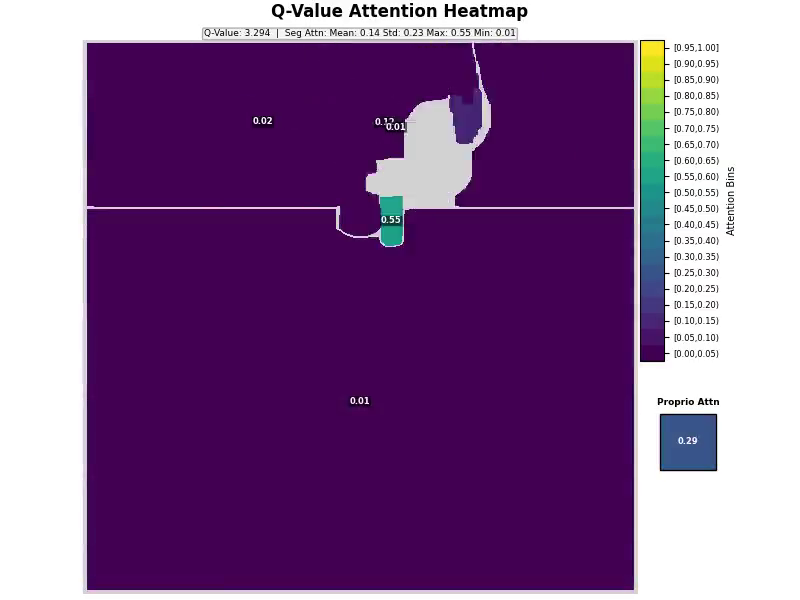}
    \caption{Average segment attention from the critic when predicting Q-values in the LiftPegUpright-v1 task. The trajectory begins in the top-left frame. Grey regions indicate areas where SAM did not detect any segments.}
    \label{fig:appendix_seg_attn_analysis_liftpegupright}
\end{figure}
\begin{figure}[H]
    \centering
    \includegraphics[width=0.32\linewidth]{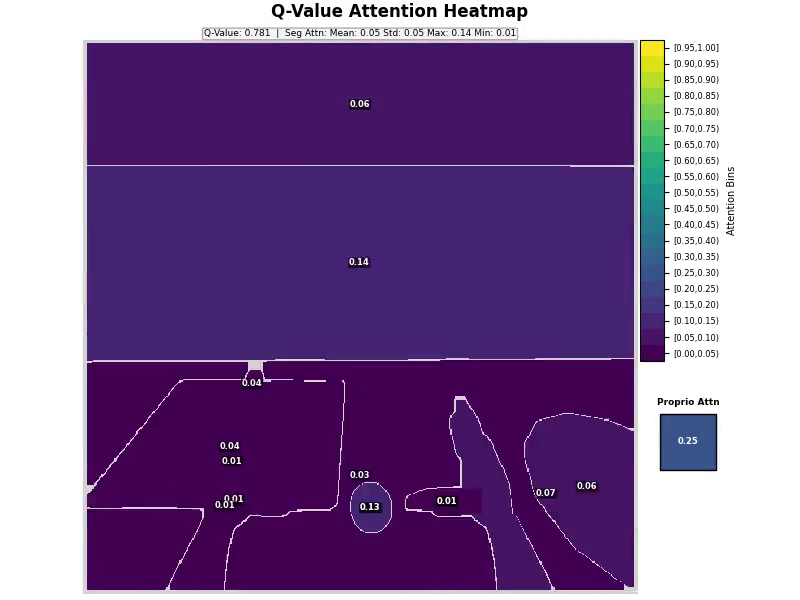}
    \includegraphics[width=0.32\linewidth]{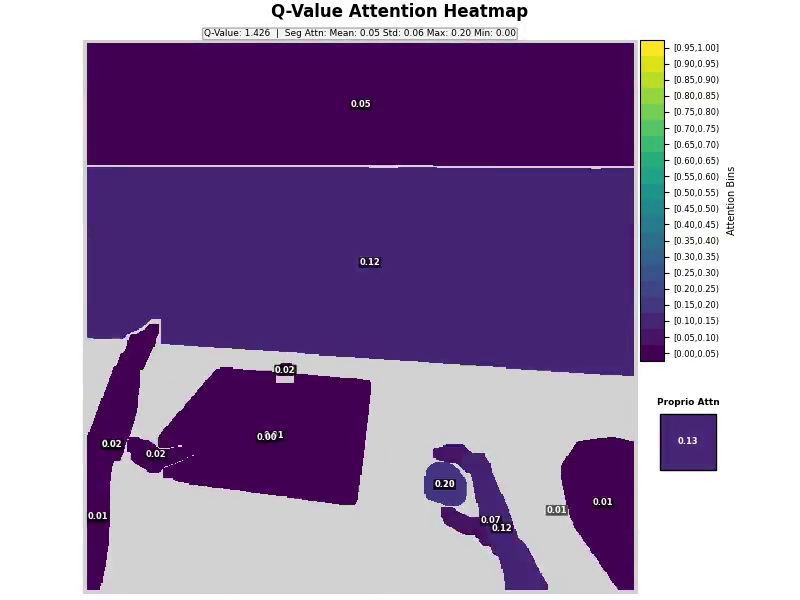}
    \includegraphics[width=0.32\linewidth]{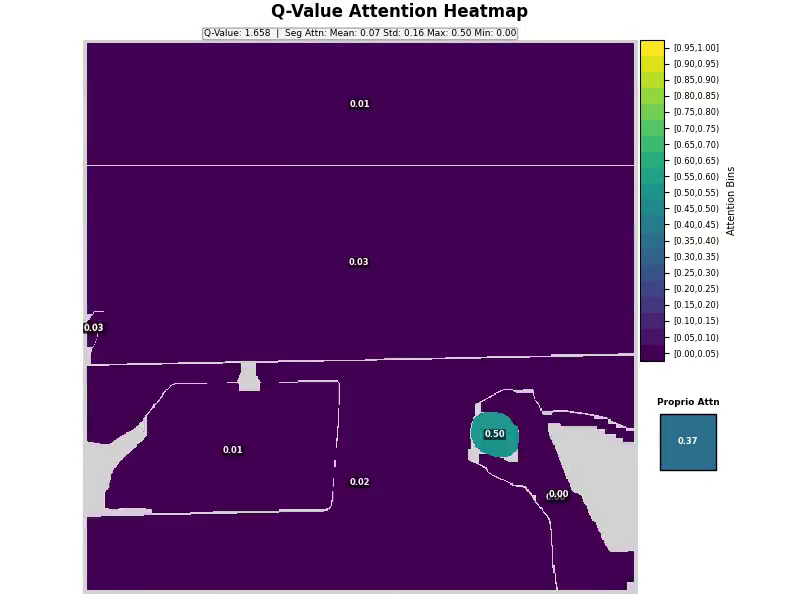}
    \includegraphics[width=0.32\linewidth]{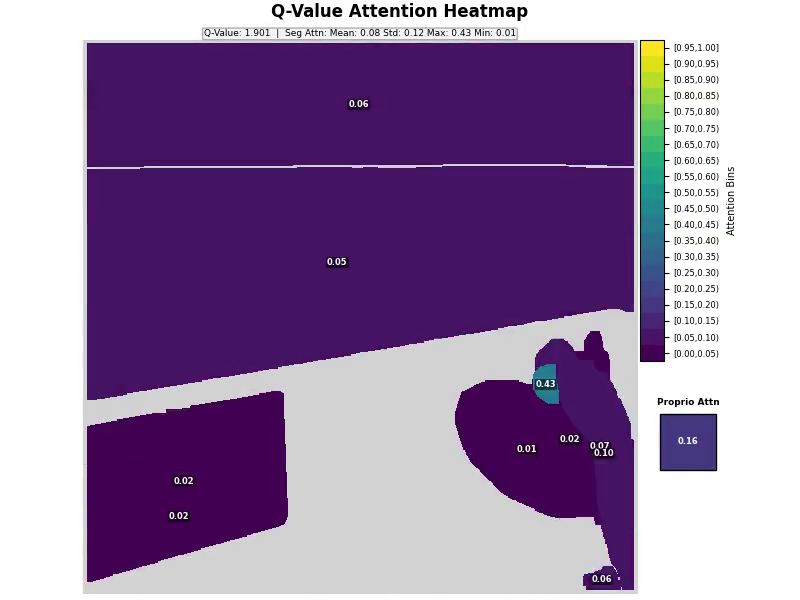}
    \includegraphics[width=0.32\linewidth]{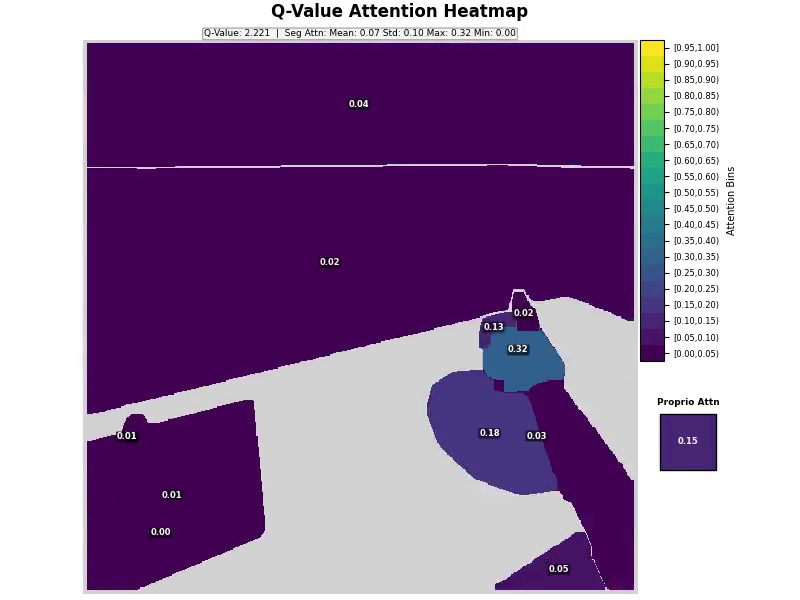}
    \includegraphics[width=0.32\linewidth]{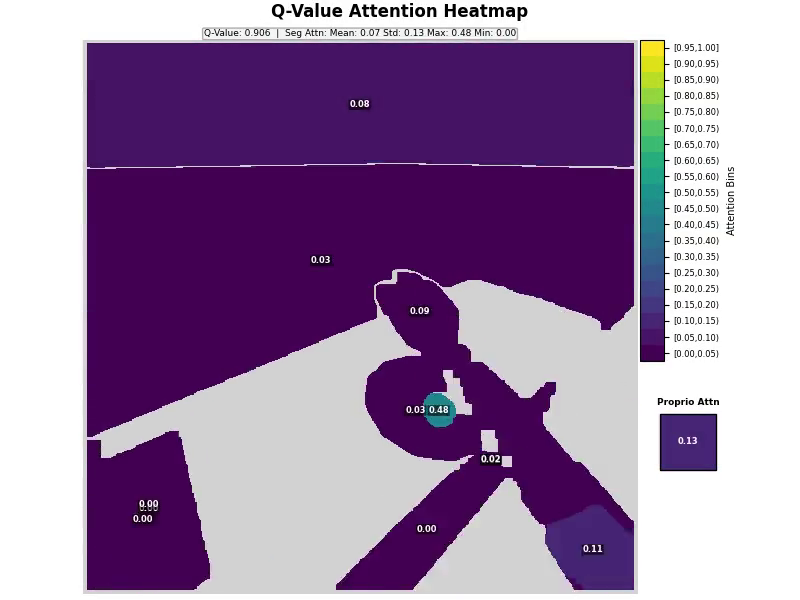}
    \caption{Average segment attention from the critic when predicting Q-values in the UnitreeG1PlaceAppleInBowl-v1 task. The trajectory begins in the top-left frame. Grey regions indicate areas where SAM did not detect any segments.}
    \label{fig:appendix_seg_attn_analysis_apple_in_bowl}
\end{figure}
\begin{figure}[H]
    \centering
    \includegraphics[width=0.24\linewidth]{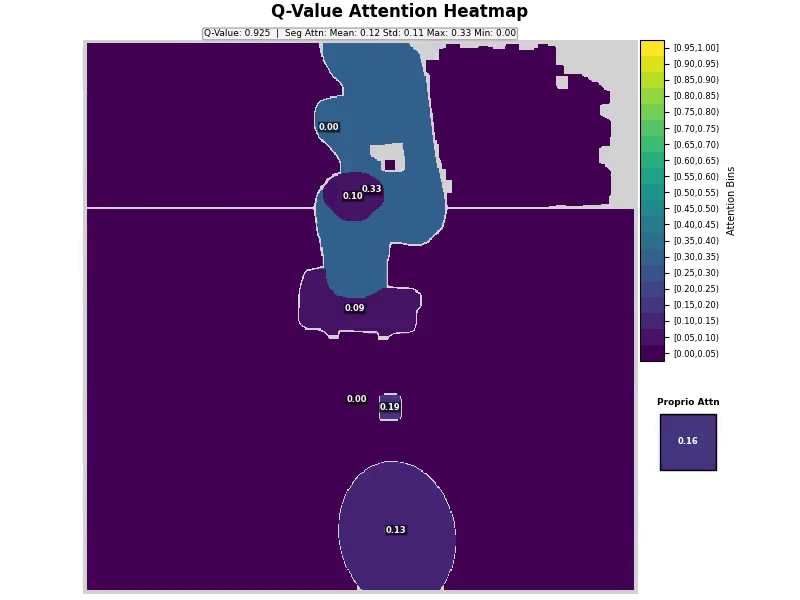}
    \includegraphics[width=0.24\linewidth]{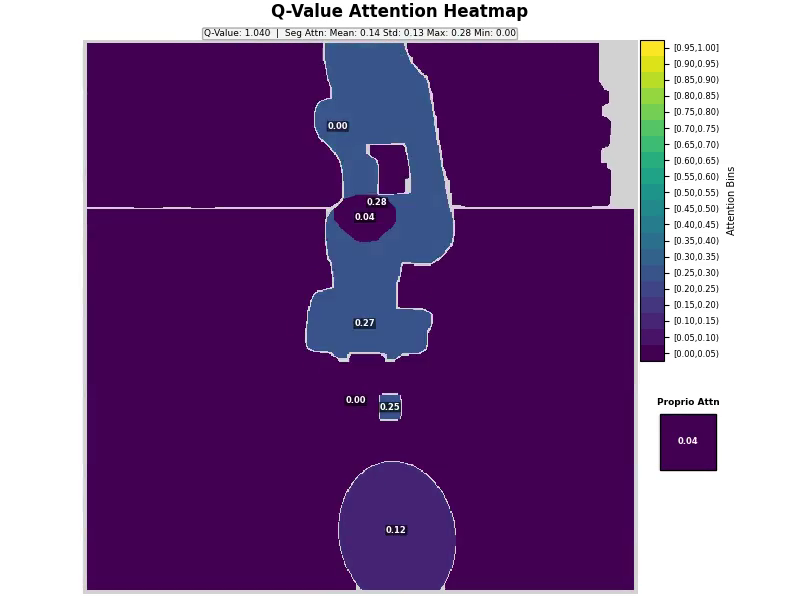}
    \includegraphics[width=0.24\linewidth]{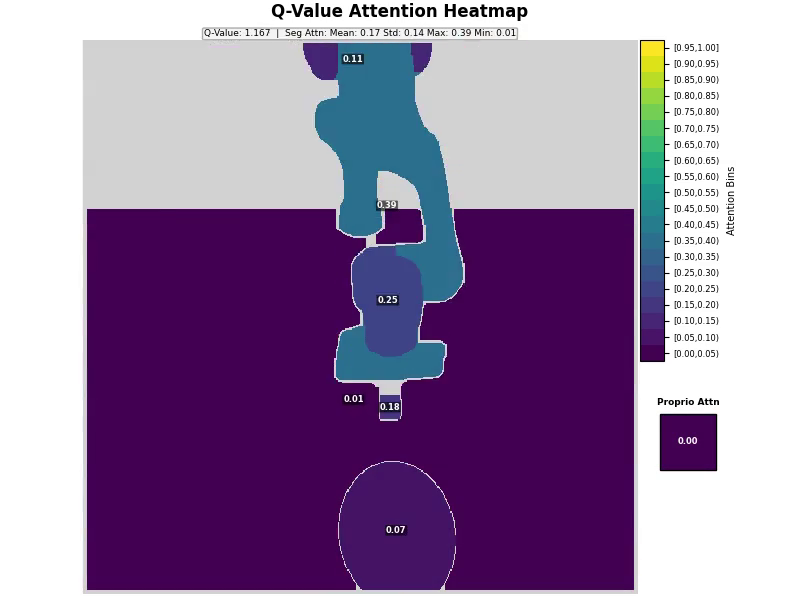}
    \includegraphics[width=0.24\linewidth]{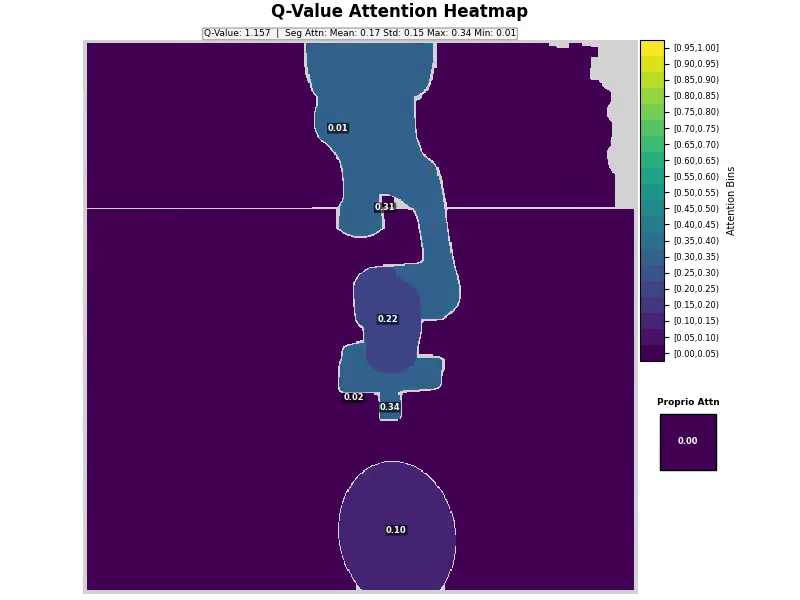}
    \includegraphics[width=0.24\linewidth]{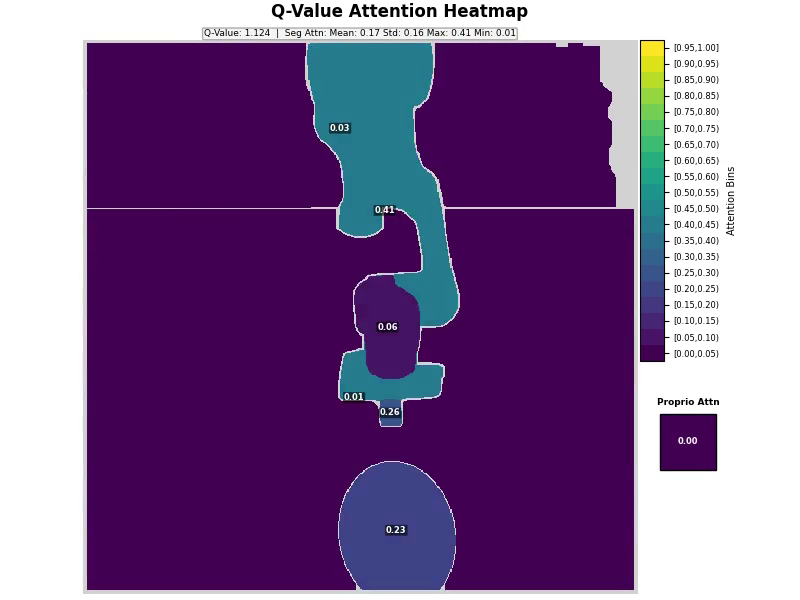}
    \includegraphics[width=0.24\linewidth]{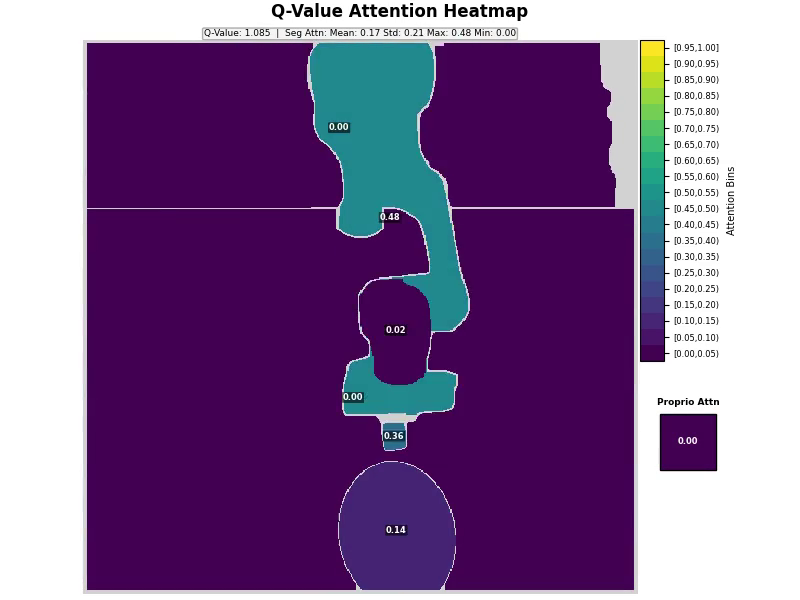}
    \includegraphics[width=0.24\linewidth]{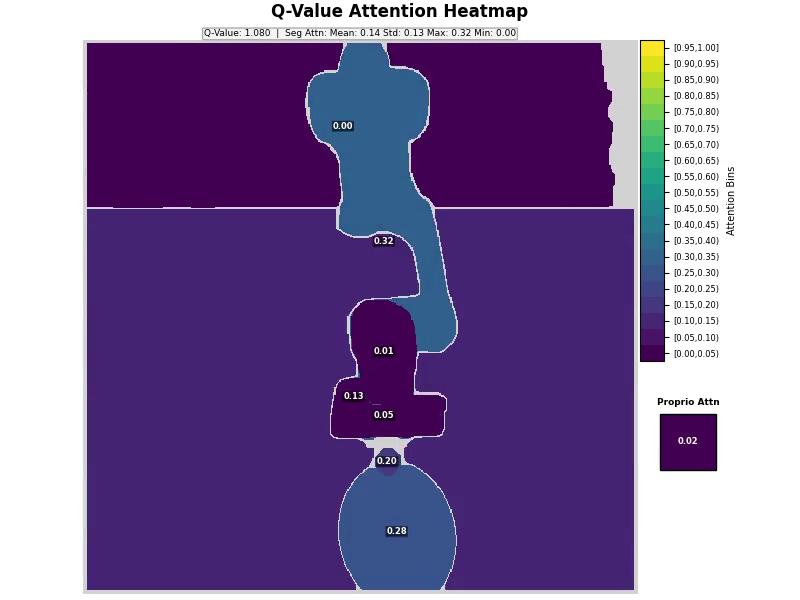}
    \includegraphics[width=0.24\linewidth]{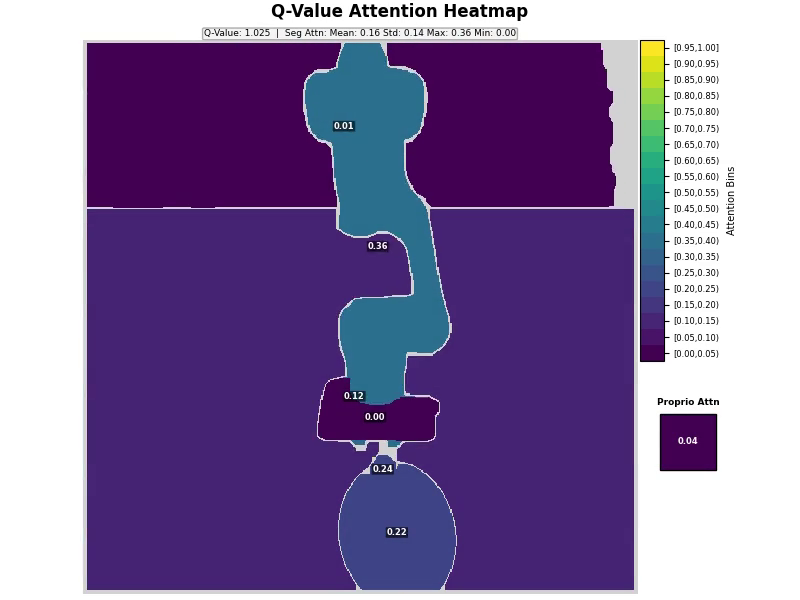}
    \caption{Average segment attention from the critic when predicting Q-values in the PushCube-v1 task. The trajectory begins in the top-left frame. Grey regions indicate areas where SAM did not detect any segments.}
    \label{fig:appendix_seg_attn_analysis_push_cube_2}
\end{figure}
\begin{figure}[H]
    \centering
    \includegraphics[width=0.24\linewidth]{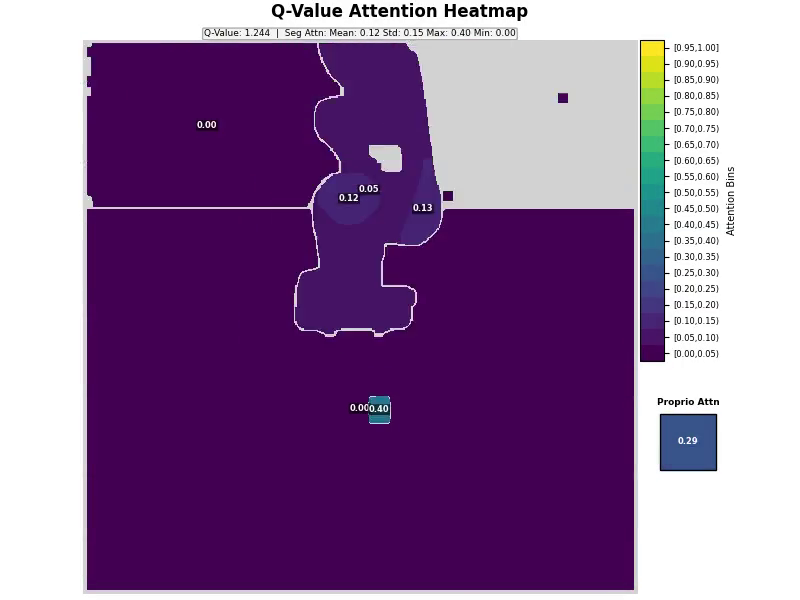}
    \includegraphics[width=0.24\linewidth]{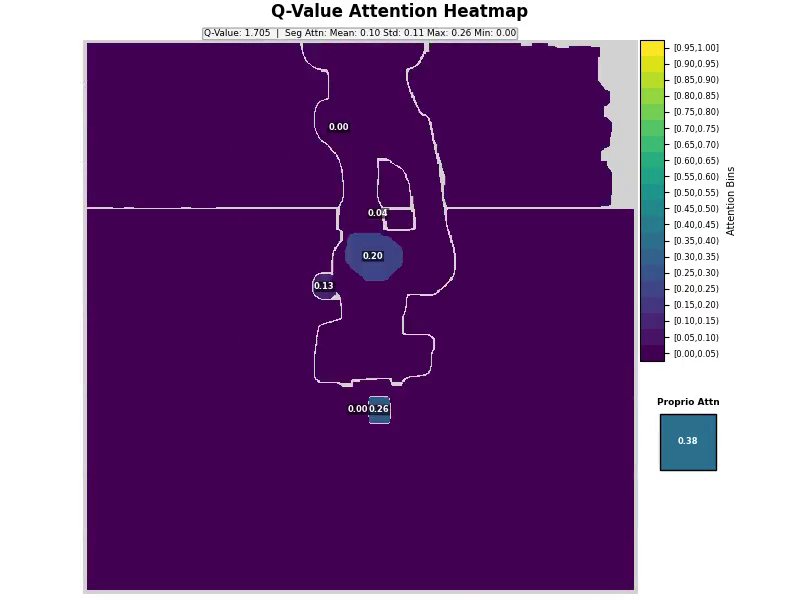}
    \includegraphics[width=0.24\linewidth]{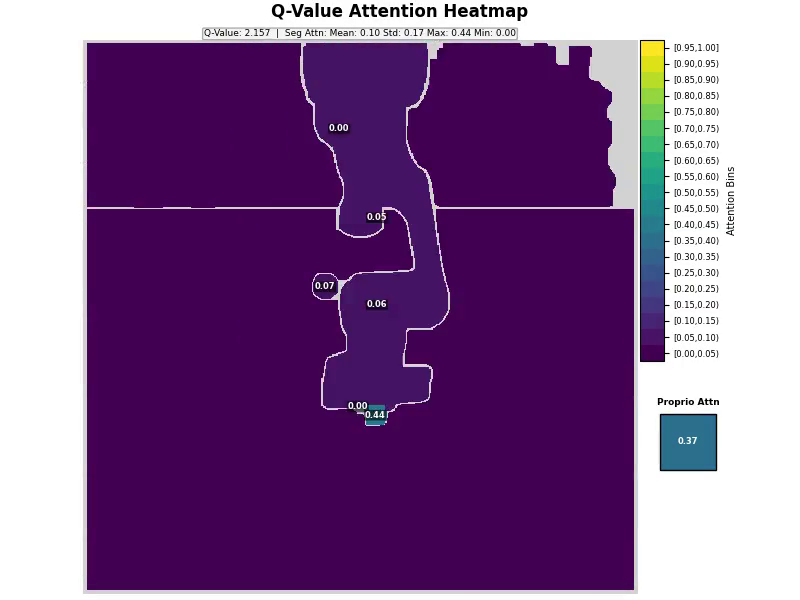}
    \includegraphics[width=0.24\linewidth]{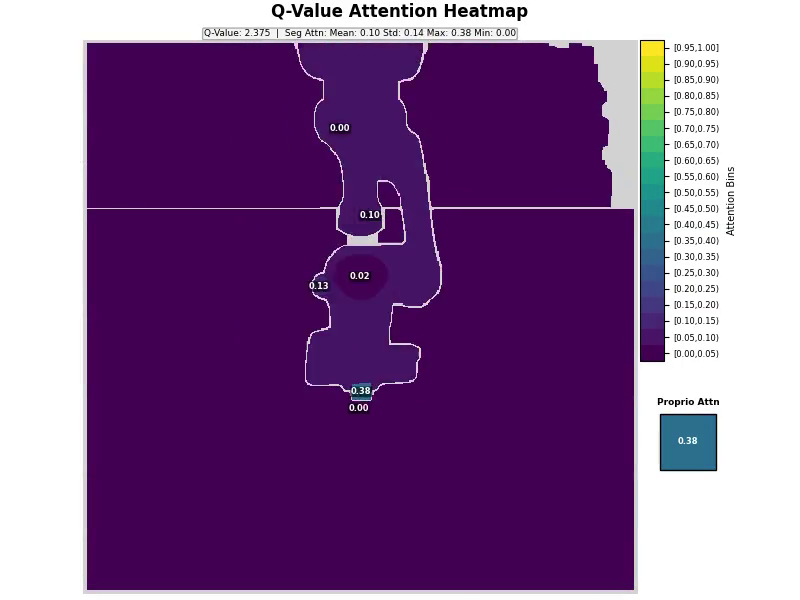}
    \includegraphics[width=0.24\linewidth]{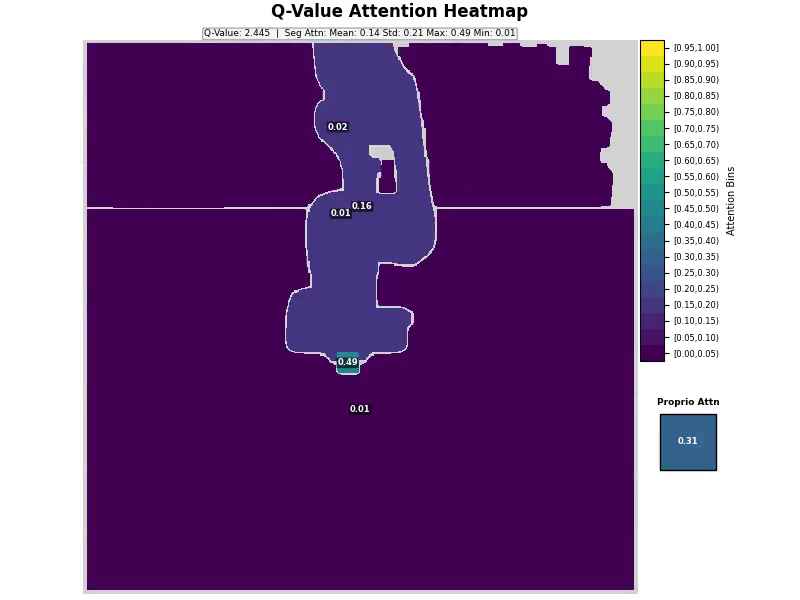}
    \includegraphics[width=0.24\linewidth]{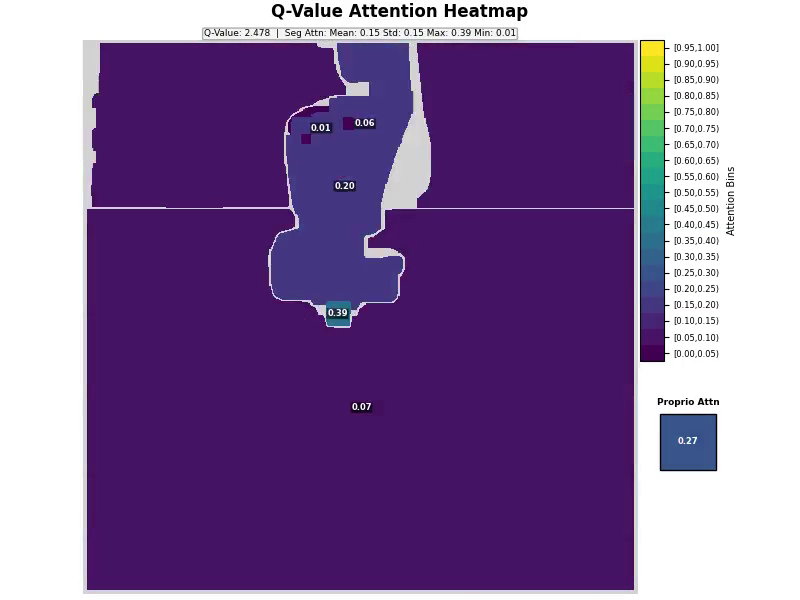}
    \includegraphics[width=0.24\linewidth]{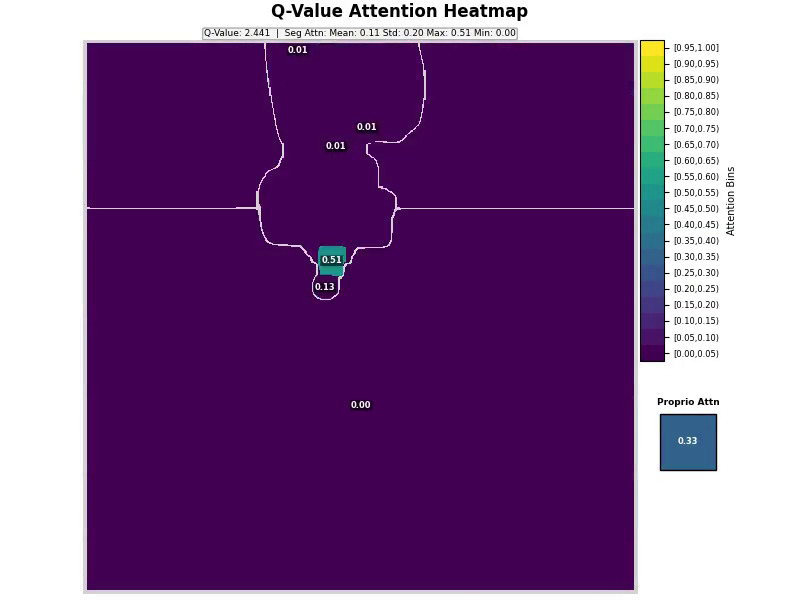}
    \caption{Average segment attention from the critic when predicting Q-values in the PickCube-v1 task. The trajectory begins in the top-left frame. Grey regions indicate areas where SAM did not detect any segments.}
    \label{fig:appendix_seg_attn_analysis_pick_cube}
\end{figure}
\begin{figure}[H]
    \centering
    \includegraphics[width=0.24\linewidth]{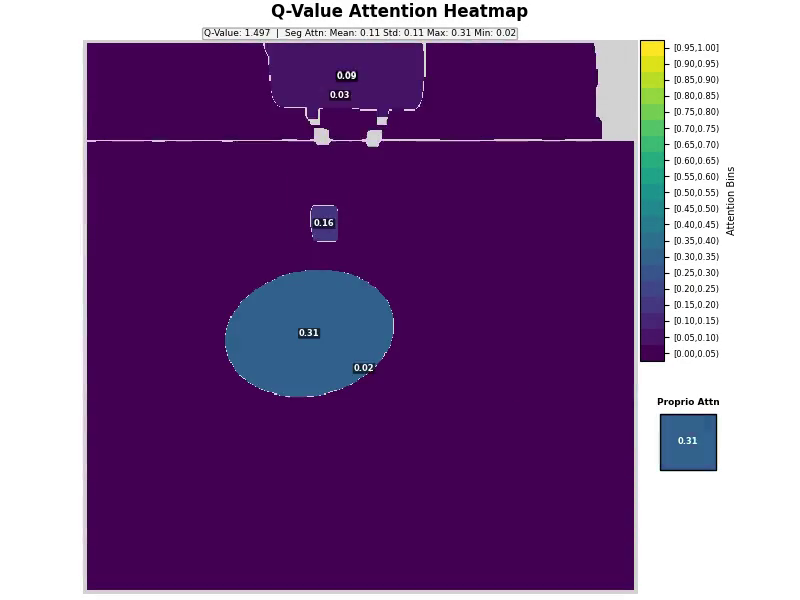}
    \includegraphics[width=0.24\linewidth]{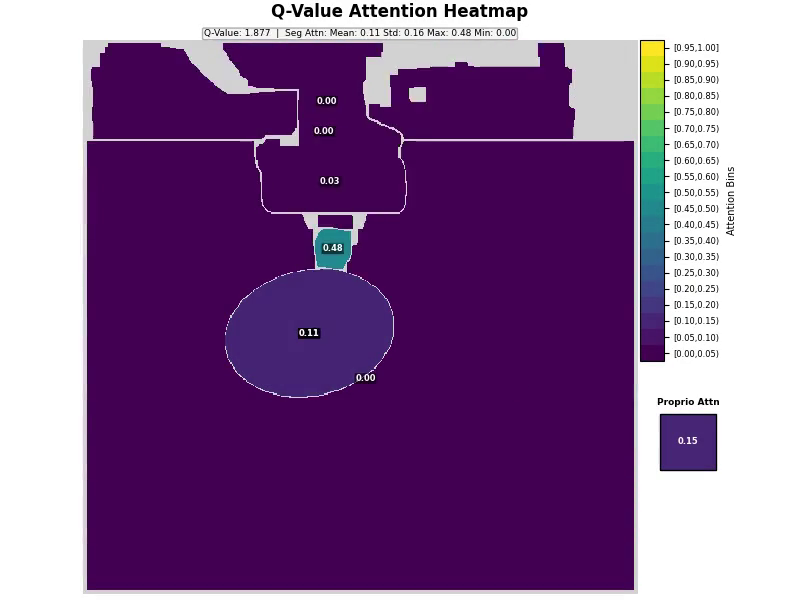}
    \includegraphics[width=0.24\linewidth]{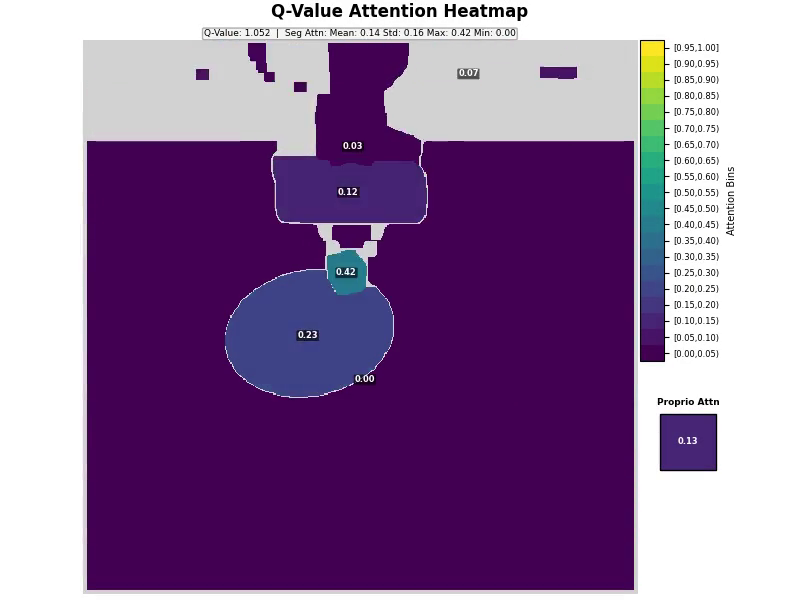}
    \includegraphics[width=0.24\linewidth]{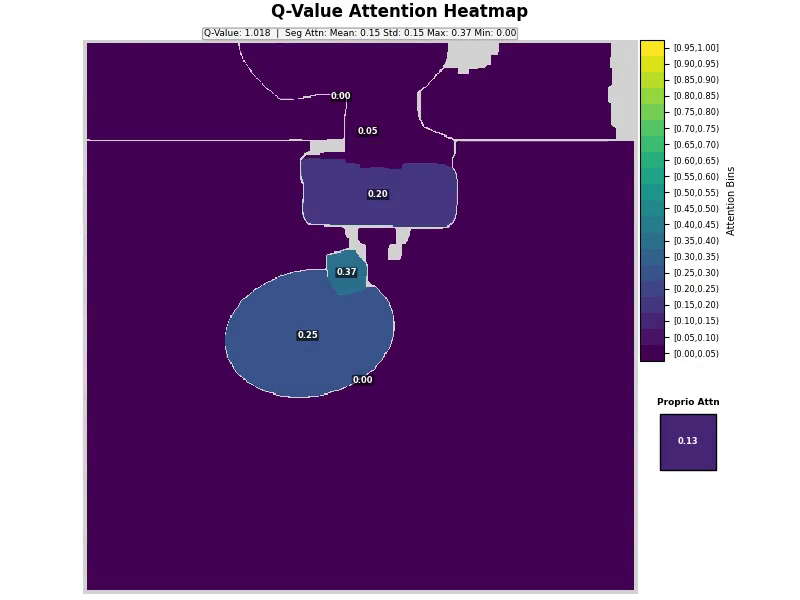}
    \caption{Average segment attention from the critic when predicting Q-values in the PullCube-v1 task. The trajectory begins on the left frame. Grey regions indicate areas where SAM did not detect any segments.}
    \label{fig:appendix_seg_attn_analysis_pull_cube}
\end{figure}
\begin{figure}[H]
    \centering
    \includegraphics[width=0.32\linewidth]{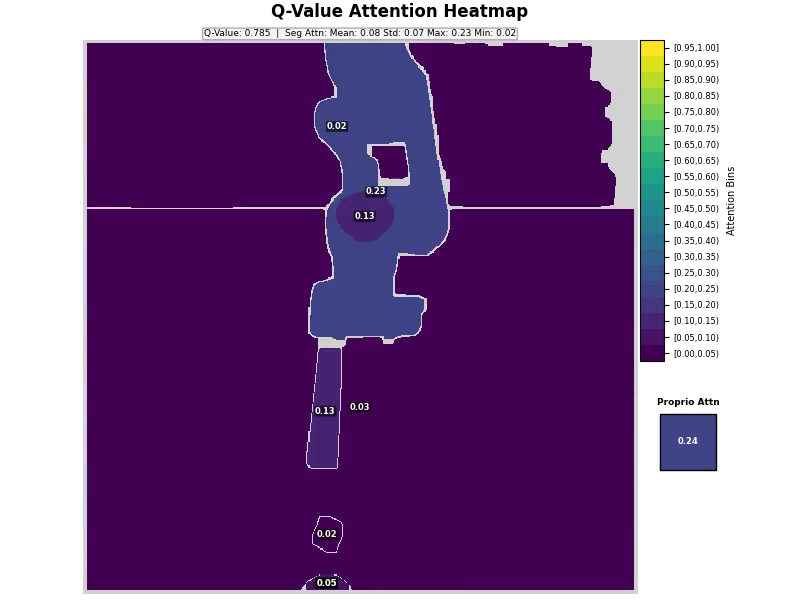}
    \includegraphics[width=0.32\linewidth]{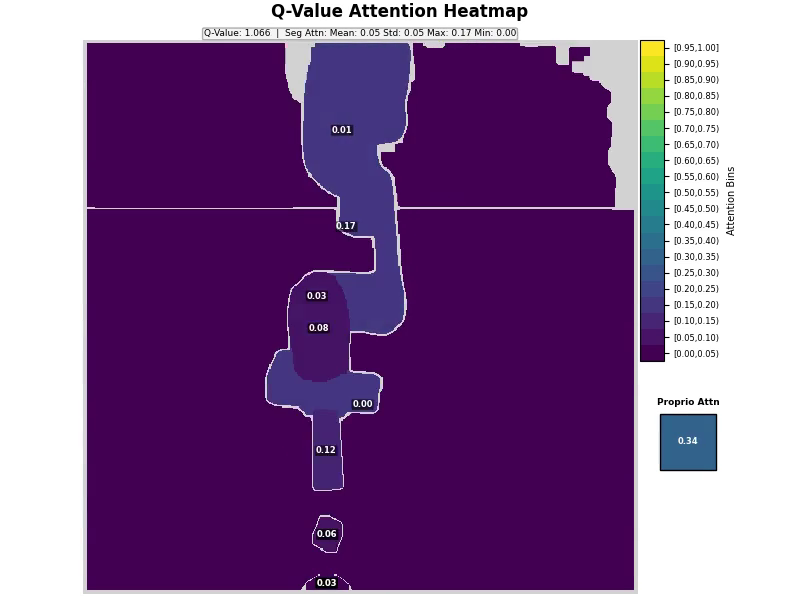}
    \includegraphics[width=0.32\linewidth]{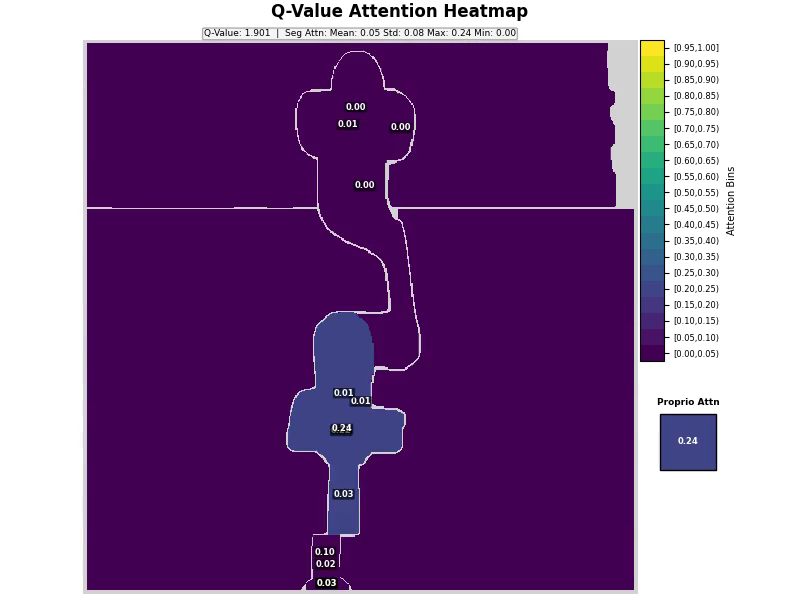}
    \includegraphics[width=0.32\linewidth]{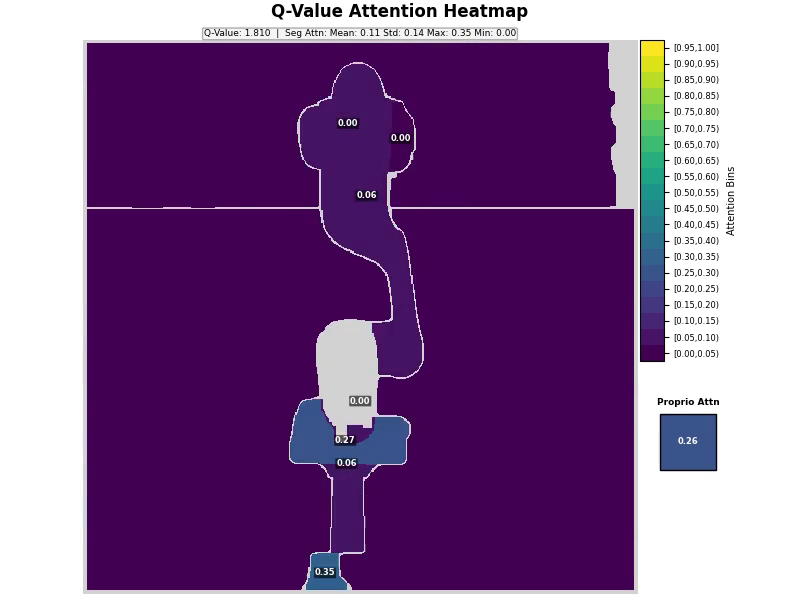}
    \caption{Average segment attention from the critic when predicting Q-values in the PokeCube-v1 task. The trajectory begins in the top-left frame. Grey regions indicate areas where SAM did not detect any segments.}
    \label{fig:appendix_seg_attn_analysis_poke_cube}
\end{figure}
\begin{figure}[H]
    \centering
    \includegraphics[width=0.32\linewidth]{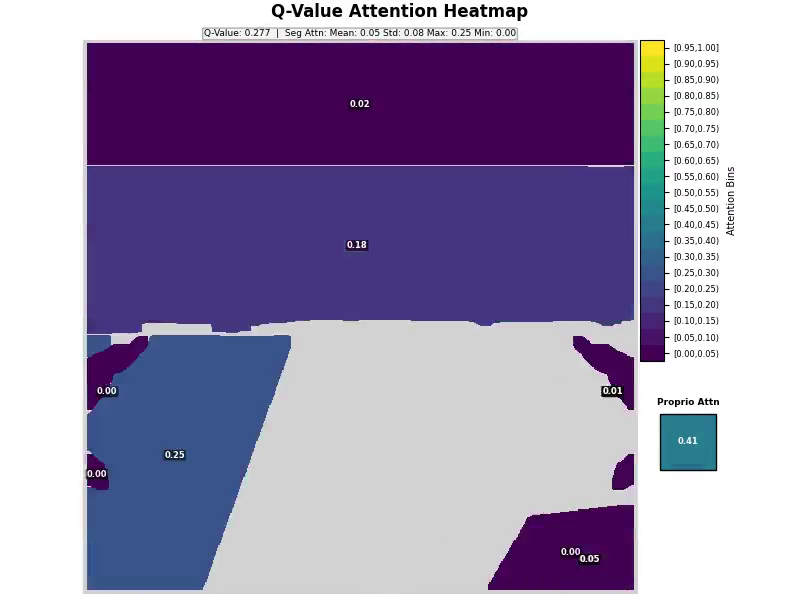}
    \includegraphics[width=0.32\linewidth]{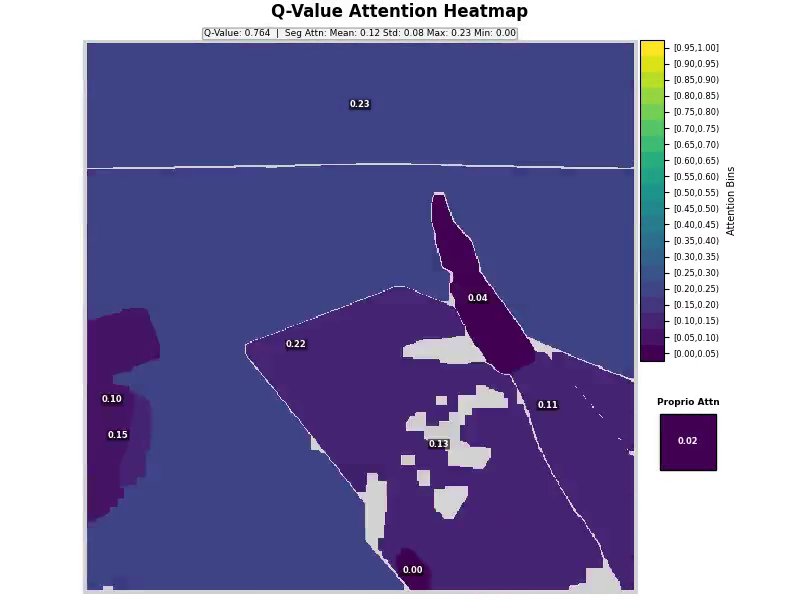}
    \includegraphics[width=0.32\linewidth]{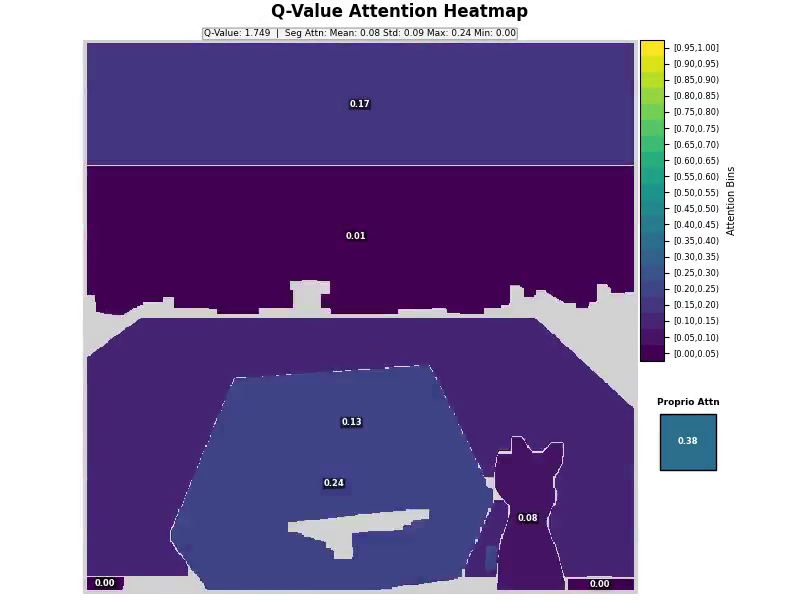}
    \includegraphics[width=0.32\linewidth]{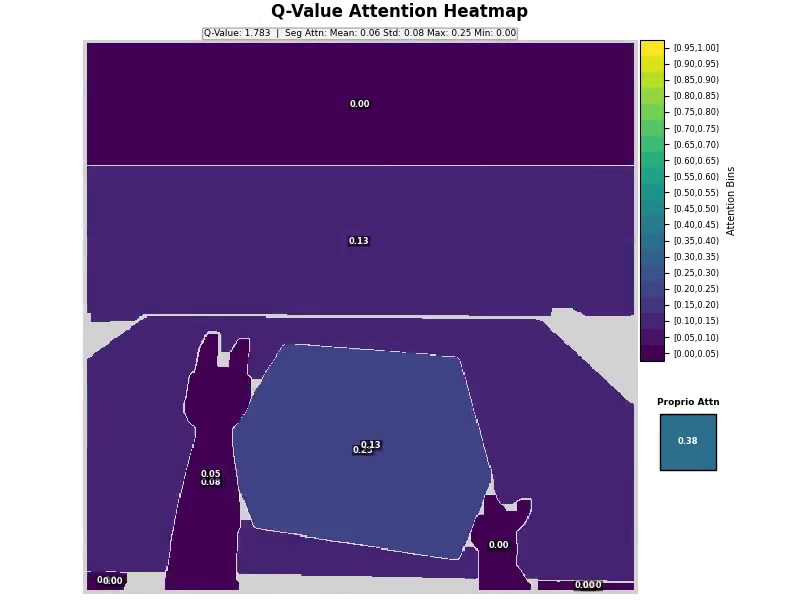}
    \includegraphics[width=0.32\linewidth]{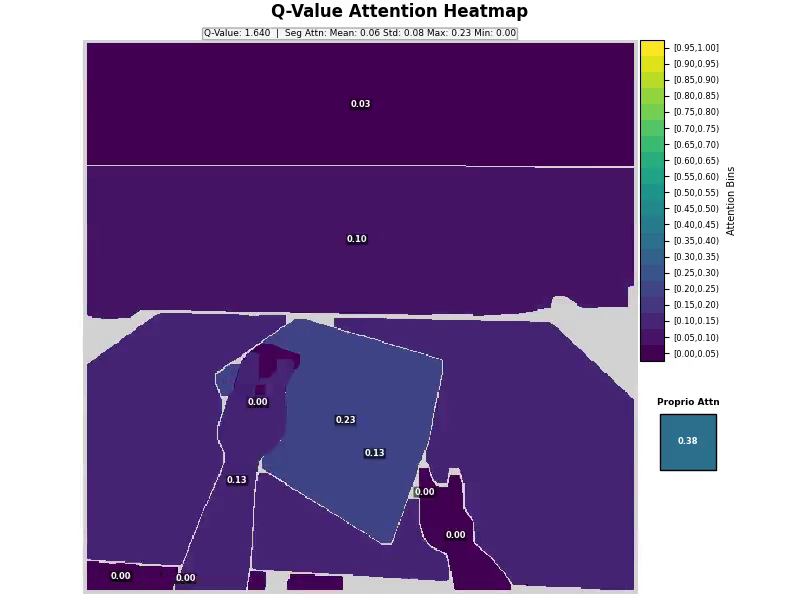}
    \caption{Average segment attention from the critic when predicting Q-values in the UnitreeG1TransportBox-v1 task. The trajectory begins in the top-left frame. Grey regions indicate areas where SAM did not detect any segments.}
    \label{fig:appendix_seg_attn_analysis_transport_box}
\end{figure}
\begin{figure}[H]
    \centering
    \includegraphics[width=0.24\linewidth]{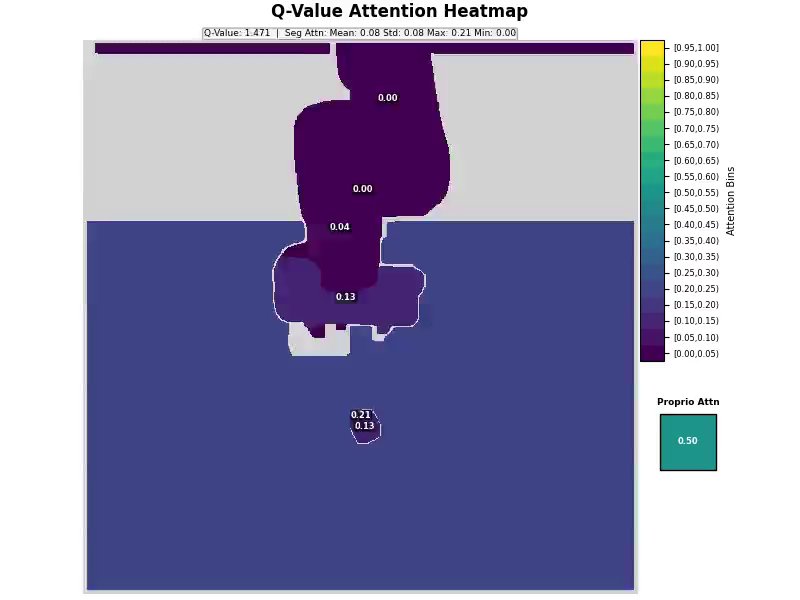}
    \includegraphics[width=0.24\linewidth]{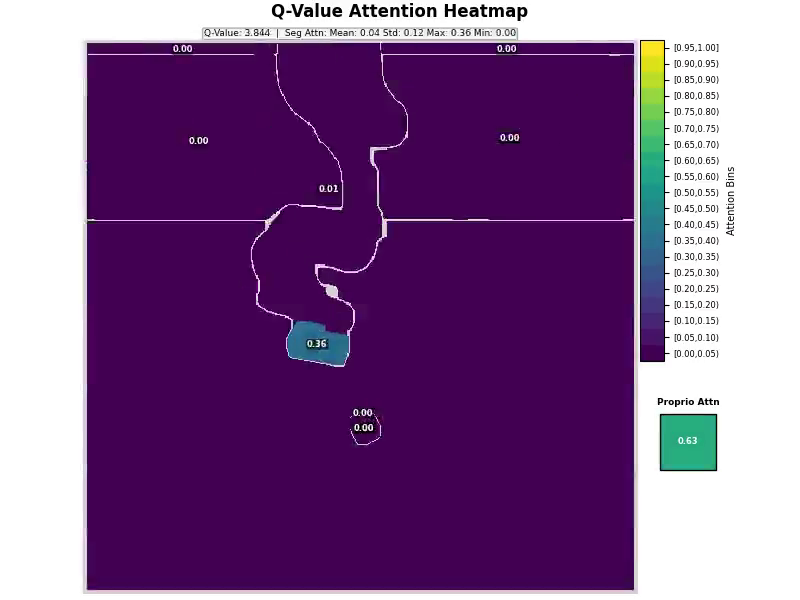}
    \includegraphics[width=0.24\linewidth]{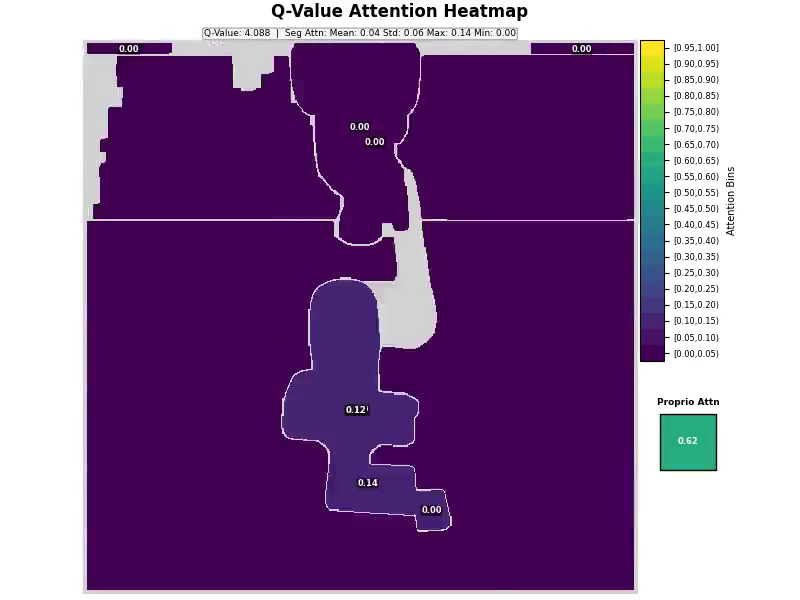}
    \includegraphics[width=0.24\linewidth]{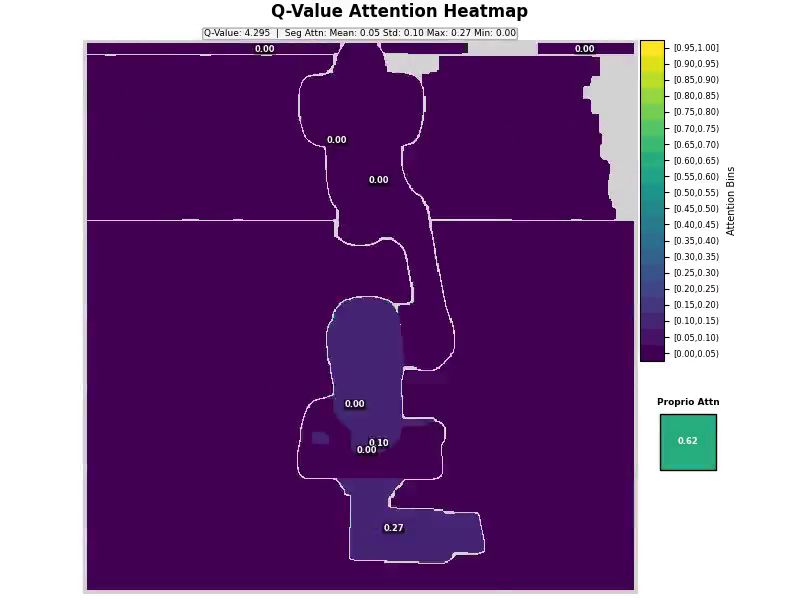}
    \includegraphics[width=0.24\linewidth]{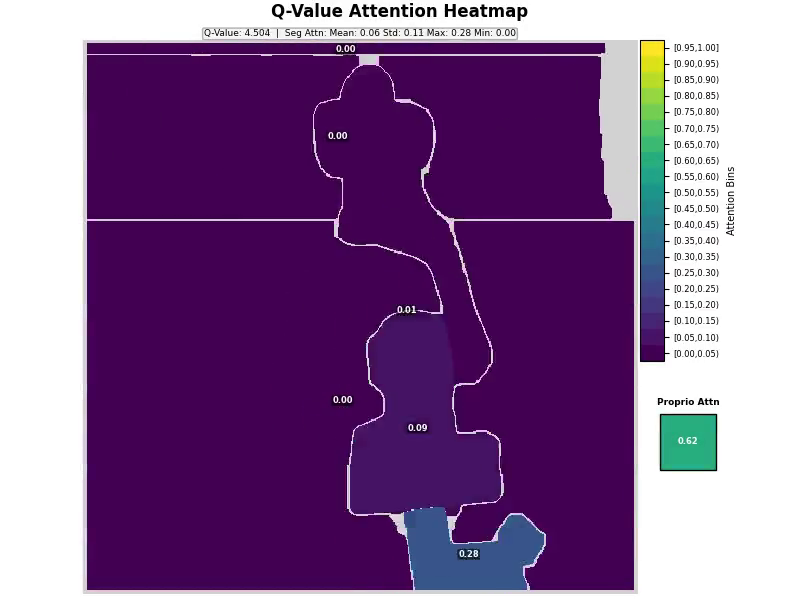}
    \includegraphics[width=0.24\linewidth]{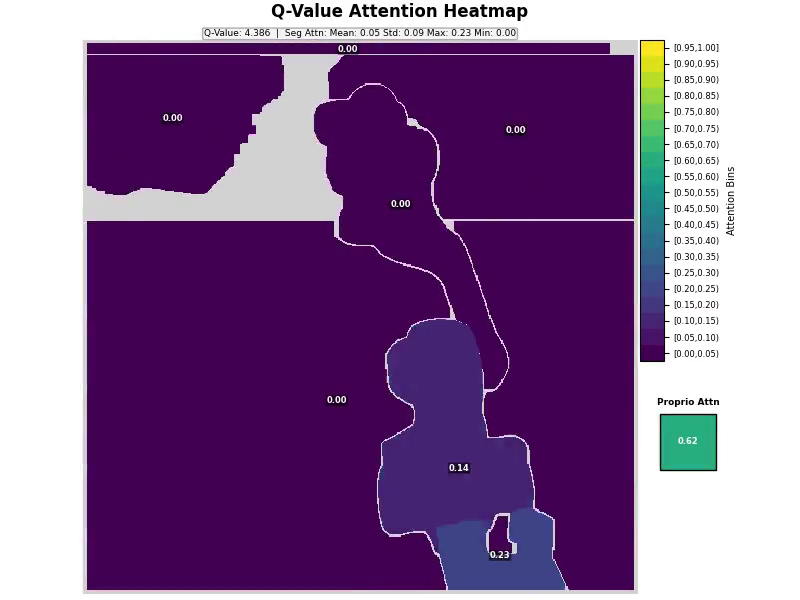}
    \includegraphics[width=0.24\linewidth]{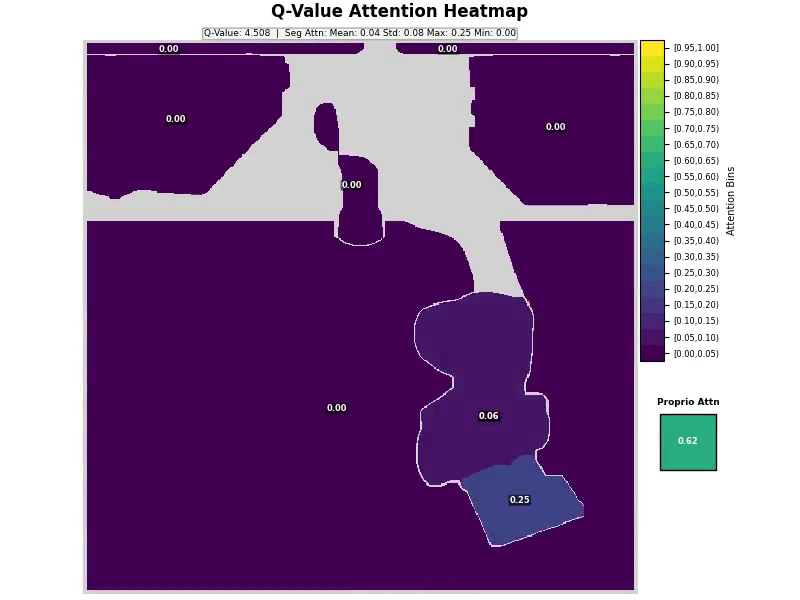}
    \includegraphics[width=0.24\linewidth]{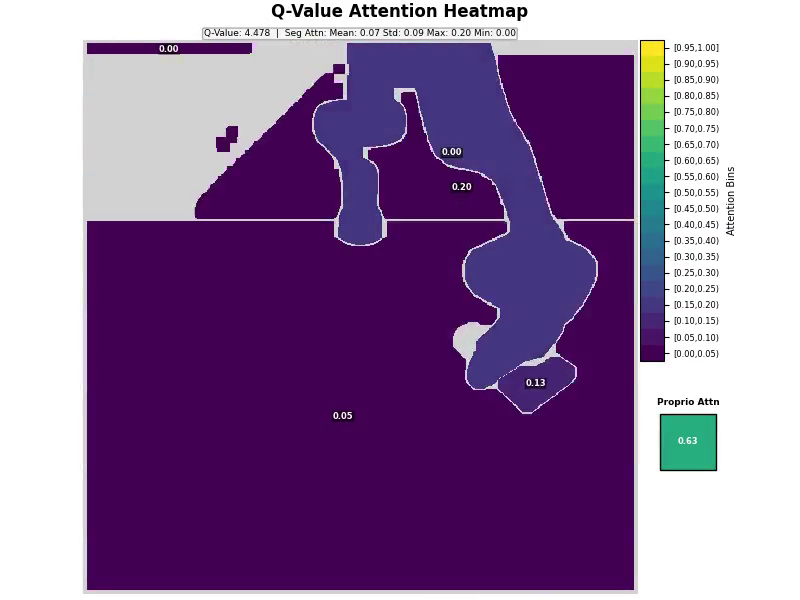}
    \caption{Average segment attention from the critic when predicting Q-values in the PullCubeTool-v1 task. The trajectory begins in the top-left frame. Grey regions indicate areas where SAM did not detect any segments.}
    \label{fig:appendix_seg_attn_analysis_pull_cube_tool}
\end{figure}

\Cref{fig:appendix_seg_attn_analysis_liftpegupright,fig:appendix_seg_attn_analysis_apple_in_bowl,fig:appendix_seg_attn_analysis_push_cube_2,fig:appendix_seg_attn_analysis_pick_cube,fig:appendix_seg_attn_analysis_pull_cube,fig:appendix_seg_attn_analysis_poke_cube,fig:appendix_seg_attn_analysis_transport_box,fig:appendix_seg_attn_analysis_pull_cube_tool} show the average segment attention from the critic when predicting Q-values for all eight tasks. Grey regions indicate areas where SAM did not detect any segments.

We observe that \methodName consistently attends to task-relevant segments, such as the peg in LiftPegUpright-v1, while ignoring background elements when they provide no useful information. For example, in LiftPegUpright-v1, the peg receives approximately 75\% of the attention early in the trajectory. Once the peg is upright, the model shifts some attention to proprioceptive inputs, possibly to stabilize the peg.

\methodName also demonstrates robustness to variability in detected segments. In PickCube-v1, early frames show the servo button segmented as a small circular region, which the critic uses to guide positioning. In later frames, when the full robot arm is detected instead, the critic adapts and reallocates attention accordingly to complete the task.

Across tasks, \methodName often allocates significant attention to the manipulation object. For instance, the apple receives about 50\% of the attention in UnitreeG1PlaceAppleInBowl-v1. Similarly, in LiftPegUpright-v1, the peg receives around 75\% early on.

Figure~\ref{fig:appendix_seg_attn_analysis_pick_cube} illustrates that \methodName is robust to partial occlusions. At the beginning of the trajectory, the goal region is fully occluded by the robot arm, and the model focuses its attention on the cube, which aligns with the sub-goal of grasping it first. As the cube is picked up and the arm moves, the goal region becomes partially visible. Even when it becomes occluded again during the motion, \methodName continues moving the arm toward the correct location. The goal region becomes visible near the end, and the model successfully completes the task. This shows that \methodName can handle some level of dynamic occlusions and maintain task-relevant behavior throughout.

\section{Implementation Details}
\label{appendix_impl_details}

\subsection{Policy Evaluation and Data Collection}
In this work, we use \textbf{evaluation} to refer to the policy’s performance under the \textit{no-perturbation} setting during training, which we use to track sample efficiency, samples used for evaluation are not seen during training. \textbf{Testing} refers to performance on the \textit{visual generalization benchmark}, which includes various visual perturbations. This distinction separates standard in-distribution evaluation from robustness testing under distribution shifts.

During both evaluation and testing, stochastic policies were run in deterministic mode by taking the mean of their action distributions. We followed the ManiSkill3 evaluation protocol\footnote{\url{https://maniskill.readthedocs.io/en/v3.0.0b21/user_guide/reinforcement_learning/setup.html\#evaluation}}, which disables early termination to ensure all trajectories have the same length. This makes performance metrics more comparable.

Evaluation metrics were averaged over 5 random seeds, each with 10 rollouts. Proper seeding was applied at the start of both training and evaluation scripts. We used 10 parallel environments to collect evaluation rollouts during training.

For training, we used 20 parallel GPU environments and performed 5 gradient updates per environment step with a batch size of 128, resulting in an update-to-data ratio of 0.25.

\subsection{Online RL Optimization}
Training online RL agents with large models such as SAM on high-resolution images (512x512) is computationally expensive. To complete experiments within one day on a single L40s GPU, we applied various key optimizations.

\paragraph{Prompt-Based Segmentation}
FTD relied on prompt-free segmentation, which we found to be significantly slower. In our tests, using bounding boxes as prompts for SAM yielded about a $10\times$ speedup compared to prompt-free mode. Prompt-based segmentation also provides semantic grounding, making it both faster and better aligned with our design.

\paragraph{Segmentation Post-Processing}
SAM-G applied iterative refinement, repeatedly invoking SAM to improve mask quality. While effective, this was too slow for online RL. We instead use a lightweight post-processing step that runs in about 1 ms on an L40s GPU. Further details are given in \Cref{appendix_morphological_mask_refinement}.

\paragraph{Efficient Vision Backbone.}
We replaced the original SAM model \citep{kirillov2023segment} with EfficientViT-SAM \citep{zhang2024efficientvitsamacceleratedsegmentmodel}, which offers comparable segmentation quality but significantly faster inference speed, approximately 48$\times$ faster according to the authors. We also tested SAM 2 \citep{ravi2024sam2segmentimages}, which improves encoder efficiency but remained significantly slower than EfficientViT-SAM in practice.

All RGB images were processed at a resolution of (3, 512, 512) in (C, H, W) format. This is significantly higher than standard visual RL settings, which often operate in the (3, 84, 84) to (3, 128, 128) range \citep{laskin2020reinforcementlearningaugmenteddata,kostrikov2021imageaugmentationneedregularizing,yarats2021masteringvisualcontinuouscontrol,grooten2023madilearningmaskdistractions}. We used the smallest checkpoint (efficientvit\_sam\_l0), which natively supports 512×512 images. Reducing the resolution further led to noticeable performance degradation with only marginal speedup, so we retained the native resolution.

We extended EfficientViT-SAM to support batched image embedding computation. This significantly improved throughput since image embedding is the most computationally expensive step. Most SAM variants, including the official implementations, do not support batching. Additionally, we tested prompt-free segmentation but found it to be too slow for our pipeline (10-100x slower), so we opted for guided segmentation using text inputs.

\paragraph{Efficient Network Computation.}
To optimize network computation, we used \texttt{torch.vmap} \citep{NEURIPS2019_bdbca288} to batch forward passes across multiple networks, such as the twin critics in SAC. This allowed us to compute their outputs in a single pass, improving efficiency.

\paragraph{Replay Buffer Design.}
We implemented a custom replay buffer using \texttt{TensorDict} \citep{bou2023torchrldatadrivendecisionmakinglibrary} to efficiently handle the variable number of segments per frame. Since the segment embedding extraction module has no trainable weights, we compute embeddings once during data collection and store them directly in the buffer rather than storing raw images.

This approach reduces both memory usage and compute load. Segment embeddings have dimension 128$\sim$256, and a typical frame has between 4 and 25 segments, this is significantly smaller than the original (3×512×512) image. This also avoids running SAM during training, which would otherwise make online RL infeasible.

\section{Sample Efficiency}
\label{sec:sample_eff_others}
\begin{figure}[h]
    \centering
    \includegraphics[width=1.0\linewidth]{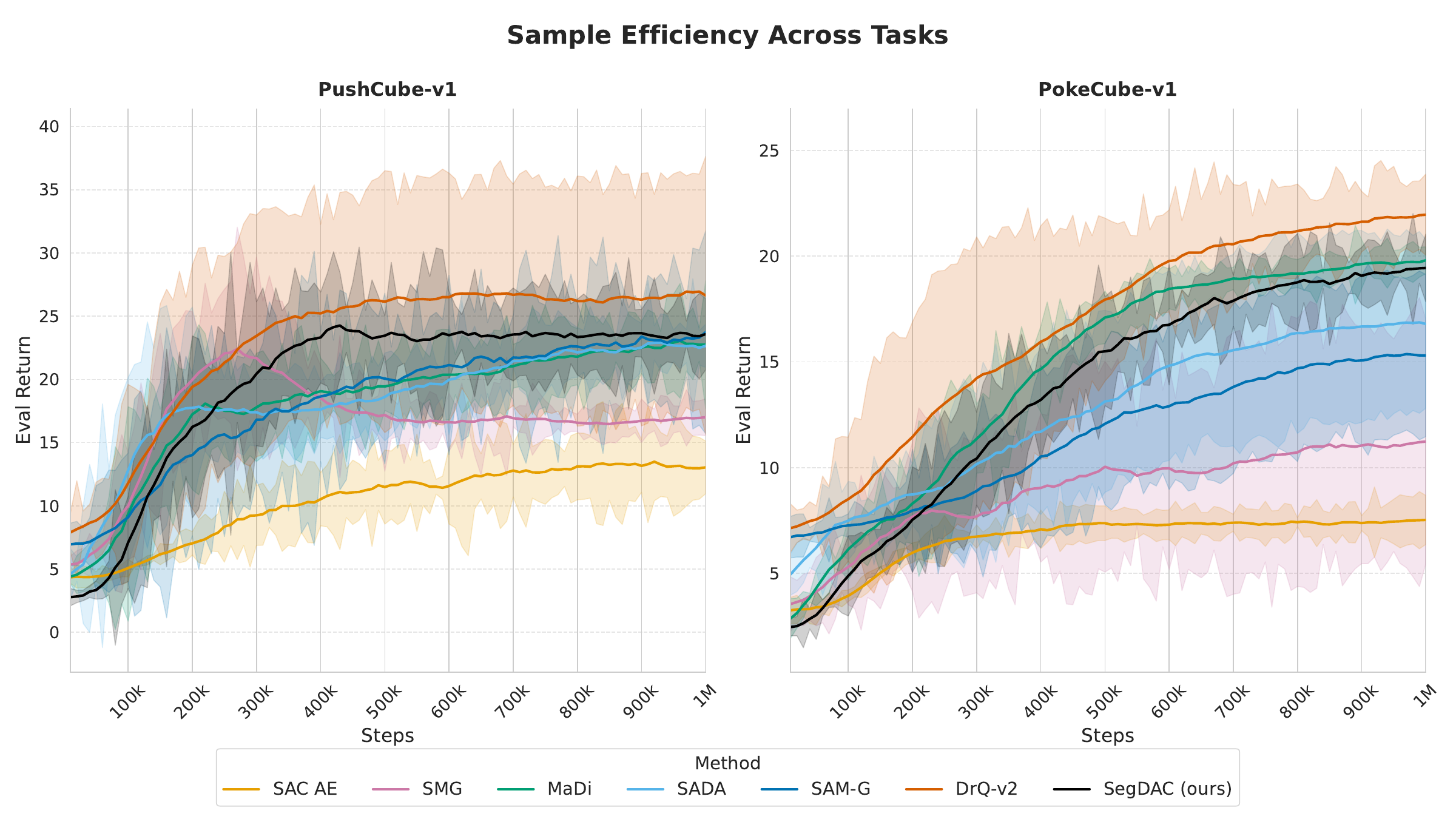}
    \caption{Evaluation return sample efficiency curves for the push/poke cube manipulation tasks.}
    \label{fig:exp_eval_return_2}
\end{figure} 
\begin{figure}[h]
    \centering
    \includegraphics[width=0.11\linewidth]{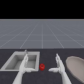}
    \includegraphics[width=0.11\linewidth]{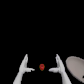}
    \includegraphics[width=0.11\linewidth]{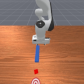}
    \includegraphics[width=0.11\linewidth]{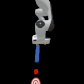}
    \includegraphics[width=0.11\linewidth]{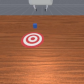}
    \includegraphics[width=0.11\linewidth]{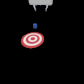}
    \includegraphics[width=0.11\linewidth]{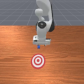}
    \includegraphics[width=0.11\linewidth]{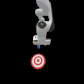}
    \caption{The reference frames and ground truth segmentation masks that SAM-G requires in order to operate (only some of the tasks are shown for brevity).}
    \label{fig:samg_g_labels}
\end{figure} 
Figure~\ref{fig:samg_g_labels} illustrates the reference frames and ground-truth masks required by SAM-G (only some of the tasks are shown for brevity but the same is true for all tasks), whereas \methodName does not rely on any ground-truth segmentation or reference frames. To produce these ground truth masks we used the first frame of a trajectory for each task then used SAM 3 playground platform and annotated each mask using 5-15 point prompts to guide SAM 3 until the segmentation mask was near perfect.

\section{Speed Analysis}
\label{sec:train_speed_analysis}
\begin{figure}[h]
    \centering
    \includegraphics[width=1.0\linewidth]{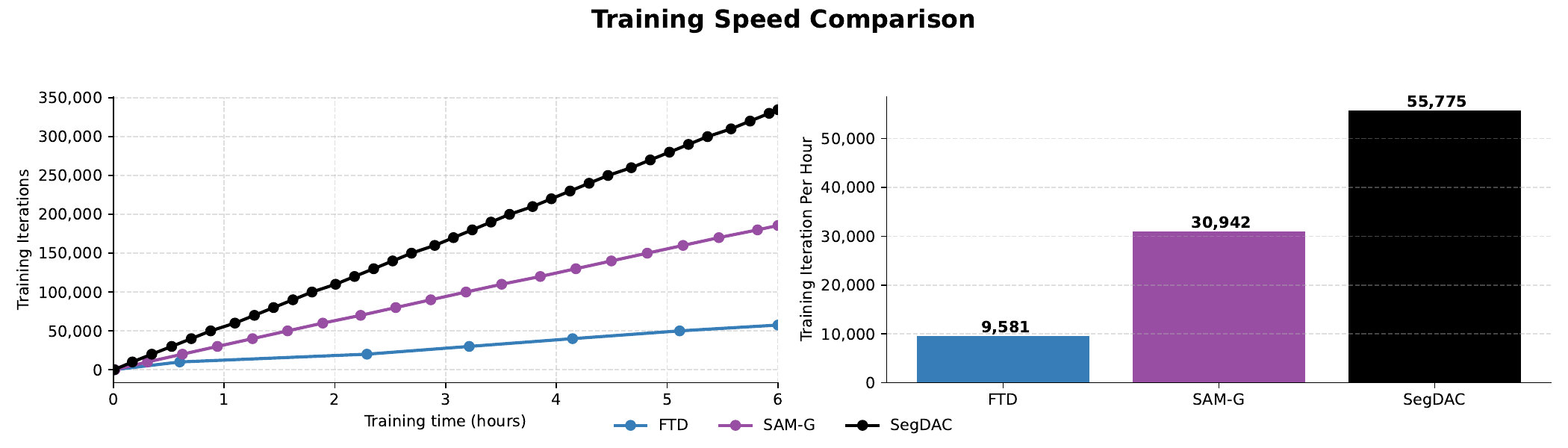}
    \caption{Training speed comparison of various methods that use SAM (includes env data collection + training).}
    \label{fig:train_iter_analysis_samg_ftd_segdac}
\end{figure} 
Figure~\ref{fig:train_iter_analysis_samg_ftd_segdac} shows the end to end training speed for methods that use SAM inside the RL loop. This includes data collection, SAM processing and all model updates (and periodic evaluation) . \methodName is noticeably faster than the previous SAM-based baselines even if we use a full transformer decoder instead of a light attention block on top of CNN features. FTD is slow because of its heavy auxiliary tasks and because it reconstructs RGB images from the segments. SAM-G also pays this cost since it learns from reconstructed images instead of working directly in latent space. FTD also relies on prompt-free segmentation which is much slower than text-guided segmentation.

\methodName keeps the pipeline simple and efficient. We use a short list of concept words for YOLO-World, and we reuse the SAM encoder features directly without any reconstruction stage. Our object-centric embeddings are extracted with a fast procedure and we store them in the replay buffer instead of RGB images. This avoids re-running SAM during training and keeps the RL loop light. The transformer then processes a small set of segment embeddings which is far cheaper than operating on hundreds of patch embeddings, and this helps address the quadratic cost of attention. Prior SAM-based methods also rely on heavy mask refinement loops to improve masks over time. We avoid this and instead use a lightweight morphological post-processing step which takes only a few milliseconds on an L40S.

These design choices make it possible to use a full transformer architecture and still achieve better end to end speed. In practice this reduces the overhead per step and leads to faster overall training. \methodName runs at almost twice the speed of SAM-G and FTD while keeping the full SAM perception stack.

\section{Frozen Weights Trade Off}
\label{sec:frozen_weights_trade_off}
We use pretrained frozen weights because full end to end updates would be too slow and potentially unstable for online RL, especially since SAM must run at every step. Freezing perception keeps the features stable and makes training practical, but the features are not guaranteed to match the RL objective and may limit performance in some cases. Full finetuning could improve representations, yet in practice it often reduces stability, lowers sample efficiency and can damage pretrained encoders, so it is not obvious that it is always better. Pretrained features are also not perfect and joint representation learning can sometimes help, but this adds a lot of complexity in our setup because we would need to store RGB images in the replay buffer instead of segment embeddings and re-run SAM during training, which would be far too slow for online RL. In our work we found that using the SAM encoder to compute segment embeddings gave representations that were strong enough to make RL training efficient even with frozen weights. Our visual generalization experiments also show that learning representations from scratch during online RL can easily overfit the training environment. Partial finetuning is a promising middle ground and we see it as a natural direction for future work.

\section{Visual Generalization Per Perturbation Results}
\label{appendix_vis_gen_tables}
\Cref{tab:appendix_camerafovtest_easy} and the subsequent tables report IQM returns and 95\% confidence intervals for each visual perturbation, across all tasks, methods, and difficulty levels. The method with the highest absolute IQM return is shown in \textbf{bold}, while the method with the best relative improvement over its no-perturbation baseline is \underline{underlined}. We also include the relative delta between the baseline (no perturbation) IQM return and the return under the perturbed setting. For example, a value of (-2.5\%) indicates a 2.5\% performance drop compared to the no perturbation case.

\begin{table}[htbp]
\centering
\scriptsize
\caption{Easy Camera Fov Test}
\label{tab:appendix_camerafovtest_easy}
\begin{adjustbox}{max width=\textwidth}
\begin{tabular}{l*{7}{c}}
\toprule
\textbf{Task} & \textbf{SAC AE} & \textbf{DrQ-v2} & \textbf{SAM-G} & \textbf{SMG} & \textbf{SADA} & \textbf{MaDi} & \textbf{SegDAC} \\
\midrule
LiftPegUpright & 0.20$\pm$0.00 \scriptsize{(-7.8\%)} & 0.23$\pm$0.09 \scriptsize{(-50.3\%)} & 0.25$\pm$0.02 \scriptsize{(-40.6\%)} & \underline{0.22$\pm$0.01 \scriptsize{(-0.1\%)}} & 0.22$\pm$0.01 \scriptsize{(-1.1\%)} & 0.22$\pm$0.01 \scriptsize{(-0.3\%)} & \textbf{0.27$\pm$0.08 \scriptsize{(-34.4\%)}} \\
PickCube & 0.07$\pm$0.01 \scriptsize{(-60.1\%)} & 0.12$\pm$0.04 \scriptsize{(-74.0\%)} & \underline{0.09$\pm$0.01 \scriptsize{(-11.5\%)}} & 0.17$\pm$0.02 \scriptsize{(-47.1\%)} & \textbf{0.26$\pm$0.08 \scriptsize{(-32.0\%)}} & 0.14$\pm$0.04 \scriptsize{(-68.9\%)} & 0.21$\pm$0.02 \scriptsize{(-36.5\%)} \\
PokeCube & 0.06$\pm$0.02 \scriptsize{(-60.8\%)} & 0.15$\pm$0.03 \scriptsize{(-66.2\%)} & 0.10$\pm$0.03 \scriptsize{(-67.7\%)} & \underline{0.12$\pm$0.04 \scriptsize{(-44.3\%)}} & 0.14$\pm$0.01 \scriptsize{(-63.4\%)} & 0.11$\pm$0.01 \scriptsize{(-72.9\%)} & \textbf{0.18$\pm$0.01 \scriptsize{(-55.0\%)}} \\
PullCube & 0.09$\pm$0.03 \scriptsize{(-76.5\%)} & 0.09$\pm$0.06 \scriptsize{(-76.0\%)} & \underline{0.14$\pm$0.01 \scriptsize{(+12.4\%)}} & 0.18$\pm$0.04 \scriptsize{(-54.0\%)} & \textbf{0.42$\pm$0.14 \scriptsize{(-13.6\%)}} & \textbf{0.42$\pm$0.17 \scriptsize{(-15.6\%)}} & 0.32$\pm$0.12 \scriptsize{(-35.7\%)} \\
PullCubeTool & 0.05$\pm$0.03 \scriptsize{(-76.8\%)} & 0.12$\pm$0.20 \scriptsize{(-80.6\%)} & 0.18$\pm$0.05 \scriptsize{(-18.8\%)} & \underline{0.22$\pm$0.09 \scriptsize{(-5.6\%)}} & 0.26$\pm$0.10 \scriptsize{(-28.3\%)} & 0.21$\pm$0.03 \scriptsize{(-65.9\%)} & \textbf{0.63$\pm$0.13 \scriptsize{(-14.2\%)}} \\
\bottomrule
\end{tabular}
\end{adjustbox}
\end{table}

\begin{table}[htbp]
\centering
\scriptsize
\caption{Easy Camera Pose Test}
\label{tab:appendix_cameraposetest_easy}
\begin{adjustbox}{max width=\textwidth}
\begin{tabular}{l*{7}{c}}
\toprule
\textbf{Task} & \textbf{SAC AE} & \textbf{DrQ-v2} & \textbf{SAM-G} & \textbf{SMG} & \textbf{SADA} & \textbf{MaDi} & \textbf{SegDAC} \\
\midrule
LiftPegUpright & 0.19$\pm$0.00 \scriptsize{(-11.9\%)} & 0.23$\pm$0.09 \scriptsize{(-50.5\%)} & \textbf{0.25$\pm$0.04 \scriptsize{(-39.9\%)}} & \underline{0.22$\pm$0.01 \scriptsize{(-0.8\%)}} & 0.22$\pm$0.04 \scriptsize{(-1.9\%)} & 0.20$\pm$0.02 \scriptsize{(-9.1\%)} & 0.19$\pm$0.00 \scriptsize{(-54.0\%)} \\
PickCube & 0.05$\pm$0.02 \scriptsize{(-74.9\%)} & 0.07$\pm$0.10 \scriptsize{(-85.2\%)} & \underline{0.08$\pm$0.01 \scriptsize{(-22.7\%)}} & 0.11$\pm$0.03 \scriptsize{(-64.5\%)} & 0.25$\pm$0.06 \scriptsize{(-34.5\%)} & \textbf{0.27$\pm$0.09 \scriptsize{(-39.5\%)}} & 0.12$\pm$0.05 \scriptsize{(-65.5\%)} \\
PokeCube & 0.03$\pm$0.01 \scriptsize{(-83.8\%)} & 0.02$\pm$0.01 \scriptsize{(-94.6\%)} & 0.12$\pm$0.01 \scriptsize{(-62.4\%)} & 0.11$\pm$0.02 \scriptsize{(-46.8\%)} & \textbf{\underline{0.22$\pm$0.04 \scriptsize{(-42.1\%)}}} & 0.16$\pm$0.01 \scriptsize{(-60.2\%)} & 0.13$\pm$0.02 \scriptsize{(-66.7\%)} \\
PullCube & 0.05$\pm$0.03 \scriptsize{(-86.1\%)} & 0.02$\pm$0.02 \scriptsize{(-94.2\%)} & \underline{0.13$\pm$0.01 \scriptsize{(+4.1\%)}} & 0.14$\pm$0.04 \scriptsize{(-62.9\%)} & 0.24$\pm$0.05 \scriptsize{(-50.1\%)} & 0.24$\pm$0.02 \scriptsize{(-51.2\%)} & \textbf{0.35$\pm$0.13 \scriptsize{(-30.4\%)}} \\
PullCubeTool & 0.02$\pm$0.02 \scriptsize{(-88.7\%)} & 0.15$\pm$0.07 \scriptsize{(-76.4\%)} & 0.17$\pm$0.04 \scriptsize{(-24.0\%)} & 0.17$\pm$0.06 \scriptsize{(-28.4\%)} & 0.22$\pm$0.04 \scriptsize{(-39.5\%)} & 0.20$\pm$0.05 \scriptsize{(-67.6\%)} & \textbf{\underline{0.59$\pm$0.18 \scriptsize{(-20.7\%)}}} \\
\bottomrule
\end{tabular}
\end{adjustbox}
\end{table}

\begin{table}[htbp]
\centering
\scriptsize
\caption{Easy Ground Color Test}
\label{tab:appendix_groundcolortest_easy}
\begin{adjustbox}{max width=\textwidth}
\begin{tabular}{l*{7}{c}}
\toprule
\textbf{Task} & \textbf{SAC AE} & \textbf{DrQ-v2} & \textbf{SAM-G} & \textbf{SMG} & \textbf{SADA} & \textbf{MaDi} & \textbf{SegDAC} \\
\midrule
LiftPegUpright & 0.21$\pm$0.00 \scriptsize{(-0.7\%)} & 0.36$\pm$0.12 \scriptsize{(-23.0\%)} & \textbf{\underline{0.43$\pm$0.15 \scriptsize{(+2.5\%)}}} & 0.22$\pm$0.00 \scriptsize{(+0.4\%)} & 0.22$\pm$0.03 \scriptsize{(+0.4\%)} & \underline{0.23$\pm$0.01 \scriptsize{(+2.5\%)}} & 0.39$\pm$0.15 \scriptsize{(-5.5\%)} \\
PickCube & 0.12$\pm$0.01 \scriptsize{(-33.7\%)} & 0.13$\pm$0.09 \scriptsize{(-72.5\%)} & 0.10$\pm$0.01 \scriptsize{(-5.4\%)} & \underline{0.33$\pm$0.07 \scriptsize{(+4.8\%)}} & 0.41$\pm$0.03 \scriptsize{(+4.7\%)} & \textbf{0.43$\pm$0.06 \scriptsize{(-2.4\%)}} & 0.35$\pm$0.04 \scriptsize{(+3.1\%)} \\
PokeCube & 0.16$\pm$0.02 \scriptsize{(-0.5\%)} & 0.20$\pm$0.12 \scriptsize{(-53.3\%)} & 0.31$\pm$0.05 \scriptsize{(-2.7\%)} & 0.21$\pm$0.14 \scriptsize{(-0.8\%)} & 0.37$\pm$0.07 \scriptsize{(-3.8\%)} & \textbf{\underline{0.41$\pm$0.01 \scriptsize{(-0.1\%)}}} & 0.34$\pm$0.06 \scriptsize{(-12.1\%)} \\
PullCube & 0.23$\pm$0.04 \scriptsize{(-38.7\%)} & 0.13$\pm$0.10 \scriptsize{(-63.6\%)} & \underline{0.13$\pm$0.01 \scriptsize{(+2.6\%)}} & 0.38$\pm$0.08 \scriptsize{(-3.2\%)} & 0.45$\pm$0.18 \scriptsize{(-8.5\%)} & 0.46$\pm$0.06 \scriptsize{(-6.8\%)} & \textbf{0.51$\pm$0.06 \scriptsize{(+1.4\%)}} \\
PullCubeTool & 0.19$\pm$0.05 \scriptsize{(-12.2\%)} & 0.36$\pm$0.35 \scriptsize{(-42.6\%)} & \underline{0.23$\pm$0.07 \scriptsize{(+5.6\%)}} & 0.23$\pm$0.10 \scriptsize{(+1.1\%)} & 0.36$\pm$0.14 \scriptsize{(-1.8\%)} & 0.61$\pm$0.20 \scriptsize{(-1.4\%)} & \textbf{0.69$\pm$0.14 \scriptsize{(-7.3\%)}} \\
PushCube & 0.21$\pm$0.03 \scriptsize{(-29.2\%)} & 0.32$\pm$0.16 \scriptsize{(-28.9\%)} & \textbf{0.48$\pm$0.09 \scriptsize{(-5.2\%)}} & 0.33$\pm$0.02 \scriptsize{(-2.8\%)} & \underline{0.46$\pm$0.02 \scriptsize{(-0.1\%)}} & 0.44$\pm$0.09 \scriptsize{(-4.1\%)} & 0.44$\pm$0.04 \scriptsize{(-1.8\%)} \\
\bottomrule
\end{tabular}
\end{adjustbox}
\end{table}

\begin{table}[htbp]
\centering
\scriptsize
\caption{Easy Ground Texture Test}
\label{tab:appendix_groundtexturetest_easy}
\begin{adjustbox}{max width=\textwidth}
\begin{tabular}{l*{7}{c}}
\toprule
\textbf{Task} & \textbf{SAC AE} & \textbf{DrQ-v2} & \textbf{SAM-G} & \textbf{SMG} & \textbf{SADA} & \textbf{MaDi} & \textbf{SegDAC} \\
\midrule
LiftPegUpright & 0.21$\pm$0.01 \scriptsize{(-1.3\%)} & \textbf{0.46$\pm$0.18 \scriptsize{(-2.8\%)}} & 0.41$\pm$0.14 \scriptsize{(-2.7\%)} & 0.22$\pm$0.00 \scriptsize{(+0.3\%)} & 0.23$\pm$0.03 \scriptsize{(+1.6\%)} & 0.22$\pm$0.01 \scriptsize{(+1.6\%)} & \underline{0.42$\pm$0.15 \scriptsize{(+1.9\%)}} \\
PickCube & 0.10$\pm$0.04 \scriptsize{(-45.5\%)} & \textbf{0.44$\pm$0.03 \scriptsize{(-4.8\%)}} & \underline{0.11$\pm$0.01 \scriptsize{(+6.4\%)}} & 0.32$\pm$0.07 \scriptsize{(+0.9\%)} & 0.40$\pm$0.03 \scriptsize{(+3.8\%)} & 0.43$\pm$0.06 \scriptsize{(-1.9\%)} & 0.35$\pm$0.05 \scriptsize{(+5.2\%)} \\
PokeCube & 0.15$\pm$0.02 \scriptsize{(-10.8\%)} & \textbf{0.43$\pm$0.06 \scriptsize{(-0.4\%)}} & \underline{0.32$\pm$0.06 \scriptsize{(+1.4\%)}} & 0.21$\pm$0.13 \scriptsize{(+0.3\%)} & 0.37$\pm$0.07 \scriptsize{(-3.9\%)} & 0.41$\pm$0.01 \scriptsize{(-1.1\%)} & 0.40$\pm$0.05 \scriptsize{(+0.8\%)} \\
PullCube & 0.25$\pm$0.03 \scriptsize{(-32.4\%)} & 0.37$\pm$0.05 \scriptsize{(-0.1\%)} & 0.12$\pm$0.01 \scriptsize{(-1.0\%)} & 0.35$\pm$0.05 \scriptsize{(-10.7\%)} & 0.42$\pm$0.15 \scriptsize{(-13.2\%)} & 0.48$\pm$0.03 \scriptsize{(-2.5\%)} & \textbf{\underline{0.53$\pm$0.06 \scriptsize{(+4.9\%)}}} \\
PullCubeTool & 0.19$\pm$0.06 \scriptsize{(-15.4\%)} & 0.62$\pm$0.17 \scriptsize{(-0.6\%)} & \underline{0.27$\pm$0.08 \scriptsize{(+20.3\%)}} & 0.23$\pm$0.10 \scriptsize{(-1.7\%)} & 0.34$\pm$0.17 \scriptsize{(-8.3\%)} & 0.61$\pm$0.21 \scriptsize{(-1.9\%)} & \textbf{0.74$\pm$0.16 \scriptsize{(-0.6\%)}} \\
PushCube & 0.21$\pm$0.03 \scriptsize{(-30.1\%)} & 0.34$\pm$0.22 \scriptsize{(-24.6\%)} & \textbf{\underline{0.56$\pm$0.09 \scriptsize{(+11.5\%)}}} & 0.34$\pm$0.03 \scriptsize{(-0.5\%)} & 0.45$\pm$0.01 \scriptsize{(-1.8\%)} & 0.46$\pm$0.07 \scriptsize{(+0.2\%)} & 0.46$\pm$0.04 \scriptsize{(+2.9\%)} \\
\bottomrule
\end{tabular}
\end{adjustbox}
\end{table}

\begin{table}[htbp]
\centering
\scriptsize
\caption{Easy Lighting Color Test}
\label{tab:appendix_lightingcolortest_easy}
\begin{adjustbox}{max width=\textwidth}
\begin{tabular}{l*{7}{c}}
\toprule
\textbf{Task} & \textbf{SAC AE} & \textbf{DrQ-v2} & \textbf{SAM-G} & \textbf{SMG} & \textbf{SADA} & \textbf{MaDi} & \textbf{SegDAC} \\
\midrule
LiftPegUpright & 0.21$\pm$0.01 \scriptsize{(-0.6\%)} & 0.21$\pm$0.02 \scriptsize{(-54.6\%)} & \textbf{0.42$\pm$0.15 \scriptsize{(-0.0\%)}} & 0.22$\pm$0.01 \scriptsize{(-0.5\%)} & 0.21$\pm$0.01 \scriptsize{(-4.2\%)} & \underline{0.22$\pm$0.02 \scriptsize{(+2.3\%)}} & 0.41$\pm$0.15 \scriptsize{(+0.2\%)} \\
PickCube & 0.16$\pm$0.04 \scriptsize{(-13.6\%)} & 0.07$\pm$0.04 \scriptsize{(-84.8\%)} & 0.10$\pm$0.01 \scriptsize{(-7.2\%)} & \underline{0.33$\pm$0.05 \scriptsize{(+5.4\%)}} & 0.21$\pm$0.07 \scriptsize{(-46.4\%)} & \textbf{0.43$\pm$0.06 \scriptsize{(-3.0\%)}} & 0.24$\pm$0.05 \scriptsize{(-28.9\%)} \\
PokeCube & 0.16$\pm$0.02 \scriptsize{(-5.7\%)} & 0.10$\pm$0.05 \scriptsize{(-76.9\%)} & 0.26$\pm$0.07 \scriptsize{(-16.6\%)} & \underline{0.22$\pm$0.12 \scriptsize{(+0.8\%)}} & 0.17$\pm$0.04 \scriptsize{(-55.1\%)} & \textbf{0.41$\pm$0.01 \scriptsize{(-0.8\%)}} & 0.36$\pm$0.03 \scriptsize{(-7.8\%)} \\
PullCube & 0.30$\pm$0.04 \scriptsize{(-19.7\%)} & 0.09$\pm$0.05 \scriptsize{(-76.1\%)} & 0.13$\pm$0.02 \scriptsize{(+0.3\%)} & \underline{0.40$\pm$0.07 \scriptsize{(+2.7\%)}} & 0.44$\pm$0.14 \scriptsize{(-9.0\%)} & 0.49$\pm$0.05 \scriptsize{(-1.0\%)} & \textbf{0.51$\pm$0.05 \scriptsize{(+0.8\%)}} \\
PullCubeTool & 0.24$\pm$0.07 \scriptsize{(+8.2\%)} & 0.09$\pm$0.07 \scriptsize{(-85.9\%)} & \underline{0.27$\pm$0.06 \scriptsize{(+22.4\%)}} & 0.23$\pm$0.11 \scriptsize{(+0.4\%)} & 0.19$\pm$0.10 \scriptsize{(-47.4\%)} & 0.61$\pm$0.19 \scriptsize{(-2.2\%)} & \textbf{0.72$\pm$0.12 \scriptsize{(-2.9\%)}} \\
PushCube & 0.26$\pm$0.02 \scriptsize{(-12.8\%)} & 0.16$\pm$0.06 \scriptsize{(-64.5\%)} & \textbf{0.49$\pm$0.13 \scriptsize{(-3.3\%)}} & 0.34$\pm$0.03 \scriptsize{(-1.7\%)} & 0.37$\pm$0.11 \scriptsize{(-18.8\%)} & 0.44$\pm$0.07 \scriptsize{(-4.7\%)} & \underline{0.48$\pm$0.02 \scriptsize{(+5.8\%)}} \\
\bottomrule
\end{tabular}
\end{adjustbox}
\end{table}

\begin{table}[htbp]
\centering
\scriptsize
\caption{Easy Lighting Direction Test}
\label{tab:appendix_lightingdirectiontest_easy}
\begin{adjustbox}{max width=\textwidth}
\begin{tabular}{l*{7}{c}}
\toprule
\textbf{Task} & \textbf{SAC AE} & \textbf{DrQ-v2} & \textbf{SAM-G} & \textbf{SMG} & \textbf{SADA} & \textbf{MaDi} & \textbf{SegDAC} \\
\midrule
LiftPegUpright & 0.21$\pm$0.01 \scriptsize{(-0.2\%)} & \textbf{0.46$\pm$0.18 \scriptsize{(-1.7\%)}} & 0.42$\pm$0.14 \scriptsize{(-0.7\%)} & 0.22$\pm$0.00 \scriptsize{(-0.4\%)} & 0.22$\pm$0.03 \scriptsize{(+0.9\%)} & \underline{0.23$\pm$0.02 \scriptsize{(+2.5\%)}} & 0.40$\pm$0.15 \scriptsize{(-2.6\%)} \\
PickCube & 0.17$\pm$0.03 \scriptsize{(-8.8\%)} & \textbf{0.42$\pm$0.04 \scriptsize{(-8.4\%)}} & 0.10$\pm$0.01 \scriptsize{(-7.6\%)} & \underline{0.32$\pm$0.05 \scriptsize{(+2.0\%)}} & 0.38$\pm$0.04 \scriptsize{(-1.0\%)} & 0.41$\pm$0.05 \scriptsize{(-6.0\%)} & 0.33$\pm$0.04 \scriptsize{(-2.9\%)} \\
PokeCube & 0.16$\pm$0.03 \scriptsize{(-1.8\%)} & 0.39$\pm$0.04 \scriptsize{(-10.5\%)} & 0.29$\pm$0.09 \scriptsize{(-8.5\%)} & 0.21$\pm$0.13 \scriptsize{(+0.0\%)} & 0.38$\pm$0.08 \scriptsize{(-1.9\%)} & \textbf{0.41$\pm$0.02 \scriptsize{(+0.8\%)}} & \underline{0.40$\pm$0.03 \scriptsize{(+2.9\%)}} \\
PullCube & 0.17$\pm$0.01 \scriptsize{(-54.7\%)} & 0.26$\pm$0.05 \scriptsize{(-30.3\%)} & 0.13$\pm$0.01 \scriptsize{(+2.5\%)} & 0.38$\pm$0.09 \scriptsize{(-1.6\%)} & \textbf{\underline{0.55$\pm$0.24 \scriptsize{(+12.0\%)}}} & 0.54$\pm$0.07 \scriptsize{(+9.6\%)} & 0.52$\pm$0.09 \scriptsize{(+4.0\%)} \\
PullCubeTool & 0.19$\pm$0.06 \scriptsize{(-11.0\%)} & 0.37$\pm$0.22 \scriptsize{(-40.4\%)} & \underline{0.22$\pm$0.08 \scriptsize{(-0.2\%)}} & 0.22$\pm$0.10 \scriptsize{(-5.2\%)} & 0.33$\pm$0.14 \scriptsize{(-9.4\%)} & 0.62$\pm$0.21 \scriptsize{(-1.2\%)} & \textbf{0.71$\pm$0.12 \scriptsize{(-4.6\%)}} \\
PushCube & 0.27$\pm$0.03 \scriptsize{(-7.7\%)} & 0.46$\pm$0.22 \scriptsize{(-0.1\%)} & \textbf{0.49$\pm$0.11 \scriptsize{(-2.8\%)}} & 0.35$\pm$0.03 \scriptsize{(+1.4\%)} & \underline{0.48$\pm$0.02 \scriptsize{(+4.7\%)}} & 0.43$\pm$0.07 \scriptsize{(-6.9\%)} & 0.45$\pm$0.04 \scriptsize{(+0.0\%)} \\
\bottomrule
\end{tabular}
\end{adjustbox}
\end{table}

\begin{table}[htbp]
\centering
\scriptsize
\caption{Easy Mo Color Test}
\label{tab:appendix_mocolortest_easy}
\begin{adjustbox}{max width=\textwidth}
\begin{tabular}{l*{7}{c}}
\toprule
\textbf{Task} & \textbf{SAC AE} & \textbf{DrQ-v2} & \textbf{SAM-G} & \textbf{SMG} & \textbf{SADA} & \textbf{MaDi} & \textbf{SegDAC} \\
\midrule
LiftPegUpright & 0.21$\pm$0.01 \scriptsize{(-1.3\%)} & 0.27$\pm$0.09 \scriptsize{(-42.1\%)} & 0.37$\pm$0.13 \scriptsize{(-12.4\%)} & 0.22$\pm$0.00 \scriptsize{(-0.5\%)} & 0.22$\pm$0.02 \scriptsize{(-0.8\%)} & \underline{0.22$\pm$0.01 \scriptsize{(-0.1\%)}} & \textbf{0.40$\pm$0.15 \scriptsize{(-2.5\%)}} \\
PickCube & 0.14$\pm$0.03 \scriptsize{(-22.0\%)} & 0.25$\pm$0.06 \scriptsize{(-45.0\%)} & \underline{0.11$\pm$0.01 \scriptsize{(+2.5\%)}} & 0.25$\pm$0.09 \scriptsize{(-19.8\%)} & 0.31$\pm$0.06 \scriptsize{(-20.1\%)} & \textbf{0.43$\pm$0.06 \scriptsize{(-2.9\%)}} & 0.28$\pm$0.06 \scriptsize{(-17.4\%)} \\
PokeCube & 0.13$\pm$0.02 \scriptsize{(-24.1\%)} & 0.17$\pm$0.01 \scriptsize{(-60.0\%)} & 0.16$\pm$0.02 \scriptsize{(-48.7\%)} & 0.19$\pm$0.12 \scriptsize{(-9.3\%)} & 0.30$\pm$0.05 \scriptsize{(-22.1\%)} & 0.39$\pm$0.01 \scriptsize{(-4.4\%)} & \textbf{\underline{0.40$\pm$0.03 \scriptsize{(+2.2\%)}}} \\
PullCube & 0.24$\pm$0.04 \scriptsize{(-34.2\%)} & 0.22$\pm$0.02 \scriptsize{(-39.7\%)} & 0.12$\pm$0.01 \scriptsize{(-4.7\%)} & 0.36$\pm$0.08 \scriptsize{(-7.4\%)} & 0.34$\pm$0.12 \scriptsize{(-31.3\%)} & 0.45$\pm$0.02 \scriptsize{(-7.6\%)} & \textbf{\underline{0.48$\pm$0.08 \scriptsize{(-3.7\%)}}} \\
PullCubeTool & 0.14$\pm$0.02 \scriptsize{(-38.2\%)} & 0.26$\pm$0.16 \scriptsize{(-58.9\%)} & 0.24$\pm$0.09 \scriptsize{(+6.2\%)} & 0.17$\pm$0.07 \scriptsize{(-24.7\%)} & 0.31$\pm$0.13 \scriptsize{(-14.2\%)} & 0.50$\pm$0.15 \scriptsize{(-19.1\%)} & \textbf{\underline{0.81$\pm$0.17 \scriptsize{(+9.8\%)}}} \\
PushCube & 0.20$\pm$0.03 \scriptsize{(-32.0\%)} & 0.23$\pm$0.16 \scriptsize{(-48.7\%)} & 0.30$\pm$0.07 \scriptsize{(-40.8\%)} & 0.26$\pm$0.04 \scriptsize{(-22.6\%)} & 0.33$\pm$0.13 \scriptsize{(-29.1\%)} & 0.43$\pm$0.07 \scriptsize{(-6.4\%)} & \textbf{\underline{0.45$\pm$0.05 \scriptsize{(-0.2\%)}}} \\
\bottomrule
\end{tabular}
\end{adjustbox}
\end{table}

\begin{table}[htbp]
\centering
\scriptsize
\caption{Easy Mo Texture Test}
\label{tab:appendix_motexturetest_easy}
\begin{adjustbox}{max width=\textwidth}
\begin{tabular}{l*{7}{c}}
\toprule
\textbf{Task} & \textbf{SAC AE} & \textbf{DrQ-v2} & \textbf{SAM-G} & \textbf{SMG} & \textbf{SADA} & \textbf{MaDi} & \textbf{SegDAC} \\
\midrule
LiftPegUpright & 0.20$\pm$0.01 \scriptsize{(-5.7\%)} & 0.22$\pm$0.02 \scriptsize{(-52.8\%)} & \textbf{0.31$\pm$0.06 \scriptsize{(-27.0\%)}} & \underline{0.22$\pm$0.01 \scriptsize{(+0.5\%)}} & 0.22$\pm$0.02 \scriptsize{(-2.6\%)} & 0.20$\pm$0.01 \scriptsize{(-6.8\%)} & \textbf{0.31$\pm$0.09 \scriptsize{(-23.8\%)}} \\
PickCube & 0.14$\pm$0.01 \scriptsize{(-22.9\%)} & \textbf{0.24$\pm$0.05 \scriptsize{(-48.0\%)}} & \underline{0.11$\pm$0.01 \scriptsize{(+2.9\%)}} & 0.06$\pm$0.07 \scriptsize{(-82.4\%)} & 0.17$\pm$0.05 \scriptsize{(-57.3\%)} & 0.23$\pm$0.08 \scriptsize{(-46.6\%)} & 0.16$\pm$0.02 \scriptsize{(-50.9\%)} \\
PokeCube & 0.12$\pm$0.01 \scriptsize{(-25.2\%)} & 0.17$\pm$0.04 \scriptsize{(-61.8\%)} & 0.17$\pm$0.03 \scriptsize{(-46.2\%)} & \underline{0.22$\pm$0.13 \scriptsize{(+4.9\%)}} & 0.22$\pm$0.03 \scriptsize{(-42.3\%)} & \textbf{0.38$\pm$0.03 \scriptsize{(-8.1\%)}} & 0.36$\pm$0.03 \scriptsize{(-8.2\%)} \\
PullCube & 0.25$\pm$0.03 \scriptsize{(-32.6\%)} & 0.27$\pm$0.05 \scriptsize{(-27.1\%)} & \underline{0.13$\pm$0.01 \scriptsize{(+1.3\%)}} & 0.35$\pm$0.05 \scriptsize{(-10.9\%)} & 0.37$\pm$0.02 \scriptsize{(-24.3\%)} & \textbf{0.47$\pm$0.05 \scriptsize{(-5.1\%)}} & 0.46$\pm$0.10 \scriptsize{(-8.5\%)} \\
PullCubeTool & 0.13$\pm$0.05 \scriptsize{(-42.0\%)} & 0.12$\pm$0.07 \scriptsize{(-80.5\%)} & \underline{0.23$\pm$0.10 \scriptsize{(+3.7\%)}} & 0.17$\pm$0.06 \scriptsize{(-27.5\%)} & 0.11$\pm$0.14 \scriptsize{(-68.9\%)} & 0.23$\pm$0.08 \scriptsize{(-63.7\%)} & \textbf{0.35$\pm$0.08 \scriptsize{(-53.1\%)}} \\
\bottomrule
\end{tabular}
\end{adjustbox}
\end{table}

\begin{table}[htbp]
\centering
\scriptsize
\caption{Easy Ro Color Test}
\label{tab:appendix_rocolortest_easy}
\begin{adjustbox}{max width=\textwidth}
\begin{tabular}{l*{7}{c}}
\toprule
\textbf{Task} & \textbf{SAC AE} & \textbf{DrQ-v2} & \textbf{SAM-G} & \textbf{SMG} & \textbf{SADA} & \textbf{MaDi} & \textbf{SegDAC} \\
\midrule
LiftPegUpright & N/A & N/A & N/A & N/A & N/A & N/A & N/A \\
PickCube & N/A & N/A & N/A & N/A & N/A & N/A & N/A \\
PokeCube & 0.17$\pm$0.03 \scriptsize{(+0.8\%)} & \textbf{0.42$\pm$0.05 \scriptsize{(-3.7\%)}} & 0.32$\pm$0.08 \scriptsize{(+0.9\%)} & \underline{0.22$\pm$0.13 \scriptsize{(+3.5\%)}} & 0.38$\pm$0.08 \scriptsize{(-1.2\%)} & 0.40$\pm$0.01 \scriptsize{(-2.3\%)} & 0.40$\pm$0.02 \scriptsize{(+1.8\%)} \\
PullCube & 0.34$\pm$0.06 \scriptsize{(-8.7\%)} & 0.29$\pm$0.08 \scriptsize{(-20.3\%)} & 0.12$\pm$0.01 \scriptsize{(-5.9\%)} & 0.39$\pm$0.08 \scriptsize{(+1.0\%)} & 0.41$\pm$0.11 \scriptsize{(-16.6\%)} & \underline{0.50$\pm$0.05 \scriptsize{(+1.1\%)}} & \textbf{0.51$\pm$0.09 \scriptsize{(+1.0\%)}} \\
PullCubeTool & 0.18$\pm$0.06 \scriptsize{(-15.4\%)} & 0.48$\pm$0.21 \scriptsize{(-22.4\%)} & 0.22$\pm$0.09 \scriptsize{(-2.9\%)} & \underline{0.23$\pm$0.10 \scriptsize{(-1.9\%)}} & 0.30$\pm$0.16 \scriptsize{(-18.6\%)} & 0.60$\pm$0.18 \scriptsize{(-3.5\%)} & \textbf{0.70$\pm$0.14 \scriptsize{(-5.9\%)}} \\
PushCube & 0.26$\pm$0.03 \scriptsize{(-13.1\%)} & 0.23$\pm$0.22 \scriptsize{(-48.8\%)} & \textbf{0.48$\pm$0.09 \scriptsize{(-4.5\%)}} & \underline{0.33$\pm$0.02 \scriptsize{(-2.3\%)}} & 0.43$\pm$0.04 \scriptsize{(-6.8\%)} & 0.43$\pm$0.07 \scriptsize{(-6.3\%)} & 0.44$\pm$0.04 \scriptsize{(-2.9\%)} \\
PlaceAppleInBowl & \underline{0.08$\pm$0.03 \scriptsize{(+6.9\%)}} & 0.11$\pm$0.02 \scriptsize{(-32.4\%)} & 0.07$\pm$0.02 \scriptsize{(+0.8\%)} & 0.10$\pm$0.05 \scriptsize{(-7.5\%)} & 0.12$\pm$0.04 \scriptsize{(+0.2\%)} & 0.12$\pm$0.05 \scriptsize{(-0.7\%)} & \textbf{0.29$\pm$0.14 \scriptsize{(-7.0\%)}} \\
TransportBox & N/A & N/A & N/A & N/A & N/A & N/A & N/A \\
\bottomrule
\end{tabular}
\end{adjustbox}
\end{table}

\begin{table}[htbp]
\centering
\scriptsize
\caption{Easy Ro Texture Test}
\label{tab:appendix_rotexturetest_easy}
\begin{adjustbox}{max width=\textwidth}
\begin{tabular}{l*{7}{c}}
\toprule
\textbf{Task} & \textbf{SAC AE} & \textbf{DrQ-v2} & \textbf{SAM-G} & \textbf{SMG} & \textbf{SADA} & \textbf{MaDi} & \textbf{SegDAC} \\
\midrule
LiftPegUpright & N/A & N/A & N/A & N/A & N/A & N/A & N/A \\
PickCube & N/A & N/A & N/A & N/A & N/A & N/A & N/A \\
PokeCube & 0.17$\pm$0.04 \scriptsize{(+0.9\%)} & 0.35$\pm$0.04 \scriptsize{(-19.5\%)} & 0.31$\pm$0.08 \scriptsize{(-2.3\%)} & \underline{0.22$\pm$0.13 \scriptsize{(+5.3\%)}} & 0.38$\pm$0.07 \scriptsize{(-2.1\%)} & \textbf{0.41$\pm$0.01 \scriptsize{(-0.0\%)}} & 0.40$\pm$0.04 \scriptsize{(+2.0\%)} \\
PullCube & 0.35$\pm$0.03 \scriptsize{(-6.0\%)} & 0.14$\pm$0.12 \scriptsize{(-63.0\%)} & 0.13$\pm$0.02 \scriptsize{(+2.1\%)} & 0.38$\pm$0.11 \scriptsize{(-3.5\%)} & \textbf{\underline{0.70$\pm$0.05 \scriptsize{(+43.4\%)}}} & 0.58$\pm$0.07 \scriptsize{(+17.4\%)} & 0.55$\pm$0.07 \scriptsize{(+9.1\%)} \\
PullCubeTool & 0.20$\pm$0.07 \scriptsize{(-6.8\%)} & 0.57$\pm$0.26 \scriptsize{(-8.2\%)} & \underline{0.24$\pm$0.07 \scriptsize{(+7.2\%)}} & 0.23$\pm$0.10 \scriptsize{(-1.1\%)} & 0.35$\pm$0.13 \scriptsize{(-3.6\%)} & 0.63$\pm$0.20 \scriptsize{(+1.2\%)} & \textbf{0.72$\pm$0.12 \scriptsize{(-2.3\%)}} \\
PushCube & 0.24$\pm$0.03 \scriptsize{(-19.4\%)} & 0.33$\pm$0.23 \scriptsize{(-28.4\%)} & 0.47$\pm$0.14 \scriptsize{(-6.9\%)} & 0.33$\pm$0.02 \scriptsize{(-3.8\%)} & \textbf{\underline{0.57$\pm$0.07 \scriptsize{(+22.9\%)}}} & 0.56$\pm$0.13 \scriptsize{(+19.8\%)} & 0.49$\pm$0.05 \scriptsize{(+10.0\%)} \\
PlaceAppleInBowl & 0.06$\pm$0.02 \scriptsize{(-20.5\%)} & 0.06$\pm$0.04 \scriptsize{(-63.9\%)} & \underline{0.06$\pm$0.01 \scriptsize{(-2.8\%)}} & 0.10$\pm$0.05 \scriptsize{(-6.5\%)} & 0.11$\pm$0.04 \scriptsize{(-3.2\%)} & 0.12$\pm$0.05 \scriptsize{(-3.2\%)} & \textbf{0.23$\pm$0.12 \scriptsize{(-25.7\%)}} \\
TransportBox & N/A & N/A & N/A & N/A & N/A & N/A & N/A \\
\bottomrule
\end{tabular}
\end{adjustbox}
\end{table}

\begin{table}[htbp]
\centering
\scriptsize
\caption{Easy Table Color Test}
\label{tab:appendix_tablecolortest_easy}
\begin{adjustbox}{max width=\textwidth}
\begin{tabular}{l*{7}{c}}
\toprule
\textbf{Task} & \textbf{SAC AE} & \textbf{DrQ-v2} & \textbf{SAM-G} & \textbf{SMG} & \textbf{SADA} & \textbf{MaDi} & \textbf{SegDAC} \\
\midrule
LiftPegUpright & 0.21$\pm$0.00 \scriptsize{(-2.3\%)} & 0.21$\pm$0.02 \scriptsize{(-55.3\%)} & \textbf{0.38$\pm$0.12 \scriptsize{(-9.2\%)}} & \underline{0.22$\pm$0.01 \scriptsize{(-0.6\%)}} & 0.22$\pm$0.03 \scriptsize{(-1.3\%)} & 0.21$\pm$0.01 \scriptsize{(-6.2\%)} & 0.28$\pm$0.08 \scriptsize{(-32.2\%)} \\
PickCube & 0.06$\pm$0.05 \scriptsize{(-68.1\%)} & 0.05$\pm$0.11 \scriptsize{(-88.7\%)} & \underline{0.11$\pm$0.02 \scriptsize{(+0.6\%)}} & 0.21$\pm$0.05 \scriptsize{(-31.7\%)} & 0.26$\pm$0.07 \scriptsize{(-32.7\%)} & \textbf{0.43$\pm$0.06 \scriptsize{(-2.1\%)}} & 0.19$\pm$0.04 \scriptsize{(-43.0\%)} \\
PokeCube & 0.14$\pm$0.02 \scriptsize{(-16.5\%)} & 0.30$\pm$0.14 \scriptsize{(-32.1\%)} & \underline{0.31$\pm$0.06 \scriptsize{(-1.5\%)}} & 0.19$\pm$0.10 \scriptsize{(-9.2\%)} & 0.28$\pm$0.07 \scriptsize{(-28.4\%)} & \textbf{0.40$\pm$0.01 \scriptsize{(-3.5\%)}} & 0.24$\pm$0.02 \scriptsize{(-38.6\%)} \\
PullCube & 0.14$\pm$0.07 \scriptsize{(-63.2\%)} & 0.30$\pm$0.22 \scriptsize{(-18.0\%)} & \underline{0.13$\pm$0.02 \scriptsize{(+2.3\%)}} & 0.25$\pm$0.10 \scriptsize{(-35.8\%)} & \textbf{0.42$\pm$0.03 \scriptsize{(-13.3\%)}} & 0.39$\pm$0.14 \scriptsize{(-21.6\%)} & \textbf{0.42$\pm$0.05 \scriptsize{(-15.7\%)}} \\
PullCubeTool & 0.19$\pm$0.06 \scriptsize{(-14.6\%)} & \textbf{0.61$\pm$0.17 \scriptsize{(-1.4\%)}} & \underline{0.28$\pm$0.09 \scriptsize{(+24.9\%)}} & 0.14$\pm$0.07 \scriptsize{(-40.8\%)} & 0.35$\pm$0.18 \scriptsize{(-5.1\%)} & 0.43$\pm$0.28 \scriptsize{(-31.6\%)} & 0.48$\pm$0.12 \scriptsize{(-35.4\%)} \\
\bottomrule
\end{tabular}
\end{adjustbox}
\end{table}

\begin{table}[htbp]
\centering
\scriptsize
\caption{Easy Table Texture Test}
\label{tab:appendix_tabletexturetest_easy}
\begin{adjustbox}{max width=\textwidth}
\begin{tabular}{l*{7}{c}}
\toprule
\textbf{Task} & \textbf{SAC AE} & \textbf{DrQ-v2} & \textbf{SAM-G} & \textbf{SMG} & \textbf{SADA} & \textbf{MaDi} & \textbf{SegDAC} \\
\midrule
LiftPegUpright & 0.21$\pm$0.00 \scriptsize{(+0.2\%)} & 0.40$\pm$0.19 \scriptsize{(-15.7\%)} & \textbf{0.41$\pm$0.14 \scriptsize{(-1.7\%)}} & 0.22$\pm$0.01 \scriptsize{(-0.3\%)} & 0.22$\pm$0.05 \scriptsize{(+0.6\%)} & \underline{0.23$\pm$0.01 \scriptsize{(+2.5\%)}} & 0.39$\pm$0.13 \scriptsize{(-5.4\%)} \\
PickCube & 0.16$\pm$0.07 \scriptsize{(-13.8\%)} & 0.33$\pm$0.15 \scriptsize{(-27.8\%)} & 0.10$\pm$0.00 \scriptsize{(-4.7\%)} & 0.29$\pm$0.07 \scriptsize{(-6.6\%)} & 0.35$\pm$0.07 \scriptsize{(-9.3\%)} & \textbf{0.42$\pm$0.05 \scriptsize{(-5.1\%)}} & \underline{0.34$\pm$0.03 \scriptsize{(+1.5\%)}} \\
PokeCube & 0.16$\pm$0.01 \scriptsize{(-5.3\%)} & 0.40$\pm$0.05 \scriptsize{(-6.9\%)} & \underline{0.32$\pm$0.08 \scriptsize{(+1.5\%)}} & 0.22$\pm$0.12 \scriptsize{(+1.0\%)} & 0.36$\pm$0.08 \scriptsize{(-6.9\%)} & \textbf{0.41$\pm$0.02 \scriptsize{(+0.1\%)}} & 0.38$\pm$0.01 \scriptsize{(-2.7\%)} \\
PullCube & 0.33$\pm$0.04 \scriptsize{(-10.9\%)} & 0.37$\pm$0.03 \scriptsize{(+1.1\%)} & 0.13$\pm$0.01 \scriptsize{(+4.3\%)} & \underline{0.41$\pm$0.13 \scriptsize{(+5.3\%)}} & \textbf{0.50$\pm$0.12 \scriptsize{(+2.3\%)}} & 0.49$\pm$0.05 \scriptsize{(+0.3\%)} & \textbf{0.50$\pm$0.08 \scriptsize{(-0.8\%)}} \\
PullCubeTool & 0.22$\pm$0.07 \scriptsize{(-1.4\%)} & 0.41$\pm$0.24 \scriptsize{(-33.8\%)} & \underline{0.26$\pm$0.07 \scriptsize{(+17.5\%)}} & 0.23$\pm$0.10 \scriptsize{(-2.4\%)} & 0.29$\pm$0.08 \scriptsize{(-20.8\%)} & 0.60$\pm$0.20 \scriptsize{(-4.0\%)} & \textbf{0.61$\pm$0.18 \scriptsize{(-17.2\%)}} \\
PushCube & 0.26$\pm$0.05 \scriptsize{(-11.7\%)} & 0.45$\pm$0.23 \scriptsize{(-2.0\%)} & \textbf{0.52$\pm$0.06 \scriptsize{(+2.8\%)}} & 0.33$\pm$0.01 \scriptsize{(-2.9\%)} & 0.45$\pm$0.09 \scriptsize{(-3.2\%)} & 0.45$\pm$0.06 \scriptsize{(-2.6\%)} & \underline{0.48$\pm$0.06 \scriptsize{(+6.9\%)}} \\
\bottomrule
\end{tabular}
\end{adjustbox}
\end{table}

\begin{table}[htbp]
\centering
\scriptsize
\caption{Medium Camera Fov Test}
\label{tab:appendix_camerafovtest_medium}
\begin{adjustbox}{max width=\textwidth}
\begin{tabular}{l*{7}{c}}
\toprule
\textbf{Task} & \textbf{SAC AE} & \textbf{DrQ-v2} & \textbf{SAM-G} & \textbf{SMG} & \textbf{SADA} & \textbf{MaDi} & \textbf{SegDAC} \\
\midrule
LiftPegUpright & 0.19$\pm$0.00 \scriptsize{(-11.7\%)} & 0.19$\pm$0.01 \scriptsize{(-59.3\%)} & 0.20$\pm$0.01 \scriptsize{(-53.6\%)} & \textbf{\underline{0.22$\pm$0.01 \scriptsize{(-1.7\%)}}} & 0.21$\pm$0.01 \scriptsize{(-3.6\%)} & 0.21$\pm$0.00 \scriptsize{(-5.5\%)} & \textbf{0.22$\pm$0.00 \scriptsize{(-47.6\%)}} \\
PickCube & 0.02$\pm$0.02 \scriptsize{(-88.9\%)} & 0.02$\pm$0.01 \scriptsize{(-96.7\%)} & 0.03$\pm$0.01 \scriptsize{(-74.9\%)} & 0.04$\pm$0.02 \scriptsize{(-86.4\%)} & \textbf{\underline{0.11$\pm$0.02 \scriptsize{(-71.7\%)}}} & 0.08$\pm$0.02 \scriptsize{(-82.8\%)} & 0.06$\pm$0.02 \scriptsize{(-81.7\%)} \\
PokeCube & 0.01$\pm$0.01 \scriptsize{(-92.3\%)} & 0.02$\pm$0.01 \scriptsize{(-95.8\%)} & 0.05$\pm$0.03 \scriptsize{(-83.7\%)} & \underline{0.08$\pm$0.03 \scriptsize{(-64.5\%)}} & \textbf{0.11$\pm$0.02 \scriptsize{(-71.6\%)}} & 0.09$\pm$0.02 \scriptsize{(-78.7\%)} & 0.09$\pm$0.01 \scriptsize{(-75.8\%)} \\
PullCube & 0.05$\pm$0.02 \scriptsize{(-86.9\%)} & 0.03$\pm$0.02 \scriptsize{(-92.8\%)} & \underline{0.14$\pm$0.02 \scriptsize{(+8.0\%)}} & 0.09$\pm$0.03 \scriptsize{(-76.2\%)} & \textbf{0.19$\pm$0.03 \scriptsize{(-60.1\%)}} & 0.15$\pm$0.03 \scriptsize{(-69.0\%)} & 0.18$\pm$0.02 \scriptsize{(-64.9\%)} \\
PullCubeTool & 0.02$\pm$0.01 \scriptsize{(-90.0\%)} & 0.03$\pm$0.05 \scriptsize{(-95.2\%)} & 0.06$\pm$0.02 \scriptsize{(-72.4\%)} & 0.09$\pm$0.06 \scriptsize{(-60.4\%)} & \underline{0.15$\pm$0.05 \scriptsize{(-58.7\%)}} & 0.10$\pm$0.02 \scriptsize{(-83.2\%)} & \textbf{0.30$\pm$0.10 \scriptsize{(-59.3\%)}} \\
\bottomrule
\end{tabular}
\end{adjustbox}
\end{table}

\begin{table}[htbp]
\centering
\scriptsize
\caption{Medium Camera Pose Test}
\label{tab:appendix_cameraposetest_medium}
\begin{adjustbox}{max width=\textwidth}
\begin{tabular}{l*{7}{c}}
\toprule
\textbf{Task} & \textbf{SAC AE} & \textbf{DrQ-v2} & \textbf{SAM-G} & \textbf{SMG} & \textbf{SADA} & \textbf{MaDi} & \textbf{SegDAC} \\
\midrule
LiftPegUpright & 0.19$\pm$0.01 \scriptsize{(-10.9\%)} & 0.19$\pm$0.01 \scriptsize{(-59.0\%)} & 0.19$\pm$0.00 \scriptsize{(-54.2\%)} & \underline{0.21$\pm$0.02 \scriptsize{(-3.8\%)}} & 0.19$\pm$0.01 \scriptsize{(-13.3\%)} & 0.20$\pm$0.02 \scriptsize{(-10.1\%)} & \textbf{0.23$\pm$0.01 \scriptsize{(-45.3\%)}} \\
PickCube & 0.03$\pm$0.01 \scriptsize{(-84.4\%)} & 0.03$\pm$0.01 \scriptsize{(-94.5\%)} & 0.04$\pm$0.03 \scriptsize{(-65.8\%)} & 0.06$\pm$0.02 \scriptsize{(-80.0\%)} & 0.11$\pm$0.02 \scriptsize{(-71.8\%)} & 0.07$\pm$0.02 \scriptsize{(-84.8\%)} & \textbf{\underline{0.12$\pm$0.04 \scriptsize{(-64.7\%)}}} \\
PokeCube & 0.02$\pm$0.02 \scriptsize{(-86.4\%)} & 0.02$\pm$0.03 \scriptsize{(-95.9\%)} & 0.02$\pm$0.03 \scriptsize{(-92.5\%)} & \underline{0.07$\pm$0.02 \scriptsize{(-67.5\%)}} & \textbf{0.11$\pm$0.02 \scriptsize{(-72.8\%)}} & 0.08$\pm$0.02 \scriptsize{(-79.6\%)} & 0.10$\pm$0.01 \scriptsize{(-74.6\%)} \\
PullCube & 0.04$\pm$0.01 \scriptsize{(-90.5\%)} & 0.02$\pm$0.02 \scriptsize{(-95.9\%)} & \underline{0.12$\pm$0.02 \scriptsize{(-7.6\%)}} & 0.08$\pm$0.04 \scriptsize{(-78.8\%)} & \textbf{0.16$\pm$0.04 \scriptsize{(-66.9\%)}} & 0.06$\pm$0.03 \scriptsize{(-87.5\%)} & 0.07$\pm$0.02 \scriptsize{(-85.4\%)} \\
PullCubeTool & 0.02$\pm$0.02 \scriptsize{(-91.1\%)} & 0.01$\pm$0.01 \scriptsize{(-99.1\%)} & \textbf{\underline{0.05$\pm$0.01 \scriptsize{(-77.9\%)}}} & 0.04$\pm$0.04 \scriptsize{(-82.7\%)} & 0.03$\pm$0.01 \scriptsize{(-93.1\%)} & 0.01$\pm$0.01 \scriptsize{(-98.8\%)} & 0.01$\pm$0.01 \scriptsize{(-98.4\%)} \\
\bottomrule
\end{tabular}
\end{adjustbox}
\end{table}

\begin{table}[htbp]
\centering
\scriptsize
\caption{Medium Ground Color Test}
\label{tab:appendix_groundcolortest_medium}
\begin{adjustbox}{max width=\textwidth}
\begin{tabular}{l*{7}{c}}
\toprule
\textbf{Task} & \textbf{SAC AE} & \textbf{DrQ-v2} & \textbf{SAM-G} & \textbf{SMG} & \textbf{SADA} & \textbf{MaDi} & \textbf{SegDAC} \\
\midrule
LiftPegUpright & 0.21$\pm$0.00 \scriptsize{(-1.1\%)} & 0.21$\pm$0.02 \scriptsize{(-55.3\%)} & 0.38$\pm$0.15 \scriptsize{(-10.4\%)} & \underline{0.22$\pm$0.00 \scriptsize{(+0.3\%)}} & 0.21$\pm$0.04 \scriptsize{(-3.0\%)} & 0.22$\pm$0.02 \scriptsize{(-0.9\%)} & \textbf{0.39$\pm$0.16 \scriptsize{(-5.1\%)}} \\
PickCube & 0.04$\pm$0.02 \scriptsize{(-80.1\%)} & 0.02$\pm$0.01 \scriptsize{(-96.7\%)} & \underline{0.11$\pm$0.01 \scriptsize{(+0.6\%)}} & 0.27$\pm$0.04 \scriptsize{(-13.4\%)} & 0.35$\pm$0.04 \scriptsize{(-8.4\%)} & \textbf{0.43$\pm$0.06 \scriptsize{(-1.5\%)}} & 0.33$\pm$0.05 \scriptsize{(-2.2\%)} \\
PokeCube & 0.12$\pm$0.05 \scriptsize{(-28.7\%)} & 0.09$\pm$0.07 \scriptsize{(-79.9\%)} & 0.29$\pm$0.08 \scriptsize{(-7.7\%)} & 0.21$\pm$0.13 \scriptsize{(-3.7\%)} & 0.32$\pm$0.08 \scriptsize{(-18.0\%)} & \textbf{\underline{0.41$\pm$0.02 \scriptsize{(-0.3\%)}}} & 0.38$\pm$0.03 \scriptsize{(-4.3\%)} \\
PullCube & 0.18$\pm$0.05 \scriptsize{(-50.9\%)} & 0.02$\pm$0.08 \scriptsize{(-95.7\%)} & 0.13$\pm$0.01 \scriptsize{(+1.9\%)} & 0.20$\pm$0.11 \scriptsize{(-49.6\%)} & 0.33$\pm$0.10 \scriptsize{(-31.7\%)} & 0.34$\pm$0.14 \scriptsize{(-30.1\%)} & \textbf{\underline{0.55$\pm$0.06 \scriptsize{(+10.5\%)}}} \\
PullCubeTool & 0.16$\pm$0.08 \scriptsize{(-25.4\%)} & 0.09$\pm$0.12 \scriptsize{(-85.4\%)} & \underline{0.26$\pm$0.06 \scriptsize{(+19.2\%)}} & 0.24$\pm$0.10 \scriptsize{(+1.9\%)} & 0.29$\pm$0.07 \scriptsize{(-20.5\%)} & 0.47$\pm$0.16 \scriptsize{(-24.6\%)} & \textbf{0.59$\pm$0.12 \scriptsize{(-20.8\%)}} \\
PushCube & 0.23$\pm$0.02 \scriptsize{(-23.4\%)} & 0.03$\pm$0.05 \scriptsize{(-92.4\%)} & \textbf{0.52$\pm$0.09 \scriptsize{(+3.7\%)}} & 0.33$\pm$0.04 \scriptsize{(-4.6\%)} & 0.38$\pm$0.11 \scriptsize{(-17.0\%)} & 0.43$\pm$0.09 \scriptsize{(-7.3\%)} & \underline{0.47$\pm$0.03 \scriptsize{(+4.0\%)}} \\
\bottomrule
\end{tabular}
\end{adjustbox}
\end{table}

\begin{table}[htbp]
\centering
\scriptsize
\caption{Medium Ground Texture Test}
\label{tab:appendix_groundtexturetest_medium}
\begin{adjustbox}{max width=\textwidth}
\begin{tabular}{l*{7}{c}}
\toprule
\textbf{Task} & \textbf{SAC AE} & \textbf{DrQ-v2} & \textbf{SAM-G} & \textbf{SMG} & \textbf{SADA} & \textbf{MaDi} & \textbf{SegDAC} \\
\midrule
LiftPegUpright & 0.21$\pm$0.01 \scriptsize{(-1.3\%)} & \textbf{0.42$\pm$0.18 \scriptsize{(-10.7\%)}} & 0.40$\pm$0.15 \scriptsize{(-5.7\%)} & 0.22$\pm$0.00 \scriptsize{(-0.1\%)} & \underline{0.22$\pm$0.04 \scriptsize{(+1.2\%)}} & 0.22$\pm$0.01 \scriptsize{(-0.1\%)} & 0.39$\pm$0.15 \scriptsize{(-4.2\%)} \\
PickCube & 0.07$\pm$0.01 \scriptsize{(-59.5\%)} & 0.39$\pm$0.03 \scriptsize{(-15.8\%)} & 0.10$\pm$0.01 \scriptsize{(-6.0\%)} & 0.28$\pm$0.06 \scriptsize{(-9.6\%)} & 0.36$\pm$0.04 \scriptsize{(-6.7\%)} & \textbf{\underline{0.43$\pm$0.06 \scriptsize{(-1.6\%)}}} & 0.33$\pm$0.03 \scriptsize{(-3.1\%)} \\
PokeCube & 0.13$\pm$0.02 \scriptsize{(-22.4\%)} & \textbf{0.40$\pm$0.09 \scriptsize{(-7.7\%)}} & 0.32$\pm$0.08 \scriptsize{(+2.1\%)} & \underline{0.22$\pm$0.13 \scriptsize{(+3.7\%)}} & 0.25$\pm$0.09 \scriptsize{(-36.3\%)} & \textbf{0.40$\pm$0.01 \scriptsize{(-2.2\%)}} & 0.39$\pm$0.02 \scriptsize{(-1.5\%)} \\
PullCube & 0.15$\pm$0.04 \scriptsize{(-58.8\%)} & 0.31$\pm$0.02 \scriptsize{(-16.0\%)} & \underline{0.13$\pm$0.02 \scriptsize{(+2.8\%)}} & 0.19$\pm$0.10 \scriptsize{(-51.6\%)} & 0.24$\pm$0.04 \scriptsize{(-50.6\%)} & 0.26$\pm$0.06 \scriptsize{(-46.6\%)} & \textbf{0.51$\pm$0.06 \scriptsize{(+1.6\%)}} \\
PullCubeTool & 0.19$\pm$0.04 \scriptsize{(-13.7\%)} & 0.56$\pm$0.13 \scriptsize{(-9.6\%)} & \underline{0.24$\pm$0.10 \scriptsize{(+7.0\%)}} & 0.23$\pm$0.10 \scriptsize{(-0.2\%)} & 0.30$\pm$0.15 \scriptsize{(-16.8\%)} & 0.58$\pm$0.19 \scriptsize{(-7.0\%)} & \textbf{0.72$\pm$0.14 \scriptsize{(-3.3\%)}} \\
PushCube & 0.14$\pm$0.03 \scriptsize{(-53.3\%)} & 0.43$\pm$0.22 \scriptsize{(-6.7\%)} & \textbf{\underline{0.55$\pm$0.09 \scriptsize{(+9.3\%)}}} & 0.32$\pm$0.04 \scriptsize{(-7.2\%)} & 0.33$\pm$0.08 \scriptsize{(-27.7\%)} & 0.44$\pm$0.08 \scriptsize{(-4.4\%)} & 0.49$\pm$0.06 \scriptsize{(+9.1\%)} \\
\bottomrule
\end{tabular}
\end{adjustbox}
\end{table}

\begin{table}[htbp]
\centering
\scriptsize
\caption{Medium Lighting Color Test}
\label{tab:appendix_lightingcolortest_medium}
\begin{adjustbox}{max width=\textwidth}
\begin{tabular}{l*{7}{c}}
\toprule
\textbf{Task} & \textbf{SAC AE} & \textbf{DrQ-v2} & \textbf{SAM-G} & \textbf{SMG} & \textbf{SADA} & \textbf{MaDi} & \textbf{SegDAC} \\
\midrule
LiftPegUpright & 0.21$\pm$0.00 \scriptsize{(-2.0\%)} & 0.19$\pm$0.00 \scriptsize{(-60.2\%)} & 0.22$\pm$0.06 \scriptsize{(-47.2\%)} & 0.22$\pm$0.00 \scriptsize{(+0.5\%)} & 0.23$\pm$0.03 \scriptsize{(+2.5\%)} & 0.22$\pm$0.01 \scriptsize{(-0.6\%)} & \textbf{\underline{0.43$\pm$0.16 \scriptsize{(+4.3\%)}}} \\
PickCube & 0.05$\pm$0.01 \scriptsize{(-70.7\%)} & 0.01$\pm$0.01 \scriptsize{(-98.3\%)} & 0.06$\pm$0.02 \scriptsize{(-46.8\%)} & \underline{0.18$\pm$0.05 \scriptsize{(-43.7\%)}} & 0.10$\pm$0.05 \scriptsize{(-73.0\%)} & \textbf{0.19$\pm$0.02 \scriptsize{(-56.1\%)}} & 0.16$\pm$0.02 \scriptsize{(-51.9\%)} \\
PokeCube & 0.06$\pm$0.04 \scriptsize{(-60.8\%)} & 0.06$\pm$0.05 \scriptsize{(-85.8\%)} & 0.09$\pm$0.03 \scriptsize{(-70.9\%)} & 0.19$\pm$0.11 \scriptsize{(-9.1\%)} & 0.18$\pm$0.07 \scriptsize{(-53.1\%)} & \textbf{\underline{0.38$\pm$0.03 \scriptsize{(-8.3\%)}}} & 0.30$\pm$0.03 \scriptsize{(-24.1\%)} \\
PullCube & 0.20$\pm$0.06 \scriptsize{(-47.0\%)} & 0.05$\pm$0.03 \scriptsize{(-86.8\%)} & \underline{0.13$\pm$0.02 \scriptsize{(+3.6\%)}} & 0.33$\pm$0.04 \scriptsize{(-14.9\%)} & 0.29$\pm$0.11 \scriptsize{(-41.4\%)} & 0.37$\pm$0.05 \scriptsize{(-24.0\%)} & \textbf{0.49$\pm$0.08 \scriptsize{(-1.9\%)}} \\
PullCubeTool & 0.12$\pm$0.04 \scriptsize{(-45.6\%)} & 0.06$\pm$0.03 \scriptsize{(-90.4\%)} & 0.19$\pm$0.03 \scriptsize{(-15.6\%)} & 0.18$\pm$0.08 \scriptsize{(-20.4\%)} & \underline{0.32$\pm$0.15 \scriptsize{(-12.5\%)}} & 0.15$\pm$0.12 \scriptsize{(-76.3\%)} & \textbf{0.55$\pm$0.13 \scriptsize{(-26.0\%)}} \\
PushCube & 0.18$\pm$0.03 \scriptsize{(-39.7\%)} & 0.03$\pm$0.03 \scriptsize{(-92.9\%)} & 0.08$\pm$0.01 \scriptsize{(-84.2\%)} & 0.26$\pm$0.04 \scriptsize{(-22.9\%)} & 0.32$\pm$0.05 \scriptsize{(-31.0\%)} & 0.48$\pm$0.10 \scriptsize{(+2.9\%)} & \textbf{\underline{0.53$\pm$0.06 \scriptsize{(+18.1\%)}}} \\
\bottomrule
\end{tabular}
\end{adjustbox}
\end{table}

\begin{table}[htbp]
\centering
\scriptsize
\caption{Medium Lighting Direction Test}
\label{tab:appendix_lightingdirectiontest_medium}
\begin{adjustbox}{max width=\textwidth}
\begin{tabular}{l*{7}{c}}
\toprule
\textbf{Task} & \textbf{SAC AE} & \textbf{DrQ-v2} & \textbf{SAM-G} & \textbf{SMG} & \textbf{SADA} & \textbf{MaDi} & \textbf{SegDAC} \\
\midrule
LiftPegUpright & 0.21$\pm$0.01 \scriptsize{(-0.6\%)} & 0.23$\pm$0.03 \scriptsize{(-51.0\%)} & 0.40$\pm$0.15 \scriptsize{(-4.1\%)} & 0.22$\pm$0.00 \scriptsize{(-0.1\%)} & 0.22$\pm$0.04 \scriptsize{(-0.6\%)} & 0.21$\pm$0.01 \scriptsize{(-2.4\%)} & \textbf{\underline{0.42$\pm$0.16 \scriptsize{(+1.3\%)}}} \\
PickCube & 0.16$\pm$0.03 \scriptsize{(-14.0\%)} & 0.36$\pm$0.10 \scriptsize{(-21.2\%)} & \underline{0.11$\pm$0.01 \scriptsize{(+6.4\%)}} & 0.29$\pm$0.06 \scriptsize{(-6.7\%)} & 0.35$\pm$0.05 \scriptsize{(-8.7\%)} & \textbf{0.43$\pm$0.05 \scriptsize{(-2.7\%)}} & 0.24$\pm$0.04 \scriptsize{(-28.2\%)} \\
PokeCube & 0.15$\pm$0.01 \scriptsize{(-7.2\%)} & 0.30$\pm$0.09 \scriptsize{(-30.0\%)} & 0.27$\pm$0.08 \scriptsize{(-12.8\%)} & \underline{0.23$\pm$0.13 \scriptsize{(+7.0\%)}} & 0.36$\pm$0.07 \scriptsize{(-7.3\%)} & \textbf{0.41$\pm$0.02 \scriptsize{(-0.3\%)}} & 0.36$\pm$0.03 \scriptsize{(-7.0\%)} \\
PullCube & 0.28$\pm$0.01 \scriptsize{(-24.2\%)} & 0.34$\pm$0.10 \scriptsize{(-6.8\%)} & \underline{0.14$\pm$0.02 \scriptsize{(+9.4\%)}} & 0.39$\pm$0.07 \scriptsize{(+0.4\%)} & 0.49$\pm$0.07 \scriptsize{(+1.4\%)} & 0.49$\pm$0.03 \scriptsize{(-1.3\%)} & \textbf{0.50$\pm$0.07 \scriptsize{(+0.5\%)}} \\
PullCubeTool & 0.22$\pm$0.06 \scriptsize{(+1.4\%)} & 0.47$\pm$0.20 \scriptsize{(-25.1\%)} & \underline{0.23$\pm$0.05 \scriptsize{(+3.2\%)}} & 0.22$\pm$0.10 \scriptsize{(-2.9\%)} & 0.30$\pm$0.10 \scriptsize{(-18.7\%)} & 0.58$\pm$0.20 \scriptsize{(-6.3\%)} & \textbf{0.71$\pm$0.12 \scriptsize{(-4.0\%)}} \\
PushCube & 0.25$\pm$0.04 \scriptsize{(-16.8\%)} & \textbf{\underline{0.51$\pm$0.24 \scriptsize{(+12.7\%)}}} & 0.49$\pm$0.10 \scriptsize{(-2.4\%)} & 0.32$\pm$0.04 \scriptsize{(-6.9\%)} & 0.46$\pm$0.05 \scriptsize{(-1.3\%)} & 0.46$\pm$0.10 \scriptsize{(-1.1\%)} & 0.48$\pm$0.05 \scriptsize{(+6.8\%)} \\
\bottomrule
\end{tabular}
\end{adjustbox}
\end{table}

\begin{table}[htbp]
\centering
\scriptsize
\caption{Medium Mo Color Test}
\label{tab:appendix_mocolortest_medium}
\begin{adjustbox}{max width=\textwidth}
\begin{tabular}{l*{7}{c}}
\toprule
\textbf{Task} & \textbf{SAC AE} & \textbf{DrQ-v2} & \textbf{SAM-G} & \textbf{SMG} & \textbf{SADA} & \textbf{MaDi} & \textbf{SegDAC} \\
\midrule
LiftPegUpright & 0.20$\pm$0.01 \scriptsize{(-5.4\%)} & 0.23$\pm$0.04 \scriptsize{(-51.3\%)} & 0.30$\pm$0.09 \scriptsize{(-29.4\%)} & \underline{0.22$\pm$0.01 \scriptsize{(+0.7\%)}} & 0.22$\pm$0.01 \scriptsize{(-0.2\%)} & 0.21$\pm$0.01 \scriptsize{(-4.8\%)} & \textbf{0.41$\pm$0.15 \scriptsize{(-0.7\%)}} \\
PickCube & 0.11$\pm$0.03 \scriptsize{(-37.7\%)} & 0.21$\pm$0.12 \scriptsize{(-53.3\%)} & \underline{0.11$\pm$0.02 \scriptsize{(+6.1\%)}} & 0.10$\pm$0.06 \scriptsize{(-68.3\%)} & 0.22$\pm$0.09 \scriptsize{(-43.6\%)} & 0.02$\pm$0.01 \scriptsize{(-96.3\%)} & \textbf{0.27$\pm$0.05 \scriptsize{(-18.8\%)}} \\
PokeCube & 0.07$\pm$0.01 \scriptsize{(-55.0\%)} & 0.02$\pm$0.02 \scriptsize{(-96.1\%)} & 0.18$\pm$0.03 \scriptsize{(-41.8\%)} & 0.19$\pm$0.10 \scriptsize{(-11.5\%)} & 0.17$\pm$0.03 \scriptsize{(-55.4\%)} & 0.07$\pm$0.15 \scriptsize{(-83.2\%)} & \textbf{\underline{0.40$\pm$0.04 \scriptsize{(+1.5\%)}}} \\
PullCube & 0.15$\pm$0.05 \scriptsize{(-60.0\%)} & 0.08$\pm$0.02 \scriptsize{(-77.8\%)} & \underline{0.13$\pm$0.02 \scriptsize{(+2.5\%)}} & 0.30$\pm$0.07 \scriptsize{(-22.0\%)} & 0.46$\pm$0.12 \scriptsize{(-5.4\%)} & 0.10$\pm$0.06 \scriptsize{(-79.1\%)} & \textbf{0.47$\pm$0.07 \scriptsize{(-6.4\%)}} \\
PullCubeTool & 0.09$\pm$0.02 \scriptsize{(-59.0\%)} & 0.03$\pm$0.04 \scriptsize{(-94.8\%)} & \underline{0.18$\pm$0.04 \scriptsize{(-17.7\%)}} & 0.18$\pm$0.08 \scriptsize{(-21.6\%)} & 0.25$\pm$0.17 \scriptsize{(-30.6\%)} & 0.12$\pm$0.05 \scriptsize{(-80.1\%)} & \textbf{0.49$\pm$0.06 \scriptsize{(-34.1\%)}} \\
PushCube & 0.15$\pm$0.01 \scriptsize{(-48.7\%)} & 0.01$\pm$0.00 \scriptsize{(-96.7\%)} & 0.34$\pm$0.03 \scriptsize{(-31.8\%)} & 0.26$\pm$0.02 \scriptsize{(-22.7\%)} & 0.41$\pm$0.16 \scriptsize{(-11.3\%)} & 0.09$\pm$0.03 \scriptsize{(-80.1\%)} & \textbf{\underline{0.47$\pm$0.03 \scriptsize{(+4.7\%)}}} \\
\bottomrule
\end{tabular}
\end{adjustbox}
\end{table}

\begin{table}[htbp]
\centering
\scriptsize
\caption{Medium Mo Texture Test}
\label{tab:appendix_motexturetest_medium}
\begin{adjustbox}{max width=\textwidth}
\begin{tabular}{l*{7}{c}}
\toprule
\textbf{Task} & \textbf{SAC AE} & \textbf{DrQ-v2} & \textbf{SAM-G} & \textbf{SMG} & \textbf{SADA} & \textbf{MaDi} & \textbf{SegDAC} \\
\midrule
LiftPegUpright & 0.20$\pm$0.01 \scriptsize{(-5.5\%)} & 0.20$\pm$0.00 \scriptsize{(-58.6\%)} & 0.22$\pm$0.01 \scriptsize{(-46.4\%)} & \underline{0.22$\pm$0.01 \scriptsize{(-0.1\%)}} & 0.21$\pm$0.01 \scriptsize{(-5.2\%)} & 0.20$\pm$0.00 \scriptsize{(-10.2\%)} & \textbf{0.30$\pm$0.09 \scriptsize{(-27.5\%)}} \\
PickCube & 0.05$\pm$0.02 \scriptsize{(-71.1\%)} & 0.01$\pm$0.00 \scriptsize{(-97.3\%)} & \textbf{\underline{0.11$\pm$0.02 \scriptsize{(+6.4\%)}}} & 0.03$\pm$0.01 \scriptsize{(-89.6\%)} & 0.02$\pm$0.02 \scriptsize{(-94.3\%)} & 0.01$\pm$0.00 \scriptsize{(-98.1\%)} & 0.10$\pm$0.02 \scriptsize{(-70.2\%)} \\
PokeCube & 0.08$\pm$0.01 \scriptsize{(-52.4\%)} & 0.07$\pm$0.02 \scriptsize{(-84.4\%)} & 0.07$\pm$0.03 \scriptsize{(-77.8\%)} & \underline{0.19$\pm$0.09 \scriptsize{(-12.7\%)}} & 0.06$\pm$0.02 \scriptsize{(-85.6\%)} & 0.16$\pm$0.10 \scriptsize{(-60.8\%)} & \textbf{0.26$\pm$0.05 \scriptsize{(-32.4\%)}} \\
PullCube & 0.10$\pm$0.01 \scriptsize{(-72.0\%)} & 0.08$\pm$0.01 \scriptsize{(-77.4\%)} & \underline{0.12$\pm$0.01 \scriptsize{(-1.2\%)}} & 0.28$\pm$0.06 \scriptsize{(-29.1\%)} & 0.08$\pm$0.04 \scriptsize{(-83.1\%)} & 0.12$\pm$0.07 \scriptsize{(-76.6\%)} & \textbf{0.47$\pm$0.07 \scriptsize{(-6.8\%)}} \\
PullCubeTool & 0.07$\pm$0.02 \scriptsize{(-68.1\%)} & 0.02$\pm$0.01 \scriptsize{(-96.5\%)} & \underline{0.14$\pm$0.03 \scriptsize{(-37.4\%)}} & 0.10$\pm$0.05 \scriptsize{(-54.6\%)} & 0.04$\pm$0.07 \scriptsize{(-89.1\%)} & 0.04$\pm$0.03 \scriptsize{(-93.0\%)} & \textbf{0.20$\pm$0.08 \scriptsize{(-73.1\%)}} \\
\bottomrule
\end{tabular}
\end{adjustbox}
\end{table}

\begin{table}[htbp]
\centering
\scriptsize
\caption{Medium Ro Color Test}
\label{tab:appendix_rocolortest_medium}
\begin{adjustbox}{max width=\textwidth}
\begin{tabular}{l*{7}{c}}
\toprule
\textbf{Task} & \textbf{SAC AE} & \textbf{DrQ-v2} & \textbf{SAM-G} & \textbf{SMG} & \textbf{SADA} & \textbf{MaDi} & \textbf{SegDAC} \\
\midrule
LiftPegUpright & N/A & N/A & N/A & N/A & N/A & N/A & N/A \\
PickCube & N/A & N/A & N/A & N/A & N/A & N/A & N/A \\
PokeCube & 0.16$\pm$0.02 \scriptsize{(-5.1\%)} & 0.30$\pm$0.06 \scriptsize{(-30.2\%)} & 0.31$\pm$0.08 \scriptsize{(-1.0\%)} & 0.21$\pm$0.12 \scriptsize{(-1.5\%)} & 0.36$\pm$0.07 \scriptsize{(-8.0\%)} & \textbf{\underline{0.41$\pm$0.02 \scriptsize{(-0.9\%)}}} & 0.38$\pm$0.03 \scriptsize{(-3.0\%)} \\
PullCube & 0.30$\pm$0.05 \scriptsize{(-18.4\%)} & 0.16$\pm$0.09 \scriptsize{(-57.3\%)} & 0.12$\pm$0.01 \scriptsize{(-5.8\%)} & 0.38$\pm$0.09 \scriptsize{(-2.3\%)} & \textbf{\underline{0.53$\pm$0.19 \scriptsize{(+9.2\%)}}} & 0.51$\pm$0.11 \scriptsize{(+3.8\%)} & 0.50$\pm$0.06 \scriptsize{(+0.6\%)} \\
PullCubeTool & 0.21$\pm$0.06 \scriptsize{(-2.5\%)} & 0.54$\pm$0.26 \scriptsize{(-12.6\%)} & 0.21$\pm$0.06 \scriptsize{(-4.1\%)} & 0.23$\pm$0.10 \scriptsize{(+0.5\%)} & 0.33$\pm$0.21 \scriptsize{(-10.4\%)} & \underline{0.63$\pm$0.20 \scriptsize{(+0.6\%)}} & \textbf{0.70$\pm$0.11 \scriptsize{(-5.1\%)}} \\
PushCube & 0.25$\pm$0.03 \scriptsize{(-14.7\%)} & 0.35$\pm$0.23 \scriptsize{(-24.3\%)} & \textbf{0.49$\pm$0.13 \scriptsize{(-2.1\%)}} & 0.32$\pm$0.03 \scriptsize{(-7.6\%)} & 0.45$\pm$0.10 \scriptsize{(-2.2\%)} & \textbf{\underline{0.49$\pm$0.07 \scriptsize{(+6.1\%)}}} & 0.45$\pm$0.05 \scriptsize{(+0.4\%)} \\
PlaceAppleInBowl & \underline{0.08$\pm$0.02 \scriptsize{(+12.4\%)}} & 0.14$\pm$0.03 \scriptsize{(-17.0\%)} & 0.07$\pm$0.02 \scriptsize{(-1.0\%)} & 0.11$\pm$0.06 \scriptsize{(-4.0\%)} & 0.12$\pm$0.04 \scriptsize{(-2.9\%)} & 0.11$\pm$0.05 \scriptsize{(-6.7\%)} & \textbf{0.25$\pm$0.09 \scriptsize{(-19.5\%)}} \\
TransportBox & N/A & N/A & N/A & N/A & N/A & N/A & N/A \\
\bottomrule
\end{tabular}
\end{adjustbox}
\end{table}

\begin{table}[htbp]
\centering
\scriptsize
\caption{Medium Ro Texture Test}
\label{tab:appendix_rotexturetest_medium}
\begin{adjustbox}{max width=\textwidth}
\begin{tabular}{l*{7}{c}}
\toprule
\textbf{Task} & \textbf{SAC AE} & \textbf{DrQ-v2} & \textbf{SAM-G} & \textbf{SMG} & \textbf{SADA} & \textbf{MaDi} & \textbf{SegDAC} \\
\midrule
LiftPegUpright & N/A & N/A & N/A & N/A & N/A & N/A & N/A \\
PickCube & N/A & N/A & N/A & N/A & N/A & N/A & N/A \\
PokeCube & 0.17$\pm$0.02 \scriptsize{(+3.2\%)} & 0.32$\pm$0.04 \scriptsize{(-26.8\%)} & 0.32$\pm$0.08 \scriptsize{(+2.6\%)} & \underline{0.23$\pm$0.12 \scriptsize{(+6.9\%)}} & 0.37$\pm$0.08 \scriptsize{(-5.1\%)} & \textbf{0.41$\pm$0.01 \scriptsize{(-1.0\%)}} & 0.39$\pm$0.03 \scriptsize{(-1.4\%)} \\
PullCube & 0.35$\pm$0.04 \scriptsize{(-6.4\%)} & 0.09$\pm$0.08 \scriptsize{(-75.1\%)} & 0.10$\pm$0.03 \scriptsize{(-22.5\%)} & 0.36$\pm$0.08 \scriptsize{(-7.2\%)} & \textbf{\underline{0.60$\pm$0.11 \scriptsize{(+22.6\%)}}} & 0.43$\pm$0.09 \scriptsize{(-12.3\%)} & 0.51$\pm$0.06 \scriptsize{(+2.6\%)} \\
PullCubeTool & 0.20$\pm$0.07 \scriptsize{(-9.9\%)} & 0.50$\pm$0.23 \scriptsize{(-18.8\%)} & \underline{0.25$\pm$0.08 \scriptsize{(+11.0\%)}} & 0.23$\pm$0.10 \scriptsize{(-0.7\%)} & 0.33$\pm$0.17 \scriptsize{(-10.8\%)} & 0.67$\pm$0.19 \scriptsize{(+7.2\%)} & \textbf{0.70$\pm$0.14 \scriptsize{(-5.9\%)}} \\
PushCube & 0.24$\pm$0.04 \scriptsize{(-19.3\%)} & 0.28$\pm$0.23 \scriptsize{(-38.1\%)} & 0.45$\pm$0.08 \scriptsize{(-9.8\%)} & 0.32$\pm$0.03 \scriptsize{(-7.5\%)} & \textbf{\underline{0.52$\pm$0.08 \scriptsize{(+13.6\%)}}} & 0.46$\pm$0.07 \scriptsize{(-1.2\%)} & 0.50$\pm$0.04 \scriptsize{(+12.0\%)} \\
PlaceAppleInBowl & 0.07$\pm$0.03 \scriptsize{(-1.4\%)} & 0.14$\pm$0.03 \scriptsize{(-14.0\%)} & 0.07$\pm$0.02 \scriptsize{(-1.0\%)} & 0.10$\pm$0.06 \scriptsize{(-8.4\%)} & 0.12$\pm$0.04 \scriptsize{(+0.6\%)} & \underline{0.12$\pm$0.05 \scriptsize{(+1.3\%)}} & \textbf{0.22$\pm$0.12 \scriptsize{(-29.0\%)}} \\
TransportBox & N/A & N/A & N/A & N/A & N/A & N/A & N/A \\
\bottomrule
\end{tabular}
\end{adjustbox}
\end{table}

\begin{table}[htbp]
\centering
\scriptsize
\caption{Medium Table Color Test}
\label{tab:appendix_tablecolortest_medium}
\begin{adjustbox}{max width=\textwidth}
\begin{tabular}{l*{7}{c}}
\toprule
\textbf{Task} & \textbf{SAC AE} & \textbf{DrQ-v2} & \textbf{SAM-G} & \textbf{SMG} & \textbf{SADA} & \textbf{MaDi} & \textbf{SegDAC} \\
\midrule
LiftPegUpright & 0.21$\pm$0.01 \scriptsize{(-3.3\%)} & 0.19$\pm$0.00 \scriptsize{(-60.0\%)} & \textbf{0.39$\pm$0.10 \scriptsize{(-7.8\%)}} & 0.22$\pm$0.01 \scriptsize{(-2.5\%)} & \underline{0.22$\pm$0.03 \scriptsize{(+0.6\%)}} & 0.20$\pm$0.01 \scriptsize{(-10.9\%)} & 0.38$\pm$0.13 \scriptsize{(-6.7\%)} \\
PickCube & 0.04$\pm$0.00 \scriptsize{(-79.6\%)} & 0.00$\pm$0.00 \scriptsize{(-99.1\%)} & \underline{0.11$\pm$0.02 \scriptsize{(-0.6\%)}} & 0.05$\pm$0.02 \scriptsize{(-82.8\%)} & 0.08$\pm$0.02 \scriptsize{(-80.0\%)} & 0.06$\pm$0.04 \scriptsize{(-85.6\%)} & \textbf{0.17$\pm$0.04 \scriptsize{(-48.9\%)}} \\
PokeCube & 0.08$\pm$0.03 \scriptsize{(-53.4\%)} & 0.05$\pm$0.05 \scriptsize{(-89.3\%)} & 0.21$\pm$0.04 \scriptsize{(-33.4\%)} & 0.10$\pm$0.06 \scriptsize{(-51.1\%)} & \textbf{\underline{0.29$\pm$0.05 \scriptsize{(-25.9\%)}}} & 0.23$\pm$0.12 \scriptsize{(-44.9\%)} & 0.28$\pm$0.02 \scriptsize{(-28.4\%)} \\
PullCube & 0.12$\pm$0.09 \scriptsize{(-66.4\%)} & 0.03$\pm$0.03 \scriptsize{(-90.7\%)} & \underline{0.12$\pm$0.01 \scriptsize{(-2.5\%)}} & 0.24$\pm$0.08 \scriptsize{(-39.5\%)} & 0.36$\pm$0.06 \scriptsize{(-26.2\%)} & 0.36$\pm$0.20 \scriptsize{(-27.0\%)} & \textbf{0.42$\pm$0.06 \scriptsize{(-15.9\%)}} \\
PullCubeTool & 0.12$\pm$0.09 \scriptsize{(-44.9\%)} & 0.06$\pm$0.04 \scriptsize{(-90.9\%)} & \underline{0.28$\pm$0.09 \scriptsize{(+26.8\%)}} & 0.08$\pm$0.05 \scriptsize{(-66.7\%)} & 0.19$\pm$0.10 \scriptsize{(-48.7\%)} & 0.09$\pm$0.02 \scriptsize{(-85.6\%)} & \textbf{0.46$\pm$0.07 \scriptsize{(-37.6\%)}} \\
\bottomrule
\end{tabular}
\end{adjustbox}
\end{table}

\begin{table}[htbp]
\centering
\scriptsize
\caption{Medium Table Texture Test}
\label{tab:appendix_tabletexturetest_medium}
\begin{adjustbox}{max width=\textwidth}
\begin{tabular}{l*{7}{c}}
\toprule
\textbf{Task} & \textbf{SAC AE} & \textbf{DrQ-v2} & \textbf{SAM-G} & \textbf{SMG} & \textbf{SADA} & \textbf{MaDi} & \textbf{SegDAC} \\
\midrule
LiftPegUpright & 0.21$\pm$0.01 \scriptsize{(-2.5\%)} & 0.19$\pm$0.00 \scriptsize{(-60.2\%)} & \textbf{\underline{0.43$\pm$0.14 \scriptsize{(+3.5\%)}}} & 0.22$\pm$0.01 \scriptsize{(-1.2\%)} & 0.22$\pm$0.04 \scriptsize{(-2.3\%)} & 0.22$\pm$0.01 \scriptsize{(-1.1\%)} & 0.41$\pm$0.15 \scriptsize{(-1.3\%)} \\
PickCube & 0.11$\pm$0.04 \scriptsize{(-39.8\%)} & 0.02$\pm$0.01 \scriptsize{(-96.4\%)} & \underline{0.09$\pm$0.01 \scriptsize{(-12.7\%)}} & 0.20$\pm$0.04 \scriptsize{(-36.3\%)} & 0.15$\pm$0.05 \scriptsize{(-61.2\%)} & \textbf{0.30$\pm$0.08 \scriptsize{(-30.9\%)}} & 0.23$\pm$0.06 \scriptsize{(-32.8\%)} \\
PokeCube & 0.12$\pm$0.02 \scriptsize{(-25.6\%)} & 0.03$\pm$0.02 \scriptsize{(-92.2\%)} & \underline{0.32$\pm$0.07 \scriptsize{(+2.3\%)}} & 0.18$\pm$0.10 \scriptsize{(-14.5\%)} & 0.17$\pm$0.03 \scriptsize{(-55.3\%)} & \textbf{0.37$\pm$0.07 \scriptsize{(-8.7\%)}} & 0.32$\pm$0.04 \scriptsize{(-18.8\%)} \\
PullCube & 0.06$\pm$0.03 \scriptsize{(-82.4\%)} & 0.01$\pm$0.00 \scriptsize{(-97.0\%)} & 0.12$\pm$0.01 \scriptsize{(-3.8\%)} & 0.32$\pm$0.07 \scriptsize{(-19.2\%)} & 0.37$\pm$0.08 \scriptsize{(-23.4\%)} & 0.41$\pm$0.07 \scriptsize{(-17.0\%)} & \textbf{\underline{0.49$\pm$0.04 \scriptsize{(-2.3\%)}}} \\
PullCubeTool & 0.06$\pm$0.09 \scriptsize{(-72.1\%)} & 0.05$\pm$0.09 \scriptsize{(-92.3\%)} & \underline{0.24$\pm$0.11 \scriptsize{(+6.4\%)}} & 0.13$\pm$0.07 \scriptsize{(-45.5\%)} & 0.09$\pm$0.10 \scriptsize{(-75.0\%)} & 0.54$\pm$0.19 \scriptsize{(-13.4\%)} & \textbf{0.64$\pm$0.13 \scriptsize{(-13.3\%)}} \\
PushCube & 0.19$\pm$0.04 \scriptsize{(-36.0\%)} & 0.02$\pm$0.02 \scriptsize{(-95.3\%)} & \textbf{0.45$\pm$0.05 \scriptsize{(-10.9\%)}} & 0.28$\pm$0.06 \scriptsize{(-17.9\%)} & 0.21$\pm$0.06 \scriptsize{(-54.5\%)} & 0.40$\pm$0.08 \scriptsize{(-13.0\%)} & \underline{0.43$\pm$0.07 \scriptsize{(-3.6\%)}} \\
\bottomrule
\end{tabular}
\end{adjustbox}
\end{table}

\begin{table}[htbp]
\centering
\scriptsize
\caption{Hard Camera Fov Test}
\label{tab:appendix_camerafovtest_hard}
\begin{adjustbox}{max width=\textwidth}
\begin{tabular}{l*{7}{c}}
\toprule
\textbf{Task} & \textbf{SAC AE} & \textbf{DrQ-v2} & \textbf{SAM-G} & \textbf{SMG} & \textbf{SADA} & \textbf{MaDi} & \textbf{SegDAC} \\
\midrule
LiftPegUpright & 0.19$\pm$0.00 \scriptsize{(-12.7\%)} & 0.19$\pm$0.01 \scriptsize{(-59.9\%)} & 0.19$\pm$0.00 \scriptsize{(-55.2\%)} & \textbf{\underline{0.22$\pm$0.01 \scriptsize{(-1.7\%)}}} & 0.20$\pm$0.02 \scriptsize{(-9.5\%)} & 0.20$\pm$0.01 \scriptsize{(-10.2\%)} & 0.20$\pm$0.01 \scriptsize{(-51.8\%)} \\
PickCube & 0.02$\pm$0.01 \scriptsize{(-91.6\%)} & 0.02$\pm$0.00 \scriptsize{(-95.0\%)} & 0.01$\pm$0.00 \scriptsize{(-91.0\%)} & 0.01$\pm$0.01 \scriptsize{(-96.1\%)} & 0.06$\pm$0.03 \scriptsize{(-84.0\%)} & 0.03$\pm$0.03 \scriptsize{(-93.0\%)} & \textbf{\underline{0.07$\pm$0.02 \scriptsize{(-78.7\%)}}} \\
PokeCube & 0.02$\pm$0.02 \scriptsize{(-89.9\%)} & 0.02$\pm$0.01 \scriptsize{(-95.2\%)} & 0.05$\pm$0.03 \scriptsize{(-82.8\%)} & \textbf{\underline{0.07$\pm$0.03 \scriptsize{(-68.3\%)}}} & 0.06$\pm$0.03 \scriptsize{(-85.1\%)} & 0.02$\pm$0.02 \scriptsize{(-94.2\%)} & 0.05$\pm$0.01 \scriptsize{(-86.3\%)} \\
PullCube & 0.06$\pm$0.03 \scriptsize{(-83.6\%)} & 0.01$\pm$0.00 \scriptsize{(-96.4\%)} & \textbf{\underline{0.12$\pm$0.04 \scriptsize{(-7.4\%)}}} & 0.09$\pm$0.01 \scriptsize{(-75.9\%)} & 0.09$\pm$0.03 \scriptsize{(-81.2\%)} & 0.05$\pm$0.02 \scriptsize{(-89.7\%)} & 0.11$\pm$0.01 \scriptsize{(-77.1\%)} \\
PullCubeTool & 0.02$\pm$0.03 \scriptsize{(-92.8\%)} & 0.02$\pm$0.03 \scriptsize{(-96.3\%)} & 0.03$\pm$0.02 \scriptsize{(-86.3\%)} & \underline{0.06$\pm$0.07 \scriptsize{(-74.5\%)}} & 0.03$\pm$0.03 \scriptsize{(-91.9\%)} & 0.05$\pm$0.02 \scriptsize{(-92.2\%)} & \textbf{0.16$\pm$0.04 \scriptsize{(-78.6\%)}} \\
\bottomrule
\end{tabular}
\end{adjustbox}
\end{table}

\begin{table}[htbp]
\centering
\scriptsize
\caption{Hard Camera Pose Test}
\label{tab:appendix_cameraposetest_hard}
\begin{adjustbox}{max width=\textwidth}
\begin{tabular}{l*{7}{c}}
\toprule
\textbf{Task} & \textbf{SAC AE} & \textbf{DrQ-v2} & \textbf{SAM-G} & \textbf{SMG} & \textbf{SADA} & \textbf{MaDi} & \textbf{SegDAC} \\
\midrule
LiftPegUpright & 0.19$\pm$0.00 \scriptsize{(-11.6\%)} & 0.19$\pm$0.00 \scriptsize{(-59.7\%)} & 0.19$\pm$0.01 \scriptsize{(-53.9\%)} & \textbf{\underline{0.22$\pm$0.01 \scriptsize{(-1.0\%)}}} & 0.20$\pm$0.01 \scriptsize{(-10.0\%)} & 0.19$\pm$0.00 \scriptsize{(-13.4\%)} & 0.20$\pm$0.01 \scriptsize{(-52.1\%)} \\
PickCube & 0.01$\pm$0.01 \scriptsize{(-92.1\%)} & 0.02$\pm$0.01 \scriptsize{(-96.7\%)} & 0.01$\pm$0.00 \scriptsize{(-90.9\%)} & 0.02$\pm$0.01 \scriptsize{(-92.3\%)} & 0.03$\pm$0.01 \scriptsize{(-93.1\%)} & 0.03$\pm$0.02 \scriptsize{(-94.2\%)} & \textbf{\underline{0.07$\pm$0.01 \scriptsize{(-79.4\%)}}} \\
PokeCube & 0.01$\pm$0.01 \scriptsize{(-92.1\%)} & 0.01$\pm$0.01 \scriptsize{(-97.2\%)} & 0.03$\pm$0.01 \scriptsize{(-90.7\%)} & \underline{0.05$\pm$0.04 \scriptsize{(-76.8\%)}} & \textbf{0.06$\pm$0.02 \scriptsize{(-84.9\%)}} & 0.03$\pm$0.02 \scriptsize{(-93.1\%)} & 0.05$\pm$0.00 \scriptsize{(-87.2\%)} \\
PullCube & 0.02$\pm$0.01 \scriptsize{(-94.5\%)} & 0.01$\pm$0.01 \scriptsize{(-98.3\%)} & \textbf{\underline{0.07$\pm$0.03 \scriptsize{(-41.3\%)}}} & 0.02$\pm$0.03 \scriptsize{(-95.5\%)} & 0.05$\pm$0.02 \scriptsize{(-88.9\%)} & 0.02$\pm$0.01 \scriptsize{(-95.3\%)} & 0.05$\pm$0.04 \scriptsize{(-89.1\%)} \\
PullCubeTool & 0.02$\pm$0.03 \scriptsize{(-92.3\%)} & 0.01$\pm$0.01 \scriptsize{(-98.1\%)} & 0.00$\pm$0.01 \scriptsize{(-97.8\%)} & \underline{0.03$\pm$0.05 \scriptsize{(-86.7\%)}} & 0.03$\pm$0.02 \scriptsize{(-90.9\%)} & 0.01$\pm$0.01 \scriptsize{(-97.6\%)} & \textbf{0.09$\pm$0.05 \scriptsize{(-87.3\%)}} \\
\bottomrule
\end{tabular}
\end{adjustbox}
\end{table}

\begin{table}[htbp]
\centering
\scriptsize
\caption{Hard Ground Color Test}
\label{tab:appendix_groundcolortest_hard}
\begin{adjustbox}{max width=\textwidth}
%
\end{adjustbox}
\end{table}

\begin{table}[htbp]
\centering
\scriptsize
\caption{Hard Ground Texture Test}
\label{tab:appendix_groundtexturetest_hard}
\begin{adjustbox}{max width=\textwidth}
%
\end{adjustbox}
\end{table}

\begin{table}[htbp]
\centering
\scriptsize
\caption{Hard Lighting Color Test}
\label{tab:appendix_lightingcolortest_hard}
\begin{adjustbox}{max width=\textwidth}
%
\end{adjustbox}
\end{table}

\begin{table}[htbp]
\centering
\scriptsize
\caption{Hard Lighting Direction Test}
\label{tab:appendix_lightingdirectiontest_hard}
\begin{adjustbox}{max width=\textwidth}
%
\end{adjustbox}
\end{table}

\begin{table}[htbp]
\centering
\scriptsize
\caption{Hard Mo Color Test}
\label{tab:appendix_mocolortest_hard}
\begin{adjustbox}{max width=\textwidth}
%
\end{adjustbox}
\end{table}

\begin{table}[htbp]
\centering
\scriptsize
\caption{Hard Ro Color Test}
\label{tab:appendix_rocolortest_hard}
\begin{adjustbox}{max width=\textwidth}
%
\end{adjustbox}
\end{table}

\begin{table}[htbp]
\centering
\scriptsize
\caption{Hard Ro Texture Test}
\label{tab:appendix_rotexturetest_hard}
\begin{adjustbox}{max width=\textwidth}
%
\end{adjustbox}
\end{table}

\begin{table}[htbp]
\centering
\scriptsize
\caption{Hard Table Texture Test}
\label{tab:appendix_tabletexturetest_hard}
\begin{adjustbox}{max width=\textwidth}
%
\end{adjustbox}
\end{table}

\end{document}